\documentclass[cornellheadings,smallerheadings]{fiu}

\usepackage{graphicx}
\usepackage{hyperref}
\usepackage{algorithm}
\usepackage{algorithmic}
\usepackage{amsmath} 
\usepackage[dvipsnames]{xcolor}
\usepackage{array,booktabs}
\usepackage{multirow}
\usepackage{makecell}
\usepackage{booktabs}
\usepackage{wrapfig}
\usepackage{caption}
\usepackage{subcaption}
\usepackage{amssymb}
\usepackage{amsthm}

\usepackage{times}
\usepackage{latexsym}
\usepackage{tcolorbox}
\usepackage[table]{xcolor}
\usepackage{enumitem}
\usepackage{tikz}
\usetikzlibrary{tikzmark}
\usepackage{float}
\usepackage{amssymb}
\usepackage{dsfont}
\usepackage{pifont}
\usepackage{bbding}
\usepackage{xspace}

\definecolor{yellow}{rgb}{0.74, 0.56, 0}
\definecolor{purple}{rgb}{0.32, 0.09, 0.98} 
\definecolor{red}{rgb}{0.81, 0.09, 0.13}
\definecolor{blue}{cmyk}{0.95,0.0,0.2,0.2}
\definecolor{lightyellow}{cmyk}{0.01,0.0,0.2,0.01}
\definecolor{lightblue}{cmyk}{0.1,0.0,0.02,0.02}
\definecolor{mypurple}{RGB}{236, 203, 250}
\definecolor{viridis3}{RGB}{33, 145, 140}

\usepackage[edges]{forest}

\definecolor{mycitecolor}{RGB}{0, 0, 128}
\definecolor{myrefcolor}{RGB}{199, 29, 29}
\definecolor{hidden-draw}{RGB}{0,0,0}
\definecolor{hidden-pink}{rgb}{0.98, 0.94, 0.75}
\definecolor{level0}{rgb}{0.67, 0.88, 0.69}
\definecolor{level1}{rgb}{0.98, 0.92, 0.84}
\definecolor{level2}{rgb}{0.8, 0.8, 1.0}
\definecolor{level3}{rgb}{1.0, 0.71, 0.76}
\usepackage{xcolor}

\tikzset{mycircled/.style={circle,draw=black,fill=mypurple,inner sep=0.1em,line width=0.1em, scale=0.8}} 

\newcommand{\hyperrag}{\texttt{Hypercube-RAG}\xspace}
\newcommand{\codicast}{\textsc{CoDiCast}\xspace}
\newcommand{\walef}{\textsc{WaLeF}\xspace}
\newcommand{\WaLeF}{\textsc{WaLeF}\xspace}
\newcommand{\fidlar}{\textsc{FIDLAr}\xspace}
\newcommand{\ul}{\underline}

\title{Accurate, Efficient, and Explainable Deep Learning Approaches for Environmental Science Problems}
\author{Jimeng Shi}
\conferraldate{May}{2025}
\defensedate{June 25, 2025}
\advisor{Giri Narasimhan}
\memberone{Leonardo Bobadilla}
\membertwo{Ananda Mondal}
\memberthree{Arturo Leon}
\memberfour{Shaowen Wang}
\degreefield{Computer Science}
\college{College of Engineering and Computing}
\collegedean{Dean Inés Triay}
\gradschooldean{Andrés G. Gil}

\begin{document} 

\setcounter{page}{1}
\pagenumbering{roman}
\pagestyle{plain}


\maketitle

\makeapproval{3}

\makecopyright

\begin{dedication}
I am deeply grateful to my family for their unwavering love and support throughout every stage of my life. Their belief in me has been a constant source of strength and motivation. 
\end{dedication}

\begin{acknowledgments}
The completion of this dissertation would not have been possible without the support and encouragement of many individuals. 
I would like to express my sincere gratitude to my advisor, Dr. Giri Narasimhan, for his patience, advice, mentorship, and support throughout my Ph.D. journey. Dr. Giri has taught me how to become an excellent researcher, ranging from idea brainstorming and improvement, paper writing and polishing, research collaboration, to research dissemination.
More notably, his guidance extended beyond research discussions, offering valuable advice on both academic matters and important life decisions.
I am honored to be one of his advisees, and I hope to one day become a knowledgeable and responsible researcher, just like him.
I would also like to thank other committee members, Dr. Leonardo Bobadilla, Dr. Ananda Mondal, Dr. Arturo Leon, and Dr. Shaowen Wang, for their instrumental support and suggestions.
Appreciations should also go to brilliant collaborators, who greatly enrich my research experience.. Their names are: Dr. Kalai Mathee, Dr. Jiawei Han, Dr. Vitalii Stebliankin, Dr. Trevor Cickovski, Dr. Jayantha Obeysekera, Dr. Upmanu Lall, Dr. Zeda Yin
Dr. Zhaonan Wang, Dr. Dongsheng Luo, Dr. Yanzhao Wu, Dr. Sundararaman Gopalakrishnan, Azam Shirali, Rahuul Rangaraj, Rukmangadh Myana,  Arpit Mehta, Saad Alqarni, Parshatd Govindasami, Nimmi Arunachalam, Ricky Ruiz, Anne Nguyen,
 Bowen Jin, Wei Hu, Sizhe Zhou, Runchu Tian, Xu Zheng, Enshi Zhang, Zichuan Liu, Mengjie Zhang, Lei Yan.
Finally, I am grateful to the Institute for Geospatial Understanding through an Integrative Discovery Environment (I-GUIDE), a large research consortium funded by the National Science Foundation (NSF), which consistently and generously supports my Ph.D. research work.
THANK YOU ALL!
\end{acknowledgments}

\begin{abstract}
Environmental science plays a pivotal role in safeguarding natural ecosystems and human well-being.
This domain encompasses a wide range of critical research challenges, such as environmental prediction, monitoring, and management, often driven by large-scale, heterogeneous data.
In the era of big data, artificial intelligence (AI) has emerged as a transformative tool, offering powerful capabilities for learning patterns from data and supporting decision-making. 
This dissertation aims to leverage and develop AI-based approaches tailored to address complex problems in environmental science with a goal of \emph{Environmental Intelligence}. Three specific problems are studied.

Firstly, we focus on the prediction and management of floods in coastal river systems. 
Conventional physics-based models are often computationally intensive, limiting their effectiveness in real-time applications. 
To overcome this limitation, we propose a deep learning (DL)-based model, \texttt{WaLeF}, for water level forecasting, and a forecast-informed DL-based model, \fidlar, to manage water levels.
Evaluated in a flood-prone coastal system in South Florida, where extreme rainfall and sea level fluctuations are prevalent, \fidlar demonstrates superior performance over baseline methods in both accuracy and computational efficiency, while providing interpretable model outputs.

Secondly, we focus on global weather prediction, a task challenged by massive data scale and limited computational efficiency. 
Traditional physics-based methods are often deterministic and computationally intensive, limiting their scalability and applicability in real-world applications.
To address this, we propose \codicast, a conditional diffusion model tailored for probabilistic weather forecasting. Originally used in generative AI, diffusion models are adapted here for predictive tasks.
Experiments show that \codicast achieves accurate, efficient forecasts with explicit uncertainty quantification.

Lastly, we target the scientific question-answering (QA) in environmental science.
When answering \emph{in-domain} questions, large language models (LLMs) often suffer from hallucinations and factual inaccuracies due to out-of-date and limited domain knowledge.
To alleviate this phenomenon, retrieval-augmented generation (RAG) was designed to retrieve domain-specific knowledge. However, existing RAG methods tend to perform poorly in either accuracy, efficiency, or explainability.
We propose \hyperrag, which is an RAG built on top of a structured knowledge representation framework (\emph{text cube}).
Experiments verify that it exhibits three properties aforementioned simultaneously.

Overall, this dissertation shows how three distinct AI techniques can be effectively applied to environmental science problems, highlighting the vital role of AI in this field.


\end{abstract}

\contentspage

\tablelistpage

\figurelistpage

\normalspacing
\setcounter{page}{1}
\pagenumbering{arabic}
\pagestyle{cornell}

\chapter{INTRODUCTION}
Environmental science is essential for protecting natural ecosystems and ensuring human well-being, particularly in the context of climate change and escalating environmental risks \cite{pecl2017biodiversity}. This field involves a diverse set of critical challenges, including environmental prediction, monitoring, and management, often characterized by large-scale, heterogeneous data sources \cite{alotaibi2024artificial, sun2019can}. 
With the emergence of the big data era, artificial intelligence (AI) has been a transformative force, offering powerful tools for uncovering complex patterns and enhancing data-driven decision-making \cite{duan2019artificial}. 
It is therefore vital to investigate how AI can be effectively leveraged to address pressing issues in environmental science, a paradigm we refer to broadly as \texttt{Environmental Intelligence}.
This dissertation explores three representative in-domain problems under this framework.

\emph{Floods Prediction and Management.}
Flooding is one of the most frequent and devastating natural hazards worldwide, causing substantial damage to infrastructures, ecosystems, and human lives \cite{ikram2024flood, glago2021flood}. 
In the United States, floods have caused an average of nearly \$8 billion in physical damages and approximately 100 fatalities annually over the past 15 years \cite{van2024economic}
Timely flood prediction plays a vital role in disaster preparedness, risk mitigation, and informed decision-making.
The World Bank has estimated that early warning systems can reduce flood-related fatalities by up to 43\% and economic costs by 35–50\% \cite{rogers2010global} and save an average of 23,000 lives per year \cite{rogers2010global, nearing2024global} in developing countries.
Extensive efforts have been made by using conventional physics-based models for flood simulation, including tools such as SWMM \cite{babaei2018urban}, HEC-RAS \cite{leon2016controlling}, and MIKE \cite{bisht2016modeling}.
Given infinite time and resources, floods can sometimes be controlled by making smart use of hydraulic structures in river systems.
For instance, operating these structures to pre-release water before storms can significantly reduce the risk of flooding.
However, as shown in this dissertation, controlling those hydraulic structures optimally can be framed as an optimization problem.
Slow simulation speeds for the water level problems pose significant challenges for the much harder problem of flood control, which often requires running thousands of water level simulations to identify ``optimal'' control strategies \cite{qi2021review}.
However, these models often face limitations related to computational efficiency and adaptability \cite{shi2023deep, jenkins2023physics}.
Therefore, the critical need for fast simulation of real-time flood management is clear.
The predictions of water levels in river systems are often dictated by the accurate forecasts of weather and rain, which leads us to the next problem tackled in this dissertation.

\emph{Global Weather Prediction.}
Accurate weather forecasting is crucial for a wide range of societal activities, from daily planning to disaster preparedness \cite{merz2020impact, shi2024fidlar}. 
Traditional physics-based numerical weather prediction (NWP) methods forecast weather by modeling them as systems of differential equations that describe the complex interactions between the variables describing phenomena in the atmosphere, land, and ocean systems \cite{price2023gencast, nguyen2023climax}. 
However, they typically yield a single deterministic outcome, failing to capture uncertainty.
While such physics-based ensemble forecasts effectively model the weather uncertainty by generating multiple runs \cite{palmer2019ecmwf, leinonen2023latent}, 
ensemble-based NWP models face two major challenges: restrictive physical assumptions and high computational cost \cite{rodwell2007using}.
Despite the notable efficiency of machine learning (ML)-based methods, such as Pangu \cite{bi2023accurate}, GraphCast \cite{lam2023learning}, ClimaX \cite{nguyen2023climax}, ForeCastNet \cite{pathak2022fourcastnet}, most of them are deterministic \cite{kochkov2024neural}, falling short in capturing the uncertainty in weather forecasts \cite{jaseena2022deterministic}.
Therefore, probabilistic weather predictions are needed for making informed decisions.

Finally, we consider the problem of building accurate and efficient information retrieval systems for environmental data with which one can interact using natural language.

\emph{In-domain Scientific Question-Answering.}
We hypothesize that policy-makers and domain scientists can make more informed decisions when equipped with up-to-date domain knowledge on critical issues such as hurricanes and aging dams \cite{rokooei2022perceptions, wang2025remflow}. 
Scientific question-answering powered by large language models (LLMs) offers a valuable solution. 
However, existing LLMs often generate hallucinated or inaccurate responses due to outdated or insufficient domain-specific knowledge \cite{huang2025survey}. 
Retrieval-augmented generation (RAG) techniques have emerged as promising approaches \cite{huang2025survey, ding2024automated}.
Yet, current RAG methods often fall short in at least one of the three aspects of accuracy, efficiency, and explainability. This highlights the need for a more robust and domain-adapted RAG framework.

Overall, this dissertation aims to leverage and develop \emph{accurate}, \emph{efficient}, and \emph{interpretable} AI-based approaches tailored to address complex problems in environmental science, leading us toward \emph{Environmental Intelligence}.

\section{Motivations}
To achieve \emph{Environmental Intelligence}, we are motivated to address challenges of traditional approaches in environmental science by leveraging and developing advanced \emph{deep learning} methods.

\subsection{Flood Prediction and Mitigation}
With regard to floods, the basic need is to determine the water level at any point of interest (i.e., a control point), given all the water inflows into the system.
This general basic problem would solve the \textbf{flood prediction problem} since we can determine the location, intensity, and duration of a flood event (defined appropriately in terms of water levels) in a given river system.
The advanced problem is the \emph{inverse} problem .
In other words, we want to know the setting that would achieve a specified water level. 
Stated differently, the advanced need is to determine the settings of the hydraulic structures (i.e., interventions) needed to achieve a set of desired water levels at various control points in the river system.
Note that if the desired water levels are not possible, we want to be able to achieve as close to the desired water level as possible.
Again, solving the advanced problem can be thought of as solving the \textbf{flood mitigation problem}, if the desired water levels are to not exceed the stated flood thresholds at the control points.
Thus, in anticipation of an impending storm, we can solve the flood prediction problem to predict whether or not floods are expected to occur. 
If the answer is yes, the flood mitigation problem is to determine the precise intervention that would prevent, or at least mitigate the floods.

Traditional flood prediction/simulation approaches typically rely on physics-based hydrological and hydraulic models, which simulate water flow dynamics using equations such as the Saint-Venant or shallow water equations \cite{busto2022staggered}. 
While these models offer outputs that are consistent with the laws of physics, they often require extensive calibration, high-resolution input data, and significant computational resources. These limitations may hinder their operational scalability \cite{chen2023physics, vadyala2022review, jenkins2023physics}.
This also poses significant challenges for flood mitigation, which often requires running thousands of such simulations to identify ``optimal'' control strategies \cite{qi2021review}. 
Furthermore, rule-based methods \cite{sadler2019leveraging} formulate control schedules based on insights gained from historically observed data. 
However, the established rules represent the collective wisdom gathered over decades of experience in managing specific river systems, exposing vulnerabilities while dealing with extremely rare events and poor generalization for complex river systems in other regions \cite{schwanenberg2015open}. 
Therefore, accurate, efficient, robust flood prediction and mitigation approaches are essential.
This dissertation offers accurate, efficient, and explainable AI-based methods for these problems.

\subsection{Weather Prediction}
\emph{Weather Prediction} refers to predicting atmospheric variables of interest for future time point(s), 
given observation(s) from a window from the recent past.
Physics-based models, General Circulation Models (GCMs) \cite{ravindra2019generalized}, and Numerical Weather Prediction (NWP) models \cite{coiffier2011fundamentals}, have been the cornerstone of weather prediction. These models simulate future weather scenarios by numerically approximating solutions to the differential equations that govern the complex physical dynamics of interconnected atmospheric, terrestrial, and oceanic systems \cite{nguyen2023climax}.

Despite significant advancements, these physics-based models face notable challenges. 
Firstly, the accuracy of conventional NWP models is dictated by the variables used and by the spatial and temporal resolutions used. Finer resolutions allow for better representation of mesoscale and localized phenomena (e.g., convection, topographic effects, or coastal systems), but significantly increase the computational cost \cite{al2010review}.
Secondly, subgrid-scale parameterizations introduce significant uncertainty. Many atmospheric processes occur at scales too small to be explicitly resolved and are instead approximated using empirical or simplified physical models. These parameterizations are often region-specific and may fail to generalize across varying conditions \cite{palmer2005representing}. 
Lastly, a single physics-based model typically produces deterministic forecasts once initial conditions are fixed, falling short of capturing uncertainties in the predictions \cite{bulte2024uncertainty}.
While ensemble forecasts, to some extent, help mitigate this phenomenon by generating probabilistic outputs \cite{gneiting2005weather}, the first two challenges persist.
Thus, accurate and efficient weather prediction with uncertainty quantification is critically needed.

\subsection{In-domain Environmental Question-Answering}
While current large language models (LLMs) are pre-trained on massive general-purpose datasets, they often underperform on in-domain questions that require specialized knowledge. 
To address this limitation, incorporating external knowledge into LLMs is essential. 
Retrieval-augmented generation (RAG) has emerged as a promising solution, enabling LLMs to generate more accurate and contextually grounded responses by leveraging retrieved documents \cite{huang2025survey,ding2024automated}.
However, conventional RAG systems based on semantic similarity often struggle to retrieve concise yet highly relevant information for knowledge-intensive domains.
The presence of irrelevant documents (i.e., retrieval ``noise'') can mislead LLMs, leading to hallucinated or counterfactual responses \cite{krishna2021hurdles, kang2024improving}. 
Furthermore, because these systems operate in a high-dimensional embedding space, it becomes difficult to justify why certain documents were selected, posing challenges for transparency and trust in high-stakes applications \cite{ji2019visual}.
While \textit{graph-based} RAG methods can retrieve documents by returning a subgraph, traversing such subgraphs introduces a significant risk of information overload, as the expanding neighborhoods may include substantial amounts of irrelevant themes \cite{wang2024graph}.
Graph traversal also poses significant scalability bottlenecks due to its inefficient searching process.
Therefore, an accurate, efficient, explainable RAG system is needed to address the above limitations.

We tested the proposed QA systems on datasets related to hurricanes and aging dams, as they are critical applications for informed decision-making and infrastructure planning \cite{shirzaei2025aging}. 
However, such knowledge is often buried in unstructured historical documents and reports.  
A question-answering (QA) framework with large language models (LLMs) can achieve information extraction and organization, providing policy-makers with timely, accessible, and context-specific insights to support risk assessment and proactive mitigation strategies.

\section{Research Contributions}
The contributions of this dissertation are threefold.
First, it presents deep learning–based models for both flood prediction and mitigation.
Second, it introduces a diffusion-based framework for probabilistic weather forecasting, enabling robust estimation of uncertainty in extreme weather events.
Finally, the dissertation proposes a novel retrieval-augmented generation (RAG) system designed to enhance the capability of large language models (LLMs) in answering domain-specific questions with greater precision and relevance.
Overall, the major contributions are summarized below.
\paragraph{\fidlar.} 
First, an accurate and efficient novel DL model for water level forecasting (WaLeF) is proposed, which replaces the role of physics-based models for flood prediction/simulation given key driving factors such as rainfall, tidal conditions, and hydraulic control operations. 
Then a forecast-informed DL approach for flood mitigation (\fidlar) is introduced to seek the ``optimal'' controls on hydraulic structures (i.e., gates and pumps).
Experiments were conducted for data from a flood-prone coastal area in South Florida, where floods are a result of multiple driving factors, such as rainfall, sea levels, and inappropriate hydraulic operations.
Results show that \fidlar is several orders of magnitude faster than currently used physics-based approaches while outperforming baseline methods with improved water pre-release schedules. 
Meanwhile, this framework presents the explainability of the proposed actions.
Open-source code is available from \url{https://github.com/JimengShi/FIDLAR}.

\paragraph{\codicast.} 
Current physics-based numerical weather prediction methods are computationally inefficient and cannot achieve uncertainty quantification.
To address these challenges, we propose a conditional diffusion-based model, \codicast, for global weather prediction. 
It is conditioned on recent past observations while probabilistically modeling forecast uncertainty. Although diffusion models are originally designed for generative tasks starting from random noise, we tailor this framework in an innovative manner for conditional predictive tasks, enabling accurate and uncertainty-aware weather forecasting.
In addition, the cross-attention mechanism was explored to effectively integrate the conditioning into the denoising process to guide the generation tasks.
Extensive experiments were conducted on a decade of ERA5 reanalysis data from the European Centre for Medium-Range Weather Forecasts (ECMWF). Results demonstrate that \codicast achieves a useful trade-off between accuracy, efficiency, and uncertainty when compared against several state-of-the-art models.
Open-source code is available from \url{https://github.com/JimengShi/CoDiCast}.

\paragraph{\hyperrag.} 
Conventional LLMs such as ChatGPT tend to hallucinate when asked for answers to specific questions. 
RAG-based extensions can help alleviate the problem of hallucinations, but are inefficient and error-prone.
To overcome these limitations, a multidimensional knowledge structure (\emph{text cube}) was employed. 
Text cubes help represent the information in unstructured documents in a structured manner by allocating documents into cube cells along multiple pre-defined dimensions.
Built on the cube structure, a simple yet accurate, efficient, and explainable RAG method, \hyperrag, was proposed. 
Experiments on scientific datasets, including hurricanes and aging dams, demonstrate that \hyperrag outperforms other baseline methods in terms of accuracy and efficiency when it comes to \emph{in-domain} question-answering.
More importantly, we demonstrate the explainability, scalability, and noise resilience of our method.
It provides meaningful insights during the retrieval process and performs efficiently and robustly on retrieval tasks from large-scale and noisy corpora. 
Open-source code is available from \url{https://github.com/JimengShi/Hypercube-RAG}.

\section{Organization of the Dissertation}
We conclude this introductory chapter with a roadmap of the dissertation.

Chapter 2 covers the foundational concepts relevant to this dissertation, including a literature review of ML/DL models and their applications in flood prediction, mitigation, and weather forecasting. It also surveys current retrieval-augmented generation (RAG) methods for \emph{in-domain} question-answering tasks.

Chapter 3 introduces \fidlar, a forecast-informed DL approach for flood prediction and mitigation. After reviewing the existing approaches and their key limitations, a DL-based framework consisting of two cascaded models, \texttt{Flood Manager} and \texttt{Flood Evaluator}, is described. Then, detailed training of \fidlar is discussed, followed by experimental results and analysis regarding accuracy, efficiency, and explainability.

Chapter 4 presents \codicast, a conditional diffusion model for weather forecasting with uncertainty quantification. After reviewing the existing approaches and their key limitations, the methodology of the conditional diffusion model is described. Then, detailed training is introduced, followed by experimental results and analysis in terms of accuracy, efficiency, and uncertainty.

Chapter 5 starts with the limitations of the existing RAG methods for \emph{in-domain} question-answering tasks. Then it reviews the related work on document indexing with a multidimensional knowledge structure (text cube).
Then hypercube-based RAG method, \hyperrag, is proposed, ranging from hypercube construction and its usage for retrieval. Lastly, experimental results and analysis, and case studies on accuracy, efficiency, and explainability are discussed in detail.

Finally, Chapter 6 wraps up the dissertation by providing a summary, conclusions, and recommendations for future research.

\chapter{BACKGROUND AND REVIEW}		
\label{sec:backgrond_reivew}
Machine learning (ML) and deep learning (DL) fall within the broader area of artificial intelligence (AI), focusing on developing algorithms capable of learning patterns from data and making informed decisions or predictions \cite{sarker2021machine}. 
Most of these methods operate in two primary phases, under the assumption that all data samples are independently and identically distributed (i.i.d.) \cite{janiesch2021machine}.
In the training phase, a model is trained on a dataset to learn the knowledge and patterns. 
In the test (inference) phase, the trained model created in the training phase is used to test its learning or to make a decision on new, previously unseen inputs \cite{sarker2021deep}.

Learning models are broadly categorized based on the nature of the training/learning methods, that is, how the learning signal or feedback (e.g., labels) is provided to the system for each input sample \cite{jiang2020supervised, singh2016review}.
The primary learning paradigms include: 
(1) \emph{Supervised learning}, the model is trained on a labeled dataset, where each input is paired with a corresponding ground-truth output. This paradigm is widely used in classification and regression tasks.
(2) \emph{Unsupervised learning}, the model is provided only with input data without explicit labels, and must discover underlying patterns or structures. Typical tasks include clustering, dimensionality reduction, and anomaly detection.
(3) \emph{Semi-supervised learning} lies between supervised and unsupervised learning. It utilizes a small amount of labeled data along with a large amount of unlabeled data to improve learning performance.
(4) \emph{Self-supervised learning} is an emerging paradigm where the model generates supervisory signals from the input data itself, often by learning to predict masked parts of the input as part of its training. The resulting model is often referred to as a pre-trained model and may be further prompted or fine-tuned for performing specialized tasks. 
In this chapter, we mainly focus on the regression tasks due to their high relevance in dealing with environmental questions. 

\section{Machine Learning Models}
%
In statistics and ML, \emph{regression analysis} is the task of modeling the relationship between a dependent variable and a set of independent variables, making it possible to infer or predict the value of the dependent variable based on the values of the independent variables.

\subsection{Linear Regression}
Let us assume a dataset consisting of $m$ samples from an $n$-dimensional feature space. 
Each sample $\mathbf{x}^i = (x_1^i, x_2^i, \dots, x_n^i) \in \mathbb{R}^n$ is associated with a scalar output $y^i \in \mathbb{R}$. 
The goal is to learn a weight vector $\mathbf{w} = (w_1, w_2, \dots, w_n) \in \mathbb{R}^n$ such that every predicted output is given as a linear combination of its input features:
\begin{equation}
y = w_1 \cdot x_1 + w_2 \cdot x_2 + \dots + w_n \cdot x_n = \mathbf{w} \cdot \mathbf{x}^T,
\end{equation}
where $\mathbf{x}$ denotes an input vector and $\mathbf{w} \cdot \mathbf{x}^T$ represents the inner product of the weights and the input features.

\subsection{Logistic Regression} 
Logistic regression is a widely used statistical model for binary classification problems, where the target variable takes on one of two possible outcomes (e.g., 0 or 1). 
The model estimates the probability that a given input vector belongs to the positive class by applying the logistic sigmoid function to a linear combination of the input features, as shown in Eq.~(\ref{eq:sigmoid}). 
This probability is then thresholded to output the final label. 
Despite its name, logistic regression performs classification rather than regression, as it models the likelihood of class membership rather than predicting a continuous output.
\begin{equation}
    \sigma(y) = \frac{1}{1 + e^{-y}}.
\label{eq:sigmoid}
\end{equation}

\subsection{Support Vector Machines}
Support Vector Machines (SVMs) are powerful supervised learning models used primarily for classification tasks. The core idea of an SVM is to find an optimal hyperplane that maximally separates data points of different classes in a high-dimensional feature space. 
For linearly separable data, the SVM aims to find the hyperplane with the largest margin between the two classes.

Given a dataset of $m$ points $\{(\mathbf{x}^i, y^i)\}_{i=1}^{m}$, where $\mathbf{x}^i \in \mathbb{R}^n$ in a $n$-dimensional space and class labels $y^i \in \{-1, 1\}$, the problem is formulated to find a hard-margin SVM (i.e., hyperplane) to separate these points:
\begin{equation}
\min_{\mathbf{w}, b} \frac{1}{2} \|\mathbf{w}\|^2
\end{equation}
$$
\text{subject to } y^i(\mathbf{w}^\top \mathbf{x}^i + b) \geq 1, \quad \forall i = 1, \dots, m,
$$
where $\mathbf{w}$ is the weight vector orthogonal to the hyperplane, and $b$ is the bias term. The objective is to minimize the norm of $\mathbf{w}$, which corresponds to maximizing the margin between the classes.

In practice, data is often not linearly separable, so the soft-margin SVM introduces slack variables $\xi^i$ to allow for some misclassifications:

\begin{equation}
\min_{\mathbf{w}, b, \boldsymbol{\xi}} \frac{1}{2} \|\mathbf{w}\|^2 + C \sum_{i=1}^{n} \xi^i
\end{equation}
$$
\text{subject to } y^i(\mathbf{w}^\top \mathbf{x}^i + b) \geq 1 - \xi^i, \quad \xi^i \geq 0, \quad \forall i,
$$
where $C > 0$ is a regularization parameter that balances the trade-off between maximizing the margin and minimizing classification errors.

To handle non-linear classification problems, SVMs utilize kernel functions $\mathcal{K}(\mathbf{x}^i, \mathbf{x}^j)$ to implicitly map the input features into a higher-dimensional space after which the separating hyperplane is sought. Common kernels include:

\begin{itemize}
    \item \textbf{Linear kernel} \cite{ghosh2019study}: $\mathcal{K}(\mathbf{x}^i, \mathbf{x}^j) = (\mathbf{x}^i)^\top \mathbf{x}^j$
    \item \textbf{Polynomial kernel} \cite{vinge2019understanding}: $\mathcal{K}(\mathbf{x}^i, \mathbf{x}^j) = ((\mathbf{x}^i)^\top \mathbf{x}^j + c)^d$
    \item \textbf{Radial basis function (RBF) kernel} \cite{ding2021random}: $\mathcal{K}(\mathbf{x}^i, \mathbf{x}^j) = \exp(-\gamma \|\mathbf{x}^i - \mathbf{x}^j\|^2)$
\end{itemize}

SVMs are valued for their robustness and strong generalization capabilities, especially in high-dimensional nonlinear spaces \cite{erfani2016high}.

\subsection{Decision Trees} 
Decision Trees are a widely used class of supervised learning algorithms suitable for both classification and regression tasks \cite{quinlan1986induction}. The model represents decisions and their possible consequences in a tree-like structure, where each internal node corresponds to a feature-based test, each branch represents an outcome of the test, and each leaf node denotes a predicted label or value.
The learning process involves recursively partitioning the input space based on feature values to maximize the separation between classes (in classification) or minimize variance (in regression). Popular criteria for selecting splits include Gini impurity and information gain (based on entropy) \cite{tangirala2020evaluating}.

Formally, given a dataset $\mathcal{D}$ with input features, the goal at each node is to select the feature that maximizes the information gain:
\begin{equation}
\text{Information Gain} = \mathcal{I}(\mathcal{D}) - \sum_{j \in \{L, R\}} \frac{|\mathcal{D}_j|}{|\mathcal{D}|} \mathcal{I}(\mathcal{D}_j),
\end{equation}
where $\mathcal{I}$ is an impurity function (e.g., entropy), and $\mathcal{D}_L$ and $\mathcal{D}_R$ are the left and right subsets resulting from the split.

Decision Trees are interpretable and easy to visualize, making them useful in applications where model transparency is important. However, they are prone to overfitting, especially when deep trees are grown. This limitation is often addressed by pruning strategies \cite{mohamed2012comparative, esposito1997comparative} or ensemble methods such as Random Forests \cite{breiman2001random} and Gradient Boosted Trees \cite{sachdeva2021comparison}.


\subsection{K-Nearest Neighbors} 
K-Nearest Neighbors (KNN) \cite{peterson2009k, kramer2013k} is a simple yet effective instance-based learning algorithm used for both classification and regression tasks. 
In classification, it assigns a class label to a query point by identifying the $k$ nearest neighbors in the feature space and selecting the majority class among them. KNN is a \textit{non-parametric} method, meaning it does not make explicit assumptions about the underlying data distribution and instead relies entirely on the training data for predictions.

However, the performance of KNN is highly dependent on the choice of the \textit{distance metric} (e.g., Euclidean, Manhattan, cosine) and the value of $k$. A small $k$ may lead to overfitting and sensitivity to noise, while a large $k$ can smooth out decision boundaries, potentially resulting in underfitting. 
Furthermore, because KNN requires computing the distance between the query and all training points at inference time, it can be computationally expensive for large datasets.
Despite these limitations, KNN remains popular due to its simplicity, interpretability, and strong performance in well-structured, low-dimensional data spaces \cite{zhang2016introduction}.

\section{Deep Learning Models}
Deep learning (DL) models typically refer to neural network architectures composed of multiple layers, where each layer consists of interconnected computational units called neurons. These models are designed to automatically learn hierarchical representations of data through successive transformations. 
Unlike traditional machine learning algorithms that often rely on manually engineered features, DL models can capture complex and nonlinear relationships directly from raw input data, such as images, text, or time series.

\subsection{Multilayer Perceptron}
\begin{figure}[ht]
\centering
    \includegraphics[width=0.75\columnwidth]{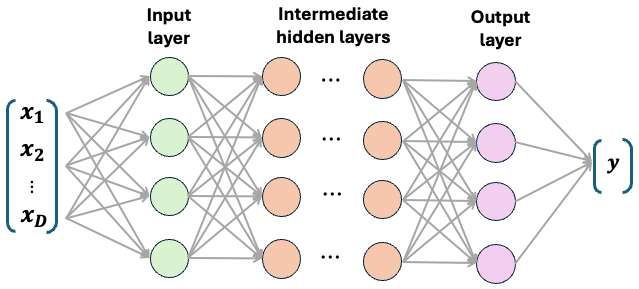} 
\caption{Schematic representation of the MLP architecture with $l$ layers, including one input layer, one output layer, and 
$l-2$ intermediate hidden layers.}
\label{fig:mlp}
\end{figure}

Multilayer Perceptron (MLP) is part of a family of fully connected feedforward neural networks. MLP usually consists of at least three layers: one input layer, one or more intermediate hidden layers, and one output layer. 
By stacking multiple layers of interconnected neurons (Figure \ref{fig:mlp}, circles), where each layer applies a non-linear activation function (e.g., sigmoid function \cite{han1995influence}), MLP can learn complex, non-linear mappings from inputs to outputs \cite{del2021auto}. 
The composition of these non-linear functions through the network allows it to approximate highly intricate functions. 
Mathematically, the computation of Figure \ref{fig:mlp} can be described as follows:
\begin{align*}
    \mathbf{h}_1 &= \sigma(W_1 \cdot \mathbf{x} + \mathbf{b}_1); \\
    \mathbf{h}_2 = \sigma(W_2 \cdot \mathbf{h}_1 + \mathbf{b}_2); \ 
    \mathbf{h}_3 &= \sigma(W_3 \cdot \mathbf{h}_{2} + \mathbf{b}_3); \dots; \
    \mathbf{h}_l = \sigma(W_l \cdot \mathbf{h}_{l-1} + \mathbf{b}_l); \\
    \mathbf{y}   &= f(\mathbf{h}_l)
\end{align*}
\noindent where $\mathbf{x} \in \mathbb{R}^D$ is the D-dimensional input vector,
$\mathbf{h}_i \in \mathbb{R}^{n(i)}$ represents the output of the neurons of the $i^{th}$ layer ($1\le i \le l$),
$W_j \in \mathbb{R}^{n(j) \times n(j-1)}$ is the weight matrix from the ${j-1}^{th}$ layer to $j^{th}$ layer, 
$\mathbf{b}_j \in \mathbb{R}^{n(j)}$ is the bias vector in the $j^\text{th}$ layer, 
$n(j)$ is the number of neurons of the $j^{th}$ layer,
and $\sigma$ stands for the sigmoid function. 
Alternatively, other activation functions are the hyperbolic tangent \cite{zamanlooy2013efficient} or the rectified linear unit activation function \cite{petersen2018optimal}.
Finally, $\mathbf{y}$ denotes the output with the function $f$, whose range depends on whether the network is designed for a regression or classification task.


However, since MLPs are fully connected neural networks, they are prone to overfit the training data, leading to poor generalization on new, unseen data \cite{zeleny2023multi}.  Regularization and Dropout techniques are generally used to mitigate overfitting by simplifying the model structure.

\subsection{Recurrent Neural Networks}
Unlike traditional feedforward neural networks, recurrent neural networks (RNNs) have connections that form directed cycles, allowing them to maintain a ``memory'' of previous inputs. This is particularly useful for tasks where the context from previous inputs is important, such as time series, text, speech, or video. The temporal dependencies can be learned by recurrently training and updating the transitions of an internal (hidden) state from the last timestep to the current timestep. 
The computational process of each hidden state (unit or cell) in an RNN is described as follows:
\begin{align*}
    \mathbf{S}_t &= \tanh(W_{xs} \cdot (\mathbf{x}_t \oplus \mathbf{S}_{t-1}) + \mathbf{b}_s), \\
    \mathbf{y}_t &= \sigma(W_y \cdot \mathbf{S}_t + \mathbf{b}_y),
\end{align*}

\noindent where $\mathbf{x}_t \in \mathbb{R}^m$ is the input vector of $m$ input features at time $t$;  
$W_{xs} \in \mathbb{R}^{n \times (m+n)}$ and $W_y \in \mathbb{R}^{n \times n}$ are parameter matrices;  
$n$ is the number of neurons in the RNN layer;  
$\mathbf{b}_s \in \mathbb{R}^n$ and $\mathbf{b}_y \in \mathbb{R}^n$ are bias vectors for the internal state and output, respectively;  
$\sigma$ is the sigmoid activation function;  
$\mathbf{S}_t$ is the internal (hidden) state;  
and $\mathbf{x}_t \oplus \mathbf{S}_{t-1}$ denotes the concatenation of vectors $\mathbf{x}_t$ and $\mathbf{S}_{t-1}$.

\begin{figure}[ht]
\centering
    \includegraphics[width=0.85\columnwidth]{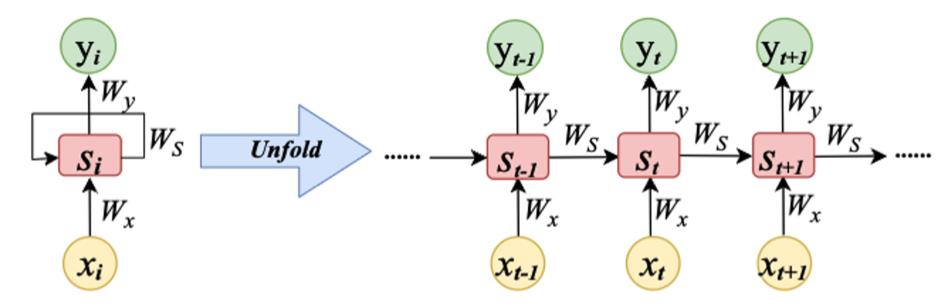} 
\caption{Schematic representation of the RNN architecture with its unfolded representation on the right shown with the recurring layers.}
\label{fig:rnn}
\end{figure}

Nevertheless, RNNs have the following limitations. Since RNNs learn knowledge iteratively over time steps, training RNNs can be challenging due to the vanishing or exploding gradient problem, especially for long sequences \cite{kag2021training}. Specifically, the gradients at the last time steps are difficult to backpropagate to the time steps at the beginning. This phenomenon is also called \emph{memory forgetting}, caused by the inherent sequential modeling, which is characteristic of RNNs.

\subsection{Convoluntional Neural Networks}
Convolutional neural networks (CNNs), originally developed for image and video processing, have also demonstrated strong performance in time series forecasting. When applied to sequential data, CNNs capture temporal patterns and dependencies by treating the time series as a one-dimensional sequence. Convolutional layers use learnable filters (kernels) that slide over the input to detect localized patterns \cite{borovykh2017conditional}. These layers are typically followed by max-pooling operations, which retain the most salient features while reducing dimensionality. For multivariate time series, the input can be structured as a two-dimensional array (time steps $\times$ features), enabling CNNs to learn both temporal dependencies and cross-variable interactions \cite{keren2016convolutional}.

\begin{figure}[ht]
\centering
    \includegraphics[width=0.95\columnwidth]{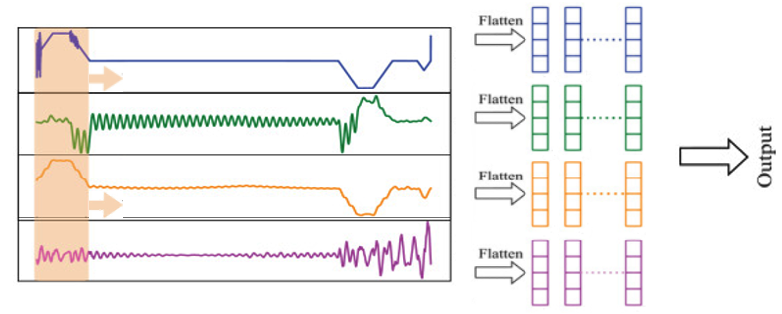} 
\caption{Schematic representation of the convoluntion operation in CNNs \cite{canizo2019multi}.}
\label{fig:cnn}
\end{figure}

CNNs offer several advantages for time series forecasting. Their strength lies in capturing local patterns through filters with limited receptive fields, allowing the model to efficiently detect short-term dependencies and localized features. A key benefit is that these filters are applied uniformly across the sequence, making CNNs invariant to input length and well-suited for handling long sequences without degradation in performance \cite{alzubaidi2021review}. Additionally, CNNs do not rely on the propagation of hidden states, unlike RNNs, which can help maintain stable prediction quality across varying sequence lengths.
However, this local focus also presents a limitation. Because CNNs primarily operate within fixed-size receptive fields, they may struggle to model long-range or sequential dependencies as effectively as recurrent architectures. This can be particularly restrictive in time series tasks where the temporal order and long-term interactions between time steps are critical for accurate forecasting.

\subsection{Graph Neural Networks}
Graph Neural Networks (GNNs) \cite{scarselli2008graph} are designed to work on graph-structured data, $\mathcal{G}=(\mathcal{V}, \mathcal{E})$, consisting of a set of nodes $\mathcal{V}$ and a set of edges $\mathcal{E}$ connecting them. The nodes and edges represent the entities and their relationships in a graph (the graph is undirected if not specified).
Spatio-temporal Graph Neural Networks (ST-GNNs) \cite{yu2017spatio} extend the architecture of GNNs to model both spatial and temporal dependencies in dynamic graph-structured data changing over time, $\mathcal{G}_t=(\mathcal{V}, \mathcal{E}, t)$. 
Here, nodes $\mathcal{V}$ refer to spatial locations, while edges $\mathcal{E}$ refer to spatial relationships between the locations. Each node $v_t^{i}$ represents the feature vector at the corresponding location $i$ and time $t$. 
For each node, the message-passing technique \cite{gilmer2017neural} is often employed to capture the spatial dependencies on its neighbors. The temporal dependencies between graph snapshots can be modeled with the sequential models aforementioned.
For the message passing, hidden states $h_t^i$ at each node are updated based on messages (feature vectors) $v_{t+1}^{i}$ according to:
\begin{equation}
    \begin{aligned}
        v_{t+1}^{i} &= \sum_{j \in N(i)} M_t(h_t^i, h_t^j, e_{ij}), \\
        h_{t+1}^{i} &= \sigma(h_t^i, v_{t+1}^{i}),
    \end{aligned}
\end{equation}
where in the sum, $N(i)$ denotes the neighbors of $i^{th}$ node in graph $\mathcal{G}$. After iterative updates $k$ time steps, the final output of the whole graph at time $t+k$ can be computed with a readout function $\mathcal{O}$:
\begin{equation}
    y_{t+k} = \mathcal{O}(\{h_{t+k}^i \mid i \in \mathcal{G}\}).
\end{equation}

\subsection{Transformers}
To overcome the limitations of RNNs, which stem from their inherent sequential processing, the Transformer model \cite{vaswani2017attention} has emerged as a powerful alternative. Using an encoder-decoder architecture, its core innovation lies in the use of the \emph{attention} mechanism, enabling it to capture dependencies between any parts of a sequence without the need for sequential steps \cite{wen2022transformers}. The \emph{attention} mechanism is described as follows: 
\begin{equation}
    \text{Attention}(\mathbf{Q}, \mathbf{K}, \mathbf{V}) = \text{softmax}\left(\frac{\mathbf{Q} \mathbf{K}^T}{\sqrt{d_k}}\right)\mathbf{V},
\end{equation}
where the $d_k$ denotes the dimension of the key, $\mathbf{Q} \in \mathbb{R}^{n \times d_k}$, $\mathbf{K} \in \mathbb{R}^{m \times d_k}$, and $\mathbf{V} \in \mathbb{R}^{m \times d_v}$ are the query matrix, key matrix, and value matrix, respectively. These three matrices are computed by linear transformations from the original input sequence $\mathbf{X} \in \mathbb{R}^{n \times d}$ with learnable weight matrices $\mathbf{W}_q \in \mathbb{R}^{d \times d_k}$, $\mathbf{W}_k \in \mathbb{R}^{d \times d_k}$, $\mathbf{W}_v \in \mathbb{R}^{d \times d_v}$, as
\begin{equation}
    \mathbf{Q} = \mathbf{X} \mathbf{W}_q, \mathbf{K} = \mathbf{X} \mathbf{W}_k, \mathbf{V} = \mathbf{X} \mathbf{W}_v.
\end{equation}

\begin{figure}[ht]
\centering
    \includegraphics[width=0.9\columnwidth]{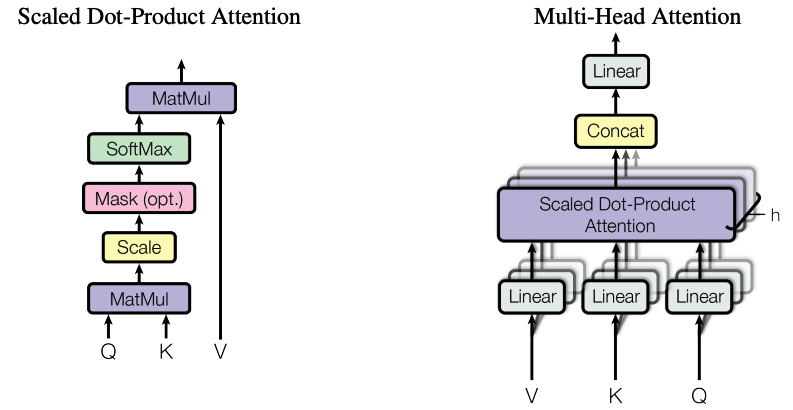} 
\caption{Schematic representation of the Attention mechanism \cite{vaswani2017attention}.}
\label{fig:attention}
\end{figure}

\subsection{Diffusion Models}
Denoising Diffusion Probabilistic Model (DDPM) \cite{ho2020denoising,song2020denoising} is a generative model that has gained significant popularity in computer vision \cite{saharia2022palette, croitoru2023diffusion}, natural language processing \cite{hertz2022prompt, li2023diffusion}, due to its ability to produce high-quality, realistic synthetic samples with similarity to the input. Diffusion models work in two processes: \emph{forward diffusion process} and \emph{reverse denoising process}. 
In the forward process, data (e.g., an image) is gradually “noised” by adding small amounts of Gaussian noise over multiple steps until it becomes nearly pure noise. This process is parameterized, where each step incrementally adds a fixed amount of noise.
A denoising model learns to denoise using pairs of original data (e.g., image before adding an increment of noise) and their diffused versions (i.e., image after adding the increment of noise), which are generated during the forward diffusion process. 
The reverse process uses the denoising model 
to recover a realistic sample from a noisy starting point. 

Mathematically, the \textit{forward process} transforms an input $\mathbf{x}_0$ with a data distribution of $q(\mathbf{x}_0)$ to a white Gaussian noise vector $\mathbf{x}_N$ in $N$ diffusion steps. It can be described as a Markov chain that gradually adds Gaussian noise to the input according to a variance schedule $\{\beta_1, \dots, \beta_N\} \in (0, 1)$:
\begin{equation}
    q(\mathbf{x}_{1:N} \mid \mathbf{x}_0) = \prod_{n=1}^N q(\mathbf{x}_n \mid \mathbf{x}_{n-1}),
\end{equation}
where at each step $n \in [1, N]$, the diffused sample $\mathbf{x}_n$ is obtained with $q(\mathbf{x}_n \mid \mathbf{x}_{n-1}) = \mathcal{N} \left( \mathbf{x}_n; \sqrt{1 - \beta_n} \mathbf{x}_{n-1}, \beta_n \mathbf{I} \right)$.

In the \textit{reverse process}, the denoising model, $p_\theta(\cdot)$, is used to recover $\mathbf{x}_0$ by gradually denoising $\mathbf{x}_n$ starting from a Gaussian noise $\mathbf{x}_N$ sampled from $\mathcal{N}(0, \mathbf{I})$. This process is presented as:
\begin{equation}
    p_\theta(\mathbf{x}_{0:N}) = p(\mathbf{x}_N) \prod_{n=1}^N p_\theta(\mathbf{x}_{n-1} \mid \mathbf{x}_n).
\end{equation}

\begin{figure}[ht]
\centering
    \includegraphics[width=0.9\columnwidth]{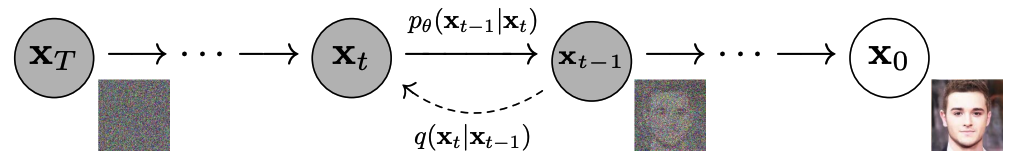} 
\caption{Schematic representation of the DDPM \cite{ho2020denoising}.}
\label{fig:ddpm}
\end{figure}

\subsection{Large Language Models}
Large language models (LLMs) are a class of deep learning models pre-trained on massive text corpora to perform a wide range of natural language processing (NLP) tasks. 
Built primarily using the Transformer architecture \cite{vaswani2017attention} and self-supervised learning methods, LLMs learn contextual representations by modeling dependencies between tokens across long sequences. 
During the inference phase, they autoregressively output the next token $\hat{w}_{n+1}$ with the highest probability based on the preceding ones $w_1, w_2, \ldots, w_n$, as described below:
\begin{equation}
    \hat{w}_{n+1} = \arg\max_{w_{n+1}} P(w_{n+1} \mid w_1, w_2, \ldots, w_n).
\end{equation}

\begin{figure}[ht]
\centering
    \includegraphics[width=0.95\columnwidth]{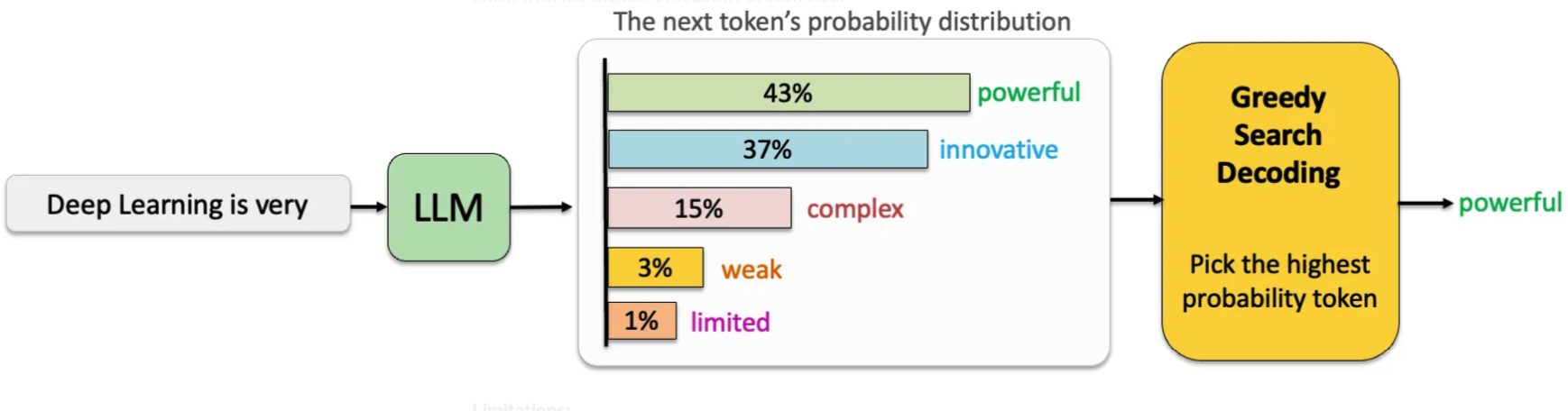} 
\caption{Schematic representation of the output process of LLMs.}
\label{fig:llm_decode}
\end{figure}

As mentioned above, LLMs are typically pre-trained on large-scale unlabeled data using self-supervised objectives and can be fine-tuned or prompted for downstream applications. Their generalization ability, scalability, and emergent reasoning capabilities have made them foundational tools across both academic research and real-world deployments. Examples include  BERT \cite{devlin2018bert}, GPT \cite{brown2020language}, ChatGPT \cite{achiam2023gpt}, and DeepSeek \cite{guo2025deepseek, liu2024deepseek}, which have demonstrated state-of-the-art performance in tasks such as question answering \cite{zhuang2023toolqa}, text summarization \cite{zhang2024comprehensive}, translation \cite{enis2024llm}, and much more.

\section{Flood Prediction and Mitigation}
Given that this dissertation deals with applying ML techniques to environmental problems, we review the areas of environmental Science relevant to us. We review water resource management in this section and weather prediction in Section \ref{subsec:weather}.

Flood prediction strategies are needed to determine the location, intensity, and duration of a flood event in a river system, given the possible driving factors, such as rainfall, sea level, and data on controllable hydraulic structures such as gates, pumps, dams, and reservoirs in the river system.
Flood management refers to determining the appropriate interventions needed to prevent or mitigate a flood that may be predicted in anticipation of an impending storm, which is often achieved by applying the ``optimal'' control of the hydraulic structures \cite{leon2020matlab, leon2021comparison}.

\subsection{Physics-based Approaches}
Conventional flood prediction methods over the last decade or more have been achieved by starting with a detailed physical model of the entire watershed and applying physical laws of conservation of water from first principles.
HEC-RAS 2-D models have assessed urban flood risks \cite{rangari2019assessment, shaikh2023application}, while a coupled 1D–2D HEC-RAS model was used for both channel and overbank flow representation \cite{dasallas2019case}. \cite{mo2023simulation} employed the HEC-RAS 1D model for floods caused by dam breaks \cite{mo2023simulation}.
Others \cite{ansarifard2024simulation,shrestha2020understanding} have studied floodplain modeling using HEC-RAS and MIKE 21 tools, and analyzed their suitability in hydraulic and hydrological models.
Khatooni et al. \cite{khatooni2025new} assessed urban flood risks by coupling SWMM and HEC-RAS-2D models.
Despite these advancements, physics-based models offer relatively accurate performance but with the following notable limitations.
(1) Computationally Intensive: For large watersheds, these models are computationally demanding \cite{chen2023physics, vadyala2022review} because they use detailed grid representations and often need to solve a large number of complex partial differential equations (PDEs) \cite{jenkins2023physics}.
(2) Lack of Flexibility: They lack flexibility and require rerunning the entire model for each new simulation \cite{shen2023differentiable}, even for relatively small changes in initial or boundary conditions \cite{guo2022machine}. This prevents the exploration of a large number of hypothetical scenarios, often employed in real-life scenarios with massive economic consequences for the model predictions.
(3) Accuracy Challenges: Physics-based models may compromise accuracy by using simplifications and approximations that reduce model complexity \cite{guo2022machine}. 
Such changes may limit the ability of the models to comprehensively cover the entire spectrum of process variability \cite{guo2023digital}.
Higher accuracies can be achieved by using finer grids, which in turn makes the models more computationally intensive.
(4) Explainability Issues: Despite elucidating complex dynamics, they struggle to provide detailed explanations, particularly in identifying specific variables and time steps influencing simulation outcomes.

Once floods are predicted, flood mitigation strategies are needed.
Water management agencies have built controllable hydraulic structures such as dams, gates, pumps, and reservoirs in river systems \cite{kerkez2016smarter}. 
On one hand, rule-based methods \cite{sadler2020exploring, shishegar2019integrated, monrat2018belief} formulate control schedules based on insights gained from historically observed data. 
However, these schedules are vulnerable to rare extreme events and sensitive to watersheds.
On the other hand, researchers have tackled this challenge as an optimization problem \cite{karimanzira2016model}. 
By optimizing variables such as water discharges, the algorithm generates control schedules for hydraulic structures to effectively mitigate flood risks  \cite{delaney2020forecast, zarei2021machine}. 
Random initialization followed by soft-computing techniques such as genetic algorithms has been employed to perform the optimization. Subsequently, physics-based models such as HEC-RAS and SWMM are used to assess the generated control schedules \cite{leon2020matlab, sadler2019leveraging, vermuyten2018combining, chen2016dimension, leon2014dynamic}. 
However, such physics-based simulators are prohibitively slow since they require thousands of simulations of water level predictions. 
Therefore, it is important to develop accurate and efficient approaches for flood prediction and mitigation.

\subsection{Data-driven ML and DL Approaches}
With the rise of big data, data-driven approaches have become increasingly promising for flood modeling and prediction. A variety of machine learning (ML) and deep learning (DL) models have been explored in this domain. 
For example, in South Florida, artificial neural networks (ANNs) have been used to forecast river flow in the Apalachicola River \cite{huang2004forecasting} and net inflow volumes into Lake Okeechobee \cite{obeysekera2000use}. 
Sadler et al.\cite{sadler2018modeling} applied a Random Forest model to predict flood severity in Norfolk, Virginia. 
Similarly, Sapitang et al.\cite{sapitang2020machine} evaluated four ML models, Boosted Decision Tree Regression, Decision Forest Regression, Bayesian Linear Regression, and Neural Network Regression, to improve daily forecasts of reservoir water levels, highlighting intra-model comparisons. 
Ayus et al. \cite{ayus2023prediction} employed Random Forest, XGBoost, and bidirectional LSTM models on three decades of data to predict daily water levels.
More recently, Kow et al.\cite{kow2024advancing} proposed a hybrid Transformer–LSTM model for accurate multi-step-ahead forecasting of flood storage pond and river water levels. 
Yin et al.\cite{yin2024fast} leveraged U-Net and generative adversarial networks (GANs) to reconstruct high-resolution flood maps from coarse-resolution numerical simulations. Additionally, there is growing interest in physics-informed ``hybrid'' models that integrate neural networks with physical constraints such as the shallow water equations \cite{yin2023physic, yin2025physics}.

\section{Weather Forecasting}
\label{subsec:weather}
\subsection{Physics-based Numerical Weather Prediction}
Physics-based models, including General Circulation Models (GCMs) \cite{ravindra2019generalized} and Numerical Weather Prediction (NWP) models \cite{coiffier2011fundamentals}, have been the cornerstone of weather prediction. These models simulate future weather scenarios by numerically approximating solutions to the set of differential equations that govern the complex physical dynamics of interconnected atmospheric, terrestrial, and oceanic systems \cite{nguyen2023climax}.
As in the case of water resource management, these physics-based models face notable challenges. 
First, the accuracy of conventional NWP models is highly dependent on the spatial and temporal resolution. Finer resolutions allow for better representation of mesoscale and localized phenomena (e.g., convection, topographic effects, or coastal systems). However, achieving higher resolution significantly increases the computational cost \cite{al2010review}.
Second, subgrid-scale parameterizations introduce significant uncertainty. Many atmospheric processes occur at scales too small to be explicitly resolved and are instead approximated using empirical or simplified physical models. These parameterizations are often region-specific and may fail to generalize across varying conditions \cite{palmer2005representing}. 
Moreover, the governing equations rely on simplified atmospheric dynamics assumptions, limiting their ability to capture rare or complex phenomena. 
Finally, a single physics-based model typically produces deterministic forecasts once initial conditions are fixed, falling short of capturing uncertainties in weather evolution, even though perturbations of initial conditions have been tried \cite{bulte2024uncertainty}. Ensemble-based NWP forecasts help address this last issue by generating probabilistic outputs \cite{gneiting2005weather}, but the first two challenges persist.

\subsection{Machine Learning Weather Prediction}
\tikzstyle{my-box}=[
    rectangle,
    draw=hidden-draw,
    rounded corners,
    text opacity=1,
    minimum height=1.5em,
    minimum width=5em,
    inner sep=2pt,
    align=center,
    fill opacity=.4,
    line width=0.8pt,
]
\tikzstyle{leaf}=[my-box, minimum height=1.5em,
    fill=hidden-pink!80, text=black, align=left, font=\normalsize,
    inner xsep=2pt,
    inner ysep=4pt,
    line width=0.8pt,
]
\begin{figure}[t]
\centering
    \resizebox{0.98\textwidth}{!}{
        \begin{forest}
            forked edges,
            for tree={
                fill=level0!80,
                grow=east,
                reversed=true,
                anchor=base west,
                parent anchor=east,
                child anchor=west,
                base=left,
                font=\large,
                rectangle,
                draw=hidden-draw,
                rounded corners,
                align=left,
                minimum width=6em,
                edge+={darkgray, line width=1pt},
                s sep=3pt,
                inner xsep=2pt,
                inner ysep=3pt,
                line width=0.8pt,
                ver/.style={rotate=90, child anchor=north, parent anchor=south, anchor=center},
            },
            where level=1{text width=11em,font=\normalsize,fill=level1!90,}{},
            where level=2{text width=10em,font=\normalsize,fill=level2!80,}{},
            where level=3{text width=5.5em,font=\normalsize,fill=level3!60,}{},
            where level=4{text width=21.5em,font=\normalsize,fill=level1!40,}{},
            [AI Models \\ for Weather \\ Prediction
                [
                    Deterministic Predictive \\ Learning 
                    [
                        General-Purpose \\
                        \textbf{Large Models}
                        [
                            Transformer
                            [
                                FourCastNet~\cite{pathak2022fourcastnet}{,} 
                                FuXi~\cite{chen2023fuxi}{,} \\
                                FengWu~\cite{chen2023fengwu}{,} 
                                FengWu-4DVar~\cite{xiao2023fengwu}{,}\\
                                SwinVRNN~\cite{hu2023swinvrnn}{,}
                                SwinRDM~\cite{chen2023swinrdm}{,} \\
                                Pangu-Weather~\cite{bi2023accurate}{,}
                                Stormer~\cite{nguyen2023scaling}{,}\\
                                HEAL-ViT~\cite{ramavajjala2024heal}{,}
                                TianXing~\cite{yuan2025tianxing}{,}\\
                                ArchesWeather~\cite{couairon2024archesweather}{,}
                                AtmoRep~\cite{lessig2023atmorep}
                            ]
                        ]
                        [
                            GNN
                            [   
                                GraphCast~\cite{lam2022graphcast}{,} GnnWeather~\cite{keisler2022forecasting}{,}\\
                                AIFS~\cite{lang2024aifs}{,}
                                GraphDOP~\cite{alexe2024graphdop}
                            ]
                        ]
                        [
                            PhysicsAI
                            [
                                ClimODE~\cite{verma2024climode}{,}
                                WeatherODE~\cite{liu2024mitigating}{,}\\
                                NeuralGCM~\cite{kochkov2024neural}{,}
                                Conformer~\cite{saleem2024conformer}
                            ]
                        ]
                    ]
                    [
                        Domain-Specific \\ Models
                        [
                            Transformer
                            [
                                SwinUnet~\cite{bojesomo2021spatiotemporal}{,}
                                Earthformer~\cite{gao2022earthformer}{,}\\
                                Rainformer~\cite{bai2022rainformer}{,}
                                U-STN~\cite{chattopadhyay2022towards}{,}\\
                                OMG-HD~\cite{zhao2024omg}{,}
                                PFformer~\cite{xu2024pfformer}
                            ]
                        ]
                        [
                            GNN
                            [
                                HiSTGNN~\cite{ma2023histgnn}{,} 
                                $w$-GNN~\cite{chen2024coupling}{,}\\
                                WeatherGNN~\cite{wu2024weathergnn}{,} MPNNs~\cite{yang2024multi}
                            ]
                        ]
                        [
                            RNN\&CNN
                            [   
                                MetNet~\cite{sonderby2020metnet,espeholt2022deep}{,}\\
                                MetNet-3~\cite{andrychowicz2023deep}{,}
                                PredRNN~\cite{wang2022predrnn}{,}\\
                                MM-RNN~\cite{ma2023mm}{,}
                                ConvLSTM~\cite{shi2015convolutional}
                            ]
                        ]
                        [
                            Mamba
                            [
                                MetMamba~\cite{qin2024metmamba}{,}
                                MambaDS~\cite{liu2024mambads}{,}\\
                                VMRNN~\cite{tang2024vmrnn}{,}
                                WPMamba~\cite{liu2024wpmamba}
                            ]
                        ]
                        [
                            PhysicsAI
                            [
                                NowcastNet~\cite{zhang2023skilful}{,}
                                PhysDL~\cite{de2019deep}{,}\\
                                PhyDNet~\cite{guen2020disentangling}{,}
                                DeepPhysiNet~\cite{li2024deepphysinet}
                            ]
                        ]
                    ]
                ]
                [
                    Probabilistic Generative \\ Learning
                    [
                        General-Purpose \\ \textbf{Large Models}
                        [
                            Diffusion
                            [ 
                                GenCast~\cite{price2023gencast}{,} CoDiCast~\cite{shi2024codicast}{,}\\
                                SEEDs~\cite{li2023seeds}{,}
                                ContinuousEnsCast~\cite{andrae2024continuous}{,}\\ 
                                ArchesWeatherGen~\cite{couairon2024archesweather}
                            ]
                        ]
                    ]
                    [
                        Domain-Specific \\ Models
                        [
                            Diffusion
                            [
                                LDMRain~\cite{leinonen2023latent}{,} PreDiff~\cite{gao2024prediff}{,}\\
                                CasCast~\cite{gong2024cascast}{,} SRNDiff~\cite{ling2024srndiff}{,}\\
                                DiffCast~\cite{yu2024diffcast}{,} 
                                GEDRain~\cite{asperti2023precipitation}
                            ]
                        ]
                        [
                            GANs
                            [
                                GANrain~\cite{ravuri2021skilful}{,}
                                MultiScaleGAN~\cite{luo2022experimental}{,}\\
                                STGM~\cite{wang2023physical}{,}
                                PCT-CycleGAN~\cite{choi2023pct}
                            ]
                        ]
                    ]
                ]
                [   
                    Pretraining \& Finetuning  
                    [
                        \textbf{Foundation Models}
                        [
                            Transformer
                            [
                                ClimaX~\cite{nguyen2023climax}{,} W-MAE~\cite{man2023w}{,}\\
                                Aurora~\cite{bodnar2024aurora}{,} Prithvi WxC~\cite{schmude2024prithvi}
                            ]
                        ]
                    ]
                ]
            ]
        \end{forest}
    }
\caption{Taxonomy of deep learning models for weather prediction across training paradigms (dark yellow), model scopes (purple), and model architectures (pink).}
\vspace{-2mm}
\label{fig:taxonomy}
\end{figure}
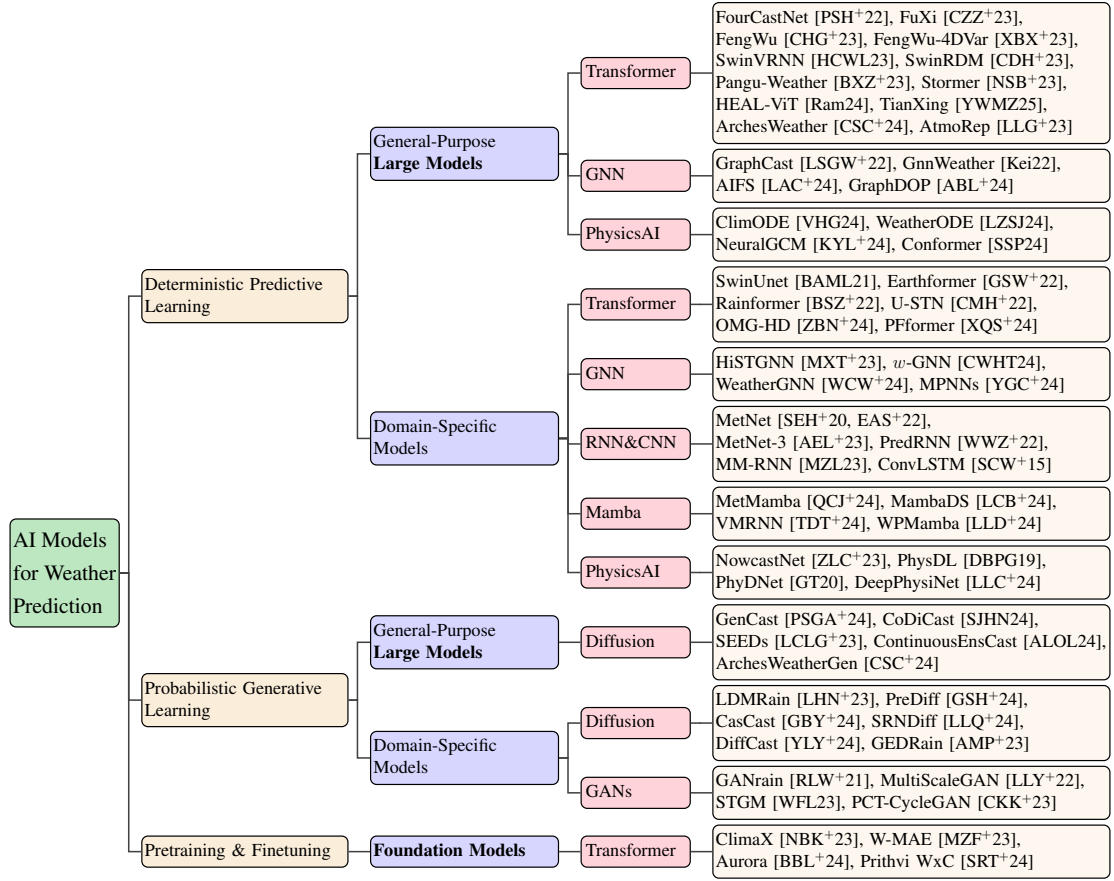

In recent years, data-driven machine learning (ML) and deep learning (DL) models have been increasingly applied to weather and climate modeling, demonstrating remarkable advances in precision, computational efficiency, and uncertainty quantification \cite{chen2023artificial, nguyen2023climatelearn}.
They have proven increasingly adept at capturing complex atmospheric dynamics in an end-to-end fashion, eliminating the reliance on explicit prior knowledge of physical relationships. 
For example, deterministic models such as Pangu \cite{bi2023accurate} and GraphCast \cite{lam2022graphcast} have achieved state-of-the-art performance in medium-range (10-day) global weather prediction, surpassing or matching traditional methods in terms of accuracy on some benchmark datasets (e.g., ERA5) while dramatically reducing computational costs (up to three orders of magnitude). However, their predictions are often blurry since they are trained by minimizing point-wise loss functions.
To overcome this limitation, probabilistic generative models have emerged as powerful tools for weather prediction while achieving uncertainty quantification in those predictions.
They consider weather prediction as probabilistic sampling (i.e., generation) conditioning on necessary constraints.
Models like CasCast \cite{gong2024cascast}, CoDiCast \cite{shi2024codicast}, and Gencast \cite{price2023gencast} leverage diffusion models for precipitation nowcasting and weather prediction, delivering both probabilistic outputs and calibrated uncertainty estimates.
More recently, foundation models have gained traction in climate and weather modeling as an emerging paradigm \cite{bodnar2024aurora, schmude2024prithvi}. These models are pre-trained on massive historical weather datasets to learn generalizable and comprehensive knowledge, which can then be fine-tuned for diverse downstream tasks, e.g., weather forecasting and climate downscaling \cite{chen2023foundation}. 
Foundation models offer two key advantages: (1) the ability to learn robust and transferable weather representations from large-scale data, and (2) the flexibility to adapt to downstream applications without the need for task-specific models trained from scratch \cite{miller2024survey, zhu2024foundations}.
We provide a comprehensive taxonomy in Figure \ref{fig:taxonomy}, including training paradigms, model architectures, and scopes.

\section{LLMs for Scientific Question-Answering}
Traditional information retrievers include BM25 \cite{robertson1994some, trotman2014improvements}, a classical sparse retriever based on TF-IDF principles. It ranks documents by matching query terms, adjusted for frequency and document length. Though efficient and interpretable, BM25 struggles with semantic understanding, often missing paraphrased or contextually relevant content.

An important approach to address retrieval is the use of embeddings, especially approaches that incorporate semantics in the embedding space.
Dense embedding retrievers, such as DPR \cite{karpukhin2020dense}, Contriever \cite{izacard2022contriever}, e5 \cite{wang2022text}, and ANCE \cite{xiong2020approximate} improve performance in open-domain QA by capturing deeper semantics. However, semantically similar documents may be topically irrelevant, leading to noise and hallucinations in the retrievals. Furthermore, their lack of interpretability limits their use in sensitive domains \cite{ji2019visual}. 

Retrieval-Augmented Generation (RAG) is a technique that depends on embeddings of entities from knowledge sources as part of a larger retrieval system.
Many structured RAG methods have been explored. GraphRAG \cite{edge2024local} builds entity graphs from source documents, while LightRAG \cite{guo2024lightrag} uses dual-level retrieval for richer knowledge access. HippoRAG \cite{gutierrez2024hipporag} and HippoRAG 2 \cite{gutierrez2025rag} leverage knowledge graphs inspired by cognitive theories to enhance retrieval and integration. SELF-RAG \cite{asai2023self} incorporates self-reflection to improve factuality, and RA-DIT \cite{lin2023ra} fine-tunes LLMs with up-to-date knowledge. Despite their promise, graph-based methods face scalability challenges due to the complexity of graph construction and reasoning.

\chapter{\fidlar: FORECAST-INFORMED DEEP LEARNING APPROACHES FOR FLOOD PREDICTION AND MITIGATION IN COASTAL RIVER SYSTEMS}	
\label{sec:fidlar}

\section{Background}
Floods are a significant threat to both lives and property \cite{ikram2024flood,rangaraj2025effective}, and pose environmental and public health hazards \cite{glago2021flood}. 
Anticipating the timing and location of floods could empower water management agencies to utilize hydraulic structures and implement timely flood mitigation strategies \cite{kumar2023state}, thus enabling citizens and local governments to enhance preparedness for potential emergencies \cite{yin2024strategic}.
Thus, predicting floods in advance and solving optimization problems related to flood mitigation are vitally significant endeavors.
In this chapter, we focus on floods in coastal river systems, with special emphasis on the South Florida river system involving the Miami river and the associated canals as it flows into the Atlantic Ocean.
The work presented in this chapter is highlighted in multiple manuscripts and archival publications, including \cite{shi2022time,shi2023deep,shi2023explainable,shi2023graph,shi2023power}.

Flood prediction methods are tasked with determining the location, intensity, and duration of a flood event in a river system. Classical methods over the last decade or more have been achieved by starting with a detailed physics-based model and applying physical laws of conservation of water from first principles.
Despite their remarkable advances, physics-based classical methods have the following limitations.
\begin{enumerate}
\item \textbf{Computationally Intensive:} For large watersheds, these models are computationally demanding \cite{chen2023physics, vadyala2022review} because they use detailed grid representations and often need to solve many complex partial differential equations (PDEs) for each cell of the grid \cite{jenkins2023physics}.
\item \textbf{Accuracy Challenges:} Physics-based models may compromise accuracy by using simplifications and approximations that reduce model complexity \cite{guo2022machine}. 
Such changes may limit the ability of the models to comprehensively cover the entire spectrum of process variability \cite{guo2023digital}.
\item \textbf{Explainability Issues:} Despite elucidating complex dynamics, they struggle to provide detailed explanations, particularly in identifying specific variables and time steps influencing simulation outcomes.
\end{enumerate}

Flood management involves going one step above flood prediction. It takes proactive steps to determine appropriate interventions to prevent or mitigate flooding, especially in anticipation of an impending storm.
The primary tool for flood management is the use of hydraulic structures constructed in the river systems for the purpose of controlling water levels.
Flood mitigation is typically achieved by pre-releasing sufficient water in advance so that water levels remain within a safe range when the storm event occurs \cite{mishra2022overview}. Pre-release is carried out by controlling hydraulic structures such as dams, gates, pumps, and reservoirs, and determining a \textit{control schedule} for these structures. 
However, determining the optimal control schedules of these hydraulic structures is a challenging problem \cite{bowes2021flood}. 
Most water management agencies have comprehensive prespecified \emph{Rule-based methods} \cite{sadler2019leveraging} that help formulate the control schedules. Such rules have been designed based on human insights gained from having managed river systems over decades and the experience gained by the decision made in the past.
Nevertheless, these rules may expose vulnerabilities while dealing with extremely rare events and may not offer effective solutions for different river systems \cite{schwanenberg2015open}. 
Another method for determining good control schedules is to formulate it as an \emph{optimization} problem.
Previous approaches include the use of soft computing techniques such as genetic algorithms (GA) and pattern search (PS) combined with a detailed physics-based simulator to evaluate the solutions \cite{debbarma2024simulation, mounce2020optimisation, elmorshedy2021recent, schwenzer2021review}.
However, because of the use of detailed simulators, they are typically too inefficient to be used for real-time flood control. 

With the rapid advancements in artificial intelligence (AI) \cite{alam2021possibilities}, deep learning (DL) models have emerged as powerful tools across numerous domains, demonstrating impressive performance \cite{choudhary2022recent, xu2021artificial}. 
DL models possess a range of characteristics, including their ability to effectively learn complex non-linear relationships \cite{wang2020recent}, deliver ultra-fast responses once trained \cite{oikonomou2022robust}, utilize feedback signals via the backpropagation algorithm during training \cite{eshraghian2023training}, and the availability of tools for model explainability \cite{angelov2020towards}. 

In this chapter, we first propose a novel DL model for water level forecasting (\WaLeF).
Then we present a forecast-informed deep learning approach for flood mitigation (\fidlar).
We perform extensive experiments to evaluate the performance of \walef and \fidlar, and discuss their performance relative to baseline methods.

\section{Related Work}
\paragraph{Flood Prediction.} Physics-based models (e.g., HEC-RAS, SWMM) have been widely used to simulate water levels and flows in river systems \cite{peker2024integration,rahman2024drivers,gomes2023modeling,rivett2022acute}. 
However, these models are computationally inefficient and fall short of capturing precise knowledge of study domains \cite{bentivoglio2022deep}. Therefore, diverse machine learning (ML) and deep learning (DL) models have been studied as surrogates to simulate water levels and flows \cite{rangaraj2025effective,rangaraj2025retrieval,zheng2025sf}.
For example, support vector machines (SVMs) have been used to predict the urban flash floods \cite{yan2018urban,choubin2019ensemble}, multivariate regression models were adopted to estimate flood volumes and peak flows \cite{yang2020regional}, random forests and K-nearest neighbors were explored for urban flood inundation mapping \cite{castro2014flood}, Gaussian process learning models for fast and accurate flood inundation simulation \cite{fraehr2023development}. Furthermore, deep learning models, such as recurrent neural networks, convolutional neural networks, and transformers, have been employed for flood inundation \cite{zhou2021rapid} and flood prediction \cite{shi2023graph,shi2023deep}.

\paragraph{Flood Mitigation.} Flood mitigation can be achieved by designing control schedules for managing all the hydraulic structures present in river systems to avoid or mitigate flood risks. However, this process may require flood prediction models as simulators to evaluate the control schedules.
Researchers have attempted to leverage the genetic algorithm and the pattern search to generate control schedules and physics-based models (e.g., EPA-SWMM5, HEC-RAS) as water simulators \cite{sadler2019leveraging, leon2014dynamic, leon2020matlab}. The Lake Mendocino Operations (LMO) model was developed to simulate operations of Lake Mendocino such as release constraints for flood control and water supply operations \cite{delaney2020forecast}.
However, such methods are computationally intensive since they require thousands of time-consuming simulation trials with physics-based models \cite{jafarzadegan2023recent}.
Additionally, the genetic algorithm and pattern search techniques are usually heuristic, without positive feedback or guidance to better control those hydraulic structures.

\section{Methodology}
\subsection{Problem Formulation}
\paragraph{Flood Prediction.} It refers to forecasting water levels at designated points of interest within river systems, given the input as known water levels from the recent past, rainfall, and control schedules of gates and pumps. The underlying transfer function is:
\begin{equation}
\label{eq:WaLeF}
    \mathcal{E}_{\theta_E}: (X^{all}_{t-w+1:t}, X^{cov}_{t+1:t+k}, X^{gate, pump}_{t+1:t+k}) \rightarrow X^{water}_{t+1:t+k},
\end{equation}
where the subscripts represent the time ranges, and the superscripts refer to the variables under consideration. The superscript $all$ represents all possible variables in a watershed (e.g., rainfall, tide, gate and pump operation, water levels), while $cov$ refers specifically to covariates that can be reliably predicted (e.g., rain, tides). 

Once trained, it plays the role of a ``referee'' who evaluates those control schedules by outputting the resulting water levels. We use the trained models for water level prediction as \texttt{Flood Evaluator} in the following sections.

\paragraph{Flood Mitigation.} It aims to manage water levels before extreme weather events by predicting control schedules for hydraulic structures such as gates and pumps within the river system, denoted as $X^{gate, pump}_{t+1:t+k}$, spanning $k$ time points into the future from $t+1$ to $t+k$.
Flood prediction and mitigation present a significant challenge due to the intricate interplay of meteorological, hydrological, and oceanographic factors. We take as input historical data on all possible factors, $X$, from the preceding $w$ time points, in conjunction with reliably forecasted covariates (such as rainfall and tide) for the next $k$ time points.
Then we could train a deep learning (DL) model, $\mathcal{M}_{\theta_M}$, with parameters $\theta_M$:
\begin{equation}
\label{eq:problem_formulation}
    \mathcal{M}_{\theta_M}: (X^{all}_{t-w+1:t}, X^{cov}_{t+1:t+k}) \rightarrow  X^{gate, pump}_{t+1:t+k}.
\end{equation}
%

\subsection{Method Overview}
\begin{figure}[ht]
\centering
    \includegraphics[width=\columnwidth]{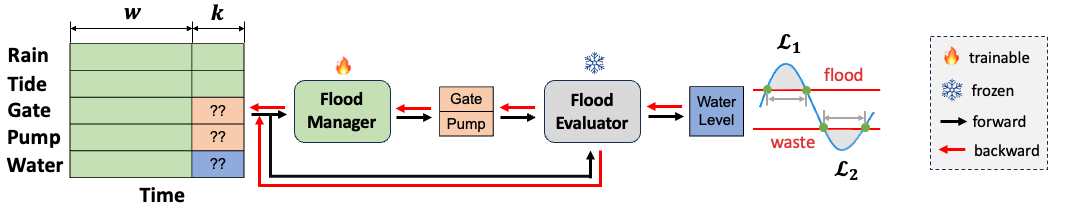} 
\caption{Forecast-Informed Deep Learning Architecture (\fidlar). 
Input data consists of five categories of variables as shown in the left table. The variables $w$ and $k$ are the lengths of the past and prediction windows, respectively.
The parts colored green are provided as inputs, while the orange and blue parts (with question marks) are outputs. 
The \texttt{Flood Manager} and \texttt{Flood Evaluator} represent deep learning (DL) models, the former to predict control schedules of controllable hydraulic structures (e.g., gates and pumps) to pre-release water, and the latter to predict the resulting water levels for those control schedules. 
Loss functions, $\mathcal{L}_{1}$ and $\mathcal{L}_{2}$, penalize the \textit{flooding} and \textit{water wastage} beyond pre-specified thresholds, respectively.}
\label{fig:fidla_framework}
\end{figure}

Intuitively, an ML model can be trained to learn the function $\mathcal{M}_{\theta_M}$ (\texttt{Flood  Manager}) to output the control schedules of gates and pumps directly.
However, a key challenge lies in the historical data, which often reflects control schedules that led to flooding or other suboptimal outcomes, making it unsuitable as ground-truth data for traditional supervised learning.
To overcome this, we first train an independent and accurate simulator model (\texttt{Flood Evaluator}), $\mathcal{M}_{\theta_E}$, by using extensive historical data to model the \emph{consequences} (e.g., water levels) of various \emph{actions} (e.g., control schedules).
We then frame the control schedule planning in \texttt{Flood Manager} as an optimization problem, seeking actions that minimize undesirable outcomes - floods or water wastage. Both the \texttt{Evaluator} and \texttt{Manager} are implemented as neural networks, with the framework illustrated in Figure \ref{fig:fidla_framework}.

\subsection{Flood Evaluator}
\label{sec:evaluator}
%
\noindent The \texttt{Evaluator} is trained independently using large-scale historical data to achieve highly accurate water level predictions for any given set of conditions and control schedules.
Therefore, once the \texttt{Evaluator} is well trained, its parameters are frozen while training the \texttt{Manager}, where it plays the role of a trained ``referee'' - scoring control schedules generated by the \texttt{Manager} by predicting the resulting water levels. 
It also serves to backpropagate the gradient descent feedback, guiding the \texttt{Manager} to produce more effective control schedules of gates and pumps.

\begin{figure}[ht]
\centering
    \includegraphics[width=0.6\columnwidth]{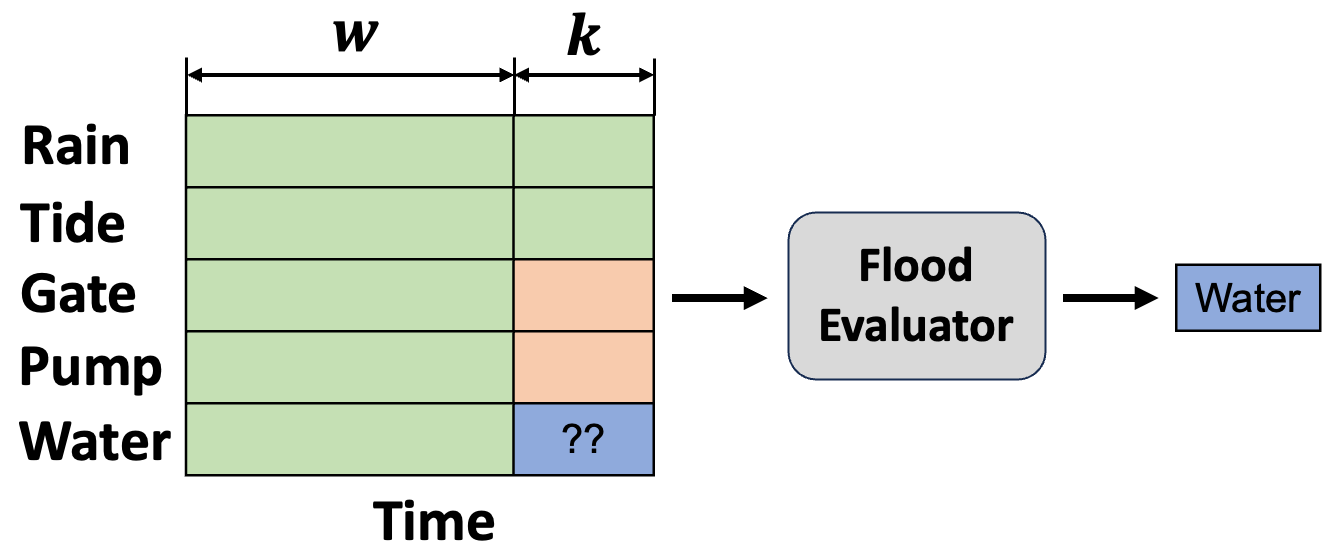} 
\caption{{\bf Flood Evaluator.} The parts shaded green are used as inputs (i.e., historical data and covariates predicted from the near future) and orange (control schedule for the gates and pumps)
Water levels (blue) are the outputs.}
\label{fig:flood_evaluator}
\end{figure}

\subsection{Flood Manager}
\label{sec:manager}
\texttt{Flood Manager} is to produce control schedules for hydraulic structures (i.e., gates and pumps), taking as inputs reliably predictable future information (rain, tide) and all historical data.
Since no ground truth is available, it is trained with the differentiability of the learned \texttt{Evaluator} model.
Therefore, we connect the \texttt{Manager} with the \texttt{Evaluator} where the output of Eq. (\ref{eq:problem_formulation}) is injected into Eq. (\ref{eq:WaLeF}):
\begin{equation}
\label{eq:predicted_gate_in_WaLeF}
    \mathcal{E}_{\theta_E}(X^{all}, X^{cov}_{t+1:t+k}, \mathcal{M}_{\theta_M}(X^{all}, X^{cov}_{t+1:t+k})) \rightarrow X^{water}_{t+1:t+k},
\end{equation}
where $X^{all}=X^{all}_{t-w+1:t}$ and $\theta_M$ and $\theta_E$ are the parameters of \texttt{Manager} and \texttt{Evaluator}.

\begin{figure}[ht]
\centering
    \includegraphics[width=0.6\columnwidth]{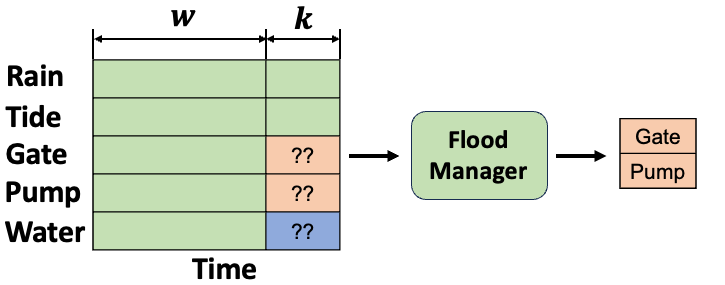}
\caption{{\bf Flood Manager.} 
The parts shaded green (historical data) are the inputs, and the parts shaded orange are the outputs. The water levels shaded blue are not predicted.}
\label{fig:flood_manager}
\end{figure}

The resulting output of water levels can be used to compute the loss in Eq. (\ref{eq:total_loss}), representing evaluation scores for generated control schedules.
Gradient descent \cite{ruder2016overview} can be back-propagated as the feedback to update the parameters of the \texttt{Manager}. The parameter update is
presented:
\begin{equation}
    \theta_M := \theta_M - \alpha \cdot \frac{\partial \mathcal{L}}{\partial \theta_M},
    \label{eq:gradient}
\end{equation}
\noindent where $\alpha$ is the learning rate and $\frac{\partial \mathcal{L}}{\partial \theta_M}$ is the partial derivative of the compound function $\frac{\partial \mathcal{L}}{\partial \theta_M} = \frac{\partial \mathcal{L}}{\partial \mathcal{E}} \cdot \frac{\partial \mathcal{E}}{\partial \mathcal{M}} \cdot \frac{\partial \mathcal{M}}{\partial \theta_M}$.
The training details of \fidlar~are in Algorithm \ref{alg:train_algorithm}.

\begin{algorithm}[ht!]
\small
\caption{Training algorithm of \fidlar}
\label{alg:train_algorithm}
\textbf{Input}: recent past data: ${\bf X}^{all}_{t-w+1, t}$\\ 
\textcolor{white}{\textbf{Input}:} near future data: ${\bf X}^{cov}_{t+1, t+k} = {\bf X}^{rain, tide, gate, pump}_{t+1, t+k}$ \\
\textbf{Parameter}: $\theta_E, \theta_M$: parameters of Evaluator and Manager; $w, k:$ length of past and prediction windows
\begin{algorithmic}[1] 
    \STATE \textcolor{blue}{// \texttt{Train Flood Evaluator,} $\mathcal{E}_{\theta_E}$}
    \STATE initialize learnable parameters $\theta_E$
    \FOR{$i = 1, \ldots, N$ epochs}
        \STATE $\text{MiniBatch} \gets (\{{\bf X}^{all}_{t-w+1, t}, {\bf X}^{cov}_{t+1, t+k}\}, {\bf X}^{water}_{t+1, t+k})$
        \STATE $\hat{{\bf X}}^{water}_{t+1, t+k} \gets \mathcal{E}_{\theta_E}({\bf X}^{all}_{t-w+1, t}, {\bf X}^{cov}_{t+1, t+k})$
        \STATE $\mathcal{L}_{E} \gets \frac{1}{k}\sum_{j=1}^k||\hat{{\bf X}}^{water}_{j} - {\bf X}^{water}_{j}||^2$
        \STATE $\nabla_{\theta_E} \gets \text{BackwardAD}(\mathcal{L}_{E})$
        \STATE $\theta_{E} \gets \theta_{E} - \eta \nabla_{\theta_E}$
    \ENDFOR
    \STATE \textbf{return} trained \texttt{Flood Evaluator}, $\mathcal{E}_{\theta_E}$
    \STATE \textcolor{blue}{// \texttt{Train Flood Manager,} $\mathcal{M}_{\theta_M}$, with frozen $\mathcal{E}_{\theta_E}$}
    \STATE initialize learnable parameters $\theta_M$
    \WHILE{$X^{water}_{t, t+k}$ violates either threshold}
        \STATE $\text{MiniBatch} \gets (\{{\bf X}^{all}_{t-w+1, t}, {\bf X}^{rain, tide}_{t+1, t+k}\}, {\bf X}^{gate, pump}_{t+1, t+k})$
        \STATE $\hat{{\bf X}}^{gate, pump}_{t+1, t+k} \gets \mathcal{M}_{\theta_M}({\bf X}^{all}_{t-w+1, t}, {\bf X}^{rain, tide}_{t+1, t+k})$
        \STATE $\hat{{\bf X}}^{water}_{t+1, t+k} \gets \mathcal{E}_{\theta_E}({\bf X}^{all}_{t-w+1, t}, {\bf X}^{rain, tide}_{t+1, t+k}, \hat{{\bf X}}^{gate, pump}_{t+1, t+k})$ 
        \STATE $\mathcal{L}_{E} = c_1 \cdot \mathcal{L}_{1}(\hat{{\bf X}}^{water}_{t+1, t+k}) + c_2 \cdot \mathcal{L}_{2}(\hat{{\bf X}}^{water}_{t+1, t+k})$ 
        \STATE $\nabla_{\theta_M} \gets \text{BackwardAD}(\mathcal{L}_{E})$
        \STATE $\theta_{M} \gets \theta_{M} - \eta \nabla_{\theta_M}$
    \ENDWHILE
    
    \STATE \textbf{return} trained \texttt{Flood Manager}, $\mathcal{M}_{\theta_M}$
\end{algorithmic}
\end{algorithm}

\subsection{Custom Loss Function}
\label{sec:loss}
\begin{figure}[ht]
\centering
    \begin{subfigure}[b]{0.45\textwidth}
        \centering
        \includegraphics[scale=0.45]{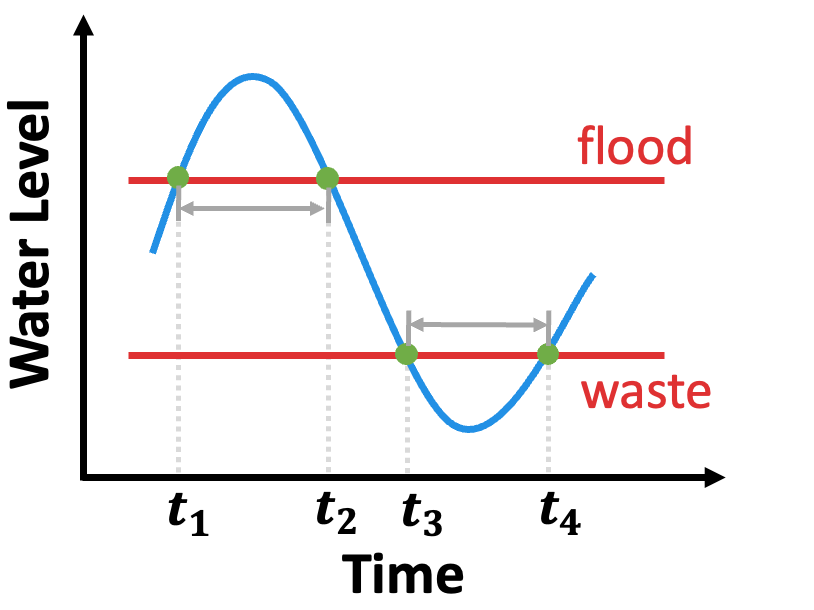}
        \caption{}
        \label{fig:loss_a}
    \end{subfigure}
    \hfill
    \begin{subfigure}[b]{0.45\textwidth}
        \centering
        \includegraphics[scale=0.45]{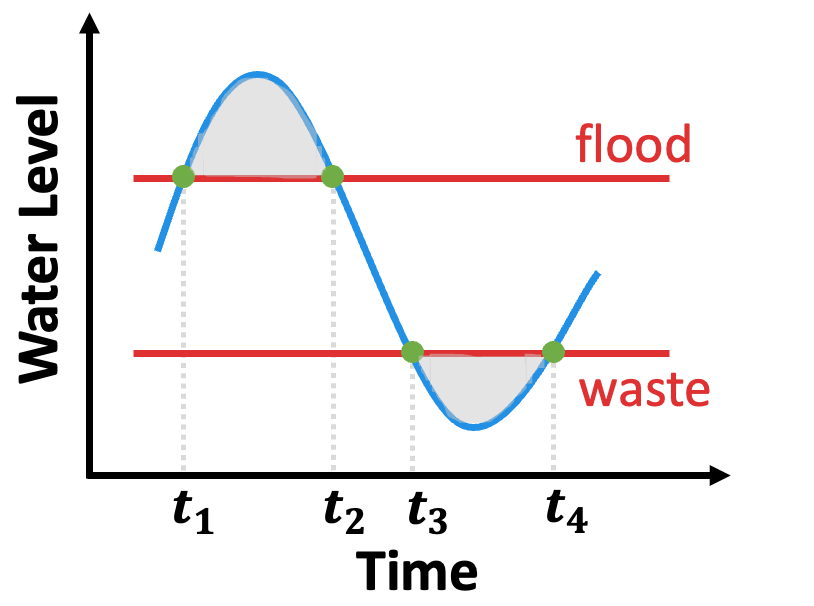}
        \caption{}
        \label{fig:loss_b}
    \end{subfigure}
\caption{The two red bars represent a threshold of flooding and a threshold of water wastage.
Shown are (a) the time spans when these thresholds are crossed, and (b) the areas between water level curves and threshold bars.
Violations of the upper and lower thresholds are captured in $\mathcal{L}_1$ and $\mathcal{L}_2$.}
\label{fig:loss}
\end{figure}

Loss functions are critical in steering the learning process. 
Our loss function penalizes the total time (Figure \ref{fig:loss_a}) for which the water levels either exceed the \textit{flooding threshold} or dip below the \textit{water wastage threshold}.
However, such a measure does not account for the severity of the flood, but only the time for which the flood occurs.
To incorporate the deviation from optimality associated with the severity of the flood or the wastage of water caused by pre-releases, our loss function also penalizes the extent to which the thresholds are exceeded to signify the severity of floods or water wastage (Figure \ref{fig:loss_b}).
The lower threshold for flood management is important in practice, since it prevents water wastage, thereby supporting irrigation, facilitating navigation, and maintaining ecological balance.
It also prevents the optimization methods from trivially recommending the depletion of valuable water resources to prevent future flooding. 
$\mathcal{L}_1$ and $\mathcal{L}_2$ represent the \textit{flooding} and \textit{water wastage} losses, respectively, and the final loss function is a balanced combination as shown in Eq. (\ref{eq:total_loss}).
\begin{equation}
\label{eq:flood_loss}
    \begin{aligned}
        \mathcal{L}_{1} = \sum^{N}_{i=1} \sum^{t+k}_{j=t+1} \Arrowvert max\{\hat{X}^{water}_{i, j} - X^{flood}_{i}, 0\} \Arrowvert^{2}, \\ 
        \mathcal{L}_{2} = \sum^{N}_{i=1} \sum^{t+k}_{j=t+1} \Arrowvert min\{\hat{X}^{water}_{i, j} - X^{waste}_{i}, 0\} \Arrowvert^{2},
    \end{aligned}
\end{equation}
where $N$ is the number of water level locations of interest; $k$ is the length of prediction horizon; $X^{flood}$ and $X^{waste}$ represent the thresholds for flooding and water wastage; and the capped version, $\hat{X}^{water}$, is obtained using the \texttt{Evaluator} module.
The combined loss function is given by:
\begin{equation}
\label{eq:total_loss}
    \mathcal{L}_{total} = c_1 \cdot \mathcal{L}_1 + c_2 \cdot \mathcal{L}_2,
\end{equation}
where $c_1/c_2$ dictates the relative importance of $\mathcal{L}_1$ and $\mathcal{L}_2$.
%

\section{Architecture of the Neural Network Model}
\label{sec:gtn}
The \texttt{Manager} and \texttt{Evaluator} modules described so far are model agnostic. We tried many existing deep learning architectures for them.
We devise the \underline{G}raph \underline{T}ransformer \underline{N}etwork (GTN) architecture by combining graph neural networks (GNNs), attention-based transformer networks, long short-term memory networks (LSTMs), and convolutional neural networks (CNNs).
GNN and LSTM modules are combined to learn the spatiotemporal dynamics of water levels, while the Transformer and CNN modules focus on extracting feature representations from the covariates. 
The \emph{attention} mechanism \cite{vaswani2017attention} is used to discern interactions between covariates and water levels, as shown in Eq. (\ref{eq:attention}) below.
Figure \ref{fig:graphtransformer} presents the GTN architecture, which is used for both \texttt{Evaluator} and \texttt{Manager}, but with minor changes accordingly of the inputs and outputs (see Figures \ref{fig:flood_evaluator} and \ref{fig:flood_manager}). We have two GNN layers with 32 and 16 channels, one LSTM layer, one CNN layer, and one Transformer encoder with 3 heads.
\begin{equation}
\label{eq:attention}
    \begin{aligned}
    Atte(Q, K, V) {} & = softmax(\frac{Q^{cov} (K^{water})^T}{\sqrt{d}}) V^{water}  \\
                     & = softmax(\frac{Q^{water} (K^{cov})^T}{\sqrt{d}}) V^{cov},  \\
    \end{aligned}
\end{equation}
where $T$ denotes the transpose operation; $water$ and $cov$ represent water levels and covariates; and $d$ is the embedding size where $d=d_q=d_k=d_v=128$.
\begin{figure}[ht]
\centering
\includegraphics[width=0.99\columnwidth]{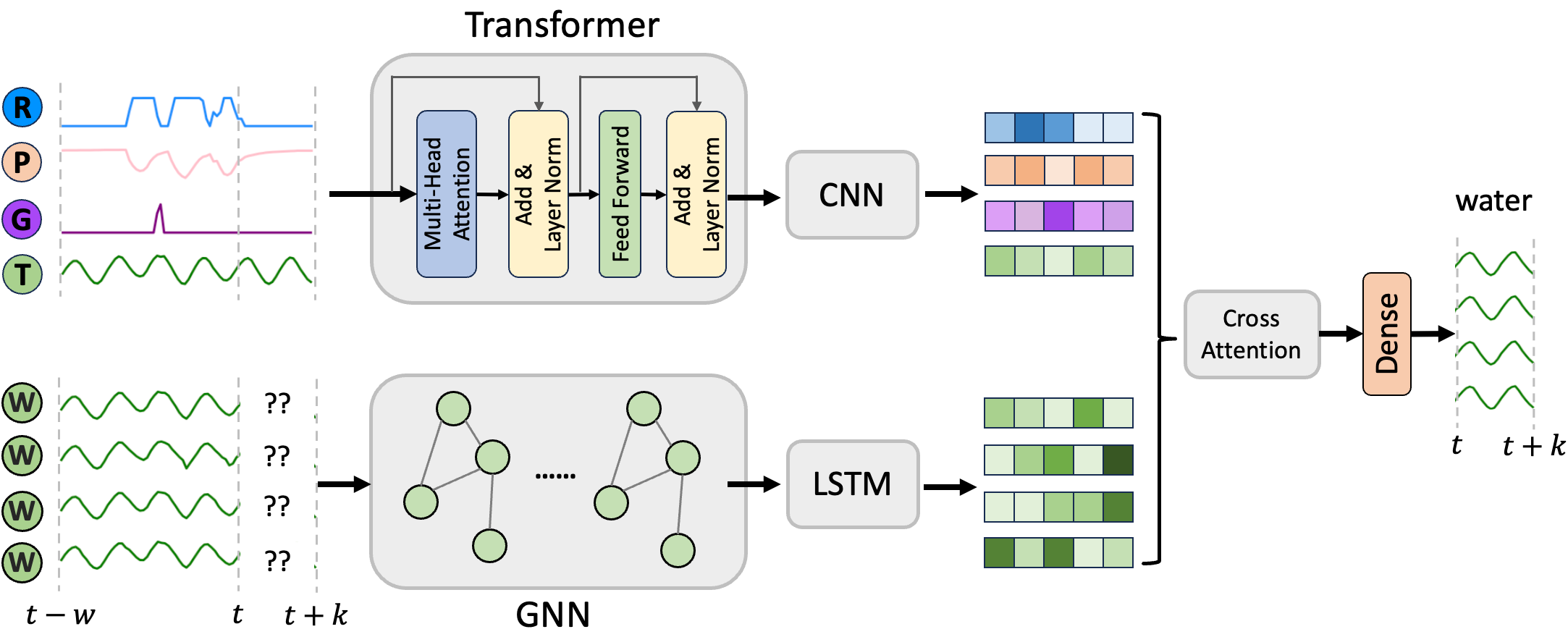} 
\caption{Graph Transformer Network for \texttt{Flood Evaluator}. Input variables include Rainfall, Pump, Gate, Tide, and Water levels as shown in Figure \ref{fig:domain}, which are generically denoted by $R, T, G, P$, and $W$. The output is water levels.}
\label{fig:graphtransformer}
\end{figure}

\section{Experiments}
\subsection{Study Domain and Dataset}
\begin{figure}[ht!]
\centering
    \includegraphics[width=0.8\columnwidth]{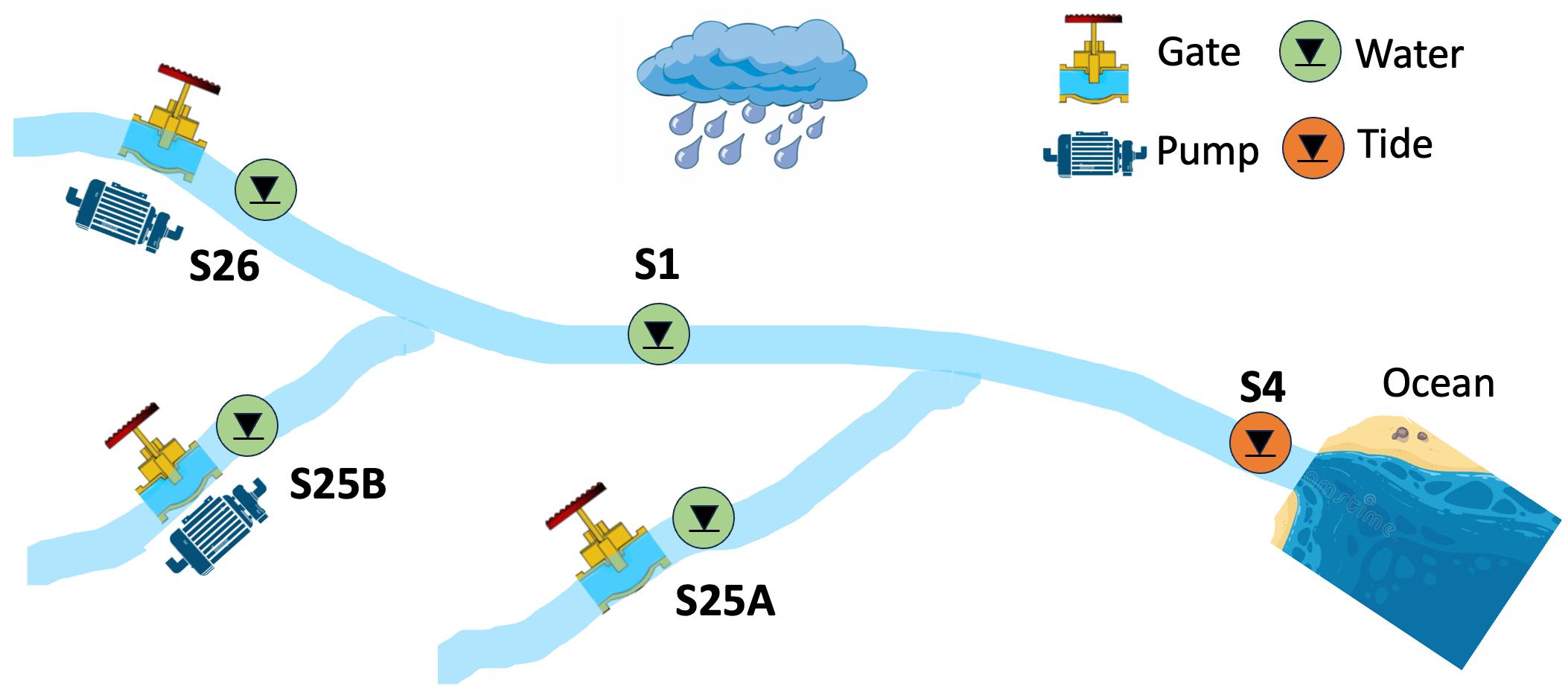} 
\caption{Schematic diagram of study domain. There are three water stations with hydraulic structures (S26, S25B, S25A), one simple water station, S1 (green circle in the middle), and a station monitoring the tide level from the ocean.}
\label{fig:domain}
\end{figure}

We obtained data from the South Florida Water Management District's (SFWMD) DBHydro database \cite{dbhydro2023sfwmd} for the coastal stretch in South Florida.
The data set consists of hourly observations for water levels and external covariates from January 1, 2010 to December 31, 2020.
As shown in Figure \ref{fig:domain}, the river system has two branches and includes several hydraulic structures (gates, pumps) to control water flows.
We aim to predict effective control schedules on hydraulic structures (gates, pumps) to minimize flood risks at four specific locations marked by green circles.
\begin{table}[ht!]
\centering
\setlength{\tabcolsep}{4pt} 
    \begin{tabular}{lcccc}
        \toprule
        \textbf{Feature}    & \textbf{Interval}  & \textbf{Unit}  & \textbf{\#Var.} & \textbf{Location}  \\
        \midrule \midrule    
        Rainfall            &Hourly         & $inch/h$      & 1    & -  \\ 
        Tide                &Hourly         & $ft$          & 1    & S4  \\ 
        Pump                &Hourly         & $ft^3/s$      & 2    & S25B, S26  \\ 
        Gate                &Hourly         & $ft$          & 3    & S25A, S25B, S26  \\
        Water               &Hourly         & $ft$          & 4    & S25A, S25B, S26, S1  \\    
        \bottomrule 
    \end{tabular}
\caption{Summary of the data set.}
\label{tab:data_summary}
\end{table}

\subsection{Experimental Design}
\label{sec:exp}
The sliding input window \cite{li2014rolling} (also known as look-back window \cite{gidea2018topological} strategy was used to process the entire dataset \cite{shi2023explainable}). For consistency, we used a look-back window of length $w=72$ hours and a prediction window of length $k=24$ hours.
The dataset was split in chronological order with the first 80\% for training and the remaining 20\% for testing. 
The eight DL methods below are used for \texttt{Flood Manager} and \texttt{Flood Evaluator}. 
We run all experiments on one NVIDIA A100 GPU with 80GB memory.
\begin{itemize}
    \item \textbf{MLP}~\cite{suykens1995artificial}: A multilayer perceptron that models non-linear dependencies through fully connected layers;
    \item \textbf{RNN}~\cite{medsker2001recurrent}: Recurrent neural networks are designed for processing sequential data by maintaining hidden states across time steps, making them suitable for time-dependent patterns;
    \item \textbf{CNN}~\cite{o2015introduction}: A 1D convolutional neural network that captures local temporal features using sliding filters;
    \item \textbf{GNN}~\cite{kipf2016semi}: A graph neural network where nodes represent variables and edges model spatial relationships;
    \item \textbf{TCN}~\cite{bai2018empirical}: A temporal convolutional network with dilated convolutions, enabling a large receptive field to capture long-range temporal dependencies;
    \item \textbf{RCNN}~\cite{zhang2020temperature}: A hybrid model that integrates RNN and CNN architectures to jointly learn local and sequential features for time series forecasting;
    \item \textbf{Transformer}~\cite{vaswani2017attention}: An attention-based model for sequence modeling (encoder-only in our setting) to capture global dependencies;
    \item \textbf{GTN} (Ours): Combining GNNs with LSTMs, CNNs, and transformers, as described in Figure \ref{fig:graphtransformer}. 
\end{itemize}

\subsection{Reproducibility}
We include all data and code in a GitHub repository\footnote{Link: \url{https://github.com/JimengShi/FIDLAR}}.

\section{Results}
\subsection{Flood Prediction} 
The role of the \texttt{Flood Evaluator} is to forecast flood events by predicting water levels for given input conditions. 
We set the upper threshold (flood level) at 3.5 feet and the lower threshold (wastage level) at 0.0 feet.
However, the methods remain consistent for many reasonable choices of threshold values. 
We measured accuracy using multiple metrics: (a) mean absolute error (MAE), (b) root mean squared error (RMSE) computed between the predicted and actual water levels, (c) number of time points where the upper or lower thresholds are breached, and (d) the area between water level curves and threshold bars.

\begin{table*}[ht]
\small
\centering
\resizebox{\textwidth}{!}{
    \begin{tabular}{l|cc|cccc}
    \toprule
    \textbf{Methods}    & \textbf{MAE (ft)}   & \textbf{RMSE (ft)} & \textbf{Over Timesteps}   & \textbf{Over Area}  & \textbf{Under Timesteps}   & \textbf{Under Area} \\
    \midrule\midrule
    Ground-truth      & -            & -        & \textcolor{blue}{96}      & \textcolor{blue}{14.82}     & \textcolor{blue}{1,346}    & \textcolor{blue}{385.80}  \\ 
    \midrule
    HEC-RAS           & 0.174        & 0.222    & 68       & 10.07    & 1,133    & 325.33  \\ 
    \midrule
    MLP               & 0.065        & 0.086    & 147      & 27.96    & 1,677    & 500.41  \\  
    RNN               & 0.054        & 0.072    & 110      & 17.12    & 1,527    & 441.41  \\   
    CNN               & 0.079        & 0.104    & 58       & 5.91     & 1,491    & 413.22  \\ 
    GNN               & 0.054        & 0.070    & 102      & 15.90    & 1,569    & 462.63  \\ 
    TCN               & 0.050        & 0.065    & 47       & 5.14     & 1,607    & 453.63  \\  
    RCNN              & 0.092        & 0.110    & 37       & 4.61     & 1,829    & 553.20  \\
    Transformer       & 0.050        & 0.066    & 151      & 25.95    & 1,513    & 434.13  \\ 
    \midrule
    GTN (ours)   & \textcolor{orange}{0.040}        & \textcolor{orange}{0.056}    & \textcolor{red}{100}      & \textcolor{red}{15.64}    & \textcolor{red}{1,390}    & \textcolor{red}{398.84}  \\  
    \bottomrule
    \end{tabular}
}
\caption{Comparison of model performances for the \texttt{Flood Evaluator} on the test set, specifically at time t+1 for measurement station S1. The terms ``Over Timesteps'' and ``Under Timesteps'' indicate the number of time steps during which water levels exceed the upper threshold or fall below the lower threshold, respectively. Similarly, ``Over Area'' and ``Under Area'' pertain to the area between the water level curve and upper or lower threshold, as was illustrated in Figure \ref{fig:loss}. Results in \textcolor{orange}{orange} are the lowest in that column while results in \textcolor{red}{red} are the closest to the ground truth (in \textcolor{blue}{blue}).}
\label{tab:flood_prediction_s1}
\end{table*}

Table \ref{tab:flood_prediction_s1} demonstrates that our model \texttt{GTN} outperforms other models with predictions (in \textcolor{red}{red}) most closely aligned with the ground truth (in \textcolor{blue}{blue}) while achieving the lowest MAE and RMSE (in \textcolor{orange}{orange}). 
Therefore, we choose our GTN model as \texttt{Evaluator} while training \texttt{Manager} in \fidlar.

\subsection{Flood Mitigation with \fidlar} 
\begin{figure}[ht]
\centering
    \includegraphics[width=0.7\columnwidth]{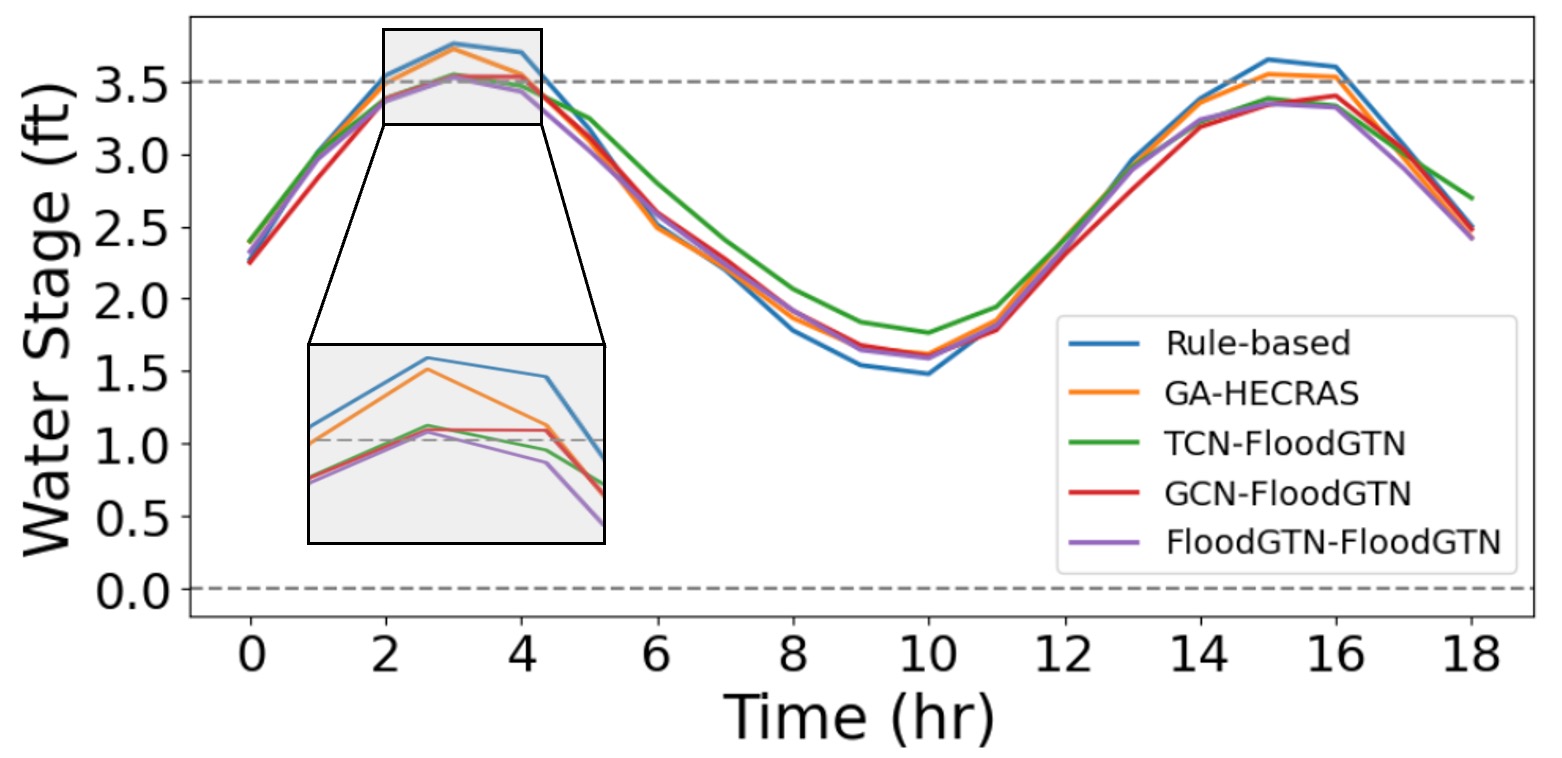} 
\caption{Visualization of water levels with various methods for flood mitigation. We zoomed in $t=2\sim4.5$ in gray. `A'--`B' in legend represents the \texttt{Manager} and \texttt{Evaluator}. Two dashed lines denote the upper (3.5 ft) and lower threshold (0.0 ft).}
\label{fig:visualize_flood_mitigation_s1}
\end{figure}

\begin{figure}[ht]
\centering
    \includegraphics[width=\columnwidth]{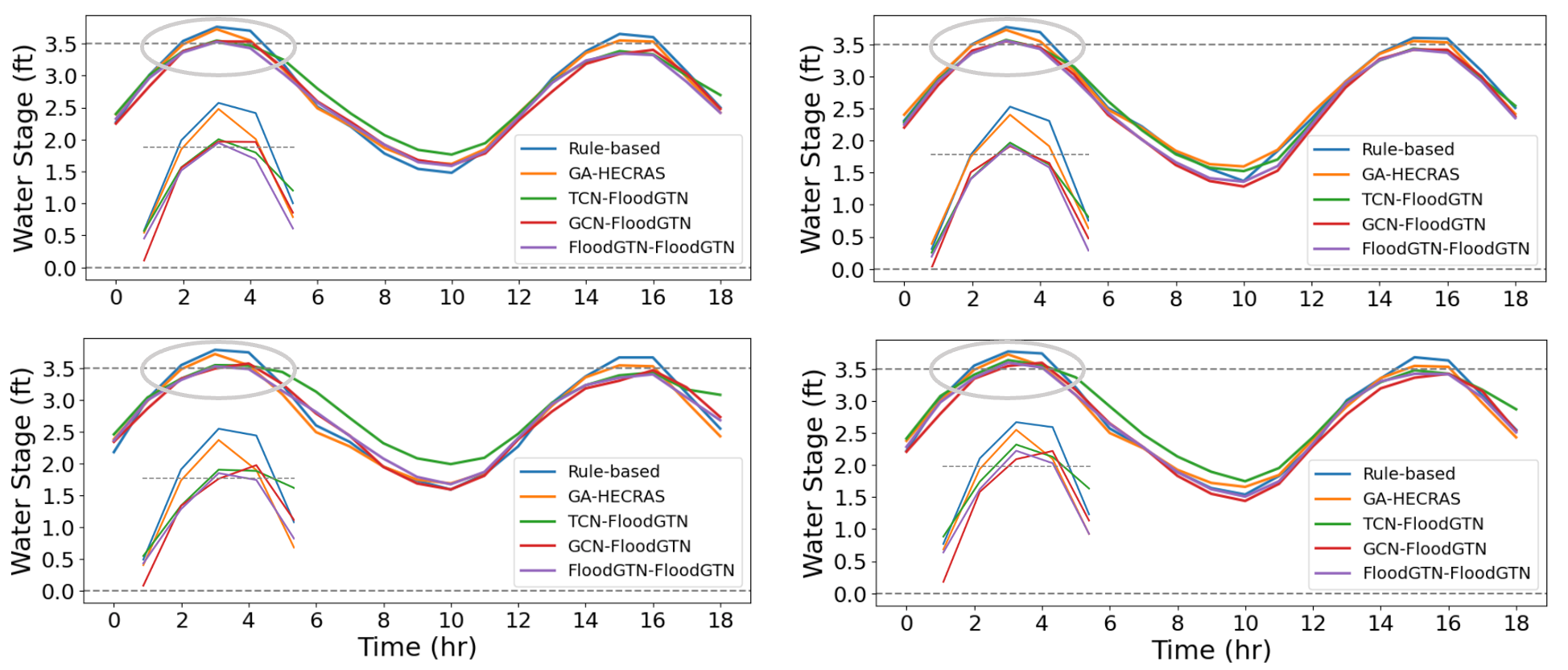} 
\caption{Visualization for flood mitigation at \textbf{all} locations. }
\label{fig:visualize_flood_mitigation_all_locations}
\end{figure}

\fidlar\ requires both \texttt{Evaluator} and \texttt{Manager} components.
For the \texttt{Manager} model, we experimented with one rule-based method, and two genetic algorithms -- one with a physics-based HEC-RAS evaluator \cite{leon2020matlab} and one with our DL-based GTN evaluator, and several DL-based managers using MLP, RNN, CNN, GNN, TCN, RCNN, Transformer, and GTN.
\fidlar\ was measured using (a) the number of time steps where the upper/lower thresholds are exceeded for the water levels, and (b) the area between the water level curves and the threshold bars.
Table \ref{tab:flood_mitigate_s1} shows that all DL-based methods consistently performed better for site S1 than rule-based and GA-based approaches. 
Furthermore, GTN has the best performance under all four metrics, whether it is to control floods or water wastage.
\begin{table*}[ht]
\small
\centering
\resizebox{\textwidth}{!}{
    \begin{tabular}{l|c|cccc}
    \toprule
    \multirow{1}{*}{\textbf{Method}}
        & \textbf{Manager}        & \textbf{Over Timesteps}    & \textbf{Over Area}   & \textbf{Under Timesteps}    & \textbf{Under Area}  \\
    \midrule \midrule 
    \multirow{1}{*}{Rule-based}
        &                   & 96         & 14.82     & 1,346     & 385.8  \\  
    \midrule
    \multirow{2}{*}{GA-based}    
        & Genetic Algorithm$^{*}$     & -          & -         & -         & -  \\
        & Genetic Algorithm$^{\dag}$     & 86         & 16.54     & 454       & 104  \\
    \midrule
    \multirow{8}{*}{DL-based}       
        & MLP                       & 91         & 13.31     & 1,071     & 268.35 \\  
        & RNN                       & 35         & 3.97      & 351       & 61.05  \\   
        & CNN                       & 81         & 11.22     & 1,163     & 314.37  \\ 
        & GNN                       & 31         & 3.72      & 429       & 84.31  \\ 
        & TCN                       & 39         & 3.77      & 306       & 55.12  \\  
        & RCNN                      & 29         & 3.28      & 328       & 58.68  \\
        & Transformer               & 85         & 11.54     & 1,180     & 310.16  \\ 
        & GTN (Ours)                &\textbf{22} &\textbf{2.23} &\textbf{299}  &\textbf{53.34}  \\
    \bottomrule
    \end{tabular}
}
\caption{Comparison of model performances for the \texttt{Flood Manager} on the test set, specifically at time t+1 for measurement station S1. The $^{*}$ denotes that the GA method was used with a physics-based (HEC-RAS) evaluator. The $-$ denotes that the experiments were timed out. The $^{\dag}$ denotes the GA method was used with the GTN as the evaluator. All other rows are DL-based flood managers with a DL-based GTN as the evaluator. 
Results in \textbf{bold} are the best in that column.} 
\label{tab:flood_mitigate_s1}
\end{table*}
We visualize water levels for a short period spanning 18 hours from September 3rd (09:00) to September 4th (03:00) in 2019 for the S1 location.
Figure \ref{fig:visualize_flood_mitigation_s1} indicates that \fidlar\ equipped with GTN model (purple curve) has led to water levels within the upper and lower thresholds, satisfying pre-defined requirements.
Moreover, \fidlar\ presents the best control (i.e., the least water levels beyond thresholds) for flood mitigation and water waste compared to other baselines.
We zoomed in on a 2.5-hour period of the resulting decreased water levels. Visualizations at all locations are in Figure \ref{fig:visualize_flood_mitigation_all_locations}.

\subsection{Ablation Study}
\label{sec:fidlar_ablation}
As introduced in Figure \ref{fig:graphtransformer} in Section \ref{sec:gtn}, the graph transformer architecture (GTN) consists of multiple components.
The \emph{ablation} study in Table \ref{tab:ablation_s1} quantifies the contribution of each component of our GTN model by measuring the performance of GTN after removing the individual components and a part of their combination. 
Removing the combination of ``GNN \& LSTM \& Attention'' (bottom branch in Figure \ref{fig:graphtransformer}) or ``Transformer \& CNN \& Attention'' (upper branch in Figure \ref{fig:graphtransformer}) results in the most significant performance degradation.
For each component, the removal of the GNN module decreases the performance the most, underscoring the importance of spatial dependency modeling in this task. 
Removing the CNN or LSTM components also leads to considerable deterioration, particularly in under-timesteps and under-area, indicating that both local temporal feature extraction and sequential modeling are essential.
Interestingly, the exclusion of the Transformer and attention mechanisms results in moderate drops, suggesting that while global context is helpful, the model remains somewhat resilient without it.
However, the full integration of all components, CNN, LSTM, GNN, Transformer, and attention, yields the best overall performance, validating the architectural design of GTN as a synergistic combination of spatiotemporal modeling techniques.
\begin{table}[H]
\small
\centering
\resizebox{0.85\textwidth}{!}{
    \begin{tabular}{l|cccc}
    \toprule
    \textbf{}          & \textbf{Over}    & \textbf{Over}   & \textbf{Under}    & \textbf{Under}  \\
    \textbf{Method}    & \textbf{Timesteps}    & \textbf{Area}   & \textbf{Timesteps}    & \textbf{Area}  \\
    \midrule \midrule     
     w/o CNN           & 37         & 4.37     & 476         & 85.54  \\  
     w/o Transformer   & 32         & 3.57     & 325         & 57.42  \\   
     w/o GNN           & 56         & 5.90     & 479         & 86.22  \\ 
     w/o LSTM          & 35         & 4.34     & 329         & 56.74  \\ 
     w/o Attention     & 32         & 3.59     & 341         & 60.48  \\  
     w/o Transformer, CNN, Attention & 69  & 11.24    & 828         & 215.94  \\  
     w/o GNN, LSTM, Attention & 78         & 9.65     & 761         & 198.33  \\  
    \midrule
    GTN (Ours)         &\textbf{22} & \textbf{2.23}  & \textbf{299}  & \textbf{53.34}  \\  
    \bottomrule
    \end{tabular}
}
\caption{Ablation study for flood mitigation for the entire test set (for time point t+1 at S1). The last row indicates the performance of the \fidlar\ system with GTN as proposed in Figure \ref{fig:graphtransformer}. The best results in the last row are in bold.}
\label{tab:ablation_s1}
\end{table}

\subsection{Analysis of Computational Time} 
Since \fidlar\ was designed for real-time flood control, we measured the running times of the models used in this work and the previous genetic algorithm (GA) associated with physics-based flood simulators.
Table \ref{tab:flood_prediction_time} shows the running times for the whole flood prediction and mitigation system in its training and test phases. 
All the DL-based approaches in the test phase are several orders of magnitude faster than the currently used physics-based and GA-based approaches for the flood mitigation task. 
Rapid inference is a critical property of data-driven DL methods.
The table also shows the training times for the DL-based approaches, although they are not necessary for the deployment in reality.
\begin{table}[ht]
\small
\centering
\resizebox{0.65\textwidth}{!}{
    \begin{tabular}{l|cccc}
        \toprule
        \multirow{2}{*}{\textbf{Model}}  & \multicolumn{2}{c}{\textbf{Flood Prediction}}        & \multicolumn{2}{c}{\textbf{Flood 
 Mitigation}} \\
                           & \textbf{Train}     & \textbf{Test}  & \textbf{Train}   & \textbf{Test} \\
        \midrule\midrule
        HEC-RAS            & $\times$       & 45 min      & $\times$       & $\times$   \\ 
        Rule-based         & /              & /           & /              & /   \\
        GA$^{*}$           & $\times$       & $\times$    & $\times$              & --     \\ 
        GA$^{\dag}$        & $\times$       & $\times$    & $\times$              & est. 30 h  \\ 
        \midrule
        MLP                & 35 min         & 1.88 s      & 58 min         & 6.13 s \\
        RNN                & 243 min        & 8.57 s      & 54 min         & 12.75 s \\
        CNN                & 37 min         & 1.93 s      & 17 min         & 5.84 s \\
        GNN                & 64 min         & 3.13 s      & 29 min         & 7.26 s \\
        TCN                & 60 min         & 4.57 s      & 45 min         & 9.06 s \\
        RCNN               & 136 min        & 8.61 s      & 61 min         & 13.27 s \\
        Transformer        & 43 min         & 2.38 s      & 23 min         & 6.76 s \\
        GTN (Ours)         & 119 min        & 2.95 s      & 35 min         & 4.90 s \\
        \bottomrule
    \end{tabular}%
}
\caption{Running time for flood prediction and mitigation. The running time for the rule-based method was not reported since historical data was directly used, denoted with ``$/$''.
``$\times$'' represents that the methods are not applicable for the tasks. 
GA$^{*}$, which combines a GA-based tool and HEC-RAS for flood mitigation (test phase), took too long and was not reported, represented with ``--''. 
GA$^{\dag}$, which combines the GA-based tool with GTN for flood mitigation (test phase), also took too long but was estimated using a smaller sample.}
\label{tab:flood_prediction_time}
\end{table}

\subsection{Explainability}
\label{sec:explain}
\begin{figure}[ht]
\centering
    \includegraphics[width=0.9\columnwidth]{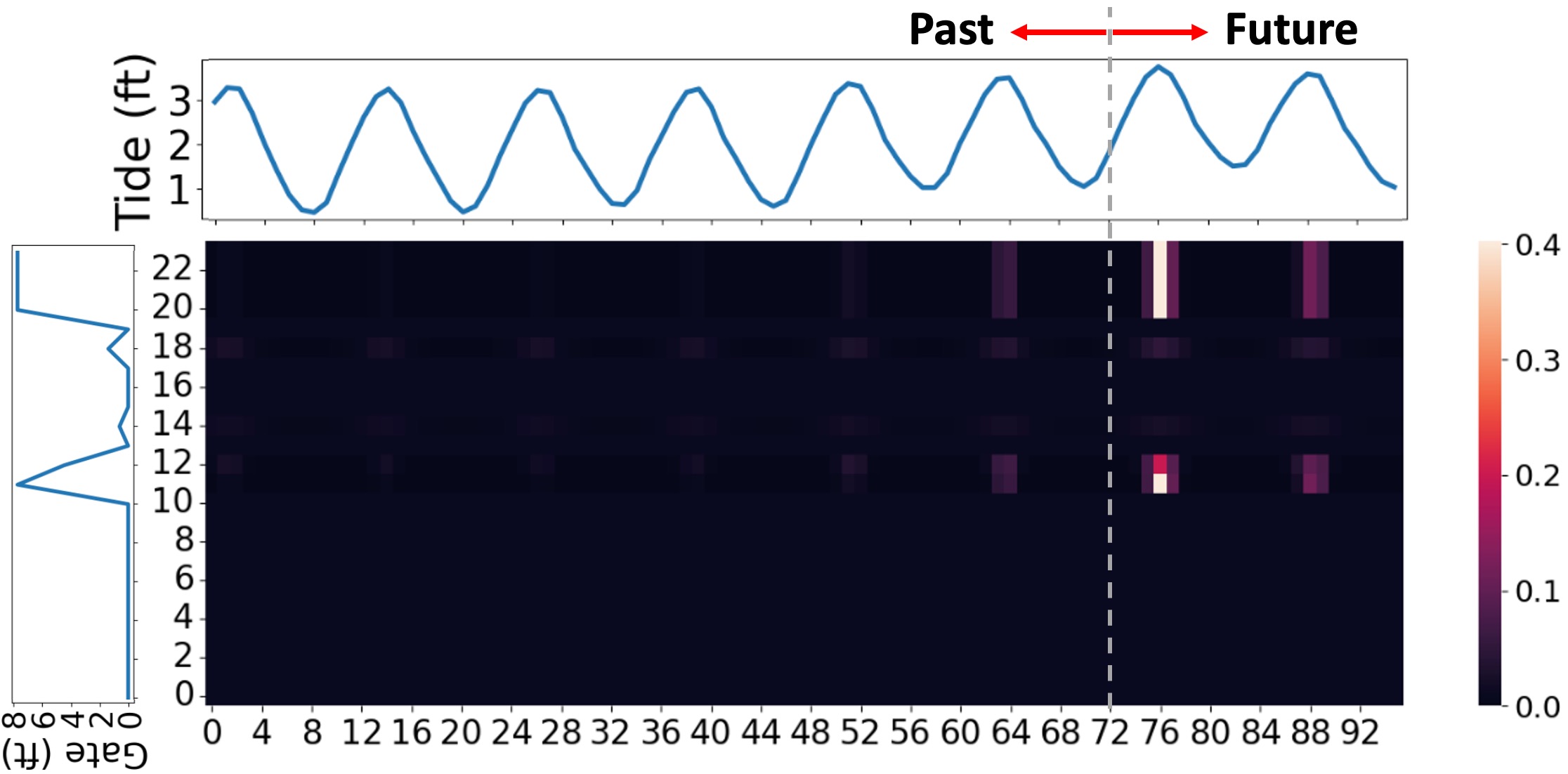} 
\caption{Importance scores of tide input. $x$ and $y$ axes are the tide and control schedule of the gate over time.}
\label{fig:gate_tide_attention}
\end{figure}

It is also important to figure out how and why the model generates the outputs.
Attention-based methods allow us to calculate the ``attention scores'' assigned to an input variable to compute a specific output variable.
The heatmap in Figure \ref{fig:gate_tide_attention} presents the attention scores assigned to the tide (columns) to compute the gate schedule output (rows) for 24 hours into the future.
Note that there are 96 columns and 24 rows because we use 72 hours of past tidal observations and 24 hours of future predicted tidal data to predict 24 hours of the gate schedule into the future.
The rows $[0,23]$ correspond to the 24 hours into the future, while the columns $[0,95]$ also include 72 hours of the recent past and 24 hours of the future predicted tidal data.
Therefore, $t=72$ corresponds to the ``current'' time point, and the columns $[72,95]$ correspond to the same time points as rows $[0,23]$.

\section{Discussion}
\subsection{Model Explainability}
The explainability feature, which is shown with an example in Figure \ref{fig:gate_tide_attention}, can provide significant insights into our results.

Firstly, we point out that the brightest patches are in the last 24 columns of the heatmap.
Thus \fidlar\ pays greater attention to the 24 hours of future predicted tidal data than the past 72 hours, highlighting the importance of forecast information to a DL-based approach to flood mitigation.
While tides may have a more predictable pattern over time, the contribution of rain to the water levels can also be seen for other time points.
Since our study domain is a coastal river system, tidal information dominates the changes of water levels, and rainfall events have relatively less impact. More results on model explainability are referred to \cite{shi2023power, shi2023explainable}.

A second critical insight is that the brightest attention patches are in columns where the tide is at its highest is critical to the prediction of gate schedules.
Additionally, the water level at the first high tide peak after the ``current'' time is more significant than the other two.
Third, the gate schedule has peaks at times $t=11$ hour and $t=22$ hour into the future, which correspond to the lowest points of the tide.
This implies that the optimal time for pre-releasing water is during low tide phases. 
Opening gates during high tide periods in coastal river systems is less advisable, as it may lead to water flowing back upstream from the ocean.

Finally, we observe that there is a light patch around column $t=65$ hour, suggesting mild attention for the previous high tide peak, but almost no attention to any of the peaks before that.
This again suggests that we could have chosen to use a smaller window for the past input.
Doing this analysis could provide evidence for the right value of $w$, the size of the look-back window.

\subsection{Spatio-temporal Modeling}
We developed the graph transformer model with a combination of GNN and LSTM on one arm, and a Transformer and CNN on another arm, both combined using an cross-attention module. 
Specifically, a GNN was used to capture the spatial relationship of the water levels between different water stations. Each station is considered as one node and the directed edges represent the direction of water flow from one station to another. In contrast, the transformer was used to model the temporal dependencies of the covariates in the time series (e.g., rainfall, tidal information, control schedules of gates and pumps).
We conjecture that this combination of spatial and temporal modeling is a powerful way to model complex environmental systems.
The ablation study in Section \ref{sec:fidlar_ablation} quantifies the contribution of individual components to the overall performance of \fidlar.

\subsection{Generalization to Other River Systems}
The models for flood prediction and mitigation introduced in this chapter can be readily applied to other single-region river systems, but they need to be re-trained using the domain-specific data (e.g., water levels, rainfall, control schedules for gates and pumps, etc.).
The limitations to more complex river systems with multiple cascading gates and pumps have not been addressed in this dissertation and are discussed below in Section \ref{sec:limitations}.

\subsection{Deep Learning for Optimization}
Traditional optimization methods require precise physical modeling. If the models involve nonlinear functions, as may be the case in complex river systems, then they are likely to be computationally intensive.
Deep learning models can approximate these complex relationships and serve as fast surrogates for physical simulations, enabling more efficient and adaptive optimization strategies.
In this chapter, accurate predictions of the water levels under different control schedules for gates and pumps was used to solve the optimization problem of predicting a good control schedule that would minimize flood risks.
The question of whether a prediction tool in tandem with an evaluation tool is an appropriate architecture for other optimization problems remains an \emph{open problem}.

\section{Limitations} 
\label{sec:limitations}
\subsection{Simple Regional Domain}
All the experiments reported in this dissertation involve one section of the river, meaning that none of the measuring stations had a gate, pump, or some hydraulic structure separating them. 
However, real-world river systems are more complex with many hydraulic structures and water stations. The solutions in this chapter are not expected to work for river systems with multiple sections without considerable changes in the architecture of the model. 
This is because the models have to account for the simple fact that water levels in one section are independent of the levels in other sections if the gates and pumps are closed. 
Therefore, extending our methodology to solve flood management in more complex river systems is viewed as future work.

\subsection{Human Intervention}
The implemented DL-based methods have not been tested in real-life scenarios.
Given that human response times for gate adjustments are typically much longer (around 6 hours) than the computation time of machine learning model outputs, we could manage control strategies by maintaining a fixed schedule over each 6-hour window while preserving the overall area under the control curve. 
Specifically, ML models first generate the fine-grained, high-resolution, fluctuating control signals, and then we can easily convert them into coarser, human-executable ones by smoothing techniques in the post-processing. 
This approach will strike a balance between the model’s responsiveness and the feasibility of manual intervention, making the implementation remain practical and aligned with real-world operational constraints. 
We point out that this reconciliation is crucial in translating ML-based recommendations into actionable flood control strategies, particularly in emergency scenarios where clarity, stability, and operational simplicity are essential.


\section{Conclusions}
In this chapter, we discuss the shortcomings of the current approaches for flood mitigation. To address the challenges, we propose \fidlar, a DL-based tool to address the problem. \fidlar\ can compute water ``pre-release'' schedules for hydraulic structures in a river system to achieve effective and efficient flood mitigation, while ensuring that water wastage is avoided.
This was made possible by the use of well-crafted loss functions for the DL models.
The dual component design (with a \texttt{Manager} and an \texttt{Evaluator}) is a strength of \fidlar. 
It exploits the gradient-based planning and the differentiability of the trained \texttt{Evaluator} model for better optimization.
During training, the gradient-based back-propagation from the \texttt{Evaluator} helps to reinforce the \texttt{Manager}.

All the DL-based versions of \fidlar\ are several orders of magnitude faster than the (physics-based or GA-based) competitors while achieving improvement over other methods in flood mitigation. 
These characteristics allow us to entertain the possibility of real-time flood management, which was challenging for previous approaches.

\chapter{\codicast: CONDITIONAL DIFFUSION MODEL FOR GLOBAL WEATHER PREDICTION WITH UNCERTAINTY QUANTIFICATION}	
\label{sec:codicast}

\section{Background}
Accurate weather forecasting is crucial for a wide range of societal activities, from daily planning to disaster preparedness \cite{merz2020impact, shi2024fidlar}. 
For example, governments, organizations, and individuals rely heavily on weather forecasts to make informed decisions that can significantly impact safety, economic efficiency, and overall well-being. 
However, weather predictions are intrinsically uncertain largely due to the complex and chaotic nature of atmospheric processes \cite{slingo2011uncertainty}. 
Therefore, assessing the range of probable weather scenarios is significant, enabling informed decision-making.
The work presented in this chapter is highlighted in the following publication \cite{shi2024codicast}.

Traditional physics-based numerical weather prediction (NWP) methods achieve weather forecasting by approximately solving the differential equations representing the integrated system between the atmosphere, land, and ocean \cite{price2023gencast, nguyen2023climax}.
However, running such an NWP model can produce only one possibility of the forecast, which ignores the weather uncertainty.
To solve this problem, \textit{Ensemble forecast}\footnote{Generating a set of forecasts, each of which represents a single possible scenario.} of multiple models is often employed to model the probability distribution of different future weather scenarios \cite{palmer2019ecmwf, leinonen2023latent}.
While such physics-based ensemble forecasts effectively model the weather uncertainty, they have two primary limitations: physics-based models inherently make restrictive assumptions of atmospheric dynamics, and running multiple NWP models requires extreme computational costs \cite{rodwell2007using}.

In recent years, machine learning (ML)-based weather predictions (MLWP) have been proposed to challenge NWP-based prediction methods \cite{ben2024rise, bulte2024uncertainty}. 
They have achieved enormous success with comparable accuracy and a much (usually three orders of magnitude) lower computational overhead. 
They are typically trained to learn weather patterns from a huge amount of historical data and predict the mean of the probable trajectories by minimizing the mean squared error (MSE) of model forecasts \cite{hewage2021deep}. 
Representative work includes Pangu \cite{bi2023accurate}, GraphCast \cite{lam2023learning}, ClimaX \cite{nguyen2023climax}, ForeCastNet \cite{pathak2022fourcastnet}, Fuxi \cite{chen2023fuxi}, Fengwu and \cite{chen2023fengwu}.
Despite the notable achievements of these MLWP methods, most of them are deterministic \cite{kochkov2024neural}, falling short in capturing the uncertainty in weather forecasts \cite{jaseena2022deterministic}.

To compute the uncertainty for ML models, two methods exist. 
Perturbing initial conditions \cite{morley2018perturbed} helps estimate the aleatoric uncertainty (data noise), while the Monte Carlo Dropout approach \cite{gal2016dropout} estimates epistemic uncertainty (model uncertainty) \cite{siddique2022survey}.
However, neither approach fully captures uncertainty in both the input conditions and the evolution of weather models.
Additionally, these methods require manual tuning of perturbations and dropout rates, which can negatively impact model accuracy. 
These limitations motivate us to explore an approach for comprehensive uncertainty quantification without the loss of accuracy.

Denoising diffusion probabilistic models (DDPMs) \cite{ho2020denoising} stand out as a probabilistic type of generative model, which can generate high-quality images.
By explicitly and iteratively modeling the noise additive and its removal, DDPMs can capture intricate details and textures of images.
Furthermore, controllable diffusion models \cite{rombach2022high,zhang2023adding} enable the generation process to be guided by specific attributes or conditions, e.g., class labels, textual descriptions, or other auxiliary information. By doing so, the models can generate images that adhere to the specified conditions.
This inspires us to consider the weather ``prediction'' tasks as ``generation'' tasks - generating plausible weather scenarios with conditional diffusion models. Promising potentials could be the following: 
(1) Weather numerical data is usually a 2-D grid over latitude and longitude, sharing a similar modality with the image. Diffusion models can capture the intricate weather distribution with iterative denoising.
(2) Weather states from the recent past (i.e., initial conditions) can be injected into diffusion models to guide the generation of future weather evolution.
(3) More notably, the starting noise sampling from the Gaussian distribution can mimic the aleatoric uncertainty, while iteratively adding and removing noise captures the epistemic uncertainty.
These features prompt probabilistic diffusion models to generate a set of diverse weather scenarios rather than a single deterministic one. 
This capability makes them well-suited for modeling the uncertain nature of weather evolution.

In this chapter, we identify the shortcomings of current weather prediction methods. Physics-based NWP methods are limited to restrictive assumptions and are computationally intensive. Moreover, a single deterministic NWP- and MLWP-based method cannot achieve uncertainty quantification.
To address these problems, we propose \codicast, a conditional diffusion model for global weather prediction conditioning on observations from the recent past while probabilistically modeling the uncertainty. In addition, we use the cross-attention mechanism to effectively integrate conditions into the denoising process to guide the generation tasks.
We conduct extensive experiments on the ERA5 reanalysis data from the European Centre for Medium-Range Weather Forecasts (ECMWF), and demonstrate that \codicast achieves an essential trade-off among accuracy, efficiency, and uncertainty against state-of-the-art baselines.

\section{Related Work}
\paragraph{Physics-based Numerical Weather Prediction.} 
Physics-based numerical Weather Prediction (NWP) methods achieve weather forecasts by modeling the system of the atmosphere, land, and ocean with complex differential equations \cite{bauer2015quiet}. 
For example,the  High-Resolution Forecasts System (HRES) \cite{HRES2023ECMWF} forecasts possible weather evolution out to 10 days ahead. 
However, it is a deterministic NWP method that only provides a single forecast. 
To overcome the limitation of deterministic methods, the ensemble forecast suite (ENS) \cite{buizza2008comparison} was developed as an ensemble of 51 forecasts by the European Centre for Medium-Range Weather Forecasts (ECMWF). ENS provides a range of possible future weather states, allowing for investigation of the detail and uncertainty in the forecast.
Even if NWP ensemble forecasts effectively model the weather evolution, they exhibit sensitivity to structural discrepancies across models and high computational demands \cite{balaji2022general}.

\paragraph{ML-Based Weather Prediction.} 
Machine learning-based weather prediction (MLWP) approaches have challenged NWP methods for weather forecasting. 
Pangu \cite{bi2023accurate} employed three-dimensional transformer networks and Earth-specific priors to deal with complex patterns in weather data.
GraphCast \cite{lam2023learning} achieved medium-range weather prediction by utilizing an ``encode-process-decode'' configuration with each part implemented by graph neural networks (GNNs). GNNs perform effectively in capturing the complex relationship between a set of surface and atmospheric variables. A similar GNN-based work is \cite{keisler2022forecasting}. 
Fuxi \cite{chen2023fuxi} and Fengwu \cite{chen2023fengwu} also employ the ``encode-decode'' strategy but with the transformer-based backbone.
FourCastNet \cite{pathak2022fourcastnet} applied Vision Transformer (ViT) and Adaptive Fourier Neural Operators (AFNO), while ClimaX \cite{nguyen2023climax} also uses a ViT backbone but the trained model can be fine-tuned to various downstream tasks.
However, these models fall short in modeling the uncertainty of weather evolution \cite{jaseena2022deterministic} even though perturbations are added to initial conditions \cite{bulte2024uncertainty} and dropout methods \cite{gal2016dropout} are used to mimic the uncertainty.
Additionally, ClimODE \cite{verma2024climode} incorporated the physical knowledge and developed a continuous-time neural advection PDE weather model.

\paragraph{Diffusion Models.} 
Diffusion models \cite{ho2020denoising} have shown their strong capability in computer vision tasks, including image generation \cite{li2022srdiff}, image editing \cite{nichol2021glide}, semantic segmentation \cite{brempong2022denoising} and point cloud completion \cite{luo2021diffusion}.
Conditional diffusion models \cite{ho2022classifier} were later proposed to make the generation step conditioned on the current context or situation.
However, not many efforts have adopted diffusion models in global medium-range weather forecasting.
More recent research has focused on precipitation nowcasting \cite{asperti2023precipitation, gao2024prediff}, and is localized in its predictions. 
GenCast \cite{price2023gencast} is a recently proposed conditional diffusion-based ensemble forecasting for medium-range weather prediction.
However, their conditioning is to directly use the observations from the recent past, which is shown to be insufficient (see the last case in the ablation study). More related work can be found in Figure \ref{fig:taxonomy}.

\section{Preliminary Knowledge}

\subsection{Problem Formulation}
\paragraph{Deterministic Global Weather Predictions.} 
Given the input consisting of the weather state(s), $X^t \in \mathbb{R}^{H \times W \times C}$ at time $t$, the problem is to predict a point-valued weather state, $X^{t+\Delta t} \in \mathbb{R}^{H \times W \times C}$ at a future time point $t+\Delta t$.
Here $H \times W$ refers to the spatial resolution of the data, which depends on how densely we grid the globe over latitudes and longitudes, $C$ refers to the number of channels (i.e., weather variables), and the superscripts $t$ and $t+\Delta t$ refer to the current and future time points. The long-range multiple-step forecasts could be achieved by autoregressive modeling or direct predictions. 

\paragraph{Probabilistic Global Weather Predictions.} 
Unlike the deterministic models that output point-valued predictions, probabilistic methods model the probability of future weather state(s) as a distribution $P(X^{t+\Delta t} \mid X^t)$, conditioned on the state(s) from the recent past. Probabilistic predictions are appropriate for quantifying the forecast uncertainty and making informed decisions.
\begin{figure}[ht]
\centering
    \includegraphics[width=0.7\textwidth]{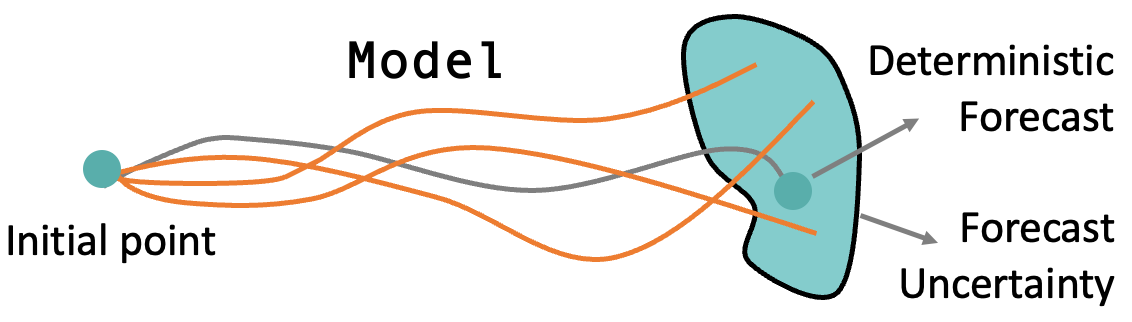}
\caption{Deterministic vs Probabilistic Models.}
\label{fig:prob_prediction}
\end{figure}

\subsection{Denoising Diffusion Probabilistic Models}
\label{sec:ddpm}
A denoising diffusion probabilistic model (DDPM) \cite{ho2020denoising} generates target samples by learning a distribution $p_\theta(x_0)$ that approximates the target distribution $q(x_0)$.
DDPM comprises a \emph{forward diffusion} process and a \emph{reverse denoising} process. 
The \emph{forward process} 
transforms an input $x_0$ with a data distribution of $q(x_0)$ to a Gaussian noise vector $x_N$ in $N$ diffusion steps. It can be described as a Markov chain that gradually adds Gaussian noise to the input according to a variance schedule $\{\beta_1, \dots, \beta_N\}$:
\begin{equation*}
    q(x_n \mid x_{n-1}) = \mathcal{N}\left(x_n; \sqrt{1 - \beta_n} x_{n-1}, \beta_n \textbf{I}\right), \mathrm{ and}
\end{equation*}
\begin{equation*}
    q(x_{1:N} \mid x_0) = \prod_{n=1}^{N} q(x_n \mid x_{n-1}), n \in [1, N],
\end{equation*}
where at each step $n$, the diffused sample $x_n$ is obtained from $x_{n-1}$ as described above. 
Multiple steps of the forward process can be described as follows in a closed form:
\begin{equation*}
    q(x_n \mid x_0) = \mathcal{N}\left(x_n; \sqrt{\bar{\alpha}_n}x_0, (1-\bar{\alpha}_n) \textbf{I}\right), 
    \label{eq:forward}
\end{equation*}
where $\alpha_n = 1 - \beta_n$ and $\bar{\alpha}_n = \prod_{s=1}^{n} \alpha_s$. Thus, $x_n = \sqrt{\bar{\alpha}_n} x_0 + \sqrt{1-\bar{\alpha}_n} \epsilon$, with $\epsilon$ sampled from $\mathcal{N}(\textbf{0}, \textbf{I})$.

In the \emph{reverse process}, the \emph{denoiser} network is used to recover $x_0$ by stepwise denoising starting from the pure noise sample, $x_N$. This process is defined as:
\begin{equation}
    p_\theta(x_{0:N}) = p(x_N)\prod_{n=1}^{N} p_\theta(x_{n-1} \mid x_{n}),
    \label{eq:reverse}
\end{equation}
where $p_\theta(x_n)$ is the distribution at step $n$ parameterized by $\theta$.

For each iteration, $n \in [1, N]$, diffusion models are trained to minimize the following KL-divergence between the true and generated descriptions:
\begin{equation}
    \mathcal{L}_n = D_{KL}\left(q(x_{n-1} \mid x_n) \mid\mid p_\theta(x_{n-1} \mid x_n) \right).
    \label{eq:kl_loss}
\end{equation}

\section{Methodology}

\subsection{Method Overview}
\begin{figure}[ht!]
\centering
    \includegraphics[width=0.95\textwidth]{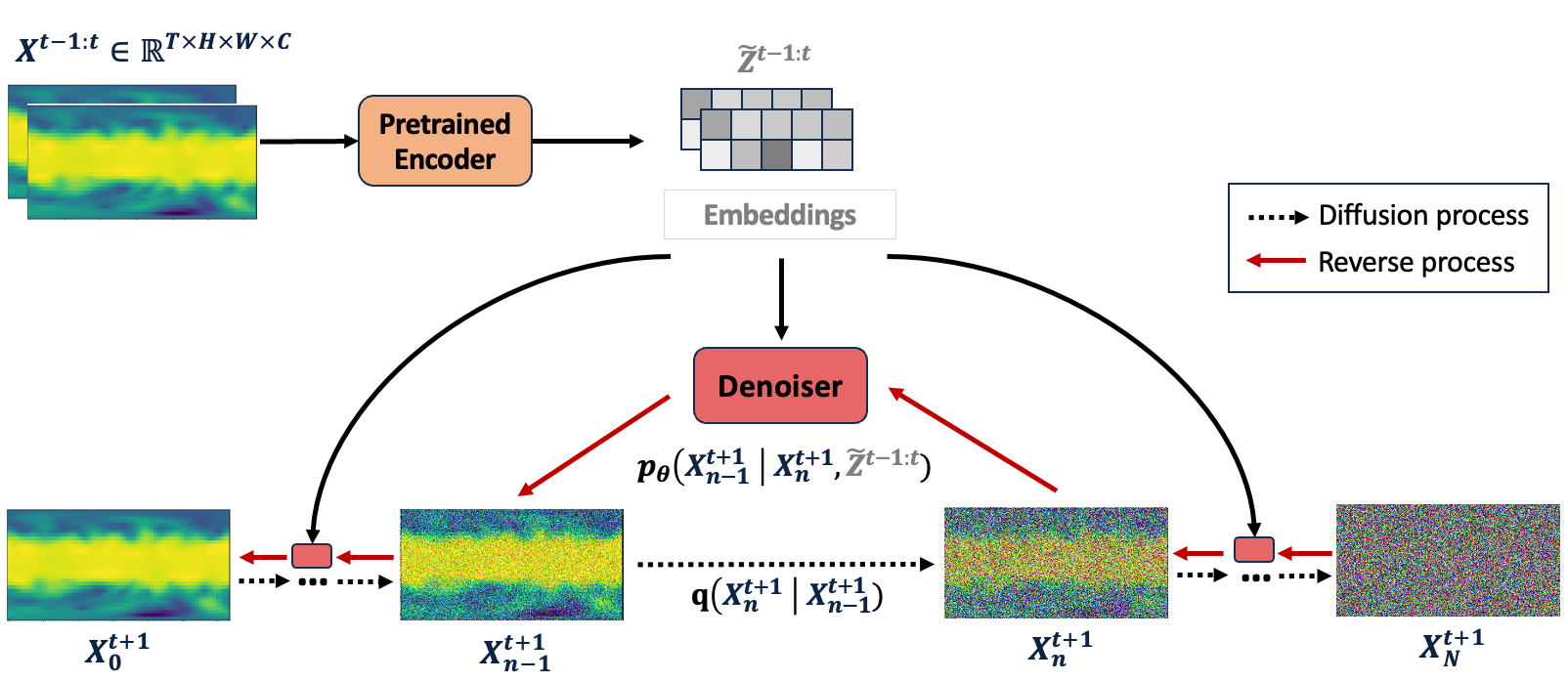}
    \caption{Framework of our conditional diffusion model for global weather forecast \codicast. The superscript $T$ and the subscript $N$ denote the time point and iteration step of adding/denoising noise. $H$ and $W$ represent the height (\#latitude) and width (\#longitude) of grid data. $C$ is the number of variables of interest. $X$ is the observation data and $\tilde{Z}$ is the feature representation in the embedding space.}
    \label{fig:framework}
\end{figure}

Figure~\ref{fig:framework} illustrates the overall framework of our proposed approach, \codicast, designed for global weather prediction using a conditional diffusion model. The core idea is to reformulate the forecasting problem as a conditional generation task, where the model learns to generate future states conditioned on recent past observations.

Given a sequence of past observations $X^{t-1:t} \in \mathbb{R}^{T \times H \times W \times C}$, a pretrained encoder first extracts spatiotemporal embeddings that serve as contextual conditioning information. 
The diffusion process progressively corrupts the future target state $X^{t+1}_0$ with noise through a forward stochastic process, resulting in a highly noisy representation $X^{t+1}_N$. The denoising model, guided by both the noisy input and the encoded context, then iteratively reverses the diffusion process to reconstruct the future weather state step-by-step. 
This formulation enables the model to capture uncertainty and generate probabilistic forecasts in a principled manner, leveraging the strengths of diffusion models while incorporating weather temporal context.

\subsection{Forward Diffusion Process}
At time point $t$, the forward diffusion process assumes that a pure noise sample $X^{t+1}_N$ is generated from $X^{t+1}_0 \in \mathbb{R}^{H \times W \times C}$ by adding noise $N$ times (see the dotted lines in Figure \ref{fig:framework}):
\begin{equation}
    X^{t+1}_n = \sqrt{\bar{\alpha}_n} \cdot X^{t+1}_0 + \sqrt{1-\bar{\alpha}_n} \epsilon,
\end{equation}
where $\epsilon$ is sampled from $\mathcal{N}(\mathbf{0}, \mathbf{I})$ with the same dimensions as $X^{t+1}_{0}$, and $\bar{\alpha}$ is as described in Section~\ref{sec:ddpm}).

\subsection{Reverse Conditional Denoising Process}
\codicast models the probability distribution of the future weather state, conditioned on the current and previous weather states. More specifically, we exploit a pre-trained encoder to learn conditions as embedding representations of the past observations $X^{t-1}$ and $X^t$, which are used to control and guide the synthesis process. 
Working in the latent space of embeddings works better than the original space of the observations.
\begin{equation}
    p_\theta(X_{0:N}^{t+1} \mid \tilde{Z}^{t-1:t}) = p(X_N^{t+1})\prod_{n=1}^{N} p_\theta(X_{n-1}^{t+1} \mid X_{n}^{t+1}, \tilde{Z}^{t-1:t}),
    \label{eq:condition_reverse}
\end{equation}
where $X_N^{t+1} \sim \mathcal{N}(\textbf{0}, \textbf{I})$, $\tilde{Z}^{t-1:t}$ is the embedding representation as shown in Eq. (\ref{eq:encoder}).

After prediction at the first time point is obtained, a forecast trajectory, $X^{1:T}$, of length $T$, can be auto-regressively modeled by conditioning on the predicted ``previous'' states.
\begin{equation}
    p_\theta(X_{0:N}^{1:T}) = \prod_{t=1}^{T} p(X_N^{t}) \prod_{n=1}^{N} p_\theta(X_{n-1}^{t} \mid X_{n}^{t}, \tilde{Z}^{t-2:t-1}).
    \label{eq:multi_prediction}
\end{equation}

\subsection{Pre-trained Encoder}
\label{sec:pretrain_encoder}
We learn an encoder by training an autoencoder network \cite{baldi2012autoencoders}.
An \texttt{Encoder} transforms the input at each time point into a latent-space representation, while \texttt{Decoder} reconstructs the input from the latent representation. 
After the encoder, $\mathcal{F}$, is trained, it can serve as a pre-trained representation learning model to project the original data into a latent embedding in Eq. (\ref{eq:encoder}).
\begin{equation}
    \tilde{Z}^{t-1:t}= \mathcal{F}(X^{t-1}, X^t).
\label{eq:encoder}
\end{equation}
\begin{figure}[ht]
\centering
    \includegraphics[width=0.7\textwidth]{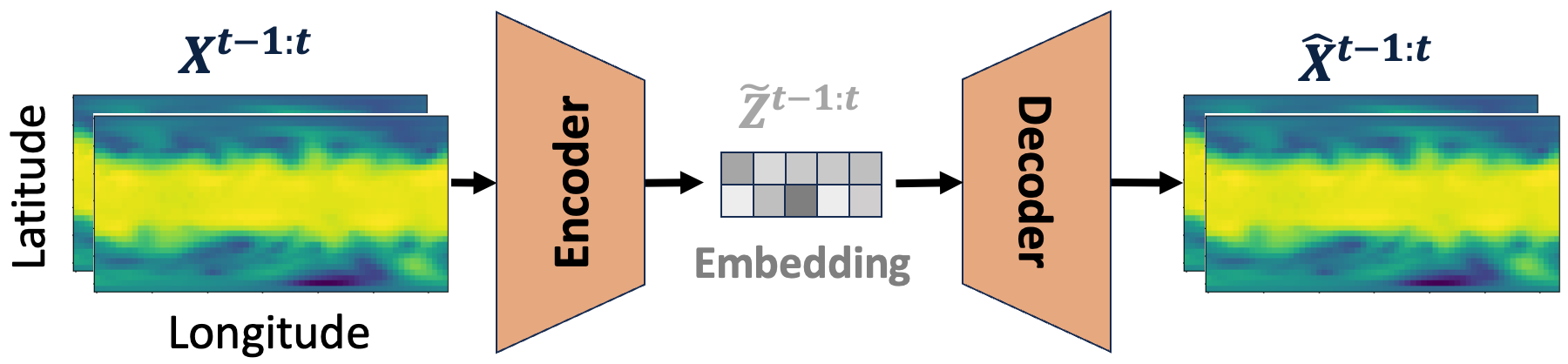}
\caption{Autoencoder structure.}
\label{fig:autoencoder}
\end{figure}

\subsection{Attention-based Denoiser Network}
\begin{figure}[ht]
\centering
    \includegraphics[width=0.55\textwidth]{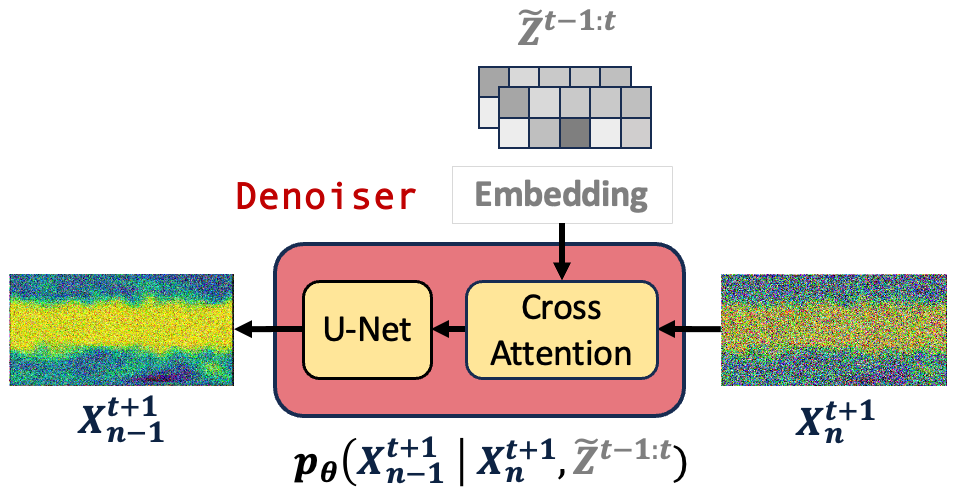}
\caption{Attention-based denoiser structure.}
\label{fig:denoiser}
\end{figure}

Our denoiser network consists of two blocks: cross-attention and U-net (as shown in Figure \ref{fig:denoiser}). 
Cross-attention mechanism \cite{hertz2022prompt} is employed to capture how past observations can contribute to the generation of future states.
The embedding of past observations, $\tilde{Z}^{t-1:t}$, and the noise data $X_n^{t+1}$ at diffusion step $n$, are projected to the same hidden dimension $d$ with the following transformation: 
%
\begin{equation}
    Q=W_q \cdot X_n^{t+1}, K=W_k \cdot \tilde{Z}^{t-1:t}, V=W_v \cdot \tilde{Z}^{t-1:t},
\end{equation}
%
\noindent where $X_n^{t+1} \in \mathbb{R}^{(H \times W) \times C}$ and $\tilde{Z}^{t-1:t} \in \mathbb{R}^{(H \times W) \times d_z}$. $W_q \in \mathbb{R}^{d \times C}, W_k \in \mathbb{R}^{d \times d_z}, W_v \in \mathbb{R}^{d \times d_z}$ are learnable projection matrices. Then we implement the cross-attention mechanism by $\texttt{Attention(Q, K, V)} = \texttt{softmax}(\frac{QK^T}{\sqrt{d}})V$. 

\begin{figure*}[ht]
\centering
    \includegraphics[width=0.98\textwidth]{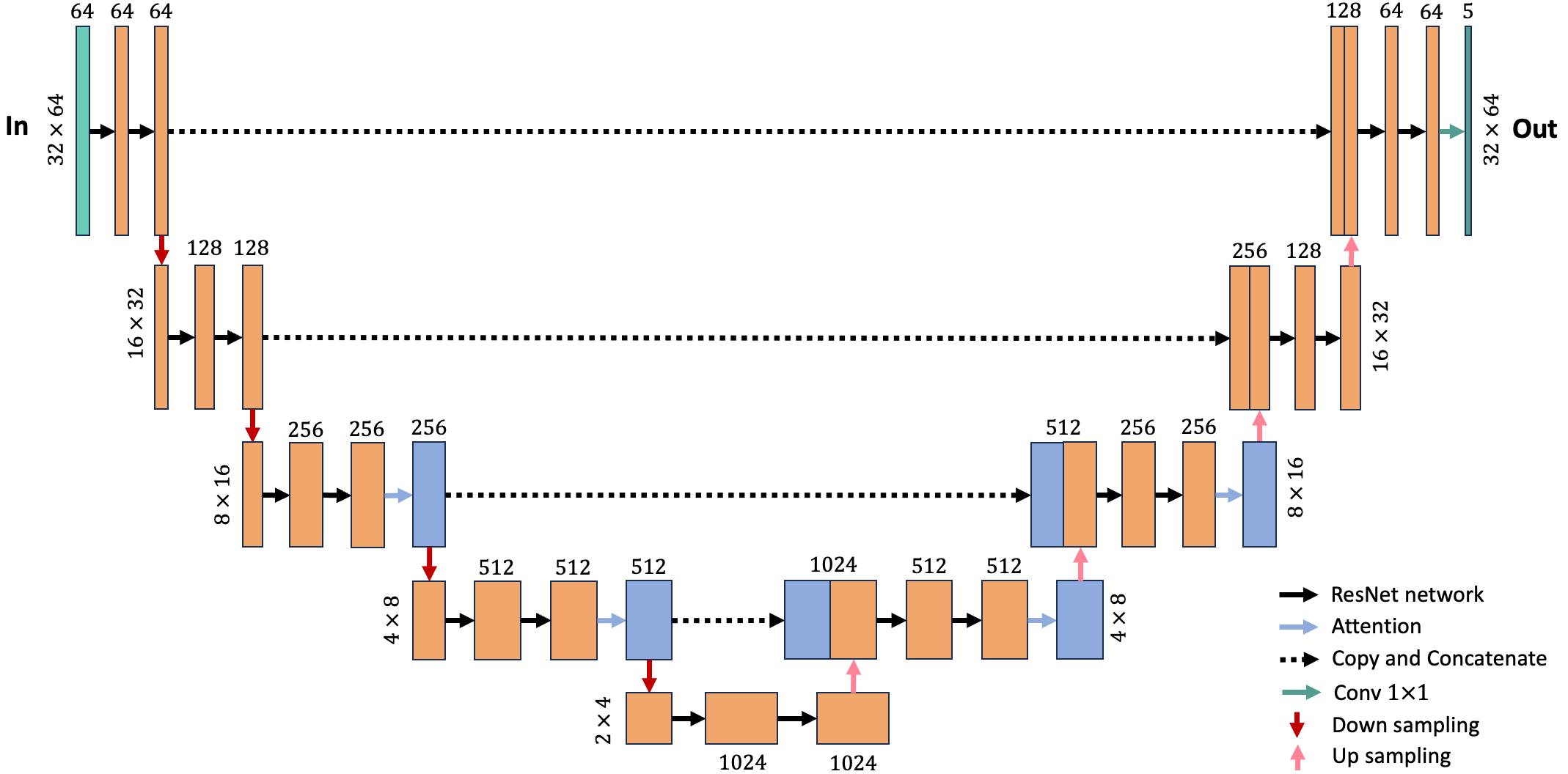}
\caption{Architecture of the U-Net model.}
\label{fig:unet_arch}
\end{figure*}

U-Net \cite{ronneberger2015u} is utilized to recover the data by removing the noise added at each diffusion step.
The \emph{skip connection} technique in U-Net concatenates feature maps from the encoder to the corresponding decoder layers, allowing the network to retain fine-grained information that might be lost during downsampling.
The detailed U-Net architecture is presented in Figure \ref{fig:unet_arch}.

\subsection{Training Process}
The training procedure is shown in the arXiv version in Algorithm \ref{alg:training}. 
Firstly, we pre-train an \texttt{encoder} to learn the condition embedding of the past observations. Subsequently, we inject it into our conditional diffusion model and train \codicast with the devised loss function:
\begin{equation}
    \mathcal{L}_{cond}(\theta) = \mathbb{E}_{X_0, \epsilon, n} \left\| \epsilon - \epsilon_\theta \left(X_{n}^{t+1}, n, \texttt{cond} \right) \right\|^2,
    \label{eq:condition_loss}
\end{equation}
where $X_{n}^{t+1} = \sqrt{\bar{\alpha}_n} X_{0}^{t+1} + \sqrt{1-\bar{\alpha}_n} \epsilon$, $\texttt{cond}=\mathcal{F}(X^{t-1:t})$, and $\epsilon_\theta$ is the denoiser in Figure \ref{fig:denoiser}.

\begin{algorithm}[H]
    \caption{Pseudocode for Training Process}
    \label{alg:training}
    \begin{small}
    \begin{algorithmic}[1]
    \STATE \textbf{Input}: Number of diffusion steps $N$, pre-trained encoder $\mathcal{F}$ \\
    \STATE \textbf{Output}: Trained denoising function $\epsilon(\cdot)$  \\
    \REPEAT
        \STATE $X^{t+1}_0 \sim q(X^{t+1}_0)$
        \STATE $n \sim \texttt{Uniform}({1, 2, \dots , N})$
        \STATE $\epsilon \sim \mathcal{N}(\mathbf{0}, \mathbf{I})$
        \STATE Get the past observations $X^{t-1}, X^t$
        \STATE Get embedding $\tilde{Z}^{t-1:t}=\mathcal{F}(X^{t-1}, X^t)$
        \STATE Take gradient descent step on:
        $$\nabla_\theta \left\| \epsilon - \epsilon_\theta \left( X_{n}^{t+1}, n, \tilde{Z}^{t-1:t} \right) \right\|^2$$
    \UNTIL{converged}
    \end{algorithmic}
    \end{small}
    \end{algorithm}

\subsection{Inference Process}
We first extract the conditional embedding representations, $\tilde{Z}^{t-1:t}$, by the pre-trained encoder, and then randomly generate a noise vector $X_N \sim \mathcal{N}(\mathbf{0}, \mathbf{I})$. 
The sampled noise vector, $X_N$, is autoregressively denoised along the reversed chain to predict the target until $n$ equals 1, we obtain the weather prediction $\hat{X}_0$ at the time $t+1$. 
Later, multi-step prediction can be implemented autoregressively - the output from the previous time step is the input while predicting the next step, as shown in Eq. (\ref{eq:multi_prediction}).

\begin{algorithm}[H]
\caption{Pseudocode for Inference Process}
\label{alg:inference}
\begin{small}
\begin{algorithmic}[1]
\STATE \textbf{Input}: Number of diffusion steps $N$, pre-trained encoder $\mathcal{F}$, trained denoising network $\epsilon(\cdot)$, past observations $X^{t-1}, X^t$ \\
\STATE \textbf{Output}: Inference target $X_0^{t+1}$  \\
\STATE Get embedding $\tilde{Z}^{t-1:t}=\mathcal{F}(X^{t-1}, X^t)$
\STATE $X_N \sim \mathcal{N}(\mathbf{0}, \mathbf{I})$
\FOR{$n = N, \dots, 1$}
    \STATE $\mathbf{\zeta} \sim \mathcal{N}(\mathbf{0}, \mathbf{I}) \text{ if } n \geq 1, \text{ else } \mathbf{\zeta} = 0$
    \STATE $X_{n-1}^{t+1} = \frac{1}{\sqrt{\alpha_n}} \left( X_n^{t+1} - \frac{1 - \alpha_n}{\sqrt{1 - \bar{\alpha}_n}} \epsilon_\theta (X_n^{t+1}, n, \tilde{Z}^{t-1:t} \right) + \sigma_n \mathbf{\zeta}$
\ENDFOR
\RETURN $X_0^{t+1}$
\end{algorithmic}
\end{small}
\end{algorithm}

\subsection{Ensemble Forecast}
\label{sec:ensemble_forecast}
To enhance the reliability of weather forecasts, \emph{ensemble forecast} strategy is often employed to capture the variability among forecasts by separately running multiple deterministic models \cite{buizza2008comparison}.
\codicast is a probabilistic model that can generate a distribution of future weather scenarios rather than a single prediction. Following \cite{price2023gencast}, we run the trained \codicast multiple times to get the ensembles. 
More specifically, by integrating initial conditions and noise sampled from a Gaussian distribution, we implement the ensemble forecast through multiple stochastic samplings during inference, capturing a range of possible forecasts.

\section{Experiments}
\subsection{Dataset}
\label{sec:data_intro}
ERA5 \cite{hersbach2020era5} is a publicly available atmospheric reanalysis dataset provided by the European Centre for Medium-Range Weather Forecasts (ECMWF). 
Following the existing work \cite{verma2024climode}, we use the preprocessed $5.625^\circ$ resolution ($32\times64$) and 6-hour increment ERA5 dataset from WeatherBench \cite{rasp2020weatherbench}. 
We downloaded 5 variables for the globe: \begin{enumerate}
    \item geopotential at 500 hPa pressure level (\texttt{Z500}), 
    \item atmospheric temperature at 850 hPa pressure level (\texttt{T850}), 
    \item ground temperature (\texttt{T2m}), 
    \item 10 meter U wind component (\texttt{U10}), and 
    \item 10 meter V wind component (\texttt{V10}).
\end{enumerate}
\begin{table}[ht]
\small
\centering
\resizebox{\textwidth}{!}{
\begin{tabular}{llcccccc}
    \toprule
    Type        & Variable              & Abbrev.   & ECMWF ID  & Levels    & Range                 & Unit  \\
    \midrule
    Single      & 2 metre temperature   & \texttt{T2m}       & 167       &           & $[193.1, 323.6]$      & $K$   \\
    Single      & 10 metre U wind       & \texttt{U10}       & 165       &           & $[-37.3, 30.2]$       & $m/s$ \\
    Single      & 10 metre V wind       & \texttt{V10}       & 166       &           & $[-31.5, 32.5]$       & $m/s$ \\
    Atmospheric & Geopotential          & \texttt{Z500}         & 129       & 500       & $[43403.6, 59196.9]$  & $m^2/s^2$ \\
    Atmospheric & Temperature           & \texttt{T850}         & 130       & 850       & $[217.9, 313.3]$      & $K$   \\
    \bottomrule
\end{tabular}
}
\caption{Information on variables used in our experiments.}
\label{tab:our_data}
\end{table}

\subsection{Baselines}
Our experiments use the following baseline tools: 
\begin{enumerate}
    \item ClimODE \cite{verma2024climode}: a spatiotemporal continuous-time model that incorporates the physical knowledge of atmospheric \emph{advection} over time; 
    \item ClimaX \cite{nguyen2023climax}: a state-of-the-art vision Transformer-based method trained on the same dataset (without pre-training that is used in the original paper); 
    \item FourCastNet \cite{pathak2022fourcastnet}: a global data-driven weather model using adaptive Fourier neural operators;
    \item Neural ODE \cite{chen2018neural}: an ODE network that learns the time derivatives as neural networks by solving an ordinary differential equation;
    \item Integrated Forecasting System IFS \cite{rasp2020weatherbench}: a global numerical weather prediction (NWP) system, integrating multiple advanced physics-based models to deal with more meteorological variables across multiple altitudes.
\end{enumerate}
Our study focuses solely only on a subset of all the available variables due to our limited computational resources, with IFS serving as the gold standard.
For a fair comparison, all ML models use the same data set described in Section \ref{sec:data_intro}.

\subsection{Experimental Design}
We use data from 2006 through 2015 as the training set, the data from 2016 as the validation set, and data from 2017 through 2018 as the test set. 
We assess the global weather forecasting capabilities of our method \codicast by predicting the weather at a future time $t+\Delta t$ ($\Delta t$ = 6 to 144 hours) based on the past two time units. 
To quantify the uncertainty in weather prediction, we generate an ``ensemble'' forecast by running \codicast five times during the inference phase. 
\paragraph{Training.} We first pretrain an \texttt{encoder} model with the \texttt{Autoencoder} architecture. For the diffusion model, we used U-Net as the denoiser network with 1000 diffusion/denoising steps. The architecture is similar to that of DDPM \cite{ho2020denoising} work.
We employ four U-Net units for both the downsampling and upsampling processes. 
Each U-Net unit comprises two ResNet blocks and a convolutional up/downsampling block. 
Before training, we apply Max-Min normalization \cite{ali2014data} to scale the input data within the range $[0, 1]$, mitigating potential biases stemming from varying scales \cite{shi2023explainable}. 
\texttt{Adam} was used as the optimizer, where the learning rate $=2e^{-4}$, decay steps $=10000$, decay rate $=0.95$. 
The batch size and number of epochs were set to 64 and 800, respectively.
More training details and model configurations are in the arXiv version.
We conduct all experiments on one NVIDIA A100 GPU with 80GB memory.

\paragraph{Evaluation Metrics.} 
Following \cite{verma2024climode}, we use latitude-weighted Root Mean Square Error (RMSE) and Anomaly Correlation Coefficient (ACC) as deterministic metrics.
RMSE measures the average difference between values predicted by a model and the actual values. 
ACC is the correlation between prediction anomalies relative to climatology and ground truth anomalies relative to climatology. It is a critical metric in climate science to evaluate the model's performance in capturing unusual weather or climate events.

Following \cite{verma2024climode}, we assess the model performance using latitude-weighted Root Mean Square Error (RMSE).
RMSE measures the average difference between values predicted by a model and the actual values. 
\begin{equation*}
\text{RMSE} = \frac{1}{M} \sum_{m=1}^{M} \sqrt{\frac{1}{H \times W} \sum_{h=1}^{H} \sum_{w=1}^{W} L(h) (\tilde{X}_{m,h,w} - X_{m,h,w})^2},
\end{equation*}
where $L(h)=\frac{1}{H}cos(h)\sum_{h'}^{H} cos(h')$ is the latitude weight and $M$ represents the number of test samples.

Anomaly Correlation Coefficient (ACC) is the correlation between prediction anomalies $\tilde{X}'$ relative to climatology and ground truth anomalies $\hat{X}$ relative to climatology.
ACC is a critical metric in climate science to evaluate the model's performance in capturing unusual weather or climate events.
\begin{equation*}
\text{ACC} = \frac{\sum_{m,h,w} L(h) \tilde{X}'_{m,h,w} {X'}_{m,h,w}}{\sqrt{\sum_{m,h,w} L(h) \tilde{X}_{m,h,w}^{'2} \cdot \sum_{m,h,w} L(h) {X}_{m,h,w}^{'2}}},
\end{equation*}
where observed and forecasted anomalies $X' = X - C, \tilde{X}' = \tilde{X} - C$, and climatology $C=\frac{1}{M} \sum_m X$ is the temporal mean of the ground truth over the entire test set.

\subsection{Reproducibility}
We include all data and code in a GitHub repository\footnote{Link: \url{https://github.com/JimengShi/CoDiCast}}.

\section{Results}

\subsection{Quantitative Evaluation}
\label{sec:performance}
\paragraph{Accuracy.} 
We compare different models in forecasting five primary meteorological variables as described in Section \ref{sec:data_intro}. 
Tables \ref{tab:global_score_rmse} and \ref{tab:global_score_acc} show \codicast presents superior performance over other MLWP baselines across Latitude-weighted RMSE and ACC results, demonstrating diffusion models can capture the weather dynamics and make predictions accurately. 
However, the performance becomes poor as the lead time increases, even though \codicast still outperforms other benchmarks. \emph{Error accumulation} is a hurdle to autoregressive forecasting methods.
Furthermore, while we consider the gold-standard \texttt{IFS} model as a reference model due to the different experimental settings, we observed that there is still room to be improved for \codicast in terms of accuracy. Integrating more meteorological variables is a possibility to enhance its performance further. 
\begin{table}[ht!]
\centering
\small
\resizebox{0.85\textwidth}{!}{
    \begin{tabular}{lc|lllll|l}
    \toprule
        \multirow{2}{*}{Variable} & \multirow{2}{*}{\makecell[c]{Lead time\\(Hours)}}  & \multicolumn{6}{c}{RMSE ($\downarrow$)} \\
        \cmidrule(lr){3-8} 
        & & NODE & ClimaX & ForeCastNet & ClimODE & CoDiCast & IFS               \\
    \midrule
        \multirow{5}{*}{Z500} 
         & 6   & 300.6 & 247.5 & 222.7\textcolor{gray}{$\pm18.1$}    &102.9\textcolor{gray}{$\pm9.3$}   &\textbf{73.1}\textcolor{gray}{$\pm6.7$}   & 26.9   \\
         & 12  & 460.2 & 265.3 & 310.9\textcolor{gray}{$\pm22.7$}    &134.8\textcolor{gray}{$\pm12.3$}  &\textbf{114.2}\textcolor{gray}{$\pm8.9$}  & 33.8    \\
         & 24  & 877.8 & 364.9 & 402.6\textcolor{gray}{$\pm27.3$}    &193.4\textcolor{gray}{$\pm16.3$}  &\textbf{186.5}\textcolor{gray}{$\pm11.8$} & 51.0    \\ 
         & 72  & N/A   & 687.0 & 755.3\textcolor{gray}{$\pm45.8$}    &478.7\textcolor{gray}{$\pm48.5$}  &\textbf{451.6}\textcolor{gray}{$\pm39.5$} & 123.2   \\
         & 144 & N/A   & 801.9 & 956.1\textcolor{gray}{$\pm59.1$}    &783.6\textcolor{gray}{$\pm37.3$}  &\textbf{757.5}\textcolor{gray}{$\pm42.8$} & 398.7   \\
    \midrule
        \multirow{5}{*}{T850} 
         & 6   & 1.82 & 1.64 & 1.75\textcolor{gray}{$\pm0.16$}    &1.16\textcolor{gray}{$\pm0.06$}  &\textbf{1.02}\textcolor{gray}{$\pm0.05$}      & 0.69  \\
         & 12  & 2.32 & 1.77 & 2.15\textcolor{gray}{$\pm0.20$}    &1.32\textcolor{gray}{$\pm0.13$}  &\textbf{1.26}\textcolor{gray}{$\pm0.10$}      & 0.75 \\
         & 24  & 3.35 & 2.17 & 2.51\textcolor{gray}{$\pm0.27$}    &1.55\textcolor{gray}{$\pm0.18$}  &\textbf{1.52}\textcolor{gray}{$\pm0.16$}      & 0.87 \\
         & 72  & N/A  & 3.17 & 3.69\textcolor{gray}{$\pm0.34$}    &2.58\textcolor{gray}{$\pm0.16$}  &\textbf{2.54}\textcolor{gray}{$\pm0.14$}      & 1.15 \\
         & 144 & N/A  & 3.97 & 4.29\textcolor{gray}{$\pm0.42$}    &3.62\textcolor{gray}{$\pm0.21$}  &\textbf{3.61}\textcolor{gray}{$\pm0.19$}      & 2.23 \\
    \midrule
        \multirow{5}{*}{T2m} 
         & 6   & 2.72 & 2.02 & 2.05\textcolor{gray}{$\pm0.18$}    &1.21\textcolor{gray}{$\pm0.09$}  &\textbf{0.95}\textcolor{gray}{$\pm0.07$}      & 0.69 \\
         & 12  & 3.16 & 2.26 & 2.49\textcolor{gray}{$\pm0.21$}    &1.45\textcolor{gray}{$\pm0.10$}  &\textbf{1.21}\textcolor{gray}{$\pm0.07$}      & 0.77 \\
         & 24  & 3.86 & 2.37 & 2.78\textcolor{gray}{$\pm0.26$}    &1.40\textcolor{gray}{$\pm0.09$}  &1.45\textcolor{gray}{$\pm0.07$}               & 1.02 \\
         & 72  & N/A  & 2.87 & 3.77\textcolor{gray}{$\pm0.32$}    &2.75\textcolor{gray}{$\pm0.49$}  &\textbf{2.39}\textcolor{gray}{$\pm0.37$}      & 1.26 \\
         & 144 & N/A  & 3.38 & 4.39\textcolor{gray}{$\pm0.41$}    &3.30\textcolor{gray}{$\pm0.23$}  &3.45\textcolor{gray}{$\pm0.22$}               & 1.78 \\
    \midrule
        \multirow{5}{*}{U10} 
         & 6   & 2.30 & 1.58 & 1.98\textcolor{gray}{$\pm0.17$}    &1.41\textcolor{gray}{$\pm0.07$}  &\textbf{1.24}\textcolor{gray}{$\pm0.06$}      & 0.61 \\
         & 12  & 3.13 & 1.96 & 2.58\textcolor{gray}{$\pm0.21$}    &1.81\textcolor{gray}{$\pm0.09$}  &\textbf{1.50}\textcolor{gray}{$\pm0.08$}      & 0.76 \\
         & 24  & 4.10 & 2.49 & 3.02\textcolor{gray}{$\pm0.27$}    &2.01\textcolor{gray}{$\pm0.10$}  &\textbf{1.87}\textcolor{gray}{$\pm0.09$}      & 1.11 \\
         & 72  & N/A  & 3.70 & 4.17\textcolor{gray}{$\pm0.36$}    &3.19\textcolor{gray}{$\pm0.18$}  &\textbf{3.15}\textcolor{gray}{$\pm0.19$}      & 1.57 \\
         & 144 & N/A  & 4.24 & 4.63\textcolor{gray}{$\pm0.45$}    &4.02\textcolor{gray}{$\pm0.12$}  &4.25\textcolor{gray}{$\pm0.15$}               & 3.04 \\
    \midrule
        \multirow{5}{*}{V10} 
         & 6   & 2.58 & 1.60 & 2.16\textcolor{gray}{$\pm0.19$}    &1.53\textcolor{gray}{$\pm0.08$}  &\textbf{1.30}\textcolor{gray}{$\pm0.06$}      & 0.61 \\
         & 12  & 3.19 & 1.97 & 2.73\textcolor{gray}{$\pm0.23$}    &1.81\textcolor{gray}{$\pm0.12$}  &\textbf{1.56}\textcolor{gray}{$\pm0.09$}      & 0.79 \\
         & 24  & 4.07 & 2.48 & 3.15\textcolor{gray}{$\pm0.28$}    &2.04\textcolor{gray}{$\pm0.10$}  &\textbf{1.94}\textcolor{gray}{$\pm0.14$}      & 1.33 \\
         & 72  & N/A  & 3.80 & 4.26\textcolor{gray}{$\pm0.34$}    &3.30\textcolor{gray}{$\pm0.22$}  &\textbf{3.18}\textcolor{gray}{$\pm0.19$}      & 1.67 \\
         & 144 & N/A  & 4.42 & 4.64\textcolor{gray}{$\pm0.45$}    &4.24\textcolor{gray}{$\pm0.10$}  &\textbf{4.21}\textcolor{gray}{$\pm0.18$}      & 3.26 \\
    \bottomrule
    \end{tabular}
}
\caption{Latitude-weighted RMSE ($\downarrow$) comparison. The results of NODE, ClimaX, and ClimODE models are from the ClimODE paper \cite{verma2024climode}. N/A represents the values that are not available.
ForeCastNet was re-trained with their code \cite{pathak2022fourcastnet}. We employ Monte Carlo Dropout \cite{gal2016dropout} during the inference to compute the uncertainty. We mark the scores in \textbf{bold} if \codicast performs the best among MLWP methods. }
\label{tab:global_score_rmse}
\end{table}

\begin{table}[ht!]
\centering
\small
\resizebox{0.85\textwidth}{!}{
    \begin{tabular}{lc|ccccc|c}
    \toprule
        \multirow{2}{*}{Variable} & \multirow{2}{*}{\makecell[c]{Lead time\\(Hours)}}  & \multicolumn{6}{c}{ACC ($\uparrow$)} \\
        \cmidrule(lr){3-8}
        &     & NODE & ClimaX & ForeCastNet & ClimODE & CoDiCast & IFS \\
    \midrule
        \multirow{5}{*}{Z500} 
         & 6      & 0.96 & 0.97 & 0.97   & 0.99 &\textbf{0.99}   & 1.00\\
         & 12     & 0.88 & 0.96 & 0.95   & 0.99 &\textbf{0.99}   & 0.99\\
         & 24     & 0.70 & 0.93 & 0.92   & 0.98 &\textbf{0.98}   & 0.99\\ 
         & 72     & N/A  & 0.73 & 0.75   & 0.88 &\textbf{0.92}   & 0.98\\
         & 144    & N/A  & 0.58 & 0.64   & 0.61 &\textbf{0.78}   & 0.86\\
    \midrule
        \multirow{5}{*}{T850} 
         & 6     & 0.94 & 0.94 & 0.94   & 0.97 &\textbf{0.99}    & 0.99\\
         & 12    & 0.85 & 0.93 & 0.92   & 0.96 &\textbf{0.99}    & 0.99\\
         & 24    & 0.72 & 0.90 & 0.89   & 0.95 &\textbf{0.97}    & 0.99\\
         & 72    & N/A  & 0.76 & 0.77   & 0.85 &\textbf{0.93}    & 0.96\\
         & 144   & N/A  & 0.69 & 0.71   & 0.77 &\textbf{0.85}    & 0.81\\
    \midrule
        \multirow{5}{*}{T2m} 
         & 6     & 0.82 & 0.92 & 0.94   & 0.97 &\textbf{0.99}    & 0.99\\
         & 12    & 0.68 & 0.90 & 0.92   & 0.96 &\textbf{0.99}    & 0.99\\
         & 24    & 0.79 & 0.89 & 0.91   & 0.96 &\textbf{0.99}    & 0.99\\
         & 72    & N/A  & 0.83 & 0.85   & 0.85 &\textbf{0.96}    & 0.96\\
         & 144   & N/A  & 0.83 & 0.81   & 0.79 &\textbf{0.91}    & 0.82\\
    \midrule
        \multirow{5}{*}{U10} 
         & 6     & 0.85 & 0.92 & 0.87   & 0.91 &\textbf{0.95}    & 0.98\\
         & 12    & 0.70 & 0.88 & 0.78   & 0.89 &\textbf{0.93}    & 0.98\\
         & 24    & 0.50 & 0.80 & 0.71   & 0.87 &\textbf{0.89}    & 0.97\\
         & 72    & N/A  & 0.45 & 0.41   & 0.66 &\textbf{0.71}    & 0.94\\
         & 144   & N/A  & 0.30 & 0.28   & 0.35 &\textbf{0.42}    & 0.72\\
    \midrule
        \multirow{5}{*}{V10} 
         & 6     & 0.81 & 0.92 & 0.86   & 0.92 &\textbf{0.95}    & 1.00\\
         & 12    & 0.61 & 0.88 & 0.76   & 0.89 &\textbf{0.93}    & 0.99\\
         & 24    & 0.35 & 0.80 & 0.68   & 0.86 &\textbf{0.89}    & 1.00\\
         & 72    & N/A  & 0.39 & 0.38   & 0.63 &\textbf{0.68}    & 0.93\\
         & 144   & N/A  & 0.25 & 0.27   & 0.32 &\textbf{0.37}    & 0.71\\
    \bottomrule
    \end{tabular}
}
\caption{ACC ($\uparrow$) comparison. Other settings are the same as the Table \ref{tab:global_score_rmse}. }
\label{tab:global_score_acc}
\end{table}

\begin{figure}[H]
\centering
    \includegraphics[width=0.7\textwidth]{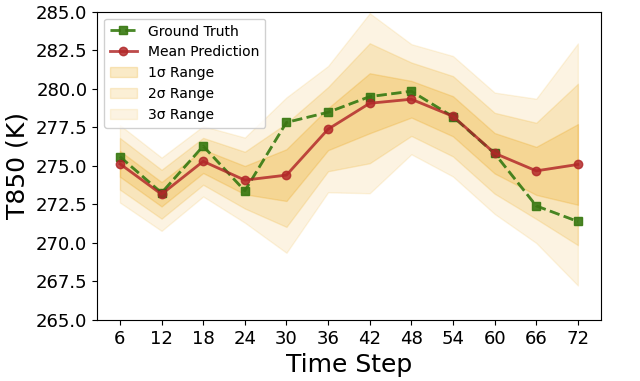}
\caption{Model Forecasts with confidence intervals.}
\label{fig:forecast_uncertainty}
\end{figure}

\paragraph{Uncertainty.} 
The gray font was used to show the magnitude of the error in Table \ref{tab:global_score_rmse},representing the model uncertainty, with fluctuations remaining within 10\% of the ground truth scale. This is a measure of the robustness of the ML models. However, \texttt{ForeCastNet} predictions have relatively larger fluctuations due to the sensitive selection of the dropout rate.
We present a case study of a 72-hour weather forecast generated by \codicast with uncertainty quantification, as shown in Figure \ref{fig:forecast_uncertainty}. The predicted mean closely follows the overall trend of the ground truth, while the uncertainty increases with lead time, consistent with the intuition that forecasts become less certain over longer horizons. 
Notably, most observed values lie within the 1$\sigma$ or 2$\sigma$ confidence intervals, demonstrating that the model provides reliable and well-calibrated predictions.

\begin{table}[ht!]
\small
\centering
\resizebox{0.8\columnwidth}{!}{
    \begin{tabular}{lc|ccccccc}
    \toprule
        \multirow{2}{*}{Variable} & \multirow{2}{*}{\makecell[c]{Lead Time\\(Hours)}} & \multicolumn{6}{c}{Diffusion Step} \\
        \cmidrule(lr){3-8}
        & & 250 & 500 & 750 & 1000 & 1500 & 2000\\
    \midrule
        \multirow{1}{*}{Z500} 
         & 24 & 696.1 & 324.8 & 190.6   &\textbf{186.5} &193.5  &191.9 \\
        \multirow{1}{*}{T850} 
         & 24 & 3.88 & 2.38 & 1.53   &\textbf{1.52}     &1.56   &1.58 \\
        \multirow{1}{*}{T2m} 
         & 24 & 5.26 & 2.79 & 1.63   &\textbf{1.44}     &1.50   &1.53 \\
        \multirow{1}{*}{U10} 
         & 24 & 2.74 & 2.05 & \textbf{1.81}  &1.87      &1.99   &2.01 \\
        \multirow{1}{*}{V10} 
         & 24 & 2.43 & 2.11 & \textbf{1.89}  &1.94      &2.04   &2.06 \\
    \midrule
        \multicolumn{2}{l|}{Inference time (min)} 
              &$\sim1.1$ &$\sim1.9$  &$\sim2.8$  &$\sim3.6$  &$\sim6.5$   &$\sim8.3$  \\
    \bottomrule
    \end{tabular}
}
\caption{RMSE with various diffusion steps at 24 hours lead time. We mark the lowest scores in \textbf{bold} font.
}
\label{tab:rmse_vs_diffusion_step}
\end{table}

\paragraph{Inference efficiency.}
Generally, numerical weather prediction models (e.g., \texttt{IFS}) require around 50 minutes for the medium-range global forecast, while deterministic ML weather prediction models take less than 1 minute \cite{rasp2020weatherbench} but cannot model the weather uncertainty. 
\codicast needs about 3.6 minutes (see the last row in Table \ref{tab:rmse_vs_diffusion_step}) for the global weather forecast, potentially balancing the efficiency and accuracy with essential uncertainty quantification. 
The efficiency also depends on the model complexity. Table \ref{tab:learning_models_comparison} presents a comparison of different methods. More details can be found in our survey paper \cite{shi2025deep}.

\begin{table}[ht]
\centering
\resizebox{.999\columnwidth}{!}{
    \begin{tabular}{l|c|c|c|c|c}
    \toprule
    \textbf{Methods}                    & \bf $\Delta x$    & \bf Train data        & \bf Train resources    & \bf Test data    & \bf Inference time \\
    \midrule
    \rowcolor{gray!10}
    \multicolumn{6}{l}{\textbf{\textit{Physics-based Models}}} \\
    IFS HRES \cite{ecmwf}              & $0.1^\circ$       &                       &                        & ERA5 2020           & $\sim$52 mins \\
    IFS ENS  \cite{ecmwf}              & $0.2^\circ$       &                       &                        & ERA5 2020           & -- \\ 
    \midrule

    \rowcolor{gray!10}
    \multicolumn{6}{l}{\textbf{\textit{Deterministic Predictive Models}}} \\
    Pangu-Weather \cite{bi2023accurate}& $0.25^\circ$      & ERA5 1979-2017        & 16 days; 192 V100 GPUs & ERA5 2020            & $\sim$secs; a GPU \\ 
    GraphCast \cite{lam2022graphcast}  & $0.25^\circ$      & ERA5 1979-2019        & 4 weeks; 32 TPU v4     & ERA5 2020            & $\sim$min; a TPU \\ 
    FuXi \cite{chen2023fuxi}           & $0.25^\circ$      & ERA5 1979-2015        & ~8 days; 8 A100 GPUs   & ERA5 2020            & $\sim$secs; a GPU \\ 
    Fengwu \cite{chen2023fengwu}       & $0.25^\circ$      & ERA5 1979-2017        & 17 days; 32 A100 GPUs  & ERA5 2020            & $\sim$secs; a GPU \\ 
    Stormer \cite{nguyen2023scaling}   & $0.25^\circ$      & ERA5 1979-2017        & ~8 days; 8 A100 GPUs   & ERA5 2020            & $\sim$secs; a GPU \\ 
    HEAL-ViT \cite{ramavajjala2024heal}    & $0.25^\circ$  & ERA5 1979-2017        & ~8 days; 8 A100 GPUs   & ERA5 2020            & $\sim$secs; a GPU \\
    GnnWeather \cite{keisler2022forecasting} & $1^\circ$   & ERA5 35 years         & 5.5 days; 1 A100 GPU   & ERA5 2020            & $\sim$secs; a GPU \\ 
    ArchesWeather\cite{couairon2024archesweather}&$1.5^\circ$&ERA5 1979-2018       & ~9 days; 1 V100 GPU    & ERA5 2020            & $\sim$secs; a GPU \\
    NeuralGCM 0.7 \cite{kochkov2024neural} & $0.7^\circ$   & ERA5 1979-2017        & 3 weeks; 256 TPUs v5   & ERA5 2020            & $\sim$min; a TPU \\ 
    NeuralGCM ENS \cite{kochkov2024neural} & $1.4^\circ$   & ERA5 1979-2017        & 10 days; 128 TPUs v5   & ERA5 2020            & $\sim$min; a TPU \\ 
    \midrule

    \rowcolor{gray!10}
    \multicolumn{6}{l}{\textbf{\textit{Probabilistic Generative Models}}} \\
    GenCast \cite{price2023gencast}    & $0.25^\circ$      & ERA5 1979-2018         & 5 days; 32 TPUs v5    & 13                   & 8 mins; a TPU \\ 
    ArchesWeatherGen\cite{couairon2024archesweather}&$1.5^\circ$ & ERA5 1979-2018   & ~45 days; 1 V100 GPU  & 13                   & -- \\
    \midrule

    \rowcolor{gray!10}
    \multicolumn{6}{l}{\textbf{\textit{Foundation Models with Pre-training and Fine-tuning}}} \\
    Aurora \cite{bodnar2024aurora} & $0.1^\circ$ & ERA5, CMIP6 &2.5 weeks; 32 A100 GPUs& HRES-T0 2022      & -- \\
    Prithvi WxC\cite{schmude2024prithvi}&$0.5^\circ$        & MERRA 1980-2019    & --; 64 A100 GPUs             & MERRA 2020-2023  & -- \\
    
    \bottomrule
    \end{tabular}
}
\caption{Comparison of Predictive, Generative, and Foundation Models for \textbf{global} weather prediction. The performance scores below are at the lead time of 6 days (except Prithvi WxC at the lead time of 5 days). These scores are either from the WeatherBench scoreboard or the original paper. ``$\Delta x$'' represents the horizontal resolution.}
\label{tab:learning_models_comparison}
\end{table}

\subsection{Qualitative Evaluations}
\begin{figure}[ht!]
    \centering
    \includegraphics[width=0.99\textwidth]{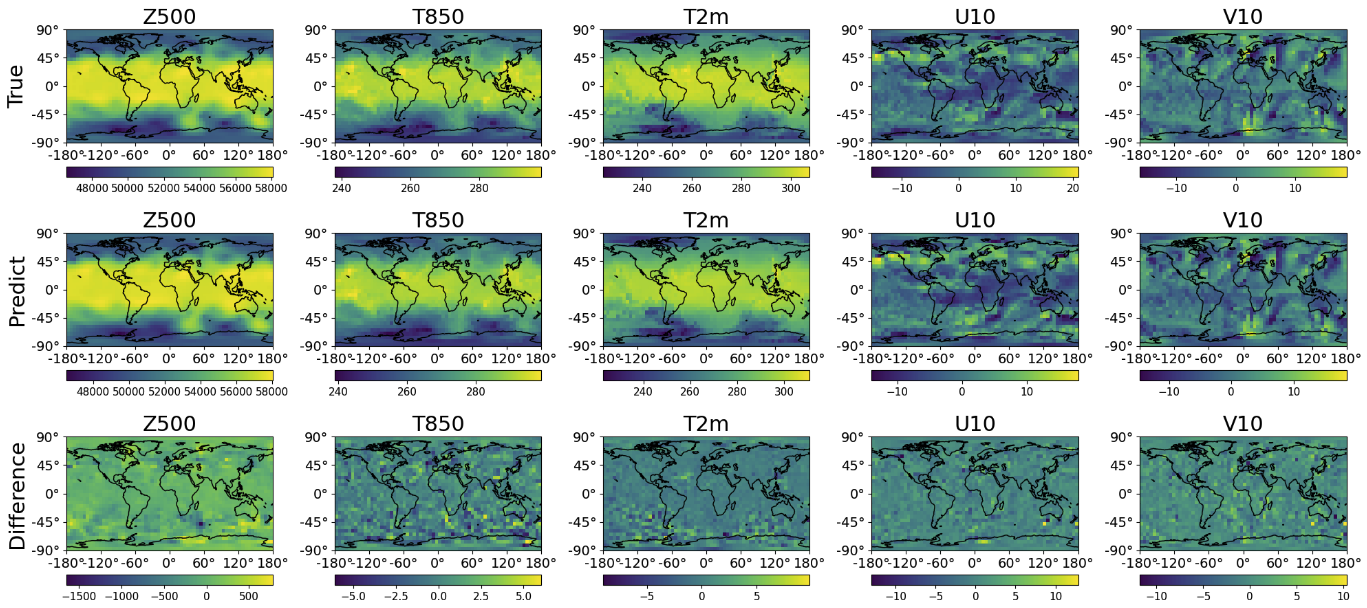}
    \caption{Visualizations of true and predicted values of all five variables at 24 hours lead time.}
    \label{fig:true_vs_pred_all_t24}
\end{figure}

In Figure \ref{fig:true_vs_pred_all_t24}, we qualitatively evaluate the performance of \codicast on global forecasting tasks for all target variables, \texttt{Z500}, \texttt{T850}, \texttt{T2m}, \texttt{U10} and \texttt{V10} at the lead time of $6$ hours.
The first row is the ground truth of the target variable, the second row is the prediction and the last row is the difference between the model prediction and the ground truth.
From the scale of color bars, we can tell that the error percentage is less than 3\% for variables \texttt{Z500}, \texttt{T850}, and \texttt{T2m}.
However, error percentages over $50\%$ exist for \texttt{U10} and \texttt{V10} even though only a few of them exist.
We also observe that most higher errors appear in the high-latitude ocean areas, probably due to the sparse data nearby.

\begin{figure}[ht!]
\centering
    \includegraphics[width=0.99\textwidth]{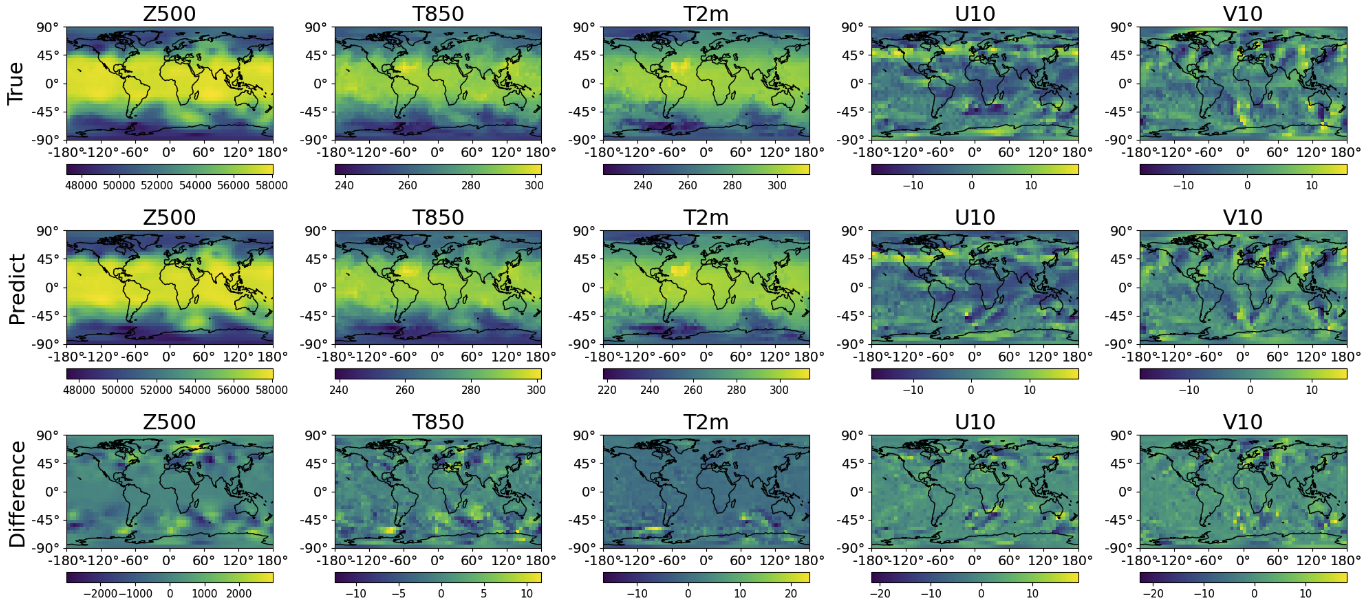}
\caption{Visualization of true and predicted values across five meteorological variables at 72 hours lead time.}
\label{fig:true_vs_pred_all_t72}
\end{figure}

\begin{figure}[ht!]
    \centering
    \includegraphics[width=0.99\textwidth]{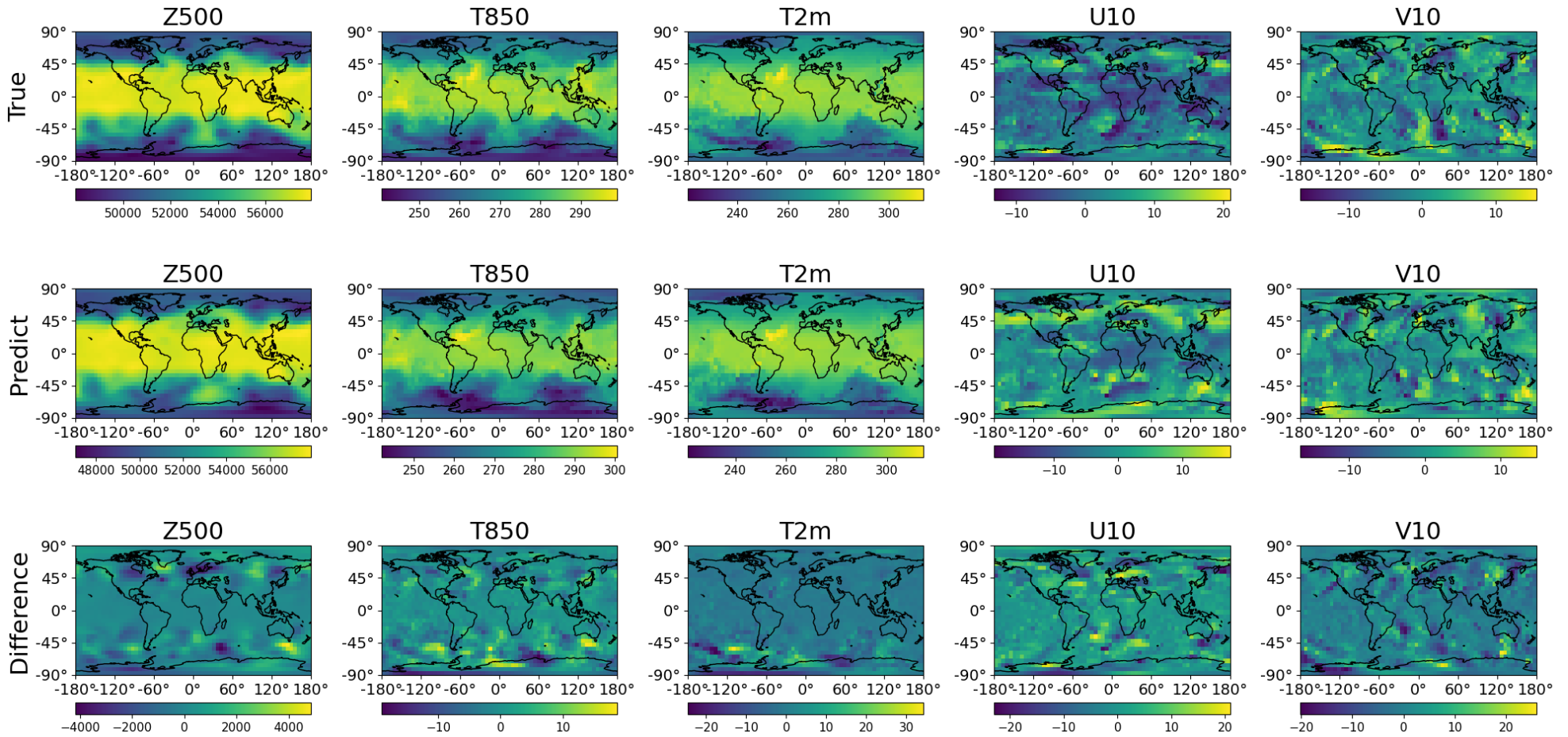}
    \caption{Visualizations of true and predicted values of all five variables at 144 hours lead time.}
    \label{fig:true_vs_pred_all_t144}
\end{figure}

\subsection{Ablation Study}
\begin{figure}[ht]
\centering
    \includegraphics[width=0.99\textwidth]{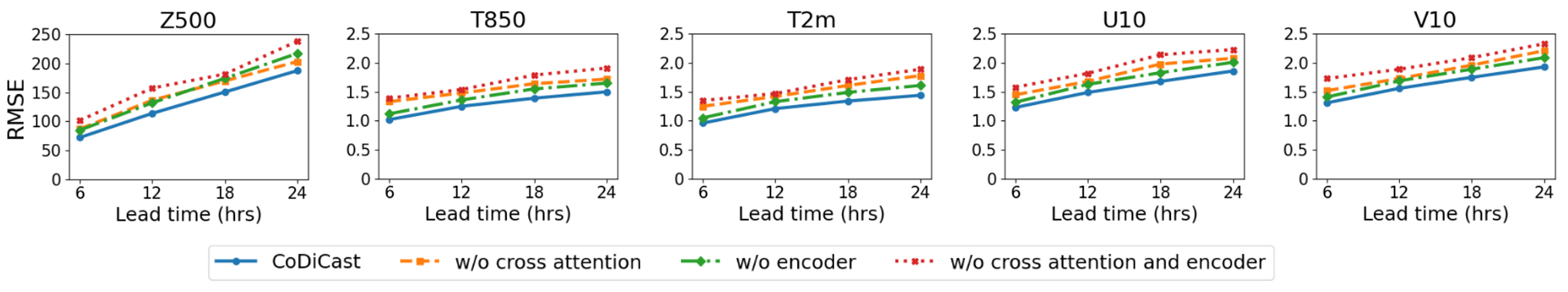}

\caption{Ablation study.}
\label{fig:ablation}
\end{figure}

\codicast includes two important components: \emph{pre-trained encoder} and \emph{cross attention}. 
To study their effectiveness, we conduct an ablation study: (a) \textbf{No-encoder} directly considers past observations as conditions to the diffusion model; (b) \textbf{No-cross-attention} simply concatenates the embedding and the noisy sample at each denoising step; (c) \textbf{No-encoder-cross-attention} concatenates the past observations and the noisy sample at each denoising step. Figure \ref{fig:ablation} shows that the full version of \codicast consistently outperforms all other variants, verifying positive contributions of both components.

\subsection{Parameter Study}
\begin{figure}[ht!]
    \centering
    \includegraphics[width=0.99\textwidth]{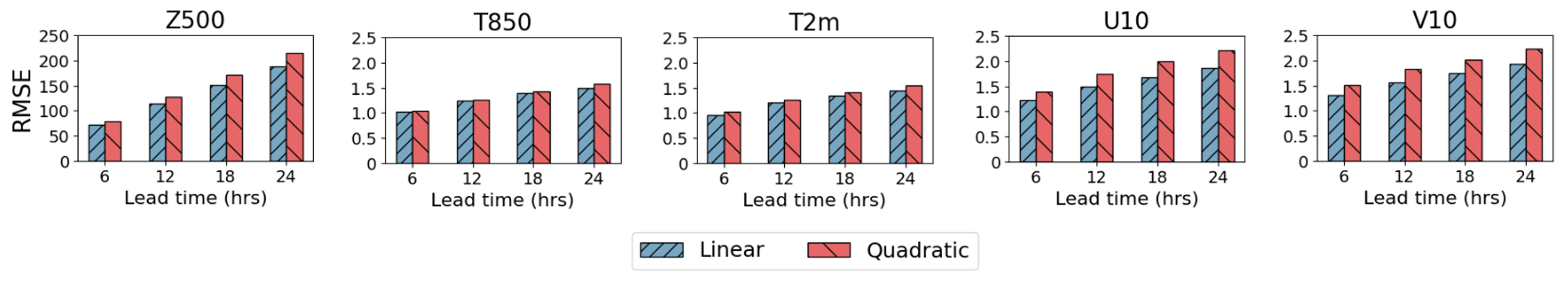}
\caption{Effect of linear and quadratic variance scheduling methods.}
\label{fig:variance_schedule}
\end{figure}

Table \ref{tab:rmse_vs_diffusion_step} shows that the accuracy improves as the number of diffusion steps, $N$, increases when $N < 1000$, indicating that more intermediate steps are more effective in learning the imperceptible attributes during the denoising process.
However, when $1000<N<2000$, the accuracy remains approximately flat but the inference time keeps increasing linearly.
Considering the trade-off between accuracy and efficiency, we finally set $N=1000$. 
Additionally, we use the same start and end variance value, $\beta$, as DDPM \cite{ho2020denoising} where $\beta \in [0.0001, 0.02]$. 
we study the effect of ``linear'' and ``quadratic'' variance scheduling $\beta$, where $\beta \in [0.0001, 0.02]$. 
Figure \ref{fig:variance_schedule} shows that the ``linear'' variance scheduling performs better than the ``quadratic'' one.

\section{Discussion}
Diffusion models, originally designed for generative tasks, have recently shown promise in predictive applications due to their transformative capabilities in conditional learning. In the context of weather forecasting, diffusion models offer several unique advantages. 
First, they inherently support probabilistic prediction by modeling a distribution over possible future states, making them particularly suitable for capturing the uncertainty of forecasts.
Second, their iterative denoising process allows for fine-grained control over the generation steps, which can be conditioned on recent observations to guide forecasts.
In this study, we adapted the diffusion framework to conditionally forecast global weather with high fidelity and quantified uncertainty. 
This positions diffusion models as a powerful tool for weather forecasting.

\paragraph{Limitations.}
We used low-resolution global weather data in our work due to limited computational resources and training time constraints. However, accurate weather forecasting—particularly for extreme events—typically requires high-resolution data, which significantly increases training time. As a result, the time complexity of such models warrants further investigation.
Moreover, our \codicast focuses on global weather prediction, without specifying non-extreme or extreme events. Therefore, the performance on extreme events has not been studied yet, which motivates us to continue exploration with a focus on extreme events.

\section{Conclusions}
In this chapter, we start with analyzing the limitations of current deterministic numerical weather prediction (NWP) and machine-learning weather prediction (MLWP) approaches. They either require substantial computational costs or lack uncertainty quantification in their forecasts. 
We propose a conditional diffusion model, \codicast, to address these limitations.
which contains a conditional \emph{pre-trained encoder} and a \emph{cross-attention} component.
Experimental results demonstrate it can \textbf{simultaneously} complete more accurate predictions than existing MLWP-based models and a faster inference than physics-based NWP models, while being capable of providing uncertainty quantification compared to deterministic methods. In conclusion, our model \codicast simultaneously achieves global weather prediction with high accuracy and efficiency while enabling crucial uncertainty quantification.
\chapter{\hyperrag: HYPERCUBE RETRIEVAL-AUGMENTED GENERATION FOR IN-DOMAIN SCIENTIFIC QUESTION-ANSWERING}	
\label{sec:hyperrag}
The work presented in this chapter is highlighted in the following publication \cite{shi2025hypercube,shi2026multicube}.

\section{Background}
Large language models (LLMs) often suffer from hallucinations and factual inaccuracies, especially for \emph{in-domain} scientific question-answering.
To address this phenomenon, retrieval-augmented generation (RAG) has emerged as the de facto approach \cite{huang2025survey,sun2026tasr}, incorporating external domain knowledge to generate contextually relevant responses \cite{huang2025survey,sun2026retrieval}.
Despite recent advancements, they exhibit key limitations in theme-specific applications on which the fine-grained topics are highly focused.
For example, retrieved documents are often semantically similar to the query, but overlook specialized themes (e.g., terminology or nuanced contextual cues) prevalent in specific literature or reports \cite{ding2024automated}.
In addition, the retrieval process often suffers from inefficiency and limited transparency, which are critical in high-stakes domains, such as environment \cite{wang2025remflow}, traffic \cite{hussien2025rag}, and healthcare \cite{pujari2023explainable}.

\begin{figure}[t!]
\centering
  \input{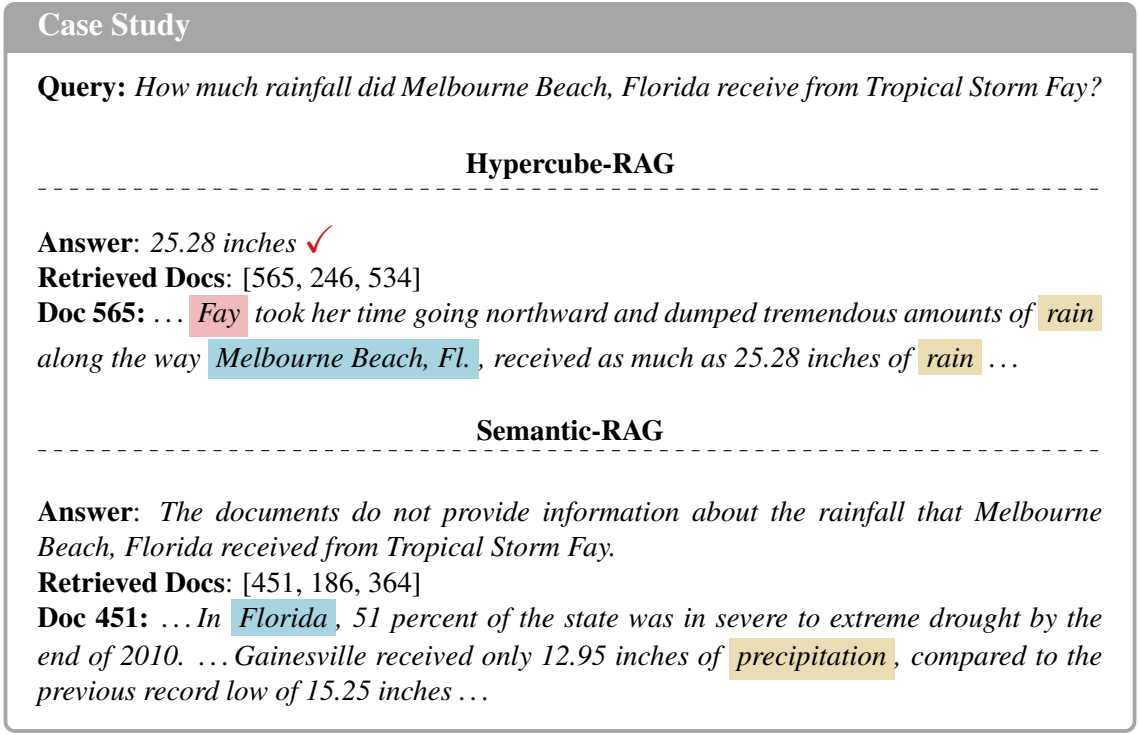}
  \vspace{-2mm}
  \caption{Hypecube- vs semantic RAG: A case study} 
  \label{fig:case_study1}
\end{figure}

The conventional \textit{sparse lexical retrieval} ranks a set of documents based on the appearance of query terms \cite{kadhim2019term}. 
Due to its reliance on the exact token overlap, it precisely captures specific themes or topics, and the retrieval process is efficient and interpretable. However, it struggles to retrieve contextually related or paraphrased documents that require semantic understanding.
%
%
To mitigate the above limitation, \textit{dense embedding retrievers} were proposed by computing the similarity between vector embeddings of query-document pairs \cite{karpukhin2020dense}. 
While contextual understanding is relatively improved, the hurdle of missing specific themes remains, causing semantically similar but off-topic retrievals \cite{krishna2021hurdles, kang2024improving}.
Figure \ref{fig:case_study1} shows an example where the Semantic-RAG returns documents related to precipitation in Florida, but overlooks the critical location-specific information for \texttt{``Melbourne Beach''}.
Moreover, retrieval in an embedding space makes it challenging to interpret why certain documents were selected \cite{ji2019visual}.
%
Additionally, a \textit{graph-based} RAG represents documents as graph structures, where nodes correspond to entities or concepts and edges capture their relationships \cite{peng2024graph}. 
They offer explanations by outputting the traversed subgraph.
However, traversing such subgraphs introduces a significant risk of information overload, as the expanding neighborhoods may include substantial amounts of irrelevant themes \cite{wang2024graph}.
Graph traversal also poses significant scalability bottlenecks due to its inefficient computational complexity of $\mathcal{O}(V + E)$, where $V$ and $E$ are the number of nodes and edges in the traversed subgraph \cite{devezas2021review}.

\begin{figure}[ht!]
\centering
  \includegraphics[width=0.95\columnwidth]{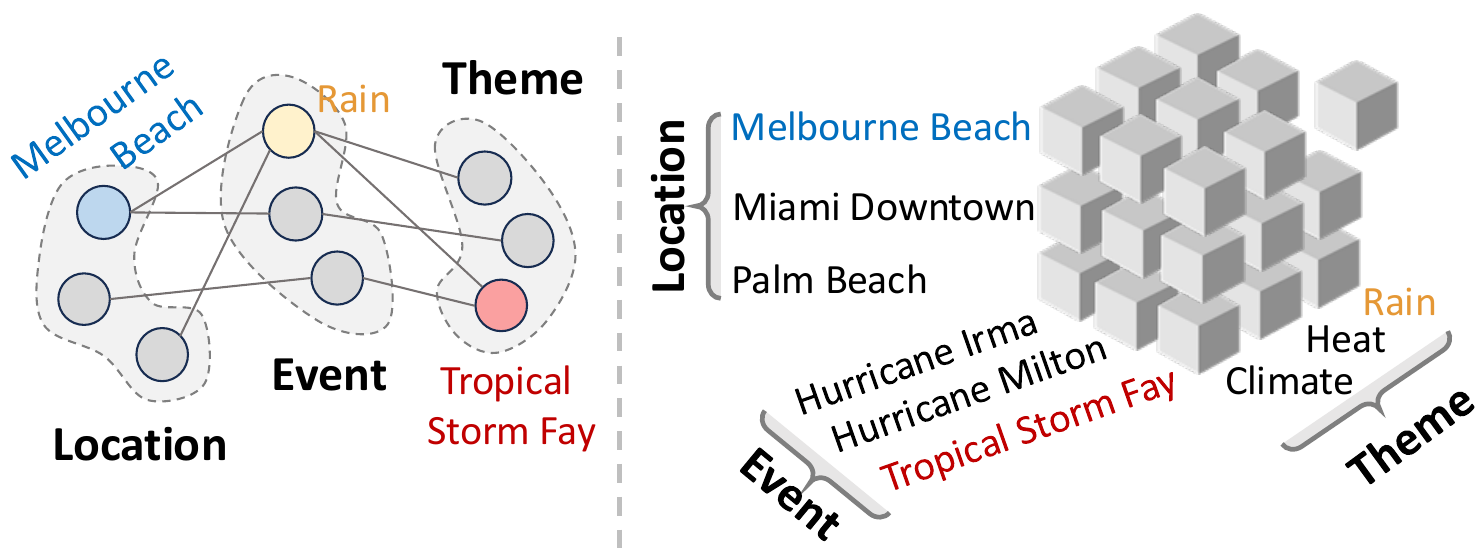}
  \caption{Graph vs.\ Hypercube.} 
  \label{fig:graph_vs_hypercube}
\end{figure}

The methods discussed above tend to \textit{perform poorly in at least one of the aspects among accuracy, efficiency, and explainability.}
This motivates us to develop a RAG system that overcomes this weakness.
In this pursuit, we identified the \emph{text cube} as a promising technique \cite{tao2018doc2cube, wang2023geospatial}. 
A text cube is an inherently explainable multidimensional structure that allocates documents into cubes along various human-defined dimensions, such as location, date, event, and theme.
Each dimension is populated with fine-grained labels (e.g., \textit{`Miami Downtown'}, \textit{`2021'}, \textit{`Hurricane Irma'}, \textit{`flooding'}) extracted from documents. 
Relevant documents can be easily indexed by these labels represented in cubes.
Furthermore, the explicit theme dimension makes the text cube particularly well-suited for \emph{theme-specific} applications where users \cite{tao2018doc2cube} primarily focus on \textit{fine-grained} inquiries, e.g., where, when, and which hurricane events hit or how much rainfall hurricanes result in certain areas.

\begin{figure}[ht!]
\centering
  \includegraphics[width=0.95\columnwidth]{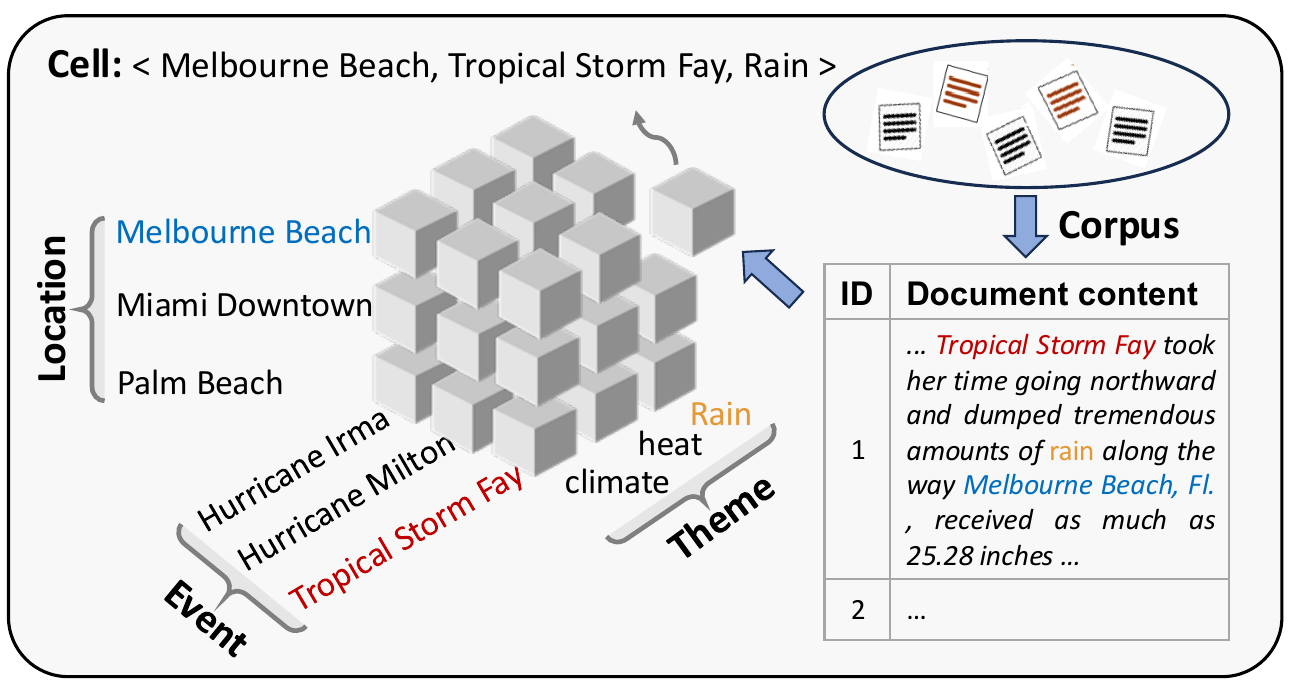}
  \caption{Hypercube construction on a corpus. We present only three dimensions for clear representation.} 
  \label{fig:hypercube}
\end{figure}

In this work, we implement a multi-dimensional (cube) structure called \textbf{Hypercube}, which indexes documents to cubes based on fine-grained labels in a human-defined, multi-dimensional space.
Building on this structure, we propose \textbf{Hypercube-RAG}, a novel RAG framework to enhance the retrieval process by retrieving relevant documents from the corresponding cube cells.
\hyperrag exhibits three characteristics simultaneously:
(1) \textit{Efficiency}. The retrieval process quickly narrows down the search to the right cube cell(s) based on the labels with the constant time $\mathcal{O}(c)$, while taking $\mathcal{O}(\kappa)$ to fetch $\kappa$ documents assigned in that cell.
(2) \textit{Explainability}. Hypercube is inherently explainable as fine-grained labels in cubes represent the compact information in documents (see Figure \ref{fig:graph_vs_hypercube}, right and Figure \ref{fig:hypercube});
(3) \textit{Accuracy}. The cube label-based search supports both sparse and dense embedding strategies, effective for capturing uncommon thematic terminology and semantically relevant information (Section \ref{sec:retrieve_hyper}).
Overall, our contributions in this chapter are summarized as follows:
\begin{enumerate}[leftmargin=0.5cm,nosep]
    \item We identify the shortcomings of conventional RAG methods, especially for \textit{theme-specific} question-answering in scientific domains.
    \item We propose \hyperrag, a simple yet accurate, efficient, and explainable RAG method.
    \item We conduct experiments on three datasets, demonstrating that our \hyperrag outperforms other baseline methods in accuracy and efficiency when it comes to \emph{in-domain} question-answering.
\end{enumerate}

\section{Related Work}
\subsection{Text-based RAG}
BM25 \cite{robertson1994some, trotman2014improvements} is a classical sparse retrieval method based on TF-IDF principles, widely used in information retrieval. It ranks documents by scoring query term matches, with adjustments for term frequency saturation and document length. While efficient and interpretable, BM25 lacks semantic understanding and often misses paraphrased or contextually related documents without exact lexical overlap, limiting its effectiveness in tasks requiring deeper language comprehension.
Dense embedding retrievers that compute contextual similarity between the given query and a document/chunk in the embedding space \cite{jiang2023active}. 
Common retrievers consist of DPR~\cite{karpukhin2020dense}, Contriever~\cite{izacard2022contriever}, e5~\cite{wang2022text}, and ANCE \cite{xiong2020approximate}.
By capturing deeper semantic relationships, these methods improve performance in open-domain question answering and generative tasks, especially when surface-form matching fails.
However, dense retrieval can return semantically close yet topically irrelevant documents, introducing noise and hallucinations. Moreover, its opaque nature also hinders interpretability \cite{ji2019visual}, limiting adoption in high-stakes or scientific domains. Recent efforts aim to mitigate these limitations by incorporating hybrid retrieval techniques~\cite{maillard2021multi} and domain-adaptive retrievers~\cite{lee2021learning}.

\subsection{Structured RAG}
Graph-based RAG methods enhance the standard RAG paradigm by introducing structured knowledge representations, such as entity or document graphs.
GraphRAG~\cite{edge2024local} derives an entity knowledge graph from the source documents and gathers summaries for all groups of closely related entities. 
LightRAG \cite{guo2024lightrag} employs a dual-level retrieval system that enhances comprehensive information retrieval from both low-level and high-level knowledge discovery.
HippoRAG \cite{gutierrez2024hipporag} is a knowledge graph-based retrieval framework inspired by the hippocampal indexing theory of human long-term memory. 
HippoRAG 2 \cite{gutierrez2025rag} builds on top of HippoRAG and enhances it with deeper passage integration and more effective online use of an LLM. 
SELF-RAG \cite{asai2023self} enhances an LLM’s quality and factuality through retrieval
and self-reflection.
RA-DIT \cite{lin2023ra} fine-tunes the pre-trained LLMs or retrievers to incorporate more up-to-date and relevant knowledge.
Despite these improvements, graph construction and reasoning can be computationally intensive due to the massive nodes and edges.

\paragraph{Text Cube.}
\cite{wang2023geospatial} designs the text cube to structure spatial-related data across geospatial dimensions, thematic categories, and diverse application semantics. 
STREAMCUBE implements the text cube to incorporate both spatial and temporal hierarchies~\cite{feng2015streamcube}.
Doc2Cube~\cite{tao2018doc2cube} automates the allocation of documents into a text cube to support multidimensional text analytics.
Despite these elaborate designs, the text cube has not yet been explored in the context of retrieval-augmented generation (RAG) systems.

\section{Methodology}
\label{sec:hrag}
\begin{figure}[ht!]
\centering
  \includegraphics[width=\columnwidth]{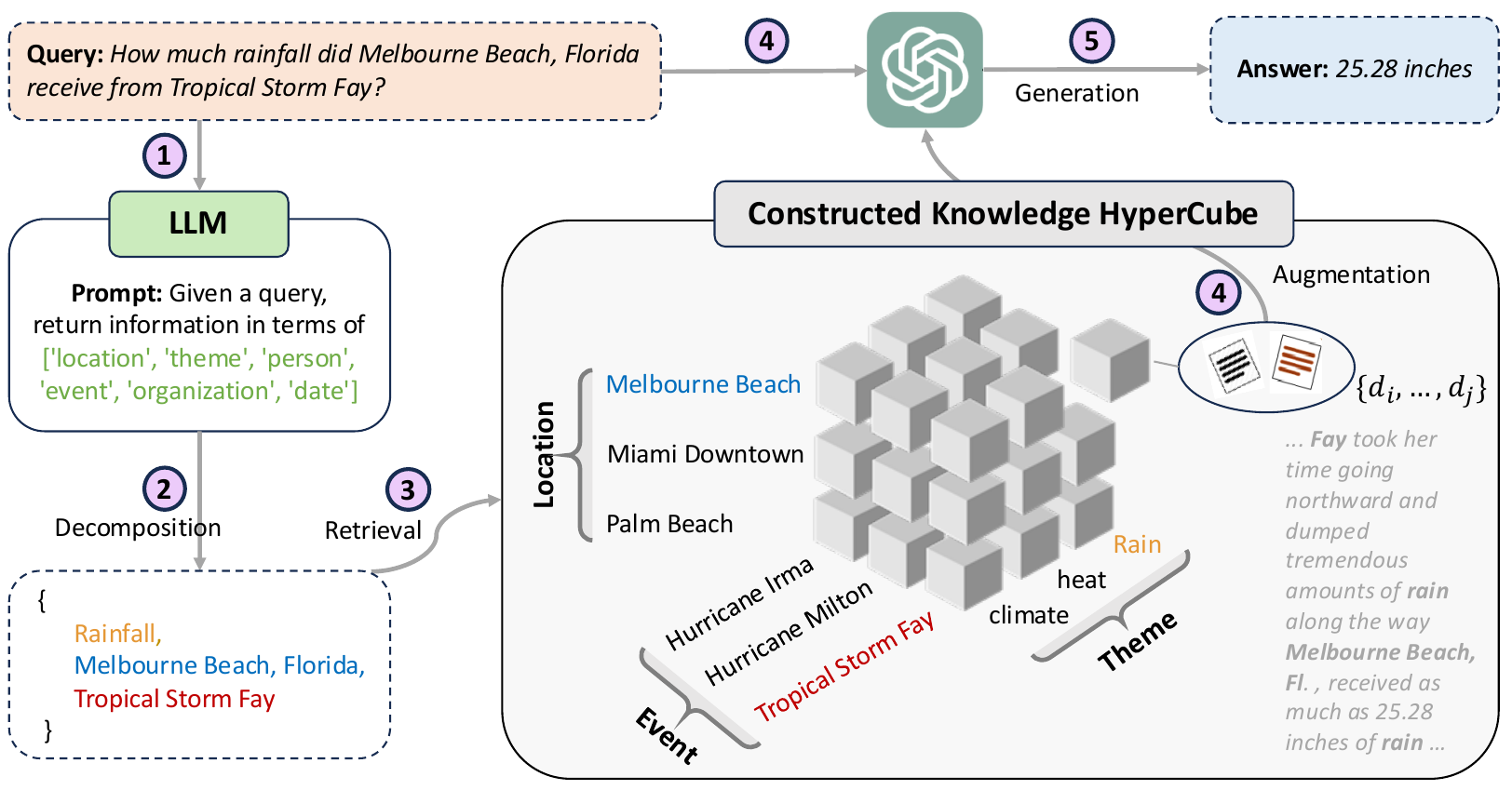}
  \caption{The illustration of \hyperrag framework. 
      \tikzmarknode[mycircled]{a1}{1} \textbf{Input:} Input the query and the prompt into a LLM; 
      \tikzmarknode[mycircled]{a1}{2} \textbf{Decomposition:} LLM decomposes the query into different dimensions; 
      \tikzmarknode[mycircled]{a1}{3} \textbf{Retrieval:} according to these dimensions, we use \hyperrag to retrieve relevant documents; \tikzmarknode[mycircled]{a1}{4} \textbf{Augmentation:} query is augmented with retrieved documents (ranked already);
      \tikzmarknode[mycircled]{a1}{5} \textbf{Generation:} LLM output with \hyperrag.
  } 
  \label{fig:hypercube_rag}
\end{figure}

Figure \ref{fig:hypercube_rag} illustrates the \hyperrag framework through a scientific question-answering example.
In the following, we define the hypercube structure (Section \ref{sec:def_hyper}), describe the process for constructing the hypercube (Section \ref{sec:construct_hyper}), explain how two sparse and dense embedding retrieval strategies are combined inside of hypercube (Section \ref{sec:retrieve_hyper}), and present how to rank documents retrieved using hypercube (Section \ref{sec:rank_hyper}).

\subsection{Hypercube Formulation}
\label{sec:def_hyper}
Given a text corpus with $k$ documents: $\mathcal{D}$, the hypercube is designed as a multidimensional data structure, $\mathcal{C}=\mathcal{L}_1\otimes \mathcal{L}_2 \otimes \dots \otimes \mathcal{L}_M$, where $M$ is the number of hypercube dimensions, and $\mathcal{L}_i$ is the set of labels used for the category chosen along the $i^{th}$ dimension.
For any document $d \in \mathcal{D}$, a text classification algorithm (e.g, TeleClass \cite{zhang2025teleclass}) will assign it to one or more cube cells. 
This process is equivalent to assigning one or more $N$-dimensional labels ($l_{t_1}, l_{t_2}, \dots, l_{t_N}$) for a document, $d$, where label $l_{t_j} \in \mathcal{L}_j$ represents the categorization of $d$ along the $j^{th}$ dimension from the set of labels, $\mathcal{L}_j$. 

\subsection{Hypercube Construction}
\label{sec:construct_hyper}
For most environmental applications, \textit{theme, location, date, event}, and/or \textit{person} are often of high interest.
The core idea of the hypercube is to use human-identified entities or topics contained in the corpus and index documents in cubes with the appropriate dimensions, thereby facilitating the subsequent retrieval. 
In this work, we employ multiple methods to construct a holistic and precise knowledge hypercube.

\paragraph{Name Entity Recognition (NER).}
We first pre-define hypercube dimensions \texttt{LOCATION}, \texttt{DATE}, \texttt{EVENT}, \texttt{PERSON} and \texttt{ORGANIZATION}. For each dimension, we utilize the pretrained language models (PLMs) \cite{devlin2019bert, liu2019roberta} to extract salient entities from documents: $\mathcal{E}=\mathcal{F}(d)$, where $\mathcal{E}(d) = \{e_1, e_2, \dots, e_{l(d)}\}$ is the set of $l(d)$ named entities identified in the document, $d$, and $\mathcal{F}$ refers to a PLM. 
For instance, those entities belonging to the \texttt{EVENT} dimension are represented:
\begin{equation}
    \mathcal{E}^{\texttt{EVENT}} = \{e\in \mathcal{E}(d) \mid \text{dim}(e)=\texttt{EVENT}\}.
\end{equation}

\paragraph{Theme-specific Extraction.}
While the above entities can be captured by NER models, abstract and intricate theme-specific information may be missed, such as `hurricane path', `rainfall intensity', and `climate change'. 
Therefore, we add another hypercube dimension \texttt{THEME}, such key theme-specific phrases can be extracted.
We implement it with KeyBERT\footnote{\url{https://github.com/MaartenGr/KeyBERT?tab=readme-ov-file}}, a language model pre-trained on large-scale text corpora, which demonstrates strong performance in capturing general topical relevance. 

Using the methods described above, we extract fine-grained entities, topics, and phrases, and assign them to different hypercube dimensions, including \texttt{LOCATION}, \texttt{DATE}, \texttt{EVENT}, \texttt{ORGANIZATION}, \texttt{PERSON}, and \texttt{THEME}. 
To enhance coverage of \textit{in-domain} semantics, we incorporate human-in-the-loop curation to expand the \texttt{THEME} dimension, ensuring it captures a comprehensive set of relevant topics. 

\subsection{Hypercube Retrieval}
\label{sec:retrieve_hyper}
A \textbf{constructed} knowledge hypercube can serve for the retrieval process based on its \textit{fine-grained labels}.
Given a query $q$, an LLM first decomposes it into a set of entities and topics, $\mathcal{E}(q) = \{e_1, e_2, \dots, e_{l(q)}\}$, aligned with various dimensions (see the prompt in Figure \ref{fig:hypercube_rag}), then retrieval inside of hypercube is performed by matching these decomposed components with cube labels.
Our hypercube supports two matching strategies: \textit{sparse exact lexical match} and \textit{dense embedding match}.
We prioritize leveraging the exact matching strategy due to its high precision, scoring documents based on the frequency with which query entities appear within them:
\begin{equation}
    \text{score}(d, q) = \sum_{e_j \in \mathcal{E}(q)} \mathds{1}_{[e_j \in \mathcal{E}(d)]},
\end{equation}
where $\mathds{1}$ is an indicator function that equals 1 if an entity of the query exactly matches a term in the document, and 0 otherwise.

Considering the example in Figure \ref{fig:hypercube_rag}, the last two components from the query, \textit{Melbourne Beach} and \textit{Tropical Storm Fay}, are handled using the exact matching strategy, as they correspond precisely to fine-grained labels in the hypercube.
However, the first component, \textit{Rainfall}, does not exactly match any label(s) in the hypercube, even though its matched label should be \textit{Rain}.

To address such cases, \hyperrag supports semantic retrieval by computing the similarity between query components and fine-grained cube labels.
When the similarity score exceeds a predefined threshold $\tau$, semantic retrieval is triggered. 
Specifically, both the query entities and cube labels are projected into an embedding space, $\mathcal{Z}$, with an encoder. It can be represented as:
\begin{align}
    \mathcal{Z}(q) &= \text{Encoder}(\mathcal{E}(q)), \\
    \mathcal{Z}(d) &= \text{Encoder}(\mathcal{E}(d)), \\ 
    \text{score}(d, q) &= \text{sim}(\mathcal{Z}(q), \mathcal{Z}(d)),
\end{align}
where $\text{sim}(\cdot)$ refers to the similarity function.

\subsection{Hypercube Ranking}
\label{sec:rank_hyper}
The retrieved documents need to be provided to the LLM as contextual input.
However, long-context RAG methods do not always improve the quality of LLM responses, since they may introduce irrelevant or noisy information \cite{liu2024lost,jin2024long}.
This raises a critical challenge: how to precisely filter and select theme-relevant documents.
To address this, we prioritize returning those documents that fully cover key components derived from the query:
\begin{equation}
    \mathcal{D}_{return} = \{ d_i \in \mathcal{D} \mid \mathcal{E}(q) \subseteq \mathcal{E}(d_i) \}.
\end{equation}

\noindent If no documents fully cover all query components (i.e., $\mathcal{D}_{return}=\emptyset$), we return the next best set of documents with the highest partial coverage:
\begin{equation}
\mathcal{D}^{*}_{\text{return}} = \arg\max_{d_i \in \mathcal{D}} |\mathcal{E}(d_i)|.
\end{equation}

\noindent Taking the example in Figure \ref{fig:hypercube_rag}, \hyperrag prioritizes returning Doc. \texttt{A} and \texttt{B} since they cover all three keywords (see the following table). 
In cases where Doc. \texttt{A} and \texttt{B} do not exist, then it returns Doc. \texttt{C}, which contains the next best subset of information based on the query decomposition.

\begin{table}[ht]
\centering
\resizebox{0.7\textwidth}{!}{
\begin{tabular}{l|l}
\toprule
\bf Doc.   & \bf Covered Information   \\
\midrule
\texttt{A}   & `Melbourne Beach', `Tropical Storm Fay', `Rainfall'    \\
\texttt{B}   & `Melbourne Beach', `Tropical Storm Fay', `Rainfall'     \\
\texttt{C}   & `Tropical Storm Fay', `Rainfall'     \\
\texttt{D}   & `Tropical Storm Fay'     \\
\bottomrule
\end{tabular}
}
\caption{An example to show the hypercube ranking algorithm.}
\end{table}

\section{Experiment}
\subsection{Datasets}
\label{sec:data}
In this work, we focus on studying \hyperrag within the context of a \textit{single} hypercube configuration.
While more general domains may benefit from modeling multiple hypercubes, our evaluation is centered on one 
scientific domain to enable a focused and controlled analysis.
We also provide a more detailed discussion in the Discussion section (Section \ref{sec:why_theme}).
We collected the text corpus from a subset of the Science Daily Climate Change dataset (SciDCC) \cite{mishra2021neuralnere}, which was created by web scraping from the ``Geography'' and ``Climate'' topics in the environmental science section of the Science Daily website\footnote{\url{https://www.sciencedaily.com/}}. 
We created an aging dam corpus by collecting news articles related to dam failures in the United States from Google News\footnote{\url{https://news.google.com/}}. Then, an LLM (GPT-4o) is employed to generate 300, 300, and 186 QA pairs, respectively. Each pair was manually validated to ensure that the corresponding question could be answered using information contained in the corpus. To ensure representativeness, each dataset generates three types of QA pairs: (a) long-form, free-style; (b) short-term factual; and (c) quantitatively factual.
We present the summary in Table \ref{tab:data_stats} and QA example of each type are as follows.

\begin{table}[ht]
\centering
\resizebox{0.8\textwidth}{!}{
\begin{tabular}{l|ccccc}
\toprule
\bf Dataset   & \# Doc.   & Doc. Length     & \# QA   & Ques. Length    & Ans. Length    \\
\midrule
\bf Hurricane & 844      & 52$\sim$1781    & 300    & 6$\sim$27       & 2$\sim$119 \\
\bf Geography & 432      & 22$\sim$1646    & 300    & 5$\sim$34       & 2$\sim$79 \\
\bf Aging Dam & 186      & 25$\sim$4782    & 186    & 6$\sim$33       & 2$\sim$150 \\
\bottomrule
\end{tabular}
}
\caption{Datasets. Length denotes the number of words.}
\label{tab:data_stats}
\end{table}

\begin{figure}[ht!]
\makebox[\textwidth][c]{%
\begin{tcolorbox}[colback=white,colframe=yellow!70,title={\bf Three types of QA pairs}, width=\textwidth] 
\footnotesize
\begin{center}
    \begin{tikzpicture}
      \draw[dashed] (0,0) -- (\linewidth,0)
      node[midway, above]{\small \textbf{Long-form, free-style}};
    \end{tikzpicture}
\end{center}
\footnotesize
\textbf{Question}: How does El Niño Modoki differ from traditional El Niño events in terms of hurricane activity?   \\
\textbf{Answer}: \textit{El Niño Modoki, a new type of El Niño that forms in the Central Pacific, is associated with a higher storm frequency and a greater potential for making landfall along the Gulf coast and the coast of Central America. This is different from traditional El Niño events, which are more difficult to forecast and can result in diminished hurricanes in the Atlantic. El Niño Modoki is more predictable, potentially providing greater warning of hurricanes by a number of months. The exact cause of the shift from traditional El Niño to El Niño Modoki is not yet clear, but it could be due to natural oscillations or El Niño's response to a warming atmosphere.} 
\begin{center}
    \vspace{-1em}
    \begin{tikzpicture}
      \draw[dashed] (0,0) -- (\linewidth,0)
      node[midway, above]{\small \textbf{Short-term factual}};
    \end{tikzpicture}
\end{center}
\footnotesize
\textbf{Question}: Which dam failure in northwestern Wisconsin triggered flooding near the Minnesota border? \\
\textbf{Answer}: \textit{Radigan Flowage Dam.} 
\begin{center}
    \vspace{-1em}
    \begin{tikzpicture}
      \draw[dashed] (0,0) -- (\linewidth,0)
      node[midway, above]{\small \textbf{Quantitative factual}};
    \end{tikzpicture}
\end{center}
\footnotesize
\textbf{Question}: How many named storms were observed in the Atlantic during the 1997 hurricane season? \\
\textbf{Answer}: \textit{7 named storms.} \\

\end{tcolorbox}
}
\label{fig:qa_type}
\end{figure}

\subsection{Baselines}
We select three types of methods for comparison: 1) \textbf{sparse
retriever} BM25 \cite{robertson1994some}; 2) \textbf{dense embedding retrieval} methods, Contriever \cite{izacard2022contriever}, e5 \cite{wang2024multilingual}, Nvidia/NV-Embedv2 \cite{lee2024nv}; 3) graph-based methods, GraphRAG \cite{edge2024local}, LightRAG \cite{guo2024lightrag}, and HippoRAG \cite{gutierrez2024hipporag} and HippoRAG 2 \cite{gutierrez2025rag}. 
For all graph-based baselines, we conducted experiments by running their official GitHub repositories. 
For dense embedding retrievers, we run these models from the Hugging Face.
For sparse retriever BM25, we implement it using a GitHub repository\footnote{\url{https://github.com/dorianbrown/rank_bm25}}.
All experiments were conducted on an A100 GPU with 80 GB memory.

In Table \ref{tab:main_result}, we also include baseline methods without RAG, such as DeepSeek-R1\footnote{\url{https://github.com/deepseek-ai/DeepSeek-R1}}, Qwen2.5-7B-Instruct\footnote{\url{https://qwenlm.github.io/blog/qwen2.5/}}, Llama-3.3-70B-Instruct\footnote{\url{https://huggingface.co/meta-llama/Llama-3.3-70B-Instruct}}, and Llama-4-Scout-17B-16E-Instruct\footnote{\url{https://huggingface.co/meta-llama/Llama-4-Scout-17B-16E-Instruct}}.
Note that DeepSeek-R1 is a reasoning model, we set 0.6 and 1 as the temperature by following their recommendations. Since its output also includes the reasoning/thinking process, we exclude it and consider the final output as the answer.
The temperatures for other models are set to 0.

\subsection{Evaluation Metrics}
Following the work \cite{cui2025timer}, we evaluate the quality of LLMs' responses using automated metrics derived from token-level representations, including BLEU \cite{papineni2002bleu} and BertScore \cite{zhang2019bertscore} to provide the standard assessment of response quality. 
BLEU scores and BERTScores are computed with the NLTK\footnote{\url{https://www.nltk.org/}} and SentenceTransformer\footnote{\url{https://sbert.net/}} Python packages, respectively.
Since the settings are open-text responses, we also employ LLM-as-a-Judge that assesses correctness and completeness. 
All LLM-based evaluations use GPT-4o as the judge. The prompts used are in Figure \ref{fig:prompt_llm_as_judge}.

\begin{figure}[t!]
\centering
  \input{tables/hypercube/prompt_llm_as_judge}
  \vspace{-2mm}
  \caption{Prompt template for LLM-as-a-judge.} 
  \label{fig:prompt_llm_as_judge}
\end{figure}

\section{Results}
In this section, we show and analyze the experimental results for accuracy, efficiency, and explainability of our \hyperrag and other baseline methods.
 
\subsection{Accuracy}
\begin{figure}[ht]
\centering
\begin{subfigure}[t]{0.48\textwidth}
  \centering
  \includegraphics[width=0.99\linewidth]{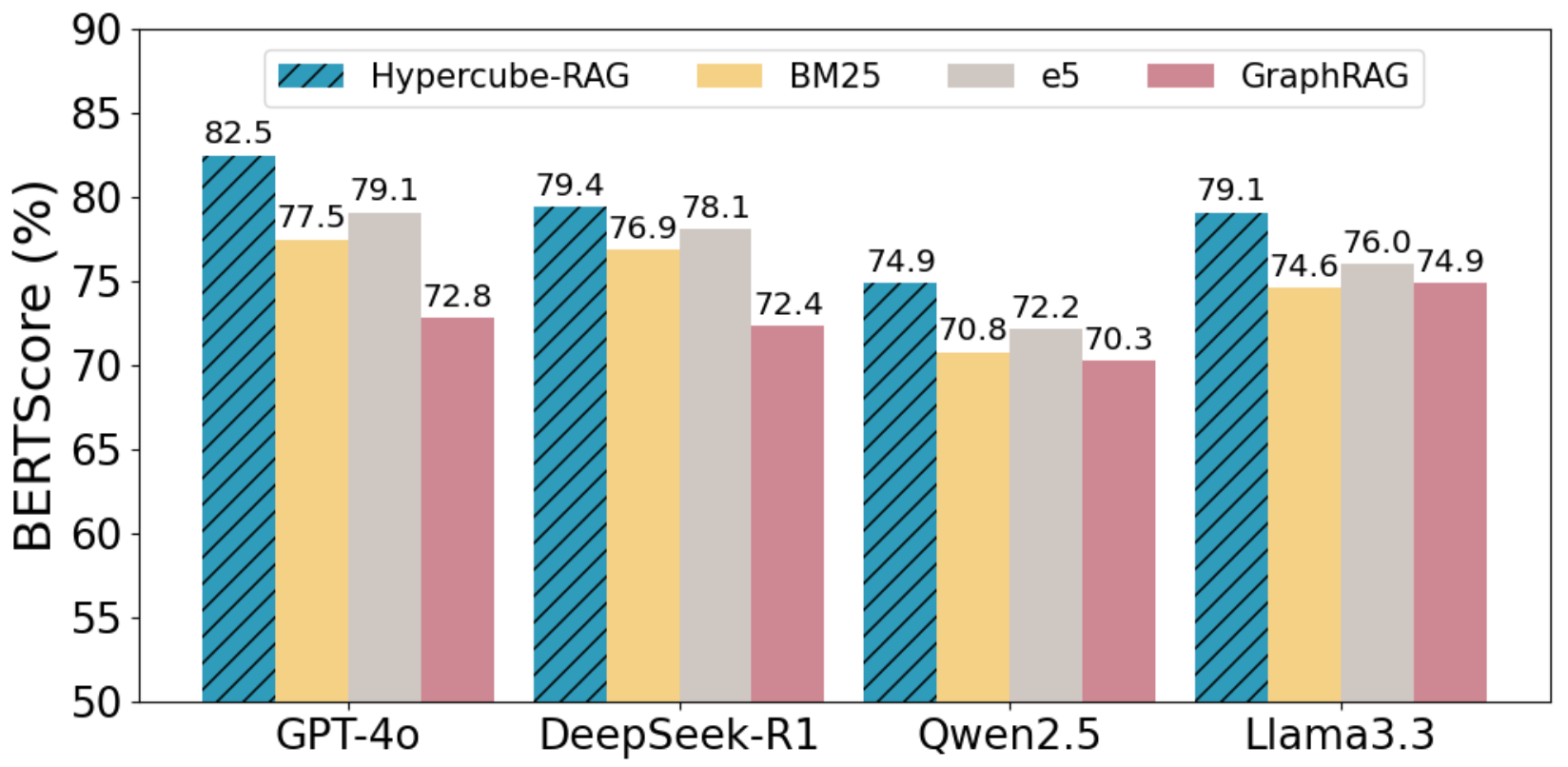}
  \caption{Hurricane}
\end{subfigure}
\hfill
\begin{subfigure}[t]{0.48\textwidth}
  \centering
  \includegraphics[width=0.99\linewidth]{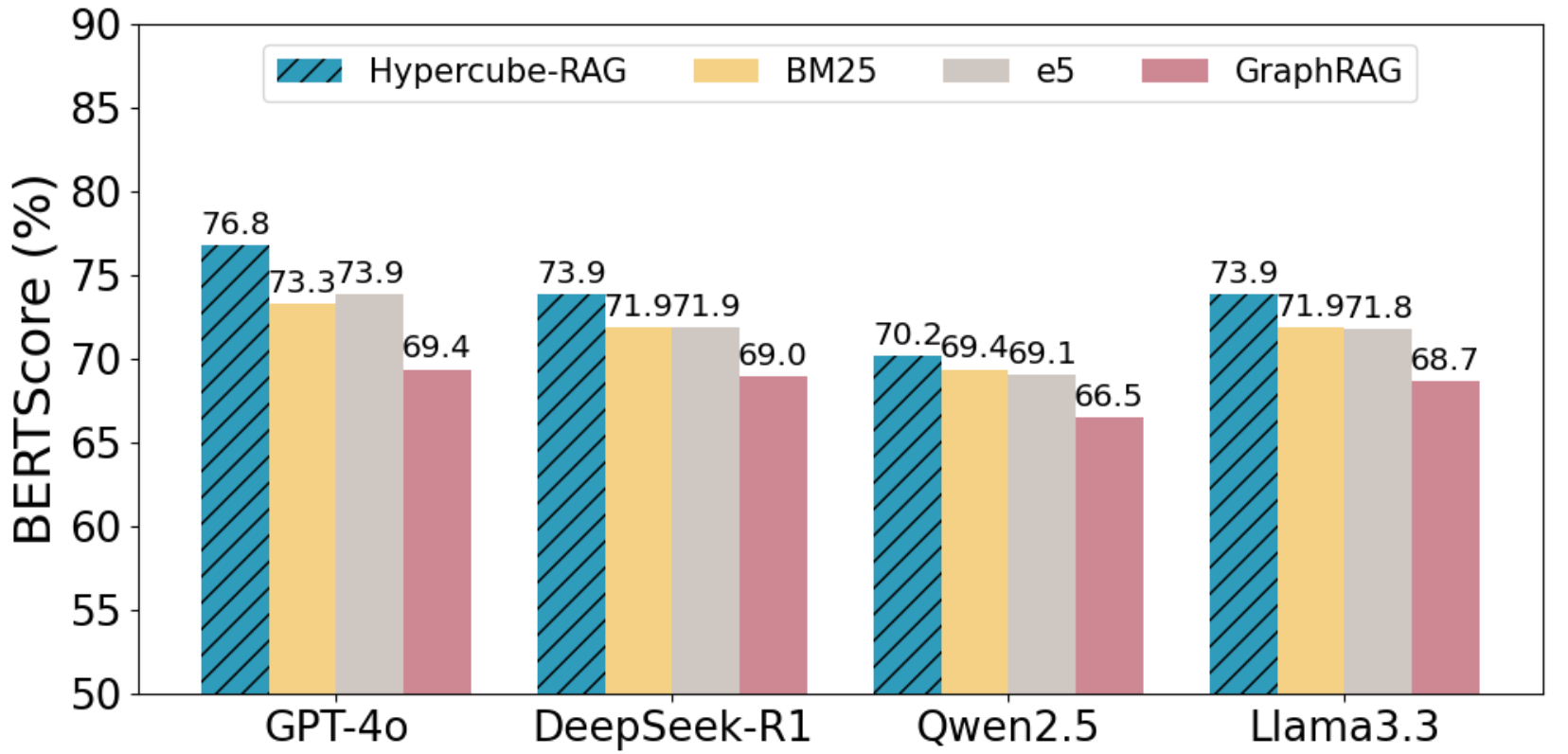}
  \caption{Geography}
\end{subfigure}
\caption{Performance comparison with \textit{various} LLMs (only the range between 50\% and 90\% shown).}
\label{fig:bar_plot}
\end{figure}

We compare our \hyperrag with semantic RAG methods, graph-based RAG methods, and LLMs without retrieval. 
Table \ref{tab:main_result} reveals three key findings.
First, our method consistently outperforms other baselines across three datasets, demonstrating its effectiveness in enhancing the capabilities of LLMs for in-domain scientific question-answering. 
Compared to the second-best method, \hyperrag improves the accuracy by $3.7\%$ measured as relative gains across three of four metrics used.
Second, all RAG methods perform much better than direct inference of LLMs without retrieval, as expected, showing the effectiveness of incorporating external knowledge for \emph{in-domain} QA tasks. 
Finally, we also evaluate \hyperrag against a range of baseline methods across multiple LLMs developed by different companies. As illustrated in Figure~\ref{fig:bar_plot}, our method achieves superior performance across all models regardless of the underlying LLM architectures.

\begin{table}[ht!]
\centering
\resizebox{\textwidth}{!}{
\begin{tabular}{l|cccc|cccc|cccc}
\toprule
\multirow{3.5}{*}{\textbf{Method}}  & \multicolumn{4}{c|}{\textbf{Hurricane}}     & \multicolumn{4}{c|}{\textbf{Geography}}   & \multicolumn{4}{c}{\textbf{Aging Dam}} \\
\cmidrule(lr){2-5}  \cmidrule(lr){6-9}  \cmidrule(lr){10-13}
& \multicolumn{2}{c}{\cellcolor{lightyellow}Automatic Metric} & \multicolumn{2}{c|}{\cellcolor{lightblue}LLM-as-Judge} & \multicolumn{2}{c}{\cellcolor{lightyellow}Automatic Metric} & \multicolumn{2}{c}{\cellcolor{lightblue}LLM-as-Judge} & \multicolumn{2}{c}{\cellcolor{lightyellow}Automatic Metric} & \multicolumn{2}{c}{\cellcolor{lightblue}LLM-as-Judge} \\
\cmidrule(lr){2-5}  \cmidrule(lr){6-9}   \cmidrule(lr){10-13}
                        & BLEU      & BERTScore       & Correct & Complete             & BLEU      & BERTScore       & Correct & Complete     & BLEU      & BERTScore       & Correct & Complete\\
\midrule

\rowcolor{gray!10}
\multicolumn{13}{l}{\textbf{\textit{No Retrieval}}} \\
GPT-4o                                      & 2.6  & 66.8 & 57.1  & 23.5    & 2.8   & 63.9   & 64.2 & 42.1    & 5.5   & 62.5    & 56.9 & 19.6 \\
DeepSeek-R1                                 & 2.2  & 69.2 & 53.4  & 28.5    & 1.3   & 61.7   & 57.9 & 45.7    & 8.7   & 71.6    & 35.4 & 18.8 \\
Qwen2.5-7B-Instruct                         & 1.2  & 46.9 & 23.5  & 6.9     & 1.7   & 47.1   & 32.1 & 12.6    & 1.9   & 43.4    & 15.5 & 6.5 \\
Llama-3.3-70B-Instruct                      & 2.0  & 60.2 & 39.8  & 14.6    & 1.6   & 51.3   & 49.3 & 21.2    & 2.4   & 46.7    & 39.7 & 17.1 \\
Llama-4-Scout-17B-16E-Instruct              & 2.3  & 62.7 & 53.9  & 39.1    & 3.1   & 61.3   & 48.8 & 21.9    & 5.2   & 64.2    & 34.4 & 15.1 \\
\midrule

\rowcolor{gray!10}
\multicolumn{13}{l}{\textbf{\textit{Sparse and Dense Embedding Retrieval}}} \\
BM25 \cite{robertson1994some}               & 10.5 & 77.5 & 74.3  & 60.8    & 7.9   & 73.3  & 79.7 & 70.8     & 18.1 & \ul{80.2}     & 85.7 & \ul{75.9} \\
Contriever \cite{izacard2021unsupervised}   & 11.3 & 78.6 & 75.7  & 60.6    & 7.5   & 74.1  & \ul{82.5} & \ul{73.6}     & 16.7 & 76.8     & 70.9 & 60.9 \\
e5 \cite{wang2024multilingual}              & 11.9 & \ul{79.1} & 78.3  & \ul{63.9}    & 7.1   & 73.9  & 81.2 & 71.2     & \ul{19.7} & 79.8     & \bf 88.9 & 75.5 \\
NV-Embed \cite{lee2024nv}                   & \ul{12.2} & \ul{79.1} & 76.9  & 59.6    & 7.5   & \ul{75.4}  & 81.9 & 72.3     & 18.8 & 79.6     & 79.4 & 73.6 \\
\midrule

\rowcolor{gray!10}
\multicolumn{13}{l}{\textbf{\textit{Graph-based RAG}}} \\
LightRAG \cite{guo2024lightrag}             & 7.6  & 72.2 & 64.1  & 47.6    & 7.0    & 67.8  & 69.8 & 59.6    & 12.8 & 75.4 & 65.9 & 46.5 \\
GraphRAG \cite{edge2024local}               & 6.5  & 72.8 & 68.8  & 50.9    & 6.9    & 69.4  & 70.6 & 63.8    & 11.8 & 72.4 & 72.5 & 58.6 \\
HippoRAG \cite{gutierrez2024hipporag}       & 6.8  & 73.6 & 70.5  & 48.8    & 7.8    & 70.3  & 76.9 & 69.7    & 12.4 & 73.8 & 80.9 & 58.6 \\
HippoRAG 2 \cite{gutierrez2025rag}          & 8.6  & 76.3 & \bf 86.7  & 53.5    & \ul{9.2}    & 74.6  & 80.6 & 67.5    & 11.5 & 74.1 & \ul{88.2} & 53.7 \\
\midrule

\rowcolor{gray!10}
\multicolumn{13}{l}{\textbf{\textit{\textcolor{red}{Our Method}}}} \\
Hypercube-RAG                               & \bf 13.2 & \bf 82.5 & \ul{80.6}  & \bf 66.2    & \bf 9.3    & \bf 76.8  & \bf 83.6 & \bf 74.8      & \bf 21.5 & \bf 82.6 & 86.9 & \bf 77.8 \\
\bottomrule

\end{tabular}
}
\caption{Performance Comparison ($\%$) of LLMs themselves without RAG, LLMs with various RAG baselines and Hypercube-RAG (\textcolor{red}{ours}). The best scores are in \textbf{bold} while the second-best scores are highlighted with \ul{underline}. All RAG methods were experimented with GPT-4o as the base.}
\label{tab:main_result}
\end{table}

\subsection{Efficiency}
We study the response speed of diverse RAG methods over the corpus size and report the retrieval time for one query in Table \ref{tab:efficiency}. 
We break down our analysis as follows.

\paragraph{Within corpus size $k$.} BM25 achieves the fastest retrieval among all evaluated methods without surprise. 
In contrast, semantic and graph-based RAGs incur significantly higher retrieval costs.
This is particularly pronounced for graph-based methods due to the computational burden of search paths in large-scale graphs, posing scalability challenges. 
Notably, our \hyperrag substantially reduces the retrieval time by one to two orders of magnitude compared to both semantic and graph-based methods, underscoring the effectiveness of the hypercube structure in optimizing retrieval efficiency. 
We attribute this to the \emph{retrieval with compact cube labels} introduced in Section \ref{sec:retrieve_hyper}. 
Although it is slightly slower than BM25, the marginal increase in retrieval time is a worthwhile trade-off for significantly improved accuracy, as shown in Table \ref{tab:main_result}.
\paragraph{Beyond corpus size $k$.}
To mimic the real-world scenario that the external knowledge database is usually large and noisy, we
expand the hurricane corpus from 844 to 11,539 documents (around 14 times).
The added documents are related to other topics, such as ``Pollution'' or ``Ozone Holes'', introducing additional noise.
\hyperrag maintains the lowest retrieval time per query (see the last column in Table \ref{tab:efficiency}), highlighting the efficiency of the cube label-based retrieval mechanism. 
This is likely because most noisy documents result in empty or sparsely populated cube cells, as they lack corresponding \emph{in-domain} fine-grained labels pre-defined in our hypercube. As a result, despite a significantly expanded corpus, the hypercube can efficiently bypass these irrelevant documents.
In contrast, other baseline methods exhibit greater sensitivity to corpus size, resulting in increased retrieval times.
\begin{table}[ht!]
\centering
\resizebox{0.6\textwidth}{!}{
\begin{tabular}{l|cccc|c}
\toprule
\bf Methods   &\bf $k/8$  &\bf $k/4$  &\bf $k/2$  &\bf $k$  &\bf $14k$\\
\midrule
BM25                    & 0.5    & 0.9      & 1.6      & 3.2      & 51.1 \\
e5                      & 14.4   & 17.4     & 19.1     & 37.8     & 82.4 \\
GraphRAG                & 114.7  & 348.2    & 1179.4   & 5260.6   & 45135.7 \\
Hypercube-RAG           & 0.7    & 1.5      & 3.3      & 7.1      & 22.1\\
\bottomrule
\end{tabular}
}
\caption{Retrieval time ($ms$) vs.\ corpus size ($k$), experimenting on the Hurricane data set (LLM base: GPT-4o).}
\label{tab:efficiency}
\end{table}

\noindent In addition, Table \ref{tab:scale} shows that the accuracy of the evaluated methods drops due to the inclusion of these noisy off-topic documents. However, \hyperrag still outperforms others, demonstrating robust resilience to noisy data.
\begin{table}[ht]
\centering
\resizebox{0.55\textwidth}{!}{
\begin{tabular}{l|cc|c}
\toprule
\bf Methods      & $k$  & $14k$     & BERTScore ($\downarrow$)\\
\midrule
BM25            & 77.5                  & 77.2                      & 0.3  \\
e5              & 79.1                  & 76.3                      & 2.8  \\
GraphRAG        & 72.8                  & 70.5                      & 2.3  \\
Hypercube-RAG   & 82.5                  & 80.3                      & 1.9  \\
\bottomrule
\end{tabular}
}
\caption{Accuracy (BERTScore) vs corpus size.}
\label{tab:scale}
\end{table}

\subsection{Explainability}
The retrieval process with our hypercube is inherently explainable through its associated cube labels. 
Considering the query \textit{``How much rainfall did Melbourne Beach, Florida receive from Tropical Storm Fay?''}, illustrated in Figure \ref{fig:hypercube_rag}, \hyperrag returns relevant documents by matching the decomposed components with fine-grained labels, \texttt{`Melbourne Beach'}, \texttt{`Tropical Storm Fay'}, and \texttt{`rainfall'}, along hypercube dimensions: \texttt{LOCATION}, \texttt{EVENT}, and \texttt{THEME}. 
Table \ref{tab:explainability} shows that three retrieved documents (Doc. \texttt{565}, \texttt{246}, \texttt{535}), which are represented with fine-grained cube labels.
These cube labels clearly explain why those documents are returned.
For example, Doc. \texttt{565} is ranked the first to return, since it contains all three key components of the query. The detailed document contents are included in Table \ref{tab:doc_content}.


\begin{table}[ht]
\centering
\resizebox{0.7\textwidth}{!}{
\begin{tabular}{l|ccc}
\toprule
\bf Doc.       & \bf Location             & \bf Event                   & \bf Theme   \\
\midrule
\texttt{565}   & `Melbourne Beach': 1  & `Tropical Storm Fay': 1   & `Rain': 5    \\
\texttt{246}   & `Florida': 1          & `Tropical Storm Fay': 1   & --    \\
\texttt{535}   & `Florida': 1          & --                        & -- \\
\bottomrule
\end{tabular}
}
\caption{Documents represented in the hypercube.}
\label{tab:explainability}
\end{table}
\begin{table}[H]
\small
\begin{tabular}{p{0.15\textwidth}|p{0.75\textwidth}}
\toprule
\textbf{Document ID} & \textbf{Document Content} \\
\midrule
\parbox[t]{0.15\textwidth}{565} & 
\parbox[t]{0.75\textwidth}{\ldots Why is September the peak month for hurricanes? NASA oceanographer, Bill Patzert at the Jet Propulsion Laboratory, Pasadena, Calif., provided the answer: ``Hurricanes are fueled by warm ocean temperatures and September is the end of the Northern Hemisphere ocean warming season. The 2008 Atlantic hurricane season started early with the formation of Tropical Storm Arthur on May 30, from the remnants of the eastern Pacific Ocean's first storm, Alma, which crossed Central America and reformed in the Gulf of Mexico. \ldots The tropical Atlantic is warm, but not unusually so.'' Once a powerful Category 3 hurricane, now a tropical depression, Gustav moved from northwest Louisiana into northeastern Texas and into Arkansas by Sept. 3. Like \textcolor{red!90}{Tropical Storm Fay} in August, Gustav's legacy will lie in large \textcolor{yellow!90}{rainfall} totals. According to the National Hurricane Center discussion on Sept. 2, "Storm total \textcolor{yellow!90}{rainfalls} are expected to be five to ten Inches with isolated maximums of 15 inches over portions of Louisiana, Arkansas and Mississippi. \textcolor{yellow!90}{Rainfall} amounts of 4-8 inches have been already reported in parts of Alabama, Mississippi and Louisiana. \ldots In August, \textcolor{red!90}{Fay}'s ten-day romp from the U.S. Southeast northward up the Appalachian Mountains seemed like a harbinger for September's storms. \textcolor{red!90}{Fay} took her time going northward and dumped tremendous amounts of \textcolor{yellow!90}{rain} along the way. \textcolor{blue}{Melbourne Beach, Florida}, received as much as 25.28 inches of \textcolor{yellow!90}{rain}. Other cities in various states reported high totals: Thomasville, Ga., reported 17.43 inches; Camden, Ala., received 6.85 inches; Beaufort, S.C., received 6.11 inches; Carthage, Tenn., reported 5.30 inches, and Charlotte, N.C., reported 5.90 inches \ldots} \\
\midrule
\parbox[t]{0.15\textwidth}{246} & 
\parbox[t]{0.75\textwidth}{\ldots Among the already apparent evidence: Dunes that historically protected Kennedy Space Center from high seas even during the worst storms were leveled during \textcolor{red!90}{Tropical Storm Fay} in 2008, Hurricane Irene in 2011 and Hurricane Sandy in 2012. A stretch of beachfront railroad track built by NASA in the early 1960s that runs parallel to the shoreline has been topped by waves repeatedly during recent storms. \ldots The problem had been occurring for years but seemed to be growing worse, beginning with the spate of hurricanes that struck \textcolor{blue}{Florida} in 2004. Jaeger said he, Adams and doctoral students Shaun Kline and Rich Mackenzie determined the cause was a gap in a near-shore sandbar \ldots} \\
\midrule
\parbox[t]{0.15\textwidth}{535} & 
\parbox[t]{0.75\textwidth}{\ldots Our theory would suggest that seepage caused by underwater flow will continue to erode and weaken the levee system around New Orleans, but the rate of this erosion should gradually slow with time, says Straub. Hopefully this research will aid the U.S. Army Corps of Engineers in identifying levees that need repair and assessing the lifespan of structures like the MRGO that are not planned for upkeep. Using fieldwork conducted in the \textcolor{blue}{Florida Panhandle}, Straub and his fellow researchers were able to better understand the process of seepage erosion, which occurs when the re-emergence of groundwater at the surface shapes the Earth's topography \ldots} \\
\bottomrule
\end{tabular}
\caption{Document contents.}
\label{tab:doc_content}
\end{table}

\subsection{Ablation Study}
\begin{figure}[ht]
\centering
  \includegraphics[width=0.7\columnwidth]{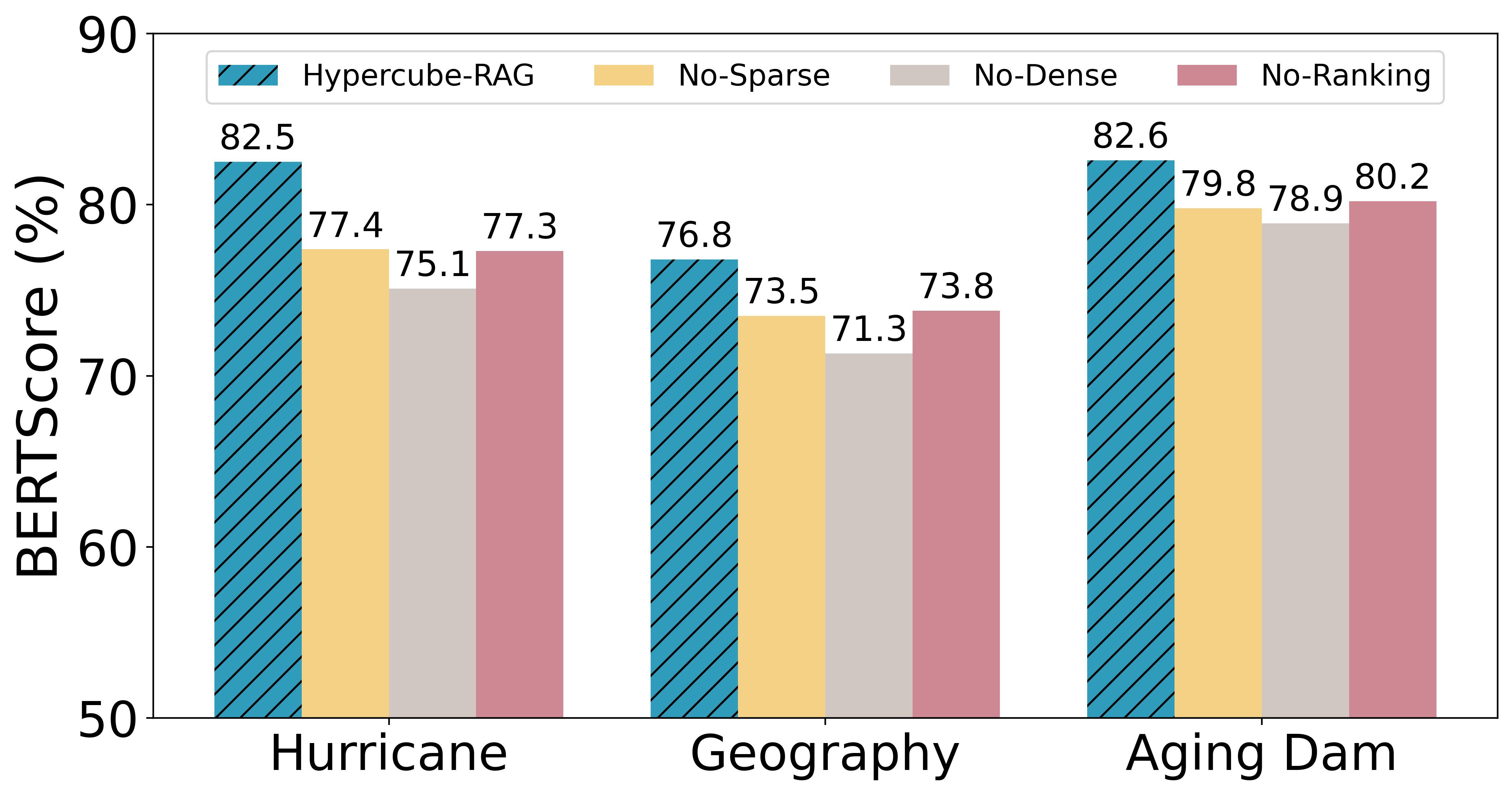}
  \caption{Ablation study on constituents of \hyperrag.} 
  \label{fig:ablation1}
\end{figure}

As shown in Section \ref{sec:retrieve_hyper}, \hyperrag combines two retrieval strategies: sparse exact matching and dense embedding retrieval. Section \ref{sec:rank_hyper} introduces the document ranking algorithm used to filter and prioritize the retrieved results. 
To study their effectiveness, we conduct an ablation study by removing each of them from \hyperrag, which are represented as \textbf{No-Sparse}, \textbf{No-Dense}, and \textbf{No-Ranking}, respectively. 
Figure \ref{fig:ablation1} shows that the full version of \hyperrag consistently outperforms all other variants, verifying positive contributions of each constituent. 
More specifically, it justifies the benefit of coupling these two strategies for in-domain theme-specific question-answering tasks, and the necessity of the ranking algorithm to filter possible irrelevant documents (``noisy'').

We also conduct an ablation study on the hypercube dimensions: \texttt{Location}, \texttt{Event}, \texttt{Date}, \texttt{Organization}, \texttt{Person}, and \texttt{Theme}. 
To study their effectiveness, we conduct an ablation study by removing each of the dimensions, which are represented as \texttt{No-Location}, \texttt{No-Event}, \texttt{No-Date}, \texttt{No-Organization}, \texttt{No-Person}, and \texttt{No-Theme}, respectively. 
Table \ref{tab:abs_dim} shows that the full version of \hyperrag consistently outperforms all other variants, verifying positive contributions of each dimension. 
More specifically, we observe that the performance drops the most when removing the \texttt{Location} and \texttt{Theme} dimensions, justifying the high importance of these two hypercube dimensions for in-domain theme-specific question-answering.

\begin{table}[ht!]
\centering
\resizebox{0.7\textwidth}{!}{
\begin{tabular}{l|ccc}
\toprule
\bf Dimension           & \bf Hurricane & \bf Geography & \bf Aging Dam \\
\midrule
No-Location             & 78.7          & 65.9          & 77.5         \\
No-Event                & 79.6          & 75.9          & 80.0        \\
No-Date                 & 79.7          & 74.5          & 80.4        \\
No-Organization         & 79.6          & 74.6          & 80.2       \\
No-Person               & 80.2          & 73.2          & 81.4         \\
No-Theme                & 77.9          & 71.7          & 81.1          \\
\midrule
FULL                    & 82.5          & 76.8          & 82.6          \\
\bottomrule
\end{tabular}
}
\caption{Ablation study on hypercube dimensions. The values are BERTScores ($\%$).}
\label{tab:abs_dim}
\end{table}

\subsection{Parameter Study}
\begin{figure}[ht]
\centering
  \includegraphics[width=0.65\columnwidth]{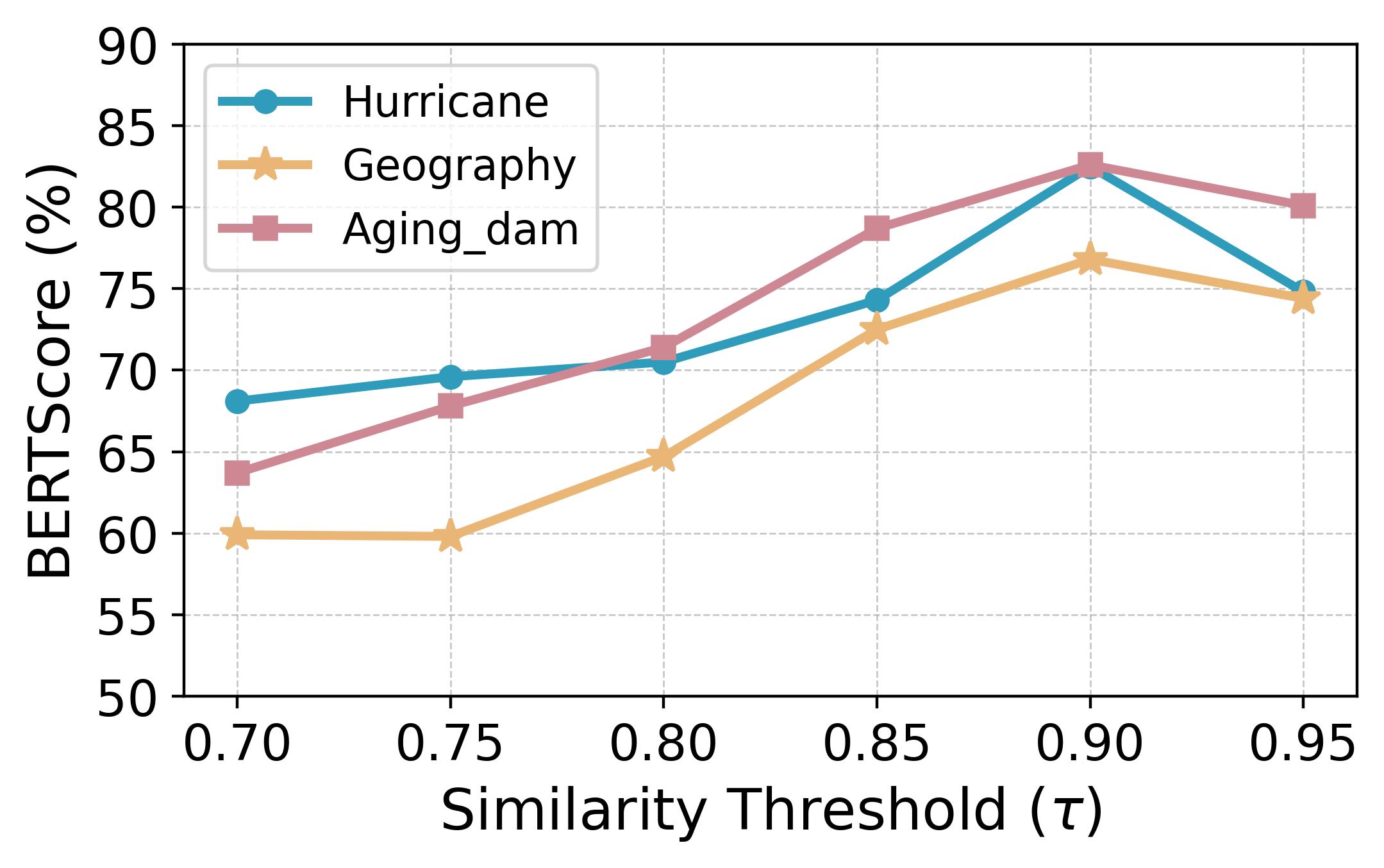}
  \caption{Performance vs similarity threshold.} 
  \label{fig:threshold}
\end{figure}

\hyperrag combines sparse exact match and semantic retrieval strategies, as introduced in Chapter \ref{sec:retrieve_hyper}. 
We investigate the effect of the semantic retrieval strategy on answer quality under varying similarity thresholds, denoted by $\tau$. 
Figure~\ref{fig:threshold} shows that performance improves with increasing threshold values when $\tau \leq 0.9$, suggesting that higher similarity scores are more effective for retrieving semantically relevant documents.
However, when $\tau > 0.9$, the behavior of semantic RAG begins to gradually degrade to the sparse exact matching strategy, resulting in a decline in performance.
This is primarily due to overly strict matching criteria, which reduce the number of retrieved documents and thus limit the available context.

\subsection{Case Study}
An additional case study in Figure \ref{fig:case_study2} 
presents that both Semantic-RAG and HippoRAG-2 fail to generate the correct answer due to inaccurate document retrieval. 
These methods tend to prioritize returning documents that are semantically similar in the embedding space.
While the retrieved documents mention general topics such as \textit{above-normal, near-normal, and below-normal hurricane season}, they miss the specific location \texttt{`Atlantic'} and organization \texttt{`Climate Prediction Center'}. 
In contrast, our \hyperrag returns the correct answer.
It accurately identifies and retrieves documents containing key elements of the query, including \textit{`Atlantic'}, \textit{`hurricane season'}, and \textit{`Climate Prediction Center'}, corresponding to the hypercube dimensions, \texttt{LOCATON}, \texttt{THEME}, and \texttt{ORGANIZATION}, respectively.
This retrieval approach intuitively prioritizes specific contextual information over abstract, ambiguous references such as \textit{above-normal, near-normal, and below-normal} alone.
\begin{figure}[t!]
\centering
  \input{tables/hypercube/case_study2}
  \vspace{-2mm}
  \caption{Comparison of three RAG methods on the same query.} 
  \label{fig:case_study2}
\end{figure}

\section{Discussion}
\subsection{Why does \hyperrag outperform other RAG methods?}
Our \hyperrag is particularly proposed for domain-focused analysis, where domain-specific information, e.g., location, date, event, event, theme, is central to the query. 
In such cases with theme-specific information, our hypercube enables precise and efficient retrieval by accessing \textbf{only one or a few cells in a theme-specific hypercube}, as each cell encapsulates multiple dimensions of information associated with a document.
These fine-grained document labels can be represented in a hypercube along different dimensions.
When a query comes in, accessing one or a few hypercubes presents a higher chance of retrieving highly relevant documents covering all key information contained in a query - \emph{Accuracy}.
Additionally, retrieval is operated with the cube labels significantly speeds up the retrieval time since those labels are in a compact space - \emph{Efficiency}. In contrast, sparse exact match and dense embedding retrieval require traversing all documents and contents inside.
Moreover, the retrieval process is inherently explainable, as the searching is based on the document's labels represented by a hypercube - \emph{Explainability}.

\paragraph{Case 1: simple query.} \textit{How much rainfall did Melbourne Beach,
Florida receive from Tropical Storm Fay?} The relevant documents can be retrieved from one cube cell in a hypercube (\textcolor{blue}{Melbourne Beach}, \textcolor{yellow!90}{rainfall}, \textcolor{red!90}{Tropical Storm Fay}).
\begin{figure}[H]
\centering  \includegraphics[width=0.65\columnwidth]{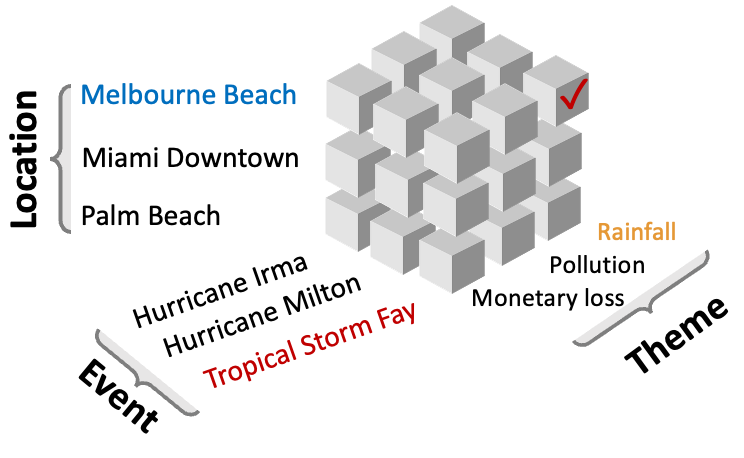}
  \caption{Access one cube cell in one hypercube. \textcolor{red!90}{\bf \checkmark} represents the touched cube cells.} 
  \label{fig:one_cube_one_cell}
\end{figure}

\paragraph{Case 2: long query with multiple topics.} In cases where a query is very diverse, they may need to access multiple cube cells such that the query information can be covered as much as possible.
For example, the diverse query could be \emph{``What consequences were caused by Tropical Storm Fay, such as how much rainfall Melbourne Beach and Palm Bay in Florida received, and how much monetary losses were caused?''}.
Those cube cells needed could be 
(\textcolor{blue}{Melbourne Beach}, \textcolor{yellow!90}{rainfall}, \textcolor{red!90}{Tropical Storm Fay}),
(\textcolor{blue}{Palm Beach}, \textcolor{yellow!90}{rainfall}, \textcolor{red!90}{Tropical Storm Fay}),
(\textcolor{blue}{Melbourne Beach}, \textcolor{yellow!90}{monetary losses}, \textcolor{red!90}{Tropical Storm Fay}),
(\textcolor{blue}{Palm Beach}, \textcolor{yellow!90}{monetary losses}, \textcolor{red!90}{Tropical Storm Fay}). 
\begin{figure}[H]
\centering  \includegraphics[width=0.65\columnwidth]{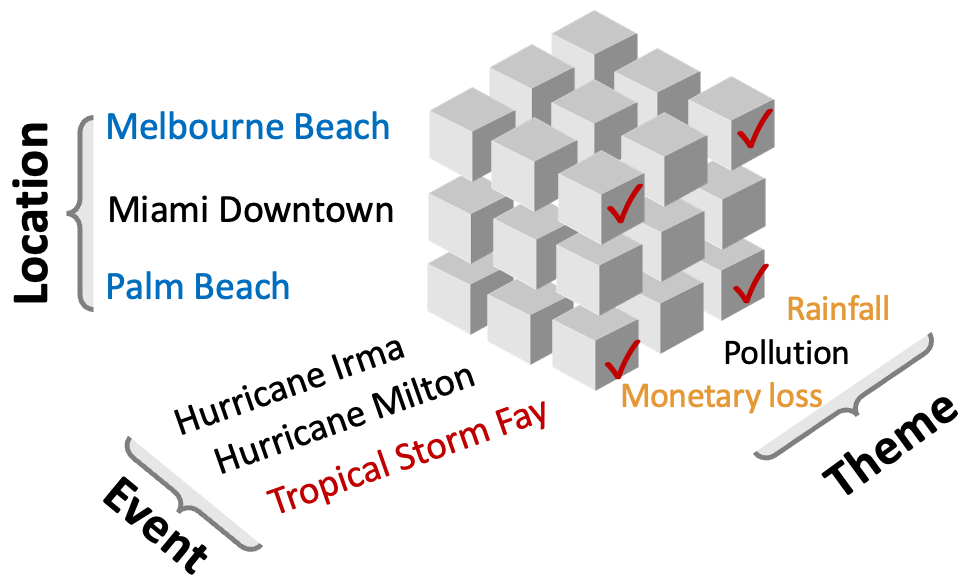}
  \caption{Access multiple cube cells in one hypercube. \textcolor{red!90}{\bf \checkmark} represents the touched cube cells.} 
  \label{fig:one_cube_multi_cells}
\end{figure}

\paragraph{Case 3: diverse query with shifted topics.} Given a more complicated query including shifted topics, e.g., \textit{``How did the monetary loss caused by Tropical Storm Fay impact the industrial layoff in Miami Downtown?''}, cube cells across two hypercubes are needed 
(\textcolor{blue}{Miami Downtown}, \textcolor{yellow!90}{Industrial Layoff}, \textcolor{red!90}{Tropical Storm Fay}) and 
(\textcolor{blue}{Miami Downtown}, \textcolor{yellow!90}{Monetary loss}, \textcolor{red!90}{Tropical Storm Fay}) to get more relevant documents.
\begin{figure}[H]
\centering  \includegraphics[width=0.98\columnwidth]{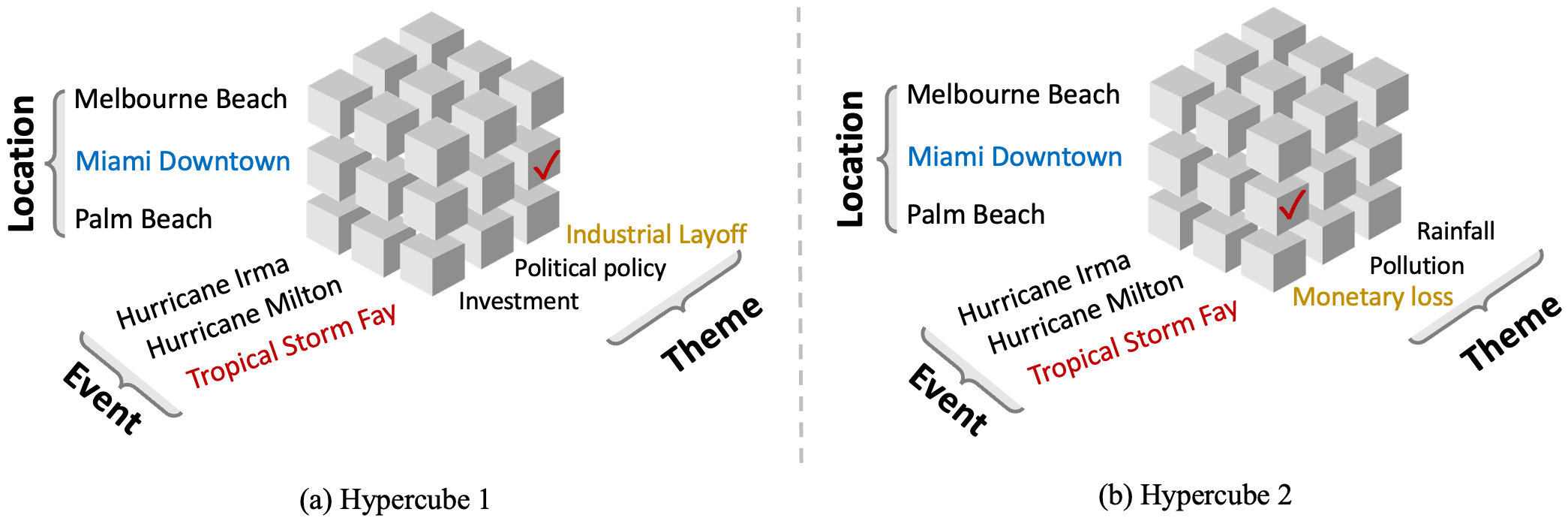}
  \caption{Access multiple cube cells in multiple hypercubes. \textcolor{red!90}{\bf \checkmark} represents the touched cube cells. } 
  \label{fig:two_cubes}
\end{figure}

\subsection{Why did we use in-domain scientific QA datasets rather than general ones?}
\label{sec:why_theme}
According to the above discussion, hypercube construction is sensitive to themes in various domains. 
While similar or related themes are allowed to be included in a hypercube, each theme-specific dataset should have a separate hypercube to maximize its power.
However, the general QA datasets, e.g., PopQA \cite{mallen2022not} and NQ \cite{kwiatkowski2019natural}, typically present articles across multiple themes.
The process of constructing multiple hypercubes will require intensive efforts (e.g., design hypercube schema for different datasets.), which is out of the scope of this study.
This is why we particularly focus on in-domain theme-specific QA datasets.

\subsection{Future work}
For general questions consisting of multiple and different themes, we plan to use large language models (LLMs) to dynamically construct multiple hypercubes. Each hypercube is used to represent documents with one specific theme, respectively. The possible related work is TeleClass \cite{zhang2025teleclass}, and TKGCon \cite{ding2024automated}.
On the other hand, we aim to explore the query-augmented technique to enhance the robustness of retrieval systems \cite{shen2024retrieval,li2022generation}.
At last, we would like to test \hyperrag on the larger corpus for industrial applications.

\section{Conclusions}
In this chapter, we start by analyzing the strengths and limitations of existing RAG methods for in-domain \emph{theme-specific} applications. 
To retain their advantages while addressing key challenges, we propose \hyperrag, innovating retrieval based on fine-grained labels with a multidimensional structure (i.e., hypercube).
The label-based retrieval integrates sparse lexical and dense semantic strategies, making it accurate, efficient, and inherently explainable.
Empirical results demonstrate that our method consistently outperforms existing RAG baselines in accuracy and efficiency, while providing interpretable provenance.
We propose future plans to generalize the \hyperrag to open-source domains and explore powerful query-augmented techniques to boost the robustness of retrieval systems.

\chapter{CONCLUSIONS}	
\label{sec:conc}

This dissertation advances the application of artificial intelligence (AI) in environmental science, aiming to achieve \emph{environmental intelligence}. 
We explored three key problems at the intersection of AI and environmental science and offer practical and high-performing AI-based solutions for them.

\section{DL-based Flood Prediction and Management}
In Chapter \ref{sec:fidlar}, we developed \walef for forecasting water levels and \fidlar, a forecast-informed DL-based architecture for managing water levels in coastal river systems, integrating diverse inputs such as rainfall, sea levels, historical water levels, and settings of hydraulic structures. Our contributions are summarized as follows:
\begin{itemize}
    \item Our graph-transformer-based model (\walef) for forecasting water levels exhibits superior performance when compared to baseline methods.
    \item \fidlar is a data-driven approach, learning flood mitigation strategies from historically observed data.
    Once trained, it offers rapid response capabilities, highlighting the advantages of DL-based models over physics-based models, particularly for real-time flood management.
    \item \fidlar seamlessly combines the two DL models in series: \texttt{Flood Manager} and \texttt{Flood Evaluator}.
    The former model is responsible for generating water pre-release schedules, while the latter model accurately forecasts the resulting water levels. Moreover, with the gradient-based planning, and differentiability of trained \texttt{Evaluator}, it can reinforce the \texttt{Manager} to generate better schedules.
    \item \fidlar is a model-agnostic framework, where both the \texttt{Manager} and \texttt{Evaluator} could be any type of DL model trained with differentiable loss functions that allow back-propagation.
    \item Demonstrated on flood-prone regions in South Florida, \fidlar offers a practical solution for real-time flood mitigation, outperforming traditional baselines in accuracy, efficiency, and interpretability.
\end{itemize}

\section{Diffusion Model for Probabilistic Weather Forecasting}
In Chapter \ref{sec:codicast}, we proposed \codicast, a conditional diffusion model repurposed from generative AI to enable accurate and efficient global weather prediction with explicit uncertainty quantification. Unlike deterministic physics-based methods, \codicast captures the stochastic nature of weather, making it suitable for anticipating extreme events and supporting informed risk assessments.
More specifically, the contributions of the chapter on \codicast are as follows:
\begin{itemize}
    \item Existing weather prediction methods are computationally intensive and often make simplifying assumptions. Our tool, \codicast, is efficient and avoids the assumptions.
    \item \codicast repurposes a ``generative'' tool as a ``predictive'' tool. It is a conditional diffusion model for global weather prediction conditioning on observations from the recent past.
    \item \codicast probabilistically models the uncertainty by generating multiple plausible weather scenarios.
    \item With ERA5 data from the European Centre for Medium-Range Weather Forecasts (ECMWF), we show that \codicast achieves good performance without suffering from the tradeoff between accuracy, efficiency, and uncertainty, while providing explainability for its answers.
\end{itemize}

\section{Question-Answering Systems for Environmental Knowledge}
In Chapter \ref{sec:hyperrag}, we focus on the knowledge retrieval from external databases to enhance the capabilities of pre-trained LLMs while answering \emph{in-domain} questions.
Our contributions are three-fold:
\begin{itemize}
    \item We analyze the limitations of current LLM-based question-answering systems, which suffer from a tradeoff between accuracy, efficiency, and explainability. 
    \item Our \hyperrag, a retrieval-augmented generation (RAG) system, is built on a multidimensional knowledge structure \emph{text cube}. 
    By allocating unstructured documents to a structured text cube based on the document labels along pre-defined dimensions, \hyperrag can perform information retrieval using fine-grained document labels organized within the cube.
    \item Experiments with scientific queries on the topics of hurricanes, aging dams, and other hazards, we show that \hyperrag enhances existing LLMs with improved accuracy, interpretability, and retrieval efficiency. \\
\end{itemize}

In summary, the contributions in this dissertation demonstrate how a diverse set of AI techniques - spanning deep learning architectures, generative diffusion models, and structured retrieval systems - can be tailored to meet the multifaceted challenges of environmental science. 
This work underscores the growing importance of \emph{Environmental Intelligence}, where advanced AI methods empower data-driven, precise, efficient, and interpretable strategies to tackle some pressing environmental problems.

\section{Future Work}
This dissertation develops and applies deep learning strategies to address key challenges in environmental science, with a focus on improving accuracy, efficiency, and explainability.
We also highlight several natural extensions that can further advance the toolkit for achieving environmental intelligence.

\subsection{Scalability}
In Chapter \ref{sec:fidlar}, although the study incorporates heterogeneous driving factors for floods, the focus remains on a regional coastal river system in South Florida. 
Extending the \fidlar methodology to more complex and large-scale river systems would be a valuable direction for future research.
In Chapter \ref{sec:codicast}, we focus on six meteorological variables at a coarse global resolution due to limited computational resources. However, accurate weather forecasting would benefit from incorporating additional atmospheric variables at higher spatial and temporal resolutions. Therefore, scaling up the dataset using more powerful computing resources represents a promising direction for future work.
Finally, in Chapter \ref{sec:hyperrag}, it would be challenging to work with a much larger body of knowledge to understand the limits of the methods used.

\subsection{Acceleration of Diffusion Models}
In Chapter \ref{sec:codicast}, while diffusion models have shown great promise in generative and probabilistic tasks, their iterative denoising process often requires hundreds or thousands of steps, posing a significant computational challenge, especially for time-sensitive applications like weather forecasting. 
To address this, recent research has focused on accelerating diffusion models through compressing the iterative process into fewer steps (e.g., 1-step or 4-step models), which enables faster sampling without compromising accuracy. 
For weather prediction, where high-resolution, multi-variable forecasts must be delivered in real time, such acceleration strategies are crucial. By integrating fast sampling techniques and optimized inference pipelines, diffusion-based models can become both a probabilistic and practical alternative to traditional numerical weather prediction systems, offering explicit uncertainty quantification with tractable computational costs. Continued research into efficient architectures and hardware-aware implementations will further enhance the viability of diffusion models for operational forecasting.

\subsection{Flexibility}
In Chapter \ref{sec:hyperrag}, our proposed \hyperrag framework is constructed atop a structured text cube defined by human-curated dimensions.
While this design proves highly effective for in-domain, theme-specific applications, it lacks adaptability when the underlying knowledge base evolves or when applied to new domains with different structures.
A promising direction for future work is to develop methods for automated or adaptive hypercube construction. This could involve leveraging large language models (LLMs), unsupervised clustering techniques, or reinforcement learning to dynamically determine optimal dimensions and hierarchies based on the corpus content.

\section{Endless Possibilities}
Advancements that once belonged to the realm of science fiction are becoming increasingly attainable. The future of \emph{AI for Science} holds vast potential, especially with the rapid emergence and evolution of large language models (LLMs) and large multimodal models (LMMs).

\bibliographystyle{alpha}
\bibliography{references}

\newcommand{\etalchar}[1]{$^{#1}$}
\begin{thebibliography}{SMRFK{\etalchar{+}}20}

\bibitem[AAA{\etalchar{+}}23]{achiam2023gpt}
Josh Achiam, Steven Adler, Sandhini Agarwal, Lama Ahmad, Ilge Akkaya, Florencia~Leoni Aleman, Diogo Almeida, Janko Altenschmidt, Sam Altman, Shyamal Anadkat, et~al.
\newblock {GPT}-4 technical report.
\newblock {\em arXiv preprint arXiv:2303.08774}, 2023.

\bibitem[ABL{\etalchar{+}}24]{alexe2024graphdop}
Mihai Alexe, Eulalie Boucher, Peter Lean, Ewan Pinnington, Patrick Laloyaux, Anthony McNally, Simon Lang, Matthew Chantry, Chris Burrows, Marcin Chrust, et~al.
\newblock Graph{DOP}: {T}owards skilful data-driven medium-range weather forecasts learnt and initialised directly from observations.
\newblock {\em arXiv preprint arXiv:2412.15687}, 2024.

\bibitem[AEk{\etalchar{+}}24]{ansarifard2024simulation}
Sara Ansarifard, Morteza Eyvazi, Mahsa kalantari, Behrooz mohseni, Mahdi Ghorbanifard, Hadi~Jafakesh Moghaddam, and Maryam Nouri.
\newblock Simulation of floods under the influence of effective factors in hydraulic and hydrological models using {HEC-RAS} and {MIKE} 21.
\newblock {\em Discover Water}, 4(1):92, 2024.

\bibitem[AEL{\etalchar{+}}23]{andrychowicz2023deep}
Marcin Andrychowicz, Lasse Espeholt, Di~Li, Samier Merchant, Alexander Merose, Fred Zyda, Shreya Agrawal, and Nal Kalchbrenner.
\newblock Deep learning for day forecasts from sparse observations, 2023.

\bibitem[AFK{\etalchar{+}}14]{ali2014data}
Peshawa Jamal~Muhammad Ali, Rezhna~Hassan Faraj, Erbil Koya, Peshawa J~Muhammad Ali, and Rezhna~H Faraj.
\newblock Data normalization and standardization: a technical report.
\newblock {\em Mach Learn Tech Rep}, 1(1):1--6, 2014.

\bibitem[Ala21]{alam2021possibilities}
Ashraf Alam.
\newblock Possibilities and apprehensions in the landscape of artificial intelligence in education.
\newblock In {\em 2021 International Conference on Computational Intelligence and Computing Applications (ICCICA)}, pages 1--8. IEEE, 2021.

\bibitem[ALOL24]{andrae2024continuous}
Martin Andrae, Tomas Landelius, Joel Oskarsson, and Fredrik Lindsten.
\newblock Continuous ensemble weather forecasting with diffusion models.
\newblock {\em arXiv preprint arXiv:2410.05431}, 2024.

\bibitem[AMP{\etalchar{+}}23]{asperti2023precipitation}
Andrea Asperti, Fabio Merizzi, Alberto Paparella, Giorgio Pedrazzi, Matteo Angelinelli, and Stefano Colamonaco.
\newblock Precipitation nowcasting with generative diffusion models.
\newblock {\em arXiv preprint arXiv:2308.06733}, 2023.

\bibitem[AN24]{alotaibi2024artificial}
Emran Alotaibi and Nadia Nassif.
\newblock Artificial intelligence in environmental monitoring: in-depth analysis.
\newblock {\em Discover Artificial Intelligence}, 4(1):84, 2024.

\bibitem[ANG23]{ayus2023prediction}
Ishan Ayus, Narayanan Natarajan, and Deepak Gupta.
\newblock Prediction of water level using machine learning and deep learning techniques.
\newblock {\em Iranian Journal of Science and Technology, Transactions of Civil Engineering}, 47(4):2437--2447, 2023.

\bibitem[AS20]{angelov2020towards}
Plamen Angelov and Eduardo Soares.
\newblock Towards explainable deep neural networks (x{DNN}).
\newblock {\em Neural Networks}, 130:185--194, 2020.

\bibitem[AWW{\etalchar{+}}23]{asai2023self}
Akari Asai, Zeqiu Wu, Yizhong Wang, Avirup Sil, and Hannaneh Hajishirzi.
\newblock Self-rag: {L}earning to retrieve, generate, and critique through self-reflection.
\newblock In {\em The Twelfth International Conference on Learning Representations}, 2023.

\bibitem[AYCG10]{al2010review}
Sultan Al-Yahyai, Yassine Charabi, and Adel Gastli.
\newblock Review of the use of numerical weather prediction ({NWP}) models for wind energy assessment.
\newblock {\em Renewable and Sustainable Energy Reviews}, 14(9):3192--3198, 2010.

\bibitem[AZH{\etalchar{+}}21]{alzubaidi2021review}
Laith Alzubaidi, Jinglan Zhang, Amjad~J Humaidi, Ayad Al-Dujaili, Ye~Duan, Omran Al-Shamma, Jos{\'e} Santamar{\'\i}a, Mohammed~A Fadhel, Muthana Al-Amidie, and Laith Farhan.
\newblock Review of deep learning: concepts, {CNN} architectures, challenges, applications, future directions.
\newblock {\em Journal of big Data}, 8:1--74, 2021.

\bibitem[Bal12]{baldi2012autoencoders}
Pierre Baldi.
\newblock Autoencoders, unsupervised learning, and deep architectures.
\newblock In {\em Proceedings of ICML workshop on unsupervised and transfer learning}, pages 37--49, Edinburgh, Scotland, 2012. JMLR Workshop and Conference Proceedings.

\bibitem[BAML21]{bojesomo2021spatiotemporal}
Alabi Bojesomo, Hasan Al-Marzouqi, and Panos Liatsis.
\newblock Spatiotemporal vision transformer for short time weather forecasting.
\newblock In {\em 2021 IEEE International Conference on Big Data (Big Data)}, pages 5741--5746. IEEE, 2021.

\bibitem[BBCM{\etalchar{+}}24]{ben2024rise}
Zied Ben~Bouall{\`e}gue, Mariana~CA Clare, Linus Magnusson, Estibaliz Gascon, Michael Maier-Gerber, Martin Janou{\v{s}}ek, Mark Rodwell, Florian Pinault, Jesper~S Dramsch, Simon~TK Lang, et~al.
\newblock The rise of data-driven weather forecasting: {A} first statistical assessment of machine learning--based weather forecasts in an operational-like context.
\newblock {\em Bulletin of the American Meteorological Society}, 105(6):E864--E883, 2024.

\bibitem[BBL{\etalchar{+}}24]{bodnar2024aurora}
Cristian Bodnar, Wessel~P Bruinsma, Ana Lucic, Megan Stanley, Johannes Brandstetter, Patrick Garvan, Maik Riechert, Jonathan Weyn, Haiyu Dong, Anna Vaughan, et~al.
\newblock Aurora: {A} foundation model of the atmosphere.
\newblock {\em arXiv preprint arXiv:2405.13063}, 2024.

\bibitem[BBO17]{borovykh2017conditional}
Anastasia Borovykh, Sander Bohte, and Cornelis~W Oosterlee.
\newblock Conditional time series forecasting with convolutional neural networks.
\newblock {\em arXiv preprint arXiv:1703.04691}, 2017.

\bibitem[BCD{\etalchar{+}}22]{balaji2022general}
V~Balaji, Fleur Couvreux, Julie Deshayes, Jacques Gautrais, Fr{\'e}d{\'e}ric Hourdin, and Catherine Rio.
\newblock Are general circulation models obsolete?
\newblock {\em Proceedings of the National Academy of Sciences}, 119(47):e2202075119, 2022.

\bibitem[BCK{\etalchar{+}}16]{bisht2016modeling}
Deepak~Singh Bisht, Chandranath Chatterjee, Shivani Kalakoti, Pawan Upadhyay, Manaswinee Sahoo, and Ambarnil Panda.
\newblock Modeling urban floods and drainage using {SWMM} and {MIKE URBAN}: a case study.
\newblock {\em Natural hazards}, 84:749--776, 2016.

\bibitem[BD22]{busto2022staggered}
Saray Busto and Michael Dumbser.
\newblock A staggered semi-implicit hybrid finite volume/finite element scheme for the shallow water equations at all {F}roude numbers.
\newblock {\em Applied Numerical Mathematics}, 175:108--132, 2022.

\bibitem[BGE18]{babaei2018urban}
Sahar Babaei, Reza Ghazavi, and Mahdi Erfanian.
\newblock Urban flood simulation and prioritization of critical urban sub-catchments using {SWMM} model and {PROMETHEE II} approach.
\newblock {\em Physics and Chemistry of the Earth, Parts A/B/C}, 105:3--11, 2018.

\bibitem[BHQL24]{bulte2024uncertainty}
Christopher B{\"u}lte, Nina Horat, Julian Quinting, and Sebastian Lerch.
\newblock Uncertainty quantification for data-driven weather models.
\newblock {\em arXiv preprint arXiv:2403.13458}, 2024.

\bibitem[BIJT22]{bentivoglio2022deep}
Roberto Bentivoglio, Elvin Isufi, Sebastian~Nicolaas Jonkman, and Riccardo Taormina.
\newblock Deep learning methods for flood mapping: a review of existing applications and future research directions.
\newblock {\em Hydrology and Earth System Sciences Discussions}, 2022:1--50, 2022.

\bibitem[BKC{\etalchar{+}}22]{brempong2022denoising}
Emmanuel~Asiedu Brempong, Simon Kornblith, Ting Chen, Niki Parmar, Matthias Minderer, and Mohammad Norouzi.
\newblock Denoising pretraining for semantic segmentation.
\newblock In {\em Proceedings of the IEEE/CVF conference on computer vision and pattern recognition}, pages 4175--4186, 2022.

\bibitem[BKK18]{bai2018empirical}
Shaojie Bai, J~Zico Kolter, and Vladlen Koltun.
\newblock An empirical evaluation of generic convolutional and recurrent networks for sequence modeling.
\newblock {\em arXiv preprint arXiv:1803.01271}, 2018.

\bibitem[BMR{\etalchar{+}}20]{brown2020language}
Tom~B Brown, Benjamin Mann, Nick Ryder, Melanie Subbiah, Jared Kaplan, Prafulla Dhariwal, Arvind Neelakantan, Pranav Shyam, Girish Sastry, Amanda Askell, et~al.
\newblock Language models are few-shot learners.
\newblock {\em Advances in Neural Information Processing Systems (NeurIPS)}, 33:1877--1901, 2020.

\bibitem[Bre01]{breiman2001random}
Leo Breiman.
\newblock Random forests.
\newblock {\em Machine learning}, 45:5--32, 2001.

\bibitem[BSZ{\etalchar{+}}22]{bai2022rainformer}
Cong Bai, Feng Sun, Jinglin Zhang, Yi~Song, and Shengyong Chen.
\newblock Rainformer: {F}eatures extraction balanced network for radar-based precipitation nowcasting.
\newblock {\em IEEE Geoscience and Remote Sensing Letters}, 19:1--5, 2022.

\bibitem[BTB15]{bauer2015quiet}
Peter Bauer, Alan Thorpe, and Gilbert Brunet.
\newblock The quiet revolution of numerical weather prediction.
\newblock {\em Nature}, 525(7567):47--55, 2015.

\bibitem[BTW{\etalchar{+}}21]{bowes2021flood}
Benjamin~D Bowes, Arash Tavakoli, Cheng Wang, Arsalan Heydarian, Madhur Behl, Peter~A Beling, and Jonathan~L Goodall.
\newblock Flood mitigation in coastal urban catchments using real-time stormwater infrastructure control and reinforcement learning.
\newblock {\em Journal of Hydroinformatics}, 23(3):529--547, 2021.

\bibitem[Bui08]{buizza2008comparison}
Roberto Buizza.
\newblock Comparison of a 51-member low-resolution ({T L 399L62}) ensemble with a 6-member high-resolution ({T L 799L91}) lagged-forecast ensemble.
\newblock {\em Monthly weather review}, 136(9):3343--3362, 2008.

\bibitem[BXZ{\etalchar{+}}23]{bi2023accurate}
Kaifeng Bi, Lingxi Xie, Hengheng Zhang, Xin Chen, Xiaotao Gu, and Qi~Tian.
\newblock Accurate medium-range global weather forecasting with 3{D} neural networks.
\newblock {\em Nature}, 619(7970):533--538, 2023.

\bibitem[CCZ{\etalchar{+}}23]{chen2023artificial}
Lin Chen, Zhonghao Chen, Yubing Zhang, Yunfei Liu, Ahmed~I Osman, Mohamed Farghali, Jianmin Hua, Ahmed Al-Fatesh, Ikko Ihara, David~W Rooney, et~al.
\newblock Artificial intelligence-based solutions for climate change: a review.
\newblock {\em Environmental Chemistry Letters}, 21(5):2525--2557, 2023.

\bibitem[CDC{\etalchar{+}}22]{choudhary2022recent}
Kamal Choudhary, Brian DeCost, Chi Chen, Anubhav Jain, Francesca Tavazza, Ryan Cohn, Cheol~Woo Park, Alok Choudhary, Ankit Agrawal, Simon~JL Billinge, et~al.
\newblock Recent advances and applications of deep learning methods in materials science.
\newblock {\em npj Computational Materials}, 8(1):59, 2022.

\bibitem[CDH{\etalchar{+}}23]{chen2023swinrdm}
Lei Chen, Fei Du, Yuan Hu, Zhibin Wang, and Fan Wang.
\newblock Swin{RDM}: integrate {S}win{RNN} with diffusion model towards high-resolution and high-quality weather forecasting.
\newblock In {\em Proceedings of the AAAI Conference on Artificial Intelligence}, pages 322--330, 2023.

\bibitem[CGPL{\etalchar{+}}14]{castro2014flood}
Mario~E Castro-Gama, Ioana Popescu, Shengyang Li, Arthur Mynett, and Arthur van Dam.
\newblock Flood inference simulation using surrogate modelling for the {Y}ellow river multiple reservoir system.
\newblock {\em Environmental modelling \& software}, 55:250--265, 2014.

\bibitem[CHG{\etalchar{+}}23]{chen2023fengwu}
Kang Chen, Tao Han, Junchao Gong, Lei Bai, Fenghua Ling, Jing-Jia Luo, Xi~Chen, Leiming Ma, Tianning Zhang, Rui Su, et~al.
\newblock Fengwu: {P}ushing the skillful global medium-range weather forecast beyond 10 days lead.
\newblock {\em arXiv preprint arXiv:2304.02948}, 2023.

\bibitem[CHIS23]{croitoru2023diffusion}
Florinel-Alin Croitoru, Vlad Hondru, Radu~Tudor Ionescu, and Mubarak Shah.
\newblock Diffusion models in vision: {A} survey.
\newblock {\em IEEE Transactions on Pattern Analysis and Machine Intelligence}, 2023.

\bibitem[CKK{\etalchar{+}}23]{choi2023pct}
Jaeho Choi, Yura Kim, Kwang-Ho Kim, Sung-Hwa Jung, and Ikhyun Cho.
\newblock {PCT}-{C}ycle{GAN}: {P}aired complementary temporal cycle-consistent adversarial networks for radar-based precipitation nowcasting.
\newblock In {\em Proceedings of the 32nd ACM International Conference on Information and Knowledge Management}, pages 348--358, 2023.

\bibitem[CLGH16]{chen2016dimension}
Duan Chen, Arturo~S Leon, Nathan~L Gibson, and Parnian Hosseini.
\newblock Dimension reduction of decision variables for multireservoir operation: {A} spectral optimization model.
\newblock {\em Water Resources Research}, 52(1):36--51, 2016.

\bibitem[CLJ{\etalchar{+}}23]{chen2023foundation}
Shengchao Chen, Guodong Long, Jing Jiang, Dikai Liu, and Chengqi Zhang.
\newblock Foundation models for weather and climate data understanding: A comprehensive survey.
\newblock {\em arXiv preprint arXiv:2312.03014}, 2023.

\bibitem[CMG{\etalchar{+}}19]{choubin2019ensemble}
Bahram Choubin, Ehsan Moradi, Mohammad Golshan, Jan Adamowski, Farzaneh Sajedi-Hosseini, and Amir Mosavi.
\newblock An ensemble prediction of flood susceptibility using multivariate discriminant analysis, classification and regression trees, and support vector machines.
\newblock {\em Science of the Total Environment}, 651:2087--2096, 2019.

\bibitem[CMH{\etalchar{+}}22]{chattopadhyay2022towards}
Ashesh Chattopadhyay, Mustafa Mustafa, Pedram Hassanzadeh, Eviatar Bach, and Karthik Kashinath.
\newblock Towards physics-inspired data-driven weather forecasting: integrating data assimilation with a deep spatial-transformer-based {U-NET} in a case study with {ERA5}.
\newblock {\em Geoscientific Model Development}, 15(5):2221--2237, 2022.

\bibitem[Coi11]{coiffier2011fundamentals}
Jean Coiffier.
\newblock {\em Fundamentals of numerical weather prediction}.
\newblock Cambridge University Press, 2011.

\bibitem[CRBD18]{chen2018neural}
Ricky~TQ Chen, Yulia Rubanova, Jesse Bettencourt, and David~K Duvenaud.
\newblock Neural ordinary differential equations.
\newblock {\em Advances in neural information processing systems}, 31, 2018.

\bibitem[CSC{\etalchar{+}}24]{couairon2024archesweather}
Guillaume Couairon, Renu Singh, Anastase Charantonis, Christian Lessig, and Claire Monteleoni.
\newblock Arches{W}eather \& {A}rchesweathergen: a deterministic and generative model for efficient {ML} weather forecasting.
\newblock {\em arXiv preprint arXiv:2412.12971}, 2024.

\bibitem[CTCO19]{canizo2019multi}
Mikel Canizo, Isaac Triguero, Angel Conde, and Enrique Onieva.
\newblock Multi-head {CNN--RNN} for multi-time series anomaly detection: {A}n industrial case study.
\newblock {\em Neurocomputing}, 363:246--260, 2019.

\bibitem[CUC{\etalchar{+}}25]{cui2025timer}
Hejie Cui, Alyssa Unell, Bowen Chen, Jason~Alan Fries, Emily Alsentzer, Sanmi Koyejo, and Nigam Shah.
\newblock {TIMER}: {T}emporal instruction modeling and evaluation for longitudinal clinical records.
\newblock In {\em Will Synthetic Data Finally Solve the Data Access Problem?}, 2025.

\bibitem[CWHT24]{chen2024coupling}
Yutong Chen, Ya~Wang, Gang Huang, and Qun Tian.
\newblock Coupling physical factors for precipitation forecast in {C}hina with graph neural network.
\newblock {\em Geophysical Research Letters}, 51(2):e2023GL106676, 2024.

\bibitem[CXL{\etalchar{+}}23]{chen2023physics}
Shengyu Chen, Yiqun Xie, Xiang Li, Xu~Liang, and Xiaowei Jia.
\newblock Physics-guided meta-learning method in baseflow prediction over large regions.
\newblock In {\em Proceedings of the 2023 SIAM International Conference on Data Mining (SDM)}, pages 217--225. SIAM, 2023.

\bibitem[CZZ{\etalchar{+}}23]{chen2023fuxi}
Lei Chen, Xiaohui Zhong, Feng Zhang, Yuan Cheng, Yinghui Xu, Yuan Qi, and Hao Li.
\newblock Fu{X}i: {A} cascade machine learning forecasting system for 15-day global weather forecast.
\newblock {\em npj Climate and Atmospheric Science}, 6(1):190, 2023.

\bibitem[DBPG19]{de2019deep}
Emmanuel De~B{\'e}zenac, Arthur Pajot, and Patrick Gallinari.
\newblock Deep learning for physical processes: {I}ncorporating prior scientific knowledge.
\newblock {\em Journal of Statistical Mechanics: Theory and Experiment}, 2019(12):124009, 2019.

\bibitem[DCLT19a]{devlin2018bert}
Jacob Devlin, Ming-Wei Chang, Kenton Lee, and Kristina Toutanova.
\newblock {BERT}: {P}re-training of deep bidirectional transformers for language understanding.
\newblock In {\em Proceedings of the 2019 Conference of the North American Chapter of the Association for Computational Linguistics (NAACL)}, pages 4171--4186, 2019.

\bibitem[DCLT19b]{devlin2019bert}
Jacob Devlin, Ming-Wei Chang, Kenton Lee, and Kristina Toutanova.
\newblock Bert: {P}re-training of deep bidirectional transformers for language understanding.
\newblock In {\em Proceedings of the 2019 conference of the North American chapter of the association for computational linguistics: human language technologies, volume 1 (long and short papers)}, pages 4171--4186, 2019.

\bibitem[DCNV{\etalchar{+}}21]{del2021auto}
Felipe~Arias Del~Campo, Mar{\'\i}a Cristina~Guevara Neri, Osslan Osiris~Vergara Villegas, Vianey Guadalupe~Cruz S{\'a}nchez, Humberto de Jes{\'u}s~Ochoa Dom{\'\i}nguez, and Vicente~Garc{\'\i}a Jim{\'e}nez.
\newblock Auto-adaptive multilayer perceptron for univariate time series classification.
\newblock {\em Expert Systems with Applications}, 181:115147, 2021.

\bibitem[DDBB24]{debbarma2024simulation}
Sagar Debbarma, Subhajit Dey, Arnab Bandyopadhyay, and Aditi Bhadra.
\newblock Simulation of flood inundation extent by integration of {HEC-HMS}, {GA}-based rating curve and cost distance analysis.
\newblock {\em Water Resources Management}, pages 1--21, 2024.

\bibitem[DED19]{duan2019artificial}
Yanqing Duan, John~S Edwards, and Yogesh~K Dwivedi.
\newblock Artificial intelligence for decision making in the era of big data--evolution, challenges and research agenda.
\newblock {\em International journal of information management}, 48:63--71, 2019.

\bibitem[DHM{\etalchar{+}}20]{delaney2020forecast}
Chris~J Delaney, Robert~K Hartman, John Mendoza, Michael Dettinger, Luca Delle~Monache, Jay Jasperse, F~Martin Ralph, Cary Talbot, James Brown, David Reynolds, et~al.
\newblock Forecast informed reservoir operations using ensemble streamflow predictions for a multipurpose reservoir in {N}orthern {C}alifornia.
\newblock {\em Water Resources Research}, 56(9):e2019WR026604, 2020.

\bibitem[Dis23]{dbhydro2023sfwmd}
South Florida Water~Management District.
\newblock {DBHYDRO} of {S}outh {F}lorida {W}ater {M}anagement {D}istrict.
\newblock \url{https://www.sfwmd.gov/science-data/dbhydro}, 2023.

\bibitem[DKA19]{dasallas2019case}
Lea Dasallas, Yeonsu Kim, and Hyunuk An.
\newblock Case study of {HEC-RAS} 1{D}--2{D} coupling simulation: 2002 {B}aeksan flood event in {K}orea.
\newblock {\em Water}, 11(10):2048, 2019.

\bibitem[DLYC21]{ding2021random}
Xiaojian Ding, Jian Liu, Fan Yang, and Jie Cao.
\newblock Random radial basis function kernel-based support vector machine.
\newblock {\em Journal of the Franklin Institute}, 358(18):10121--10140, 2021.

\bibitem[DN21]{devezas2021review}
Jos{\'e} Devezas and S{\'e}rgio Nunes.
\newblock A review of graph-based models for entity-oriented search.
\newblock {\em SN Computer Science}, 2(6):437, 2021.

\bibitem[DZXH24]{ding2024automated}
Linyi Ding, Sizhe Zhou, Jinfeng Xiao, and Jiawei Han.
\newblock Automated construction of theme-specific knowledge graphs.
\newblock {\em arXiv preprint arXiv:2404.19146}, 2024.

\bibitem[EAS{\etalchar{+}}22]{espeholt2022deep}
Lasse Espeholt, Shreya Agrawal, Casper S{\o}nderby, Manoj Kumar, Jonathan Heek, Carla Bromberg, Cenk Gazen, Rob Carver, Marcin Andrychowicz, Jason Hickey, et~al.
\newblock Deep learning for twelve hour precipitation forecasts.
\newblock {\em Nature communications}, 13(1):1--10, 2022.

\bibitem[{ECM}]{ecmwf}
{ECMWF}.
\newblock \url{https://www.ecmwf.int/en/forecasts/documentation-and-support/medium-range-forecasts}.

\bibitem[ECM23]{HRES2023ECMWF}
ECMWF.
\newblock Medium-range forecasts, 2023.

\bibitem[EH24]{enis2024llm}
Maxim Enis and Mark Hopkins.
\newblock From llm to nmt: {A}dvancing low-resource machine translation with claude.
\newblock {\em arXiv preprint arXiv:2404.13813}, 2024.

\bibitem[EMSK97]{esposito1997comparative}
Floriana Esposito, Donato Malerba, Giovanni Semeraro, and J~Kay.
\newblock A comparative analysis of methods for pruning decision trees.
\newblock {\em IEEE transactions on pattern analysis and machine intelligence}, 19(5):476--491, 1997.

\bibitem[ERKL16]{erfani2016high}
Sarah~M Erfani, Sutharshan Rajasegarar, Shanika Karunasekera, and Christopher Leckie.
\newblock High-dimensional and large-scale anomaly detection using a linear one-class {SVM} with deep learning.
\newblock {\em Pattern Recognition}, 58:121--134, 2016.

\bibitem[ETC{\etalchar{+}}24]{edge2024local}
Darren Edge, Ha~Trinh, Newman Cheng, Joshua Bradley, Alex Chao, Apurva Mody, Steven Truitt, Dasha Metropolitansky, Robert~Osazuwa Ness, and Jonathan Larson.
\newblock From local to global: {A} graph rag approach to query-focused summarization.
\newblock {\em arXiv preprint arXiv:2404.16130}, 2024.

\bibitem[EWN{\etalchar{+}}23]{eshraghian2023training}
Jason~K Eshraghian, Max Ward, Emre~O Neftci, Xinxin Wang, Gregor Lenz, Girish Dwivedi, Mohammed Bennamoun, Doo~Seok Jeong, and Wei~D Lu.
\newblock Training spiking neural networks using lessons from deep learning.
\newblock {\em Proceedings of the IEEE}, 2023.

\bibitem[EXES{\etalchar{+}}21]{elmorshedy2021recent}
Mahmoud~F Elmorshedy, Wei Xu, Fayez~FM El-Sousy, Md~Rabiul Islam, and Abdelsalam~A Ahmed.
\newblock Recent achievements in model predictive control techniques for industrial motor: A comprehensive state-of-the-art.
\newblock {\em IEEE Access}, 9:58170--58191, 2021.

\bibitem[FWWN23]{fraehr2023development}
Niels Fraehr, Quan~J Wang, Wenyan Wu, and Rory Nathan.
\newblock Development of a fast and accurate hybrid model for floodplain inundation simulations.
\newblock {\em Water Resources Research}, 59(6):e2022WR033836, 2023.

\bibitem[FZZ{\etalchar{+}}15]{feng2015streamcube}
Wei Feng, Chao Zhang, Wei Zhang, Jiawei Han, Jianyong Wang, Charu Aggarwal, and Jianbin Huang.
\newblock {STREAMCUBE}: {H}ierarchical spatio-temporal hashtag clustering for event exploration over the {T}witter stream.
\newblock In {\em 2015 IEEE 31st international conference on data engineering}, pages 1561--1572. IEEE, 2015.

\bibitem[GAC{\etalchar{+}}22]{guo2022machine}
Shenghan Guo, Mohit Agarwal, Clayton Cooper, Qi~Tian, Robert~X Gao, Weihong~Guo Grace, and YB~Guo.
\newblock Machine learning for metal additive manufacturing: {T}owards a physics-informed data-driven paradigm.
\newblock {\em Journal of Manufacturing Systems}, 62:145--163, 2022.

\bibitem[GBY{\etalchar{+}}24]{gong2024cascast}
Junchao Gong, Lei Bai, Peng Ye, Wanghan Xu, Na~Liu, Jianhua Dai, Xiaokang Yang, and Wanli Ouyang.
\newblock Cas{C}ast: {S}killful high-resolution precipitation nowcasting via cascaded modelling.
\newblock {\em arXiv preprint arXiv:2402.04290}, 2024.

\bibitem[GDS19]{ghosh2019study}
Sourish Ghosh, Anasuya Dasgupta, and Aleena Swetapadma.
\newblock A study on support vector machine based linear and non-linear pattern classification.
\newblock In {\em 2019 International conference on intelligent sustainable systems (ICISS)}, pages 24--28. IEEE, 2019.

\bibitem[GG16]{gal2016dropout}
Yarin Gal and Zoubin Ghahramani.
\newblock Dropout as a bayesian approximation: {R}epresenting model uncertainty in deep learning.
\newblock In {\em international conference on machine learning}, pages 1050--1059. PMLR, 2016.

\bibitem[GJRO{\etalchar{+}}23]{gomes2023modeling}
Marcus~N Gomes~Jr, Luis~MC R{\'a}palo, Paulo~TS Oliveira, Marcio~H Giacomoni, C{\'e}sar~AF do~Lago, and Eduardo~M Mendiondo.
\newblock Modeling unsteady and steady 1{D} hydrodynamics under different hydraulic conceptualizations: {M}odel/{S}oftware development and case studies.
\newblock {\em Environmental Modelling \& Software}, 167:105733, 2023.

\bibitem[GK18]{gidea2018topological}
Marian Gidea and Yuri Katz.
\newblock Topological data analysis of financial time series: Landscapes of crashes.
\newblock {\em Physica A: Statistical Mechanics and its Applications}, 491:820--834, 2018.

\bibitem[GKBG23]{guo2023digital}
Yuebin Guo, Andreas Klink, Paulo Bartolo, and Weihong~Grace Guo.
\newblock Digital twins for electro-physical, chemical, and photonic processes.
\newblock {\em CIRP Annals}, 72(2):593--619, 2023.

\bibitem[Gla21]{glago2021flood}
Frank~Jerome Glago.
\newblock Flood disaster hazards; causes, impacts and management: a state-of-the-art review.
\newblock {\em Natural hazards-impacts, adjustments and resilience}, pages 29--37, 2021.

\bibitem[GR05]{gneiting2005weather}
Tilmann Gneiting and Adrian~E Raftery.
\newblock Weather forecasting with ensemble methods.
\newblock {\em Science}, 310(5746):248--249, 2005.

\bibitem[GSG{\etalchar{+}}24]{gutierrez2024hipporag}
Bernal~Jim{\'e}nez Guti{\'e}rrez, Yiheng Shu, Yu~Gu, Michihiro Yasunaga, and Yu~Su.
\newblock Hipporag: {N}eurobiologically inspired long-term memory for large language models.
\newblock In {\em The Thirty-eighth Annual Conference on Neural Information Processing Systems}, 2024.

\bibitem[GSH{\etalchar{+}}24]{gao2024prediff}
Zhihan Gao, Xingjian Shi, Boran Han, Hao Wang, Xiaoyong Jin, Danielle Maddix, Yi~Zhu, Mu~Li, and Yuyang~Bernie Wang.
\newblock Pre{D}iff: {P}recipitation nowcasting with latent diffusion models.
\newblock {\em Advances in Neural Information Processing Systems}, 36, 2024.

\bibitem[GSQ{\etalchar{+}}25]{gutierrez2025rag}
Bernal~Jim{\'e}nez Guti{\'e}rrez, Yiheng Shu, Weijian Qi, Sizhe Zhou, and Yu~Su.
\newblock From {RAG} to memory: {N}on-parametric continual learning for large language models.
\newblock {\em arXiv preprint arXiv:2502.14802}, 2025.

\bibitem[GSR{\etalchar{+}}17]{gilmer2017neural}
Justin Gilmer, Samuel~S Schoenholz, Patrick~F Riley, Oriol Vinyals, and George~E Dahl.
\newblock Neural message passing for quantum chemistry.
\newblock In {\em International conference on machine learning}, pages 1263--1272. PMLR, 2017.

\bibitem[GSW{\etalchar{+}}22]{gao2022earthformer}
Zhihan Gao, Xingjian Shi, Hao Wang, Yi~Zhu, Yuyang~Bernie Wang, Mu~Li, and Dit-Yan Yeung.
\newblock Earthformer: {E}xploring space-time transformers for earth system forecasting.
\newblock {\em Advances in Neural Information Processing Systems}, 35:25390--25403, 2022.

\bibitem[GT20]{guen2020disentangling}
Vincent~Le Guen and Nicolas Thome.
\newblock Disentangling physical dynamics from unknown factors for unsupervised video prediction.
\newblock In {\em Proceedings of the IEEE/CVF conference on computer vision and pattern recognition}, pages 11474--11484, 2020.

\bibitem[GXY{\etalchar{+}}24]{guo2024lightrag}
Zirui Guo, Lianghao Xia, Yanhua Yu, Tu~Ao, and Chao Huang.
\newblock Lightrag: {S}imple and fast retrieval-augmented generation.
\newblock {\em arXiv preprint arXiv:2410.05779}, 2024.

\bibitem[GYZ{\etalchar{+}}25]{guo2025deepseek}
Daya Guo, Dejian Yang, Haowei Zhang, Junxiao Song, Ruoyu Zhang, Runxin Xu, Qihao Zhu, Shirong Ma, Peiyi Wang, Xiao Bi, et~al.
\newblock Deepseek-{R}1: {I}ncentivizing reasoning capability in {LLM}s via reinforcement learning.
\newblock {\em arXiv preprint arXiv:2501.12948}, 2025.

\bibitem[HBB{\etalchar{+}}20]{hersbach2020era5}
Hans Hersbach, Bill Bell, Paul Berrisford, Shoji Hirahara, Andr{\'a}s Hor{\'a}nyi, Joaqu{\'\i}n Mu{\~n}oz-Sabater, Julien Nicolas, Carole Peubey, Raluca Radu, Dinand Schepers, et~al.
\newblock The {ERA5} global reanalysis.
\newblock {\em Quarterly Journal of the Royal Meteorological Society}, 146(730):1999--2049, 2020.

\bibitem[HCWL23]{hu2023swinvrnn}
Yuan Hu, Lei Chen, Zhibin Wang, and Hao Li.
\newblock Swin{VRNN}: {A} data-driven ensemble forecasting model via learned distribution perturbation.
\newblock {\em Journal of Advances in Modeling Earth Systems}, 15(2):e2022MS003211, 2023.

\bibitem[HJA20]{ho2020denoising}
Jonathan Ho, Ajay Jain, and Pieter Abbeel.
\newblock Denoising diffusion probabilistic models.
\newblock {\em Advances in neural information processing systems}, 33:6840--6851, 2020.

\bibitem[HM95]{han1995influence}
Jun Han and Claudio Moraga.
\newblock The influence of the sigmoid function parameters on the speed of backpropagation learning.
\newblock In {\em International workshop on artificial neural networks}, pages 195--201. Springer, 1995.

\bibitem[HMB{\etalchar{+}}25]{hussien2025rag}
Mohamed~Manzour Hussien, Angie~Nataly Melo, Augusto~Luis Ballardini, Carlota~Salinas Maldonado, Rub{\'e}n Izquierdo, and Miguel~Angel Sotelo.
\newblock Rag-based explainable prediction of road users behaviors for automated driving using knowledge graphs and large language models.
\newblock {\em Expert Systems with Applications}, 265:125914, 2025.

\bibitem[HMT{\etalchar{+}}22]{hertz2022prompt}
Amir Hertz, Ron Mokady, Jay Tenenbaum, Kfir Aberman, Yael Pritch, and Daniel Cohen-Or.
\newblock Prompt-to-prompt image editing with cross attention control.
\newblock {\em arXiv preprint arXiv:2208.01626}, 2022.

\bibitem[HS22]{ho2022classifier}
Jonathan Ho and Tim Salimans.
\newblock Classifier-free diffusion guidance.
\newblock {\em arXiv preprint arXiv:2207.12598}, 2022.

\bibitem[HTPB21]{hewage2021deep}
Pradeep Hewage, Marcello Trovati, Ella Pereira, and Ardhendu Behera.
\newblock Deep learning-based effective fine-grained weather forecasting model.
\newblock {\em Pattern Analysis and Applications}, 24(1):343--366, 2021.

\bibitem[HXCH04]{huang2004forecasting}
Wenrui Huang, Bing Xu, and Amy Chan-Hilton.
\newblock Forecasting flows in {A}palachicola {R}iver using neural networks.
\newblock {\em Hydrological processes}, 18(13):2545--2564, 2004.

\bibitem[HYM{\etalchar{+}}25]{huang2025survey}
Lei Huang, Weijiang Yu, Weitao Ma, Weihong Zhong, Zhangyin Feng, Haotian Wang, Qianglong Chen, Weihua Peng, Xiaocheng Feng, Bing Qin, et~al.
\newblock A survey on hallucination in large language models: {P}rinciples, taxonomy, challenges, and open questions.
\newblock {\em ACM Transactions on Information Systems}, 43(2):1--55, 2025.

\bibitem[ICH{\etalchar{+}}21]{izacard2021unsupervised}
Gautier Izacard, Mathilde Caron, Lucas Hosseini, Sebastian Riedel, Piotr Bojanowski, Armand Joulin, and Edouard Grave.
\newblock Unsupervised dense information retrieval with contrastive learning.
\newblock {\em arXiv preprint arXiv:2112.09118}, 2021.

\bibitem[ICH{\etalchar{+}}22]{izacard2022contriever}
Gautier Izacard, Mathilde Caron, Lucas Hosseini, Sebastian Riedel, Piotr Bojanowski, Armand Joulin, and Edouard Grave.
\newblock Unsupervised dense information retrieval with contrastive learning.
\newblock {\em Transactions on Machine Learning Research}, 2022.

\bibitem[IJH{\etalchar{+}}24]{ikram2024flood}
Qiyamud~Din Ikram, Abdur~Rashid Jamalzi, Abdur~Rahim Hamidi, Irfan Ullah, and Muhmmad Shahab.
\newblock Flood risk assessment of the population in {A}fghanistan: a spatial analysis of hazard, exposure, and vulnerability.
\newblock {\em Natural Hazards Research}, 4(1):46--55, 2024.

\bibitem[JCT{\etalchar{+}}23]{jenkins2023physics}
Luke~T Jenkins, Maggie~J Creed, Karim Tarbali, Manoranjan Muthusamy, Robert~{\v{S}}aki{\'c} Trogrli{\'c}, Jeremy~C Phillips, C~Scott Watson, Hugh~D Sinclair, Carmine Galasso, and John McCloskey.
\newblock Physics-based simulations of multiple natural hazards for risk-sensitive planning and decision making in expanding urban regions.
\newblock {\em International Journal of Disaster Risk Reduction}, 84:103338, 2023.

\bibitem[JGR20]{jiang2020supervised}
Tammy Jiang, Jaimie~L Gradus, and Anthony~J Rosellini.
\newblock Supervised machine learning: a brief primer.
\newblock {\em Behavior therapy}, 51(5):675--687, 2020.

\bibitem[JK22]{jaseena2022deterministic}
KU~Jaseena and Binsu~C Kovoor.
\newblock Deterministic weather forecasting models based on intelligent predictors: {A} survey.
\newblock {\em Journal of king saud university-computer and information sciences}, 34(6):3393--3412, 2022.

\bibitem[JMP{\etalchar{+}}23]{jafarzadegan2023recent}
Keighobad Jafarzadegan, Hamid Moradkhani, Florian Pappenberger, Hamed Moftakhari, Paul Bates, Peyman Abbaszadeh, Reza Marsooli, Celso Ferreira, Hannah~L Cloke, Fred Ogden, et~al.
\newblock Recent advances and new frontiers in riverine and coastal flood modeling.
\newblock {\em Reviews of Geophysics}, 61(2):e2022RG000788, 2023.

\bibitem[JSR{\etalchar{+}}19]{ji2019visual}
Xiaonan Ji, Han-Wei Shen, Alan Ritter, Raghu Machiraju, and Po-Yin Yen.
\newblock Visual exploration of neural document embedding in information retrieval: {S}emantics and feature selection.
\newblock {\em IEEE transactions on visualization and computer graphics}, 25(6):2181--2192, 2019.

\bibitem[JXG{\etalchar{+}}23]{jiang2023active}
Zhengbao Jiang, Frank~F Xu, Luyu Gao, Zhiqing Sun, Qian Liu, Jane Dwivedi-Yu, Yiming Yang, Jamie Callan, and Graham Neubig.
\newblock Active retrieval augmented generation.
\newblock In {\em Proceedings of the 2023 Conference on Empirical Methods in Natural Language Processing}, pages 7969--7992, 2023.

\bibitem[JYHA25]{jin2024long}
Bowen Jin, Jinsung Yoon, Jiawei Han, and Sercan~O Arik.
\newblock Long-context {LLM}s meet {RAG}: {O}vercoming challenges for long inputs in {RAG}.
\newblock In {\em The Thirteenth International Conference on Learning Representations}, 2025.

\bibitem[JZH21]{janiesch2021machine}
Christian Janiesch, Patrick Zschech, and Kai Heinrich.
\newblock Machine learning and deep learning.
\newblock {\em Electronic markets}, 31(3):685--695, 2021.

\bibitem[Kad19]{kadhim2019term}
Ammar~Ismael Kadhim.
\newblock Term weighting for feature extraction on {T}witter: {A} comparison between bm25 and tf-idf.
\newblock In {\em 2019 international conference on advanced science and engineering (ICOASE)}, pages 124--128. IEEE, 2019.

\bibitem[KAJ{\etalchar{+}}24]{kang2024improving}
SeongKu Kang, Shivam Agarwal, Bowen Jin, Dongha Lee, Hwanjo Yu, and Jiawei Han.
\newblock Improving retrieval in theme-specific applications using a corpus topical taxonomy.
\newblock In {\em Proceedings of the ACM Web Conference 2024}, pages 1497--1508, 2024.

\bibitem[Kar16]{karimanzira2016model}
Divas Karimanzira.
\newblock Model based decision support systems.
\newblock {\em Modeling, Control and Optimization of Water Systems: Systems Engineering Methods for Control and Decision Making Tasks}, pages 185--220, 2016.

\bibitem[KAS{\etalchar{+}}23]{kumar2023state}
Vijendra Kumar, Hazi~Md Azamathulla, Kul~Vaibhav Sharma, Darshan~J Mehta, and Kiran~Tota Maharaj.
\newblock The state of the art in deep learning applications, challenges, and future prospects: {A} comprehensive review of flood forecasting and management.
\newblock {\em Sustainability}, 15(13):10543, 2023.

\bibitem[Kei22]{keisler2022forecasting}
Ryan Keisler.
\newblock Forecasting global weather with graph neural networks.
\newblock {\em arXiv preprint arXiv:2202.07575}, 2022.

\bibitem[KGL{\etalchar{+}}16]{kerkez2016smarter}
Branko Kerkez, Cyndee Gruden, Matthew Lewis, Luis Montestruque, Marcus Quigley, Brandon Wong, Alex Bedig, Ruben Kertesz, Tim Braun, Owen Cadwalader, et~al.
\newblock Smarter stormwater systems, 2016.

\bibitem[KHMN25]{khatooni2025new}
Kousha Khatooni, Farhad Hooshyaripor, Bahram MalekMohammadi, and Roohollah Noori.
\newblock A new approach for urban flood risk assessment using coupled {SWMM--HEC-RAS-2D} model.
\newblock {\em Journal of Environmental Management}, 374:123849, 2025.

\bibitem[KK13]{kramer2013k}
Oliver Kramer and Oliver Kramer.
\newblock K-nearest neighbors.
\newblock {\em Dimensionality reduction with unsupervised nearest neighbors}, pages 13--23, 2013.

\bibitem[KLY{\etalchar{+}}24]{kow2024advancing}
Pu-Yun Kow, Jia-Yi Liou, Ming-Ting Yang, Meng-Hsin Lee, Li-Chiu Chang, and Fi-John Chang.
\newblock Advancing climate-resilient flood mitigation: {U}tilizing transformer-{LSTM} for water level forecasting at pumping stations.
\newblock {\em Science of the Total Environment}, 927:172246, 2024.

\bibitem[KOM{\etalchar{+}}20]{karpukhin2020dense}
Vladimir Karpukhin, Barlas Oguz, Sewon Min, Patrick~SH Lewis, Ledell Wu, Sergey Edunov, Danqi Chen, and Wen-tau Yih.
\newblock Dense passage retrieval for open-domain question answering.
\newblock In {\em EMNLP (1)}, pages 6769--6781, 2020.

\bibitem[KPR{\etalchar{+}}19]{kwiatkowski2019natural}
Tom Kwiatkowski, Jennimaria Palomaki, Olivia Redfield, Michael Collins, Ankur Parikh, Chris Alberti, Danielle Epstein, Illia Polosukhin, Jacob Devlin, Kenton Lee, et~al.
\newblock Natural questions: a benchmark for question answering research.
\newblock {\em Transactions of the Association for Computational Linguistics}, 7:453--466, 2019.

\bibitem[KRI21]{krishna2021hurdles}
Kalpesh Krishna, Aurko Roy, and Mohit Iyyer.
\newblock Hurdles to progress in long-form question answering.
\newblock In Kristina Toutanova, Anna Rumshisky, Luke Zettlemoyer, Dilek Hakkani-Tur, Iz~Beltagy, Steven Bethard, Ryan Cotterell, Tanmoy Chakraborty, and Yichao Zhou, editors, {\em Proceedings of the 2021 Conference of the North American Chapter of the Association for Computational Linguistics: Human Language Technologies}, pages 4940--4957, Online, June 2021. Association for Computational Linguistics.

\bibitem[KS16]{keren2016convolutional}
Gil Keren and Bj{\"o}rn Schuller.
\newblock Convolutional {RNN}: an enhanced model for extracting features from sequential data.
\newblock In {\em 2016 International Joint Conference on Neural Networks (IJCNN)}, pages 3412--3419. IEEE, 2016.

\bibitem[KS21]{kag2021training}
Anil Kag and Venkatesh Saligrama.
\newblock Training recurrent neural networks via forward propagation through time.
\newblock In {\em International Conference on Machine Learning}, pages 5189--5200. PMLR, 2021.

\bibitem[KW16]{kipf2016semi}
Thomas~N Kipf and Max Welling.
\newblock Semi-supervised classification with graph convolutional networks.
\newblock {\em arXiv preprint arXiv:1609.02907}, 2016.

\bibitem[KYL{\etalchar{+}}24]{kochkov2024neural}
Dmitrii Kochkov, Janni Yuval, Ian Langmore, Peter Norgaard, Jamie Smith, Griffin Mooers, Milan Kl{\"o}wer, James Lottes, Stephan Rasp, Peter D{\"u}ben, et~al.
\newblock Neural general circulation models for weather and climate.
\newblock {\em Nature}, pages 1--7, 2024.

\bibitem[LAC{\etalchar{+}}24]{lang2024aifs}
Simon Lang, Mihai Alexe, Matthew Chantry, Jesper Dramsch, Florian Pinault, Baudouin Raoult, Mariana~CA Clare, Christian Lessig, Michael Maier-Gerber, Linus Magnusson, et~al.
\newblock {AIFS-ECMWF}'s data-driven forecasting system.
\newblock {\em arXiv preprint arXiv:2406.01465}, 2024.

\bibitem[LBT21]{leon2021comparison}
Arturo~S Leon, Linlong Bian, and Yun Tang.
\newblock Comparison of the genetic algorithm and pattern search methods for forecasting optimal flow releases in a multi-storage system for flood control.
\newblock {\em Environmental Modelling \& Software}, 145:105198, 2021.

\bibitem[LCB{\etalchar{+}}24]{liu2024mambads}
Zili Liu, Hao Chen, Lei Bai, Wenyuan Li, Wanli Ouyang, Zhengxia Zou, and Zhenwei Shi.
\newblock Mamba{DS}: {N}ear-surface meteorological field downscaling with topography constrained selective state space modeling.
\newblock {\em arXiv preprint arXiv:2408.10854}, 2024.

\bibitem[LCC{\etalchar{+}}23]{lin2023ra}
Xi~Victoria Lin, Xilun Chen, Mingda Chen, Weijia Shi, Maria Lomeli, Richard James, Pedro Rodriguez, Jacob Kahn, Gergely Szilvasy, Mike Lewis, et~al.
\newblock Ra-dit: {R}etrieval-augmented dual instruction tuning.
\newblock In {\em The Twelfth International Conference on Learning Representations}, 2023.

\bibitem[LCLG{\etalchar{+}}23]{li2023seeds}
Lizao Li, Rob Carver, Ignacio Lopez-Gomez, Fei Sha, and John Anderson.
\newblock {SEED}s: {E}mulation of weather forecast ensembles with diffusion models.
\newblock {\em arXiv preprint arXiv:2306.14066}, 2023.

\bibitem[LCT21]{lee2021learning}
Kenton Lee, Ming-Wei Chang, and Kristina Toutanova.
\newblock Learning dense representations of phrases at scale.
\newblock In {\em ACL}, 2021.

\bibitem[LFX{\etalchar{+}}24]{liu2024deepseek}
Aixin Liu, Bei Feng, Bing Xue, Bingxuan Wang, Bochao Wu, Chengda Lu, Chenggang Zhao, Chengqi Deng, Chenyu Zhang, Chong Ruan, et~al.
\newblock Deepseek-v3 technical report.
\newblock {\em arXiv preprint arXiv:2412.19437}, 2024.

\bibitem[LG16]{leon2016controlling}
Arturo~S Leon and Christopher Goodell.
\newblock Controlling hec-ras using {MATLAB}.
\newblock {\em Environmental modelling \& software}, 84:339--348, 2016.

\bibitem[LH21]{luo2021diffusion}
Shitong Luo and Wei Hu.
\newblock Diffusion probabilistic models for 3{D} point cloud generation.
\newblock In {\em Proceedings of the IEEE/CVF conference on computer vision and pattern recognition}, pages 2837--2845, 2021.

\bibitem[LHN{\etalchar{+}}23]{leinonen2023latent}
Jussi Leinonen, Ulrich Hamann, Daniele Nerini, Urs Germann, and Gabriele Franch.
\newblock Latent diffusion models for generative precipitation nowcasting with accurate uncertainty quantification.
\newblock {\em arXiv preprint arXiv:2304.12891}, 2023.

\bibitem[LKVS14]{leon2014dynamic}
Arturo~S Leon, Elizabeth~A Kanashiro, Rachelle Valverde, and Venkataramana Sridhar.
\newblock Dynamic framework for intelligent control of river flooding: {C}ase study.
\newblock {\em Journal of Water Resources Planning and Management}, 140(2):258--268, 2014.

\bibitem[LLC{\etalchar{+}}24]{li2024deepphysinet}
Wenyuan Li, Zili Liu, Keyan Chen, Hao Chen, Shunlin Liang, Zhengxia Zou, and Zhenwei Shi.
\newblock Deepphysinet: {B}ridging deep learning and atmospheric physics for accurate and continuous weather modeling.
\newblock {\em arXiv preprint arXiv:2401.04125}, 2024.

\bibitem[LLD{\etalchar{+}}24]{liu2024wpmamba}
Tengyuan Liu, Lei Liu, Xue Dong, Qiuju Chen, and Bin Li.
\newblock {WPM}amba: {E}nhanced wind power forecasting model based on {M}amba with weather forecast data.
\newblock In {\em 2024 The 9th International Conference on Power and Renewable Energy (ICPRE)}, pages 1429--1435. IEEE, 2024.

\bibitem[LLG{\etalchar{+}}23]{lessig2023atmorep}
Christian Lessig, Ilaria Luise, Bing Gong, Michael Langguth, Scarlet Stadtler, and Martin Schultz.
\newblock Atmo{R}ep: {A} stochastic model of atmosphere dynamics using large scale representation learning.
\newblock {\em arXiv preprint arXiv:2308.13280}, 2023.

\bibitem[LLH{\etalchar{+}}24]{liu2024lost}
Nelson~F Liu, Kevin Lin, John Hewitt, Ashwin Paranjape, Michele Bevilacqua, Fabio Petroni, and Percy Liang.
\newblock Lost in the middle: {H}ow language models use long contexts.
\newblock {\em Transactions of the Association for Computational Linguistics}, 12:157--173, 2024.

\bibitem[LLQ{\etalchar{+}}24]{ling2024srndiff}
Xudong Ling, Chaorong Li, Fengqing Qin, Peng Yang, and Yuanyuan Huang.
\newblock {SRND}iff: {S}hort-term rainfall nowcasting with condition diffusion model.
\newblock {\em arXiv preprint arXiv:2402.13737}, 2024.

\bibitem[LLY{\etalchar{+}}22]{luo2022experimental}
Chuyao Luo, Xutao Li, Yunming Ye, Shanshan Feng, and Michael~K Ng.
\newblock Experimental study on generative adversarial network for precipitation nowcasting.
\newblock {\em IEEE Transactions on Geoscience and Remote Sensing}, 60:1--20, 2022.

\bibitem[LNML14]{li2014rolling}
Lei Li, Farzad Noorian, Duncan~JM Moss, and Philip~HW Leong.
\newblock Rolling window time series prediction using {M}ap{Reduce}.
\newblock In {\em Proceedings of the 2014 IEEE 15th international conference on information reuse and integration (IEEE IRI 2014)}, pages 757--764. IEEE, 2014.

\bibitem[LOG{\etalchar{+}}19]{liu2019roberta}
Yinhan Liu, Myle Ott, Naman Goyal, Jingfei Du, Mandar Joshi, Danqi Chen, Omer Levy, Mike Lewis, Luke Zettlemoyer, and Veselin Stoyanov.
\newblock Roberta: {A} robustly optimized bert pretraining approach.
\newblock {\em arXiv preprint arXiv:1907.11692}, 2019.

\bibitem[LRX{\etalchar{+}}25]{lee2024nv}
Chankyu Lee, Rajarshi Roy, Mengyao Xu, Jonathan Raiman, Mohammad Shoeybi, Bryan Catanzaro, and Wei Ping.
\newblock {NV}-embed: Improved techniques for training {LLM}s as generalist embedding models.
\newblock In {\em The Thirteenth International Conference on Learning Representations}, 2025.

\bibitem[LSGW{\etalchar{+}}22]{lam2022graphcast}
Remi Lam, Alvaro Sanchez-Gonzalez, Matthew Willson, Peter Wirnsberger, Meire Fortunato, Alexander Pritzel, Suman Ravuri, Timo Ewalds, Ferran Alet, Zach Eaton-Rosen, et~al.
\newblock Graph{C}ast: {L}earning skillful medium-range global weather forecasting.
\newblock {\em arXiv preprint arXiv:2212.12794}, 2022.

\bibitem[LSGW{\etalchar{+}}23]{lam2023learning}
Remi Lam, Alvaro Sanchez-Gonzalez, Matthew Willson, Peter Wirnsberger, Meire Fortunato, Ferran Alet, Suman Ravuri, Timo Ewalds, Zach Eaton-Rosen, Weihua Hu, et~al.
\newblock Learning skillful medium-range global weather forecasting.
\newblock {\em Science}, 382(6677):1416--1421, 2023.

\bibitem[LSJ{\etalchar{+}}22]{li2022generation}
Dong Li, Yelong Shen, Ruoming Jin, Yi~Mao, Kuan Wang, and Weizhu Chen.
\newblock Generation-augmented query expansion for code retrieval.
\newblock {\em arXiv preprint arXiv:2212.10692}, 2022.

\bibitem[LTQC20]{leon2020matlab}
Arturo~S Leon, Yun Tang, Li~Qin, and Duan Chen.
\newblock A {MATLAB} framework for forecasting optimal flow releases in a multi-storage system for flood control.
\newblock {\em Environmental modelling \& software}, 125:104618, 2020.

\bibitem[LYC{\etalchar{+}}22]{li2022srdiff}
Haoying Li, Yifan Yang, Meng Chang, Shiqi Chen, Huajun Feng, Zhihai Xu, Qi~Li, and Yueting Chen.
\newblock Srdiff: Single image super-resolution with diffusion probabilistic models.
\newblock {\em Neurocomputing}, 479:47--59, 2022.

\bibitem[LZSJ24]{liu2024mitigating}
Peiyuan Liu, Tian Zhou, Liang Sun, and Rong Jin.
\newblock Mitigating time discretization challenges with weather{ODE}: {A} sandwich physics-driven neural {ODE} for weather forecasting.
\newblock {\em arXiv preprint arXiv:2410.06560}, 2024.

\bibitem[LZZW23]{li2023diffusion}
Yifan Li, Kun Zhou, Wayne~Xin Zhao, and Ji-Rong Wen.
\newblock Diffusion models for non-autoregressive text generation: {A} survey.
\newblock {\em arXiv preprint arXiv:2303.06574}, 2023.

\bibitem[MAS{\etalchar{+}}24]{miller2024survey}
John~A Miller, Mohammed Aldosari, Farah Saeed, Nasid~Habib Barna, Subas Rana, I~Budak Arpinar, and Ninghao Liu.
\newblock A survey of deep learning and foundation models for time series forecasting.
\newblock {\em arXiv preprint arXiv:2401.13912}, 2024.

\bibitem[MAZ{\etalchar{+}}22]{mallen2022not}
Alex Mallen, Akari Asai, Victor Zhong, Rajarshi Das, Daniel Khashabi, and Hannaneh Hajishirzi.
\newblock When not to trust language models: {I}nvestigating effectiveness of parametric and non-parametric memories.
\newblock {\em arXiv preprint arXiv:2212.10511}, 2022.

\bibitem[MIHA18]{monrat2018belief}
Ahmed~Afif Monrat, Raihan~Ul Islam, Mohammad~Shahadat Hossain, and Karl Andersson.
\newblock A belief rule based flood risk assessment expert system using real time sensor data streaming.
\newblock In {\em 2018 IEEE 43rd Conference on Local Computer Networks Workshops (LCN Workshops)}, pages 38--45. IEEE, 2018.

\bibitem[MJ01]{medsker2001recurrent}
Larry~R Medsker and LC~Jain.
\newblock Recurrent neural networks.
\newblock {\em Design and Applications}, 5(64-67):2, 2001.

\bibitem[MKK{\etalchar{+}}20]{merz2020impact}
Bruno Merz, Christian Kuhlicke, Michael Kunz, Massimiliano Pittore, Andrey Babeyko, David~N Bresch, Daniela~IV Domeisen, Frauke Feser, Inga Koszalka, Heidi Kreibich, et~al.
\newblock Impact forecasting to support emergency management of natural hazards.
\newblock {\em Reviews of Geophysics}, 58(4):e2020RG000704, 2020.

\bibitem[MM21]{mishra2021neuralnere}
Prakamya Mishra and Rohan Mittal.
\newblock Neuralnere: {N}eural named entity relationship extraction for end-to-end climate change knowledge graph construction.
\newblock In {\em Tackling climate change with machine learning workshop at ICML}, 2021.

\bibitem[MMM{\etalchar{+}}22]{mishra2022overview}
Ashok Mishra, Sourav Mukherjee, Bruno Merz, Vijay~P Singh, Daniel~B Wright, Gabriele Villarini, Subir Paul, D~Nagesh Kumar, C~Prakash Khedun, Dev Niyogi, et~al.
\newblock An overview of flood concepts, challenges, and future directions.
\newblock {\em Journal of hydrologic engineering}, 27(6):03122001, 2022.

\bibitem[MOY21]{maillard2021multi}
Jean Maillard, Barlas Oguz, and Wen-tau Yih.
\newblock Multi-faith retrieval for question answering.
\newblock In {\em NAACL}, 2021.

\bibitem[MSL{\etalchar{+}}23]{mo2023simulation}
Chongxun Mo, Yue Shen, Xingbi Lei, Huazhen Ban, Yuli Ruan, Shufeng Lai, Weiyan Cen, and Zhenxiang Xing.
\newblock Simulation of one-dimensional dam-break flood routing based on {HEC-RAS}.
\newblock {\em Frontiers in Earth Science}, 10:1027788, 2023.

\bibitem[MSO12]{mohamed2012comparative}
W~Nor Haizan~W Mohamed, Mohd Najib~Mohd Salleh, and Abdul~Halim Omar.
\newblock A comparative study of reduced error pruning method in decision tree algorithms.
\newblock In {\em 2012 IEEE International conference on control system, computing and engineering}, pages 392--397. IEEE, 2012.

\bibitem[MSO{\etalchar{+}}20]{mounce2020optimisation}
SR~Mounce, Will Shepherd, Sonja Ostojin, Mohamad Abdel-Aal, ANA Schellart, JD~Shucksmith, and SJ~Tait.
\newblock Optimisation of a fuzzy logic-based local real-time control system for mitigation of sewer flooding using genetic algorithms.
\newblock {\em Journal of Hydroinformatics}, 22(2):281--295, 2020.

\bibitem[MWW18]{morley2018perturbed}
SK~Morley, DT~Welling, and JR~Woodroffe.
\newblock Perturbed input ensemble modeling with the space weather modeling framework.
\newblock {\em Space Weather}, 16(9):1330--1347, 2018.

\bibitem[MXT{\etalchar{+}}23]{ma2023histgnn}
Minbo Ma, Peng Xie, Fei Teng, Bin Wang, Shenggong Ji, Junbo Zhang, and Tianrui Li.
\newblock Hi{STGNN}: {H}ierarchical spatio-temporal graph neural network for weather forecasting.
\newblock {\em Information Sciences}, 648:119580, 2023.

\bibitem[MZF{\etalchar{+}}23]{man2023w}
Xin Man, Chenghong Zhang, Jin Feng, Changyu Li, and Jie Shao.
\newblock W-mae: {P}re-trained weather model with masked autoencoder for multi-variable weather forecasting.
\newblock {\em arXiv preprint arXiv:2304.08754}, 2023.

\bibitem[MZL23]{ma2023mm}
Zhifeng Ma, Hao Zhang, and Jie Liu.
\newblock {MM-RNN}: {A} multimodal {RNN} for precipitation nowcasting.
\newblock {\em IEEE Transactions on Geoscience and Remote Sensing}, 2023.

\bibitem[NBK{\etalchar{+}}23]{nguyen2023climax}
Tung Nguyen, Johannes Brandstetter, Ashish Kapoor, Jayesh~K Gupta, and Aditya Grover.
\newblock Climax: {A} foundation model for weather and climate.
\newblock {\em arXiv preprint arXiv:2301.10343}, 2023.

\bibitem[NCD{\etalchar{+}}24]{nearing2024global}
Grey Nearing, Deborah Cohen, Vusumuzi Dube, Martin Gauch, Oren Gilon, Shaun Harrigan, Avinatan Hassidim, Daniel Klotz, Frederik Kratzert, Asher Metzger, et~al.
\newblock Global prediction of extreme floods in ungauged watersheds.
\newblock {\em Nature}, 627(8004):559--563, 2024.

\bibitem[NDR{\etalchar{+}}21]{nichol2021glide}
Alex Nichol, Prafulla Dhariwal, Aditya Ramesh, Pranav Shyam, Pamela Mishkin, Bob McGrew, Ilya Sutskever, and Mark Chen.
\newblock Glide: {T}owards photorealistic image generation and editing with text-guided diffusion models.
\newblock {\em arXiv preprint arXiv:2112.10741}, 2021.

\bibitem[NJB{\etalchar{+}}23]{nguyen2023climatelearn}
Tung Nguyen, Jason Jewik, Hritik Bansal, Prakhar Sharma, and Aditya Grover.
\newblock Climate{L}earn: {B}enchmarking machine learning for weather and climate modeling.
\newblock {\em arXiv preprint arXiv:2307.01909}, 2023.

\bibitem[NSB{\etalchar{+}}23]{nguyen2023scaling}
Tung Nguyen, Rohan Shah, Hritik Bansal, Troy Arcomano, Romit Maulik, Veerabhadra Kotamarthi, Ian Foster, Sandeep Madireddy, and Aditya Grover.
\newblock Scaling transformer neural networks for skillful and reliable medium-range weather forecasting.
\newblock {\em arXiv preprint arXiv:2312.03876}, 2023.

\bibitem[OKP{\etalchar{+}}22]{oikonomou2022robust}
A~Oikonomou, Manos Kirtas, Nikos Passalis, George Mourgias-Alexandris, Miltiadis Moralis-Pegios, Nikos Pleros, and Anastasios Tefas.
\newblock A robust, quantization-aware training method for photonic neural networks.
\newblock In {\em International conference on engineering applications of neural networks}, pages 427--438. Springer, 2022.

\bibitem[ON15]{o2015introduction}
Keiron O'Shea and Ryan Nash.
\newblock An introduction to convolutional neural networks.
\newblock {\em arXiv preprint arXiv:1511.08458}, 2015.

\bibitem[OTC{\etalchar{+}}00]{obeysekera2000use}
Jayantha Obeysekera, P~Trimble, Luis Cadavid, ER~Santee, and Cary White.
\newblock Use of climate outlook for water management in {S}outh {F}lorida, {USA}.
\newblock In {\em Proc., Engineering Jubilee Conf}, 2000.

\bibitem[PAB{\etalchar{+}}17]{pecl2017biodiversity}
Gretta~T Pecl, Miguel~B Ara{\'u}jo, Johann~D Bell, Julia Blanchard, Timothy~C Bonebrake, I-Ching Chen, Timothy~D Clark, Robert~K Colwell, Finn Danielsen, Birgitta Eveng{\aa}rd, et~al.
\newblock Biodiversity redistribution under climate change: {I}mpacts on ecosystems and human well-being.
\newblock {\em Science}, 355(6332):eaai9214, 2017.

\bibitem[Pal19]{palmer2019ecmwf}
Tim Palmer.
\newblock The {ECMWF} ensemble prediction system: Looking back (more than) 25 years and projecting forward 25 years.
\newblock {\em Quarterly Journal of the Royal Meteorological Society}, 145:12--24, 2019.

\bibitem[Pet09]{peterson2009k}
Leif~E Peterson.
\newblock K-nearest neighbor.
\newblock {\em Scholarpedia}, 4(2):1883, 2009.

\bibitem[PGD{\etalchar{+}}24]{peker2024integration}
{\.I}smail~Bilal Peker, Sezar G{\"u}lbaz, Vahdettin Demir, Osman Orhan, and Neslihan Beden.
\newblock Integration of {HEC-RAS} and {HEC-HMS} with {GIS} in flood modeling and flood hazard mapping.
\newblock {\em Sustainability}, 16(3):1226, 2024.

\bibitem[PPG23]{pujari2023explainable}
Tejaskumar Pujari, Anil~Kumar Pakina, and Anshul Goel.
\newblock Explainable ai and governance: {E}nhancing transparency and policy frameworks through retrieval-augmented generation ({RAG}).
\newblock {\em IOSR Journal of Computer Engineering}, 2023.

\bibitem[PRWZ02]{papineni2002bleu}
Kishore Papineni, Salim Roukos, Todd Ward, and Wei-Jing Zhu.
\newblock Bleu: a method for automatic evaluation of machine translation.
\newblock In {\em Proceedings of the 40th annual meeting of the Association for Computational Linguistics}, pages 311--318, 2002.

\bibitem[PSGA{\etalchar{+}}24]{price2023gencast}
Ilan Price, Alvaro Sanchez-Gonzalez, Ferran Alet, Timo Ewalds, Andrew El-Kadi, Jacklynn Stott, Shakir Mohamed, Peter Battaglia, Remi Lam, and Matthew Willson.
\newblock Gen{C}ast: {D}iffusion-based ensemble forecasting for medium-range weather.
\newblock {\em arXiv preprint arXiv:2312.15796}, May 2024.

\bibitem[PSH{\etalchar{+}}05]{palmer2005representing}
TN~Palmer, GJ~Shutts, R~Hagedorn, FJ~Doblas-Reyes, Thomas Jung, and M~Leutbecher.
\newblock Representing model uncertainty in weather and climate prediction.
\newblock {\em Annu. Rev. Earth Planet. Sci.}, 33(1):163--193, 2005.

\bibitem[PSH{\etalchar{+}}22]{pathak2022fourcastnet}
Jaideep Pathak, Shashank Subramanian, Peter Harrington, Sanjeev Raja, Ashesh Chattopadhyay, Morteza Mardani, Thorsten Kurth, David Hall, Zongyi Li, Kamyar Azizzadenesheli, et~al.
\newblock Fourcastnet: {A} global data-driven high-resolution weather model using adaptive fourier neural operators.
\newblock {\em arXiv preprint arXiv:2202.11214}, 2022.

\bibitem[PV18]{petersen2018optimal}
Philipp Petersen and Felix Voigtlaender.
\newblock Optimal approximation of piecewise smooth functions using deep relu neural networks.
\newblock {\em Neural Networks}, 108:296--330, 2018.

\bibitem[PZL{\etalchar{+}}24]{peng2024graph}
Boci Peng, Yun Zhu, Yongchao Liu, Xiaohe Bo, Haizhou Shi, Chuntao Hong, Yan Zhang, and Siliang Tang.
\newblock Graph retrieval-augmented generation: {A} survey.
\newblock {\em arXiv preprint arXiv:2408.08921}, 2024.

\bibitem[QCJ{\etalchar{+}}24]{qin2024metmamba}
Haoyu Qin, Yungang Chen, Qianchuan Jiang, Pengchao Sun, Xiancai Ye, and Chao Lin.
\newblock Met{M}amba: {R}egional weather forecasting with spatial-temporal mamba model.
\newblock {\em arXiv preprint arXiv:2408.06400}, 2024.

\bibitem[QMX{\etalchar{+}}21]{qi2021review}
Wenchao Qi, Chao Ma, Hongshi Xu, Zifan Chen, Kai Zhao, and Hao Han.
\newblock A review on applications of urban flood models in flood mitigation strategies.
\newblock {\em Natural Hazards}, 108:31--62, 2021.

\bibitem[Qui86]{quinlan1986induction}
J.~Ross Quinlan.
\newblock Induction of decision trees.
\newblock {\em Machine learning}, 1:81--106, 1986.

\bibitem[RA24]{rahman2024drivers}
Motiur Rahman and Md~Shahjahan Ali.
\newblock Drivers of tidal flow variability in the {P}ussur fluvial estuary: A numerical study by {HEC-RAS}.
\newblock {\em Heliyon}, 10(4), 2024.

\bibitem[Ram24]{ramavajjala2024heal}
Vivek Ramavajjala.
\newblock {HEAL-ViT}: {V}ision transformers on a spherical mesh for medium-range weather forecasting.
\newblock {\em arXiv preprint arXiv:2403.17016}, 2024.

\bibitem[RBL{\etalchar{+}}22]{rombach2022high}
Robin Rombach, Andreas Blattmann, Dominik Lorenz, Patrick Esser, and Bj{\"o}rn Ommer.
\newblock High-resolution image synthesis with latent diffusion models.
\newblock In {\em Proceedings of the IEEE/CVF conference on computer vision and pattern recognition}, pages 10684--10695, 2022.

\bibitem[RDS{\etalchar{+}}20]{rasp2020weatherbench}
Stephan Rasp, Peter~D Dueben, Sebastian Scher, Jonathan~A Weyn, Soukayna Mouatadid, and Nils Thuerey.
\newblock Weather{B}ench: a benchmark data set for data-driven weather forecasting.
\newblock {\em Journal of Advances in Modeling Earth Systems}, 12(11):e2020MS002203, 2020.

\bibitem[RFB15]{ronneberger2015u}
Olaf Ronneberger, Philipp Fischer, and Thomas Brox.
\newblock U-net: {C}onvolutional networks for biomedical image segmentation.
\newblock In {\em Medical image computing and computer-assisted intervention--MICCAI 2015: 18th international conference, October 5-9, 2015, proceedings, part III 18}, pages 234--241, 2015.

\bibitem[RLW{\etalchar{+}}21]{ravuri2021skilful}
Suman Ravuri, Karel Lenc, Matthew Willson, Dmitry Kangin, Remi Lam, Piotr Mirowski, Megan Fitzsimons, Maria Athanassiadou, Sheleem Kashem, Sam Madge, et~al.
\newblock Skilful precipitation nowcasting using deep generative models of radar.
\newblock {\em Nature}, 597(7878):672--677, 2021.

\bibitem[RP07]{rodwell2007using}
MJ~Rodwell and TN~Palmer.
\newblock Using numerical weather prediction to assess climate models.
\newblock {\em Quarterly Journal of the Royal Meteorological Society: A journal of the atmospheric sciences, applied meteorology and physical oceanography}, 133(622):129--146, 2007.

\bibitem[RRMA19]{ravindra2019generalized}
Khaiwal Ravindra, Preety Rattan, Suman Mor, and Ashutosh~Nath Aggarwal.
\newblock Generalized additive models: {B}uilding evidence of air pollution, climate change and human health.
\newblock {\em Environment international}, 132:104987, 2019.

\bibitem[RSP{\etalchar{+}}25]{rangaraj2025retrieval}
Rahuul Rangaraj, Jimeng Shi, Rajendra Paudel, Giri Narasimhan, and Yanzhao Wu.
\newblock Retrieval-augmented water level forecasting for everglades.
\newblock {\em arXiv preprint arXiv:2508.04888}, 2025.

\bibitem[RSS{\etalchar{+}}25]{rangaraj2025effective}
Rahuul Rangaraj, Jimeng Shi, Azam Shirali, Rajendra Paudel, Yanzhao Wu, and Giri Narasimhan.
\newblock How effective are large time series models in hydrology? a study on water level forecasting in everglades.
\newblock {\em arXiv preprint arXiv:2505.01415}, 2025.

\bibitem[RT10]{rogers2010global}
David Rogers and Vladimir Tsirkunov.
\newblock Global assessment report on disaster risk reduction: costs and benefits of early warning systems.
\newblock Technical report, The World Bank, 2010.

\bibitem[RTLC{\etalchar{+}}22]{rivett2022acute}
Michael~O Rivett, Laurent-Charles Tremblay-Levesque, Ruth Carter, Rudi~CH Thetard, Morris Tengatenga, Ann Phoya, Emma Mbalame, Edwin Mchilikizo, Steven Kumwenda, Prince Mleta, et~al.
\newblock Acute health risks to community hand-pumped groundwater supplies following {C}yclone {I}dai flooding.
\newblock {\em Science of The Total Environment}, 806:150598, 2022.

\bibitem[RUB19]{rangari2019assessment}
Vinay~Ashok Rangari, NV~Umamahesh, and CM~Bhatt.
\newblock Assessment of inundation risk in urban floods using {HEC-RAS} 2{D}.
\newblock {\em Modeling Earth Systems and Environment}, 5(4):1839--1851, 2019.

\bibitem[Rud16]{ruder2016overview}
Sebastian Ruder.
\newblock An overview of gradient descent optimization algorithms.
\newblock {\em arXiv preprint arXiv:1609.04747}, 2016.

\bibitem[RVB22]{rokooei2022perceptions}
Saeed Rokooei, Farshid Vahedifard, and Solomon Belay.
\newblock Perceptions of civil engineering and construction students toward community and infrastructure resilience.
\newblock {\em Journal of Civil Engineering Education}, 148(1):04021015, 2022.

\bibitem[RW94]{robertson1994some}
Stephen~E Robertson and Steve Walker.
\newblock Some simple effective approximations to the 2-poisson model for probabilistic weighted retrieval.
\newblock In {\em SIGIR’94: Proceedings of the Seventeenth Annual International ACM-SIGIR Conference on Research and Development in Information Retrieval, organised by Dublin City University}, pages 232--241. Springer, 1994.

\bibitem[SABA21]{schwenzer2021review}
Max Schwenzer, Muzaffer Ay, Thomas Bergs, and Dirk Abel.
\newblock Review on model predictive control: {A}n engineering perspective.
\newblock {\em The International Journal of Advanced Manufacturing Technology}, 117(5):1327--1349, 2021.

\bibitem[SAG{\etalchar{+}}23]{shen2023differentiable}
Chaopeng Shen, Alison~P Appling, Pierre Gentine, Toshiyuki Bandai, Hoshin Gupta, Alexandre Tartakovsky, Marco Baity-Jesi, Fabrizio Fenicia, Daniel Kifer, Li~Li, et~al.
\newblock Differentiable modeling to unify machine learning and physical models and advance geosciences.
\newblock {\em arXiv preprint arXiv:2301.04027}, 2023.

\bibitem[Sar21a]{sarker2021deep}
Iqbal~H Sarker.
\newblock Deep learning: a comprehensive overview on techniques, taxonomy, applications and research directions.
\newblock {\em SN computer science}, 2(6):1--20, 2021.

\bibitem[Sar21b]{sarker2021machine}
Iqbal~H Sarker.
\newblock Machine learning: {A}lgorithms, real-world applications and research directions.
\newblock {\em SN computer science}, 2(3):160, 2021.

\bibitem[SBB{\etalchar{+}}20]{shrestha2020understanding}
Alen Shrestha, Linkon Bhattacharjee, Sudip Baral, Balbhadra Thakur, Neekita Joshi, Ajay Kalra, and Ritu Gupta.
\newblock Understanding suitability of {MIKE} 21 and {HEC-RAS} for 2{D} floodplain modeling.
\newblock In {\em World environmental and water resources congress 2020}, pages 237--253. American Society of Civil Engineers Reston, VA, 2020.

\bibitem[SBX15]{schwanenberg2015open}
D~Schwanenberg, BPJ Becker, and M~Xu.
\newblock The open real-time control ({RTC})-tools software framework for modeling {RTC} in water resources sytems.
\newblock {\em Journal of Hydroinformatics}, 17(1):130--148, 2015.

\bibitem[SCC{\etalchar{+}}22]{saharia2022palette}
Chitwan Saharia, William Chan, Huiwen Chang, Chris Lee, Jonathan Ho, Tim Salimans, David Fleet, and Mohammad Norouzi.
\newblock Palette: {I}mage-to-image diffusion models.
\newblock In {\em ACM SIGGRAPH 2022 Conference Proceedings}, pages 1--10, 2022.

\bibitem[SCW{\etalchar{+}}15]{shi2015convolutional}
Xingjian Shi, Zhourong Chen, Hao Wang, Dit-Yan Yeung, Wai-Kin Wong, and Wang-chun Woo.
\newblock Convolutional {LSTM} network: {A} machine learning approach for precipitation nowcasting.
\newblock {\em Advances in neural information processing systems}, 28, 2015.

\bibitem[SDP19]{shishegar2019integrated}
Shadab Shishegar, Sophie Duchesne, and Genevi{\`e}ve Pelletier.
\newblock An integrated optimization and rule-based approach for predictive real time control of urban stormwater management systems.
\newblock {\em Journal of Hydrology}, 577:124000, 2019.

\bibitem[SEH{\etalchar{+}}20]{sonderby2020metnet}
Casper~Kaae S{\o}nderby, Lasse Espeholt, Jonathan Heek, Mostafa Dehghani, Avital Oliver, Tim Salimans, Shreya Agrawal, Jason Hickey, and Nal Kalchbrenner.
\newblock Met{N}et: {A} neural weather model for precipitation forecasting.
\newblock {\em arXiv preprint arXiv:2003.12140}, 2020.

\bibitem[SGB{\etalchar{+}}19]{sadler2019leveraging}
Jeffrey~M Sadler, Jonathan~L Goodall, Madhur Behl, Mohamed~M Morsy, Teresa~B Culver, and Benjamin~D Bowes.
\newblock Leveraging open source software and parallel computing for model predictive control of urban drainage systems using {EPA-SWMM5}.
\newblock {\em Environmental Modelling \& Software}, 120:104484, 2019.

\bibitem[SGB{\etalchar{+}}20]{sadler2020exploring}
Jeffrey~M Sadler, Jonathan~L Goodall, Madhur Behl, Benjamin~D Bowes, and Mohamed~M Morsy.
\newblock Exploring real-time control of stormwater systems for mitigating flood risk due to sea level rise.
\newblock {\em Journal of Hydrology}, 583:124571, 2020.

\bibitem[SGMS18]{sadler2018modeling}
JM~Sadler, JL~Goodall, MM~Morsy, and K~Spencer.
\newblock Modeling urban coastal flood severity from crowd-sourced flood reports using {P}oisson regression and {R}andom {F}orest.
\newblock {\em Journal of hydrology}, 559:43--55, 2018.

\bibitem[SGT{\etalchar{+}}08]{scarselli2008graph}
Franco Scarselli, Marco Gori, Ah~Chung Tsoi, Markus Hagenbuchner, and Gabriele Monfardini.
\newblock The graph neural network model.
\newblock {\em IEEE transactions on neural networks}, 20(1):61--80, 2008.

\bibitem[SHT{\etalchar{+}}26]{shi2026multicube}
Jimeng Shi, Wei Hu, Runchu Tian, Bowen Jin, Wonbin Kweon, SeongKu Kang, Yunfan Kang, Dingqi Ye, Sizhe Zhou, Shaowen Wang, et~al.
\newblock Multicube-rag for multi-hop question answering.
\newblock {\em arXiv preprint arXiv:2602.15898}, 2026.

\bibitem[SJHN24]{shi2024codicast}
Jimeng Shi, Bowen Jin, Jiawei Han, and Giri Narasimhan.
\newblock Co{D}i{C}ast: Conditional diffusion model for weather prediction with uncertainty quantification.
\newblock {\em arXiv preprint arXiv:2409.05975}, 2024.

\bibitem[SJN22]{shi2022time}
Jimeng Shi, Mahek Jain, and Giri Narasimhan.
\newblock Time series forecasting using various deep learning models.
\newblock {\em International Journal of Computer and Systems Engineering}, 16(6):224--232, 2022.

\bibitem[SK21]{sachdeva2021comparison}
Shruti Sachdeva and Bijendra Kumar.
\newblock Comparison of gradient boosted decision trees and random forest for groundwater potential mapping in {D}holpur ({R}ajasthan), {I}ndia.
\newblock {\em Stochastic Environmental Research and Risk Assessment}, 35(2):287--306, 2021.

\bibitem[SLG{\etalchar{+}}24]{shen2024retrieval}
Tao Shen, Guodong Long, Xiubo Geng, Chongyang Tao, Yibin Lei, Tianyi Zhou, Michael Blumenstein, and Daxin Jiang.
\newblock Retrieval-augmented retrieval: Large language models are strong zero-shot retriever.
\newblock In {\em Findings of the Association for Computational Linguistics ACL 2024}, pages 15933--15946, 2024.

\bibitem[SME20]{song2020denoising}
Jiaming Song, Chenlin Meng, and Stefano Ermon.
\newblock Denoising diffusion implicit models.
\newblock {\em arXiv preprint arXiv:2010.02502}, 2020.

\bibitem[SMK{\etalchar{+}}22]{siddique2022survey}
Talha Siddique, Md~Shaad Mahmud, Amy~M Keesee, Chigomezyo~M Ngwira, and Hyunju Connor.
\newblock A survey of uncertainty quantification in machine learning for space weather prediction.
\newblock {\em Geosciences}, 12(1):27, 2022.

\bibitem[SMRFK{\etalchar{+}}20]{sapitang2020machine}
Michelle Sapitang, Wanie M.~Ridwan, Khairul Faizal~Kushiar, Ali Najah~Ahmed, and Ahmed El-Shafie.
\newblock Machine learning application in reservoir water level forecasting for sustainable hydropower generation strategy.
\newblock {\em Sustainability}, 12(15):6121, 2020.

\bibitem[SMS{\etalchar{+}}23]{shi2023explainable}
Jimeng Shi, Rukmangadh Myana, Vitalii Stebliankin, Azam Shirali, and Giri Narasimhan.
\newblock Explainable parallel {RCNN} with novel feature representation for time series forecasting.
\newblock In {\em International Workshop on Advanced Analytics and Learning on Temporal Data}, pages 56--75, Torino, Italy, 2023. Springer.

\bibitem[SP11]{slingo2011uncertainty}
Julia Slingo and Tim Palmer.
\newblock Uncertainty in weather and climate prediction.
\newblock {\em Philosophical Transactions of the Royal Society A: Mathematical, Physical and Engineering Sciences}, 369(1956):4751--4767, 2011.

\bibitem[SPW{\etalchar{+}}23]{shaikh2023application}
Arbaaz~Aziz Shaikh, Azazkhan~Ibrahimkhan Pathan, Sahita~Ibopishak Waikhom, Prasit~Girish Agnihotri, Md~Nazrul Islam, and Sudhir~Kumar Singh.
\newblock Application of latest {HEC-RAS} version 6 for 2{D} hydrodynamic modeling through {GIS} framework: {A} case study from coastal urban floodplain in {I}ndia.
\newblock {\em Modeling Earth Systems and Environment}, 9(1):1369--1385, 2023.

\bibitem[SRT{\etalchar{+}}24]{schmude2024prithvi}
Johannes Schmude, Sujit Roy, Will Trojak, Johannes Jakubik, Daniel~Salles Civitarese, Shraddha Singh, Julian Kuehnert, Kumar Ankur, Aman Gupta, Christopher~E Phillips, et~al.
\newblock Prithvi {WxC}: {F}oundation model for weather and climate.
\newblock {\em arXiv preprint arXiv:2409.13598}, 2024.

\bibitem[SS19]{sun2019can}
Alexander~Y Sun and Bridget~R Scanlon.
\newblock How can big data and machine learning benefit environment and water management: a survey of methods, applications, and future directions.
\newblock {\em Environmental Research Letters}, 14(7):073001, 2019.

\bibitem[SSJ{\etalchar{+}}25]{shi2025deep}
J~Shi, A~Shirali, B~Jin, S~Zhou, W~Hu, R~Rangaraj, S~Wang, J~Han, Z~Wang, U~Lall, et~al.
\newblock Deep learning and foundation models for weather prediction: a survey (2025).
\newblock {\em arXiv preprint arXiv:2501.06907}, 2025.

\bibitem[SSN23]{shi2023power}
Jimeng Shi, Vitalii Stebliankin, and Giri Narasimhan.
\newblock The power of explainability in forecast-informed deep learning models for flood mitigation.
\newblock {\em arXiv preprint arXiv:2310.19166}, 2023.

\bibitem[SSP24]{saleem2024conformer}
Hira Saleem, Flora Salim, and Cormac Purcell.
\newblock Conformer: {E}mbedding continuous attention in vision transformer for weather forecasting.
\newblock {\em arXiv preprint arXiv:2402.17966}, 2024.

\bibitem[SSW{\etalchar{+}}23]{shi2023graph}
Jimeng Shi, Vitalii Stebliankin, Zhaonan Wang, Shaowen Wang, and Giri Narasimhan.
\newblock Graph transformer network for flood forecasting with heterogeneous covariates.
\newblock {\em arXiv preprint arXiv:2310.07631}, 2023.

\bibitem[SSX{\etalchar{+}}26]{sun2026retrieval}
Jiashuo Sun, Jimeng Shi, Yixuan Xie, Saizhuo Wang, Jash~Rajesh Parekh, Pengcheng Jiang, Zhiyi Shi, Jiajun Fan, Qinglong Zheng, Peiran Li, et~al.
\newblock Retrieval is cheap, show me the code: Executable multi-hop reasoning for retrieval-augmented generation.
\newblock {\em arXiv preprint arXiv:2605.12975}, 2026.

\bibitem[STS16]{singh2016review}
Amanpreet Singh, Narina Thakur, and Aakanksha Sharma.
\newblock A review of supervised machine learning algorithms.
\newblock In {\em 2016 3rd international conference on computing for sustainable global development (INDIACom)}, pages 1310--1315. Ieee, 2016.

\bibitem[SVDM95]{suykens1995artificial}
Johan~AK Suykens, Joos~PL Vandewalle, and Bart~L De~Moor.
\newblock {\em Artificial neural networks for modelling and control of non-linear systems}.
\newblock Springer Science \& Business Media, 1995.

\bibitem[SVS{\etalchar{+}}25]{shirzaei2025aging}
Manoochehr Shirzaei, Farshid Vahedifard, Nitheshnirmal Sadhasivam, Leonard Ohenhen, Oluwaseyi Dasho, Ashutosh Tiwari, Susanna Werth, Mohammed Azhar, Yunxia Zhao, Robert~J Nicholls, et~al.
\newblock Aging dams, political instability, poor human decisions and climate change: recipe for human disaster.
\newblock {\em npj Natural Hazards}, 2(1):5, 2025.

\bibitem[SXS{\etalchar{+}}26]{sun2026tasr}
Jiashuo Sun, Yixuan Xie, Jimeng Shi, Shaowen Wang, and Jiawei Han.
\newblock Tasr-rag: Taxonomy-guided structured reasoning for retrieval-augmented generation.
\newblock {\em arXiv preprint arXiv:2603.09341}, 2026.

\bibitem[SYL{\etalchar{+}}24]{shi2024fidlar}
Jimeng Shi, Zeda Yin, Arturo Leon, Jayantha Obeysekera, and Giri Narasimhan.
\newblock {FIDLAR}: {F}orecast-informed deep learning architecture for flood mitigation.
\newblock {\em arXiv preprint arXiv:2402.13371}, 2024.

\bibitem[SYM{\etalchar{+}}23]{shi2023deep}
Jimeng Shi, Zeda Yin, Rukmangadh Myana, Khandker Ishtiaq, Anupama John, Jayantha Obeysekera, Arturo Leon, and Giri Narasimhan.
\newblock Deep learning models for water stage predictions in {S}outh {F}lorida.
\newblock {\em arXiv preprint arXiv:2306.15907}, 2023.

\bibitem[SZJ{\etalchar{+}}25]{shi2025hypercube}
Jimeng Shi, Sizhe Zhou, Bowen Jin, Wei Hu, Shaowen Wang, Giri Narasimhan, and Jiawei Han.
\newblock Hypercube-{RAG}: {H}ypercube-based retrieval-augmented generation for in-domain scientific question-answering.
\newblock {\em arXiv preprint arXiv:2505.19288}, 2025.

\bibitem[Tan20]{tangirala2020evaluating}
Suryakanthi Tangirala.
\newblock Evaluating the impact of {GINI} index and information gain on classification using decision tree classifier algorithm.
\newblock {\em International Journal of Advanced Computer Science and Applications}, 11(2):612--619, 2020.

\bibitem[TDT{\etalchar{+}}24]{tang2024vmrnn}
Yujin Tang, Peijie Dong, Zhenheng Tang, Xiaowen Chu, and Junwei Liang.
\newblock {VMRNN}: {I}ntegrating {V}ision {M}amba and {LSTM} for efficient and accurate spatiotemporal forecasting.
\newblock In {\em Proceedings of the IEEE/CVF Conference on Computer Vision and Pattern Recognition}, pages 5663--5673, 2024.

\bibitem[TPB14]{trotman2014improvements}
Andrew Trotman, Antti Puurula, and Blake Burgess.
\newblock Improvements to {BM}25 and language models examined.
\newblock In {\em Proceedings of the 19th Australasian Document Computing Symposium}, pages 58--65, 2014.

\bibitem[TZC{\etalchar{+}}18]{tao2018doc2cube}
Fangbo Tao, Chao Zhang, Xiusi Chen, Meng Jiang, Tim Hanratty, Lance Kaplan, and Jiawei Han.
\newblock Doc2cube: {A}llocating documents to text cube without labeled data.
\newblock In {\em 2018 IEEE International Conference on Data Mining (ICDM)}, pages 1260--1265. IEEE, 2018.

\bibitem[VBMM22]{vadyala2022review}
Shashank~Reddy Vadyala, Sai~Nethra Betgeri, John~C Matthews, and Elizabeth Matthews.
\newblock A review of physics-based machine learning in civil engineering.
\newblock {\em Results in Engineering}, 13:100316, 2022.

\bibitem[VH24]{van2024economic}
George Van~Houtven.
\newblock Economic value of flood forecasts and early warning systems: {A} review.
\newblock {\em Natural Hazards Review}, 25(4):03124002, 2024.

\bibitem[VHG24]{verma2024climode}
Yogesh Verma, Markus Heinonen, and Vikas Garg.
\newblock Clim{ODE}: {C}limate and weather forecasting with physics-informed neural odes.
\newblock {\em arXiv preprint arXiv:2404.10024}, 2024.

\bibitem[VM19]{vinge2019understanding}
Rikard Vinge and Tomas McKelvey.
\newblock Understanding support vector machines with polynomial kernels.
\newblock In {\em 2019 27th European signal processing conference (EUSIPCO)}, pages 1--5. IEEE, 2019.

\bibitem[VMWW18]{vermuyten2018combining}
Evert Vermuyten, Pieter Meert, Vincent Wolfs, and Patrick Willems.
\newblock Combining model predictive control with a reduced genetic algorithm for real-time flood control.
\newblock {\em Journal of Water Resources Planning and Management}, 144(2):04017083, 2018.

\bibitem[VSP{\etalchar{+}}17]{vaswani2017attention}
Ashish Vaswani, Noam Shazeer, Niki Parmar, Jakob Uszkoreit, Llion Jones, Aidan~N Gomez, {\L}ukasz Kaiser, and Illia Polosukhin.
\newblock Attention is all you need.
\newblock {\em Advances in neural information processing systems}, 30, 2017.

\bibitem[WCW{\etalchar{+}}24]{wu2024weathergnn}
Binqing Wu, Weiqi Chen, Wengwei Wang, Bingqing Peng, Liang Sun, and Ling Chen.
\newblock Weather{GNN}: {E}xploiting meteo-and spatial-dependencies for local numerical weather prediction bias-correction.
\newblock In {\em Proceedings of the International Joint Conference on Artificial Intelligence}, pages 2433--2441, 2024.

\bibitem[WFL23]{wang2023physical}
Rui Wang, Jimmy~CH Fung, and Alexis~KH Lau.
\newblock Physical-{D}ynamic-{D}riven {AI}-synthetic precipitation nowcasting using task-segmented generative model.
\newblock {\em Geophysical Research Letters}, 50(21):e2023GL106084, 2023.

\bibitem[WHC{\etalchar{+}}24]{wang2024graph}
Shijie Wang, Jiani Huang, Zhikai Chen, Yu~Song, Wenzhuo Tang, Haitao Mao, Wenqi Fan, Hui Liu, Xiaorui Liu, Dawei Yin, et~al.
\newblock Graph machine learning in the era of large language models (llms).
\newblock {\em ACM Transactions on Intelligent Systems and Technology}, 2024.

\bibitem[WJH{\etalchar{+}}23]{wang2023geospatial}
Zhaonan Wang, Bowen Jin, Wei Hu, Minhao Jiang, Seungyeon Kang, Zhiyuan Li, Sizhe Zhou, Jiawei Han, and Shaowen Wang.
\newblock Geospatial knowledge hypercube.
\newblock In {\em Proceedings of the 31st ACM International Conference on Advances in Geographic Information Systems}, pages 1--4, 2023.

\bibitem[WLL{\etalchar{+}}25]{wang2025remflow}
Gelan Wang, Yu~Liu, Shukai Liu, Ling Zhang, and Liqun Yang.
\newblock {REMFLOW}: {RAG}-enhanced multi-factor rainfall flooding warning in sponge airports via large language model.
\newblock {\em International Journal of Machine Learning and Cybernetics}, pages 1--21, 2025.

\bibitem[WWZ{\etalchar{+}}22]{wang2022predrnn}
Yunbo Wang, Haixu Wu, Jianjin Zhang, Zhifeng Gao, Jianmin Wang, S~Yu Philip, and Mingsheng Long.
\newblock Pred{RNN}: {A} recurrent neural network for spatiotemporal predictive learning.
\newblock {\em IEEE Transactions on Pattern Analysis and Machine Intelligence}, 45(2):2208--2225, 2022.

\bibitem[WYH{\etalchar{+}}24]{wang2024multilingual}
Liang Wang, Nan Yang, Xiaolong Huang, Linjun Yang, Rangan Majumder, and Furu Wei.
\newblock Multilingual e5 text embeddings: {A} technical report.
\newblock {\em arXiv preprint arXiv:2402.05672}, 2024.

\bibitem[WYW{\etalchar{+}}22]{wang2022text}
Yi~Wang, Wenhao Yu, Yining Wang, Chenguang Li, Soroush Liu, and Hoifung Poon.
\newblock Text embeddings by weakly-supervised contrastive pre-training.
\newblock {\em arXiv preprint arXiv:2212.03533}, 2022.

\bibitem[WZP20]{wang2020recent}
Xizhao Wang, Yanxia Zhao, and Farhad Pourpanah.
\newblock Recent advances in deep learning.
\newblock {\em International Journal of Machine Learning and Cybernetics}, 11:747--750, 2020.

\bibitem[WZZ{\etalchar{+}}22]{wen2022transformers}
Qingsong Wen, Tian Zhou, Chaoli Zhang, Weiqi Chen, Ziqing Ma, Junchi Yan, and Liang Sun.
\newblock Transformers in time series: {A} survey.
\newblock {\em arXiv preprint arXiv:2202.07125}, 2022.

\bibitem[XBX{\etalchar{+}}23]{xiao2023fengwu}
Yi~Xiao, Lei Bai, Wei Xue, Kang Chen, Tao Han, and Wanli Ouyang.
\newblock Fengwu-4{DV}ar: {C}oupling the data-driven weather forecasting model with 4{D} variational assimilation.
\newblock {\em arXiv preprint arXiv:2312.12455}, 2023.

\bibitem[XLC{\etalchar{+}}21]{xu2021artificial}
Yongjun Xu, Xin Liu, Xin Cao, Changping Huang, Enke Liu, Sen Qian, Xingchen Liu, Yanjun Wu, Fengliang Dong, Cheng-Wei Qiu, et~al.
\newblock Artificial intelligence: {A} powerful paradigm for scientific research.
\newblock {\em The Innovation}, 2(4), 2021.

\bibitem[XQS{\etalchar{+}}24]{xu2024pfformer}
Luwen Xu, Jiwei Qin, Dezhi Sun, Yuanyuan Liao, and Jiong Zheng.
\newblock {PF}former: {A} time-series forecasting model for short-term precipitation forecasting.
\newblock {\em IEEE Access}, 2024.

\bibitem[XXL{\etalchar{+}}21]{xiong2020approximate}
Lee Xiong, Chenyan Xiong, Ye~Li, Kwok-Fung Tang, Jialin Liu, Paul~N. Bennett, Junaid Ahmed, and Arnold Overwijk.
\newblock Approximate nearest neighbor negative contrastive learning for dense text retrieval.
\newblock In {\em International Conference on Learning Representations}, 2021.

\bibitem[YBH{\etalchar{+}}23]{yin2023physic}
Zeda Yin, Linglong Bian, Beichao Hu, Jimeng Shi, and Arturo~S Leon.
\newblock Physic-informed neural network approach coupled with boundary conditions for solving 1{D} steady shallow water equations for riverine system.
\newblock In {\em World Environmental and Water Resources Congress 2023}, pages 280--288, 2023.

\bibitem[YC20]{yang2020regional}
Shun-Nien Yang and Li-Chiu Chang.
\newblock Regional inundation forecasting using machine learning techniques with the internet of things.
\newblock {\em Water}, 12(6):1578, 2020.

\bibitem[YGC{\etalchar{+}}24]{yang2024multi}
Qidong Yang, Jonathan Giezendanner, Daniel~Salles Civitarese, Johannes Jakubik, Eric Schmitt, Anirban Chandra, Jeremy Vila, Detlef Hohl, Chris Hill, Campbell Watson, et~al.
\newblock Multi-modal graph neural networks for localized off-grid weather forecasting.
\newblock {\em arXiv preprint arXiv:2410.12938}, 2024.

\bibitem[YJC{\etalchar{+}}18]{yan2018urban}
Jun Yan, Jiaming Jin, Furong Chen, Guo Yu, Hailong Yin, and Wenjia Wang.
\newblock Urban flash flood forecast using support vector machine and numerical simulation.
\newblock {\em Journal of Hydroinformatics}, 20(1):221--231, 2018.

\bibitem[YLY{\etalchar{+}}24]{yu2024diffcast}
Demin Yu, Xutao Li, Yunming Ye, Baoquan Zhang, Chuyao Luo, Kuai Dai, Rui Wang, and Xunlai Chen.
\newblock Diffcast: {A} unified framework via residual diffusion for precipitation nowcasting.
\newblock In {\em Proceedings of the IEEE/CVF Conference on Computer Vision and Pattern Recognition}, pages 27758--27767, 2024.

\bibitem[YSB{\etalchar{+}}25]{yin2025physics}
Zeda Yin, Jimeng Shi, Linlong Bian, William~H Campbell, Sumit~R Zanje, Beichao Hu, and Arturo~S Leon.
\newblock Physics-informed neural network approach for solving the one-dimensional unsteady shallow-water equations in riverine systems.
\newblock {\em Journal of Hydraulic Engineering}, 151(1):04024060, 2025.

\bibitem[YSH{\etalchar{+}}24]{yin2024fast}
Zeda Yin, Yasaman Saadati, Beichao Hu, Arturo~S Leon, M~Hadi Amini, and Dwayne McDaniel.
\newblock Fast high-fidelity flood inundation map generation by super-resolution techniques.
\newblock {\em Journal of Hydroinformatics}, 26(1):319--336, 2024.

\bibitem[YWMZ25]{yuan2025tianxing}
Shijin Yuan, Guansong Wang, Bin Mu, and Feifan Zhou.
\newblock Tianxing: {A} linear complexity transformer model with explicit attention decay for global weather forecasting.
\newblock {\em Advances in Atmospheric Sciences}, 42(1):9--25, 2025.

\bibitem[YYY{\etalchar{+}}24]{yin2024strategic}
Jie Yin, Yuhan Yang, Dapeng Yu, Ning Lin, Robert Wilby, Stuart Lane, Bindong Sun, Jeremy Bricker, Nigel Wright, Lili Yang, et~al.
\newblock Strategic storm flood evacuation planning for large coastal cities enables more effective transfer of elderly populations.
\newblock {\em Nature Water}, pages 1--11, 2024.

\bibitem[YYZ17]{yu2017spatio}
Bing Yu, Haoteng Yin, and Zhanxing Zhu.
\newblock Spatio-temporal graph convolutional networks: {A} deep learning framework for traffic forecasting.
\newblock {\em arXiv preprint arXiv:1709.04875}, 2017.

\bibitem[ZBHB{\etalchar{+}}21]{zarei2021machine}
Manizhe Zarei, Omid Bozorg-Haddad, Sahar Baghban, Mohammad Delpasand, Erfan Goharian, and Hugo~A Lo{\'a}iciga.
\newblock Machine-learning algorithms for forecast-informed reservoir operation ({FIRO}) to reduce flood damages.
\newblock {\em Scientific reports}, 11(1):24295, 2021.

\bibitem[ZBN{\etalchar{+}}24]{zhao2024omg}
Pengcheng Zhao, Jiang Bian, Zekun Ni, Weixin Jin, Jonathan Weyn, Zuliang Fang, Siqi Xiang, Haiyu Dong, Bin Zhang, Hongyu Sun, et~al.
\newblock {OMG-HD}: {A} high-resolution {AI} weather model for end-to-end forecasts from observations.
\newblock {\em arXiv preprint arXiv:2412.18239}, 2024.

\bibitem[ZD20]{zhang2020temperature}
Zao Zhang and Yuan Dong.
\newblock Temperature forecasting via convolutional recurrent neural networks based on time-series data.
\newblock {\em Complexity}, 2020:1--8, 2020.

\bibitem[ZF23]{zeleny2023multi}
Ond{\v{r}}ej Zelen{\`y} and Tomas Fryza.
\newblock Multi-branch multi layer perceptron: {A} solution for precise regression using machine learning.
\newblock In {\em 2023 33rd International Conference Radioelektronika (RADIOELEKTRONIKA)}, pages 1--5. IEEE, 2023.

\bibitem[Zha16]{zhang2016introduction}
Zhongheng Zhang.
\newblock Introduction to machine learning: k-nearest neighbors.
\newblock {\em Annals of translational medicine}, 4(11):218, 2016.

\bibitem[ZJM{\etalchar{+}}24]{zhang2024comprehensive}
Yang Zhang, Hanlei Jin, Dan Meng, Jun Wang, and Jinghua Tan.
\newblock A comprehensive survey on process-oriented automatic text summarization with exploration of llm-based methods.
\newblock {\em arXiv preprint arXiv:2403.02901}, 2024.

\bibitem[ZKW{\etalchar{+}}20]{zhang2019bertscore}
Tianyi Zhang*, Varsha Kishore*, Felix Wu*, Kilian~Q. Weinberger, and Yoav Artzi.
\newblock {BERTS}core: Evaluating text generation with {BERT}.
\newblock In {\em International Conference on Learning Representations}, 2020.

\bibitem[ZLC{\etalchar{+}}23]{zhang2023skilful}
Yuchen Zhang, Mingsheng Long, Kaiyuan Chen, Lanxiang Xing, Ronghua Jin, Michael~I Jordan, and Jianmin Wang.
\newblock Skilful nowcasting of extreme precipitation with nowcastnet.
\newblock {\em Nature}, 619(7970):526--532, 2023.

\bibitem[ZLC{\etalchar{+}}25]{zheng2025sf}
Xu~Zheng, Chaohao Lin, Sipeng Chen, Zhuomin Chen, Jimeng Shi, Wei Cheng, Jayantha Obeysekera, Jason Liu, and Dongsheng Luo.
\newblock Sf2bench: Evaluating data-driven models for compound flood forecasting in south florida.
\newblock {\em arXiv preprint arXiv:2506.04281}, 2025.

\bibitem[ZM13]{zamanlooy2013efficient}
Babak Zamanlooy and Mitra Mirhassani.
\newblock Efficient vlsi implementation of neural networks with hyperbolic tangent activation function.
\newblock {\em IEEE Transactions on Very Large Scale Integration (VLSI) Systems}, 22(1):39--48, 2013.

\bibitem[ZRA23]{zhang2023adding}
Lvmin Zhang, Anyi Rao, and Maneesh Agrawala.
\newblock Adding conditional control to text-to-image diffusion models.
\newblock In {\em Proceedings of the IEEE/CVF International Conference on Computer Vision}, pages 3836--3847, 2023.

\bibitem[ZWNW21]{zhou2021rapid}
Yuerong Zhou, Wenyan Wu, Rory Nathan, and Quan~J Wang.
\newblock A rapid flood inundation modelling framework using deep learning with spatial reduction and reconstruction.
\newblock {\em Environmental Modelling \& Software}, 143:105112, 2021.

\bibitem[ZXW{\etalchar{+}}24]{zhu2024foundations}
Xiao~Xiang Zhu, Zhitong Xiong, Yi~Wang, Adam~J Stewart, Konrad Heidler, Yuanyuan Wang, Zhenghang Yuan, Thomas Dujardin, Qingsong Xu, and Yilei Shi.
\newblock On the foundations of earth and climate foundation models.
\newblock {\em arXiv preprint arXiv:2405.04285}, 2024.

\bibitem[ZYW{\etalchar{+}}23]{zhuang2023toolqa}
Yuchen Zhuang, Yue Yu, Kuan Wang, Haotian Sun, and Chao Zhang.
\newblock Toolqa: A dataset for llm question answering with external tools.
\newblock {\em Advances in Neural Information Processing Systems}, 36:50117--50143, 2023.

\bibitem[ZYX{\etalchar{+}}25]{zhang2025teleclass}
Yunyi Zhang, Ruozhen Yang, Xueqiang Xu, Rui Li, Jinfeng Xiao, Jiaming Shen, and Jiawei Han.
\newblock Teleclass: {T}axonomy enrichment and llm-enhanced hierarchical text classification with minimal supervision.
\newblock In {\em Proceedings of the ACM on Web Conference 2025}, pages 2032--2042, 2025.

\end{thebibliography}


\begin{vita}
\begin{center}
JIMENG SHI \\[4ex]
\end{center}

\noindent
\begin{tabular}{ll}

2021.08-present \hspace{2in}  & Graduate Research Assistant \\
                              & Florida International University \\
		                   & Miami, FL, USA \\[2ex]

2021.08-present \hspace{2in}  & Researcher (Remote) \\
                              & NSF I-GUIDE, \\
		                   & Urbana Champaign, IL, USA \\[2ex]
              
2019.01-2021.05               & M.S., Computer Science \\
                              & Florida International University \\
		                   & Miami, FL, USA \\[2ex]

2014.09-2018.07               & B.S., Mechanical Engineering \\
	                       & Tianjin U. of Science and Technology \\
		                   & Tianjin, China \\[2ex]
                           
2024.05-2024.08               & Momentum Research Fellow \\
                              & NSF LEAP, Columbia University \\
		                   & New York City, NY, USA \\[2ex]

2020.09-2020.12               & Lid Vizion Company  \\
                              & Miami, FL, USA \\ \\

\end{tabular}

\noindent
PUBLICATIONS ($^*$ represents equal contribution.) \\[1.5ex]

\noindent
Jimeng Shi, Zeda Yin, Arturo Leon, Jayantha Obeysekera, Giri Narasimhan.
{\em FIDLAR: Forecast-Informed Deep Learning Approaches for Flood Mitigation.}
In Proceedings of the AAAI Conference on Artificial Intelligence, vol. 39, no. 27, pp. 28377-28385. 2025.  \\[1.2ex]
Jimeng Shi, Bowen Jin, Jiawei Han, Sundararaman Gopalakrishnan, Giri Narasimhan.
{\em CoDiCast: Conditional Diffusion Model for Global Weather Prediction with Uncertainty Quantification.}
The 34th International Joint Conference on Artificial Intelligence. 2025. (To appear) \\[1.2ex]
Jimeng Shi, Azam Shirali, Giri Narasimhan. 
{\em ReFine: Boosting Time Series Prediction of Extreme Events by Reweighting and Fine-tuning}. In Proceedings of 2024 IEEE International Conference on Big Data (BigData), pp. 1450-1457. IEEE, 2024. \\[1.2ex]
Jimeng Shi$^*$, Zeda Yin$^*$, Rukmangadh Myana, Khandker Ishtiaq, Anupama John, Jayantha Obeysekera, Arturo Leon, Giri Narasimhan.
Journal of Water Resources Planning and Management (In press). 2025.  \\[1.2ex]
Jimeng Shi, Sizhe Zhou, Bowen Jin, Wei Hu, Shaowen Wang, Giri Narasimhan, Jiawei Han. {\em Hypercube-RAG: Hypercube Retrieval-Augmented Generation for In-domain Scientific Question-Answering}. arXiv preprint, 2025. (Under review)  \\[1.2ex]
Jimeng Shi, Rukmangadh Myana, Vitalii Stebliankin, Azam Shirali, Giri Narasimhan.
{\em Explainable Parallel RCNN with Novel Feature Representation for Time Series Forecasting}.
In International Workshop on Advanced Analytics and Learning on Temporal Data, pp. 56-75. 2023. \\[1.2ex]
Jimeng Shi, Azam Shirali, Bowen Jin, Sizhe Zhou, Wei Hu, Rahuul Rangaraj, Shaowen Wang, Jiawei Han, Zhaonan Wang, Upmanu Lall, Yanzhao Wu, Leonardo Bobadilla, Giri Narasimhan. 
{\em Deep Learning and Foundation Models for Weather Prediction: A Survey}. arXiv preprint. 2025. (Under review) \\[1.2ex]
Jimeng Shi, Mahek Jain, Giri Narasimhan.
{\em Time Series Forecasting (TSF) Using Various Deep Learning Models.}
International Conference of Machine Learning Applications (ICMLA) 2022. \\[1.2ex]
Rahuul Rangaraj$^*$, Jimeng Shi$^*$, Azam Shirali, Rajendra Paudel, Yanzhao Wu, and Giri Narasimhan. {\em Retrieval-Augmented Water Level Forecasting for Everglades}. arXiv preprint. 2025. (Under review) \\[1.2ex]
Rahuul Rangaraj$^*$, Jimeng Shi$^*$, Azam Shirali, Rajendra Paudel, Yanzhao Wu, and Giri Narasimhan. 
{\em How Effective are Large Time Series Models in Hydrology? A Study on Water Level Forecasting in Everglades}. 
arXiv preprint. 2025. (Under review) \\[1.2ex]
Vitalii Stebliankin, Azam Shirali, Prabin Baral, Jimeng Shi, Prem Chapagain, Kalai Mathee, Giri Narasimhan.
{\em Evaluating protein binding interfaces with transformer networks.}
Nature Machine Intelligence, 5(9), 1042-1053. 2023.  \\[1.2ex]
Azam Shirali, Vitalii Stebliankin, Jimeng Shi, Giri Narasimhan.
{\em A Comprehensive Survey of Scoring Functions for Protein Docking Models.}
BMC bioinformatics, 26(1), 25. 2025. \\[1.2ex]
Zichuan Liu, Tianchun Wang, Jimeng Shi, Xu Zheng, Zhuomin Chen, Lei Song, Wenqian Dong, Jayantha Obeysekera, Farhad Shirani, Dongsheng Luo.
{\em TimeX++: Learning Time-Series Explanations with Information Bottleneck}.
In Proceedings of the 41st International Conference on Machine Learning, no. 1297, pp. 32062-32082. 2024. \\[1.2ex]
Zeda Yin, Jimeng Shi, Linlong Bian, William Campbell, Sumit Zanje, Arturo Leon.
{\em Physics-Informed Neural Network Approach for Solving the One-Dimensional Unsteady Shallow-Water Equations in Riverine Systems}.
Journal of Hydraulic Engineering, 151(1), 04024060. 2025. \\[1.2ex]
Zeda Yin, Linlong Bian, Beichao Hu, Jimeng Shi, Arturo Leon.
{\em Physic-Informed Neural Network Approach Coupled with Boundary Conditions for Solving 1D Steady Shallow Water Equations for Riverine System}.
In World Environmental and Water Resources Congress 2023, pp. 280-288. 2023. 
\\

\noindent
PRESENTATIONS ($^*$ represents equal contribution.) \\[1ex]

\noindent
Jimeng Shi et al., {\em \hyperrag: Hypercube Retrieval-augmented Generation for In-domain Scientific Question-Answering} at the I-GUIDE Forum, Chicago, June 2025. \\[1.2ex]
Jimeng Shi et al., {\em \hyperrag: Hypercube Retrieval-augmented Generation for In-domain Scientific Question-Answering} at the startup company (GRaiL), Miami, May 2025. \\[1.2ex]
Jimeng Shi et al., {\em \fidlar: Forecast-Informed Deep Learning Approaches for Flood Mitigation} at Climate AI Journal Club, Florida International University, Miami, March 2025. \\[1.2ex]
Jimeng Shi et al., {\em \fidlar: Forecast-Informed Deep Learning Approaches for Flood Mitigation} at AAAI'25, Philadelphia, Pennsylvania, Feb. 2025. \\[1.2ex]
Jimeng Shi et al., {\em AI4Science: Coastal Flood Management and Global Weather Prediction} at Hurricane Research Division/AOML/NOAA, Miami, Feb. 2025. \\[1.2ex]
Jimeng Shi et al., {\em ReFine: Boosting Time Series Prediction of Extreme Events by Reweighting and Fine-tuning} at IEEE International Conference on Big Data, Washington DC, Dec. 2024. \\[1.2ex]
Jimeng Shi et al., {\em Explainable Deep Learning Models for Flood Prediction and Mitigation} and {\em generative diffusion model for global weather forecast} at Columbia Water Center, Columbia University, July 2024. \\[1.2ex]
Jimeng Shi et al., {\em The Power of Explainability in Forecast-Informed Deep Learning Models for Flood Mitigation} at
NeurIPS 2023 workshop on Climate Change AI.  \\[1.2ex]
Jimeng Shi et al., {\em Graph Transformer Network for Flood Forecast in A Coastal System} at NSF I-GUIDE Forum, New York City, Oct. 2023.  \\[1.2ex]
Jimeng Shi et al., {\em Deep Learning Models for Water Level Prediction} at I-GUIDE All Hands Meeting (AHM), Chicago, Sep. 2022.  \\[1.2ex]
Jimeng Shi et al., {\em Time Series Forecasting (TSF) Using Various Deep Learning Models} at International Conference of Machine Learning Applications (ICMLA) 2022.  \\[1.2ex]

\end{vita}

\end{document}